\documentclass[twoside,11pt]{report}
\usepackage{marginnote}
\usepackage[left=1.1in,right=1.1in,top=1.1in,bottom=1.1in]{geometry}
\usepackage[usenames,dvipsnames,table]{xcolor}
\usepackage{graphicx,amsmath,amssymb,multirow,booktabs,color,verbatim,cite,url,placeins}
\usepackage{array,wrapfig}
\usepackage{xcolor}
\usepackage[normalem]{ulem}
\usepackage{tgpagella}
\usepackage{afterpage}
\usepackage[nottoc]{tocbibind}
\usepackage[makeroom]{cancel}
\usepackage[linktocpage]{hyperref}
\hypersetup{
    colorlinks=true,
    citecolor=teal,
    linkcolor=teal,
    filecolor=magenta
}

\usepackage{pifont}
\usepackage{enumitem}
\usepackage{algorithm, algpseudocode}
\usepackage{multirow}
\usepackage{bbm}
\usepackage[utf8]{inputenc} %
\usepackage[T1]{fontenc}
\usepackage{xspace}
\usepackage[titles]{tocloft}

\newcommand{\eg}{e.g.}
\newcommand{\ie}{i.e.}
\newcommand{\mb}[1]{\mathbf{#1}}
\newcommand{\bs}[1]{\boldsymbol{#1}}
\newcommand{\wh}[1]{\widehat{#1}}
\newcommand{\wc}[1]{\widecheck{#1}}
\newcommand{\wt}[1]{\widetilde{#1}}
\newcommand{\rot}{\gamma}

\makeatletter
\DeclareRobustCommand\onedot{\futurelet\@let@token\@onedot}
\def\@onedot{\ifx\@let@token.\else.\null\fi\xspace}
\def\wrt{w.r.t\onedot}
\def\etal{et al\onedot}

\newcommand{\cmark}{\ding{51}}
\newcommand{\xmark}{\ding{55}}
\DeclareMathOperator*{\argmax}{arg\,max}

\makeatletter
\@ifundefined{c@rownum}{%
  \let\c@rownum\rownum
}{}
\@ifundefined{therownum}{%
  \def\therownum{\@arabic\rownum}%
}{}
\makeatother

\DeclareFontFamily{U}{mathx}{\hyphenchar\font45}
\DeclareFontShape{U}{mathx}{m}{n}{
      <5> <6> <7> <8> <9> <10>
      <10.95> <12> <14.4> <17.28> <20.74> <24.88>
      mathx10
      }{}
\DeclareSymbolFont{mathx}{U}{mathx}{m}{n}
\DeclareFontSubstitution{U}{mathx}{m}{n}
\DeclareMathAccent{\widecheck}{0}{mathx}{"71}
\definecolor{lightgray}{gray}{0.9}
\definecolor{grassgreen}{RGB}{122,201,67}

\newcommand{\clearemptydoublepage}{\newpage{\pagestyle{empty}\cleardoublepage}}

\begin{document}

\title{Unified Simulation, Perception, and Generation of Human Behavior}
\author{Ye Yuan}

\thispagestyle{empty}
\date{}

\begin{center}

{\Huge \bf Unified Simulation, Perception, and Generation of Human Behavior} \\
\vspace{1cm}  
{\Large Ye Yuan} \\
\vspace{1cm} 

\vspace{-0.5cm}
{\large CMU-RI-TR-22-10}
\vspace{2cm}

{\Large April 19, 2022} \\
\vspace{2cm}
{\Large
The Robotics Institute\\
Carnegie Mellon University\\
Pittsburgh, PA\\
}
\vspace{1cm}
{\Large
{\bf Thesis Committee:}\\
Kris Kitani, Chair \\
Jessica Hodgins \\
David Held \\
Josh Merel, \textit{Meta Reality Lab} \\
Sanja Fidler, \textit{University of Toronto}\\
}
\vspace{2cm}
\par ~ \\
{\large \it Thesis submitted in partial fulfillment of the \\
    requirements for the degree of Doctor of Philosophy in Robotics}
\vfill
{\large \copyright Ye Yuan, 2022}
\end{center}

\clearpage

\pagenumbering{Roman}
\clearemptydoublepage
\setcounter{page}{1}

\begin{centering} \section*{Abstract} \end{centering}

Understanding and modeling human behavior is fundamental to almost any computer vision and robotics applications that involve humans. In this thesis, we take a holistic approach to human behavior modeling and tackle its three essential aspects --- simulation, perception, and generation. Throughout the thesis, we show how the three aspects are deeply connected and how utilizing and improving one aspect can greatly benefit the other aspects.

As humans live in a physical world, we treat physics simulation as the foundation of our approach. In the first part of the thesis, we start by developing a robust framework for representing human behavior in physics simulation. In particular, we model a human using a proxy humanoid agent inside a physics simulator and treat human behavior as the result of an optimal control policy for the humanoid. This framework allows us to formulate human behavior modeling as policy learning, which can be solved with reinforcement learning (RL). Since it is difficult and often suboptimal to manually design simulated agents such as humanoids, we further propose a transform-and-control RL framework for efficient and automatic design of agents that are more performant than those created by experts.

In the second part of the thesis, we study the perception of human behavior through the lens of human pose estimation where we utilize the simulation-based framework developed in the first part. Specifically, we learn a video-conditioned policy with RL using a reward function based on how the policy-generated pose aligns with the ground truth. For both first-person and third-person human pose estimation, our simulation-based approach significantly outperforms kinematics-based methods in terms of pose accuracy and physical plausibility. The improvement is especially evident in the challenging first-person setting where the front-facing camera cannot see the person. Besides using simulation, we also propose to use human behavior generation models for global occlusion-aware human pose estimation with dynamic cameras. Concretely, we use deep generative motion and trajectory models to hallucinate poses for occluded frames and generate consistent global trajectories from estimated body poses.

In the third part of the thesis, we focus on the generation of human behavior, leveraging our simulation-based framework and deep generative models. We first present a simulation-based generation approach that can generate a single future motion of a person from a first-person video. To address the uncertainty in future human behavior, we develop two deep generative models that can generate diverse future human motions using determinantal point processes (DPPs) and latent normalizing flows respectively. Finally, extending from the single-agent setting, we further study multi-agent behavior generation where multiple humans interact with each other in complex social scenes. We develop a stochastic agent-aware trajectory generation framework that can forecast diverse and socially-aware human trajectories.

\clearpage

\vspace*{\fill}
\begin{center}
\emph{
    Dedicated to my wife, Yanglan Ou, my daughter, Mila,\\
    and my parents, Liying Ye and Yijun Yuan
}
\end{center}
\vspace*{\fill}

\clearpage
\begin{centering} \section*{Acknowledgements} \end{centering}

I am deeply grateful to my mentors, collaborators, friends, and family. Ph.D. is a long and arduous journey. Their advice, support, friendship, and love made it much more fun and enjoyable. Without them, this thesis would not have been possible.

First and foremost, I would like to sincerely thank my advisor, Kris Kitani. When I first entered the program, I really wanted to do research in computer vision and machine learning but had little knowledge in these areas. Despite my lack of experience, Kris was able to see the potential in me and picked me up as his student. He is the best advisor I can ever dream of. He possesses all the qualities of a great advisor: passionate, knowledgable, insightful, rigorous, and always looking out for my best interests. I have learned so much from him, not just research but also many life lessons. I can never thank him enough for his faith, guidance, and support.

I would also like to thank my thesis committee members: Jessica Hodgins, David Held, Josh Merel, and Sanja Fidler, who have provided thoughtful feedback and insightful comments at various stages of my research. Their pioneering work has largely inspired my research on human behavior modeling, and it is an honor for me to have my thesis judged by them.

I am also very grateful to my previous research advisor, Stelian Coros, who has helped me develop my research skills and supported me through my Ph.D. application. I would also like to thank my undergraduate advisors: Kun Zhou, Changxi Zheng, Xiang Chen, and Tianjia Shao. They patiently led me into the fascinating world of research and supported me in pursuing graduate study abroad.

I am also very fortunate to have collaborated with many fantastic collaborators and mentors. I would like to thank Umar Iqbal, Pavlo Molchanov, and Jan Kautz for the awesome summer internship at NVIDIA Research. I would also like to thank Jason Saragih, Shih-En Wei, Tomas Simon, and Ying Yang for the wonderful summer internships at Meta Reality Lab. In addition, many thanks to Mariko Isogawa, Matthew O'Toole, Wen Sun, Jiaqi Guan, Erwin Wu, and Dong-Hyun Hwang for the productive collaborations at CMU.

My Ph.D. journey would not have been as fun without my friends and labmates at CMU. I would like to thank Yan Xu, Jack Li, Scott Sun, Yuda Song, Erica Weng, Zhengyi Luo, Xiaofang Wang, Xinshuo Weng, Peiyun Hu, Qichen Fu, Xingyu Liu, Jinkun Cao, Yunze Man, Shengcao Cao, Vivek Roy, Shun Iwase, Tanya Marwah, Rawal Khirodkar, Navyata Sanghvi, Hana Hoshino, Shin Usami and many, many others. The order is based only on the difficulty of spelling.

I am deeply grateful to my parents, Liying Ye and Yijun Yuan. Their unconditional support and love made me who I am today, for which I am forever indebted to them.

A special thanks to my wife's parents, Xiaoyan Qian and Yiming Ou, and Aunt Xiaoxia Qian, who have provided tremendous support during my Ph.D.

I would also like to thank my daughter, Mila, who lights up my world with her infectious smile and makes every day a sunny day.

Lastly, I would like to wholeheartedly thank my wife, Yanglan Ou, for her love, faith, and sacrifices. Without her, this thesis is impossible. I am incredibly lucky and grateful to have her by my side.

\tableofcontents

\clearpage \listoffigures
\clearpage \listoftables

\clearpage
\pagenumbering{arabic}
\setcounter{page}{1}

\chapter{Introduction}
\label{chap:introduction}

\begin{figure}[t]
    \centering
    \includegraphics[width=\linewidth]{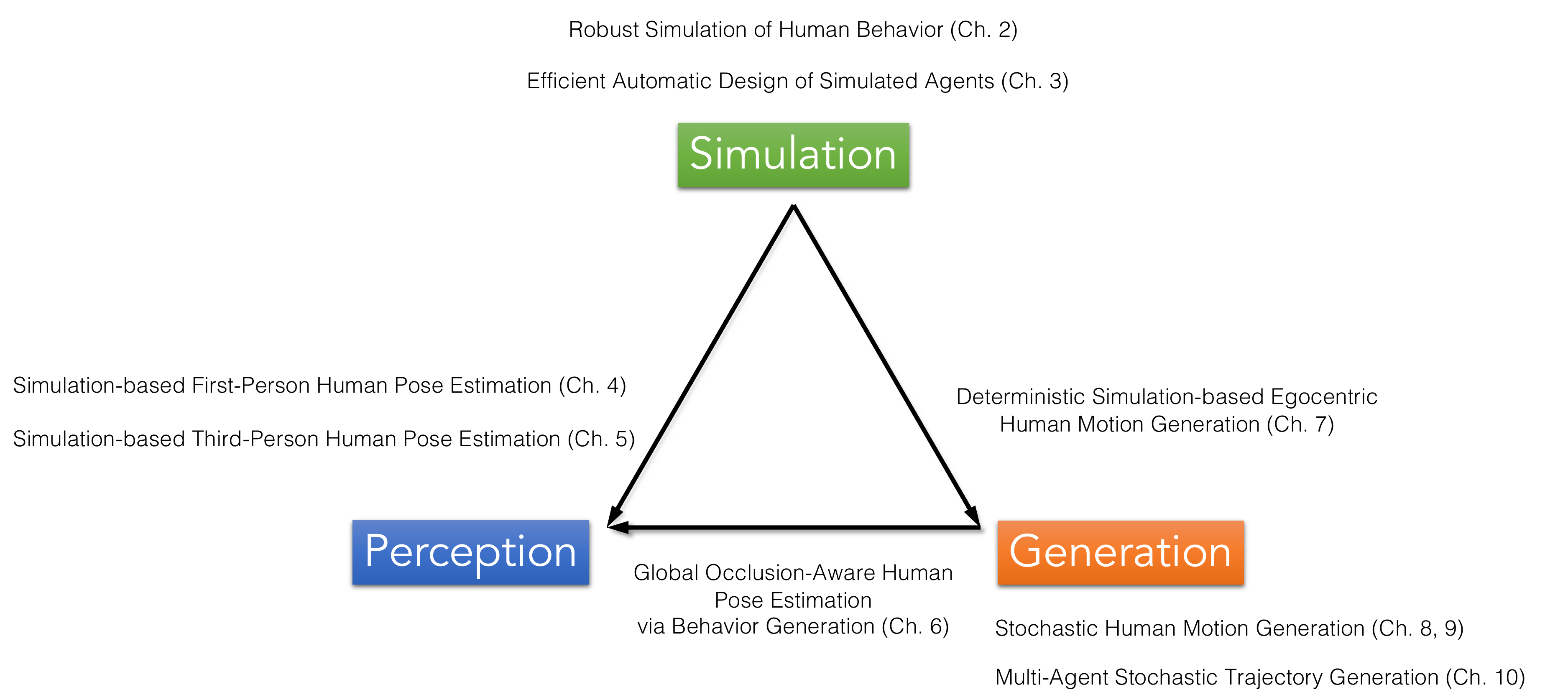}
    \caption{Overview of this thesis.  We tackle simulation, perception, and generation of human behavior, where we take a unified approach and explore the synergy between the three aspects indicated by the arrows. For example, how simulation can benefit the perception (Ch.~\ref{chap:egopose18} and \ref{chap:simpoe}) and generation (Ch.~\ref{chap:egopose19}) of human behavior, and how behavior generation can improve perception (Ch.~\ref{chap:glamr}).}
    \label{fig:thesis_overview}
\end{figure}

From motion capture studios that reconstruct actors' performance to self-driving vehicles that yield to pedestrians, many computer vision and robotics applications require accurate and effective modeling of human behavior. There are three essential aspects of human behavior modeling: (1) \emph{Perception}, which is the process of understanding human behavior from visual inputs\footnote{This is the definition used throughout the thesis, which is different from another common meaning of perception, i.e., the process of an embodied agent such as a human using its senses to perceive the world.}; (2) \emph{Generation}, which aims to generate human behavior from existing behavior data; (3) \emph{Simulation}, which intends to replicate human behavior inside a physics simulator. Most prior research on human behavior modeling focuses mainly on one of the three aspects. For instance, work on perception, such as human pose estimation~\cite{kanazawa2018end,kocabas2020vibe} or action recognition~\cite{tran2018closer,shi2019skeleton}, usually does not pay attention to the generation and simulation of human behavior. Similarly, work on humanoid control in physics simulation (e.g., \cite{merel2017learning,peng2018deepmimic}) typically does not address the perception of human behavior. In this thesis, as illustrated in Fig.~\ref{fig:thesis_overview}, we aim to show that a unified treatment of the three aspects allows us to explore the synergy between them and build intelligent systems with strong abilities to simulate, perceive, and generate human behavior.

As humans live in a physical world, we base the foundation of our approach on the physics simulation of human behavior. The physics simulation in our approach provides the proper instrument for a decision-theoretic perspective of human behavior, where humans are modeled as agents that interact with a physically-simulated environment and their behavior is the result of an optimal control policy, which is learned from the rewards they receive. This decision-theoretic perspective mirrors the situation in the real world, where we as humans are constantly making decisions according to some policy and refining the policy based on the future payoff. Through this perspective, we can formulate human behavior modeling as policy learning, which can be solved using the tools of optimal control or reinforcement learning (RL). 

However, it is quite challenging to control a humanoid agent in physics simulation, which is largely due to the complexity of simulating the dynamics of real humans. To address this problem, we first propose an approach called residual force control (RFC) that augments a control policy with external residual forces to compensate for any dynamics mismatch between the humanoid and real humans. Our approach significantly improves the robustness of controlling humanoids in physics simulation, therefore laying a solid foundation for any downstream applications such as simulation-based perception and generation of human behavior.
We also propose a second approach for enhancing the ability of simulated agents such as humanoids, called Transform2Act. Unlike RFC which adds imaginary residual forces, Transform2Act is a general approach for automatically and efficiently designing simulated agents that are more performant than manually-created agents. Its main idea is to use RL to learn a policy that can both design and control an agent.

Based on our simulation-based behavior modeling framework, the perception of human behaviors can also be framed as policy learning, where we learn a policy conditioned on visual input to produce the desired behavior. A key advantage of this approach is that the human behavior output by the system is always constrained to be physically-plausible. For 3D human pose estimation, this means the estimated human pose is free from physical artifacts such as jitter, foot sliding, and ground or object penetration, which are very important for applications like virtual reality and medical monitoring. In this thesis, we will demonstrate how simulation-based human pose estimation methods significantly outperform their kinematic counterparts in terms of pose accuracy and physical plausibility for both first-person and third-person settings. Besides leveraging simulation for perception, we also explore the synergy between generation and perception. Specifically, we propose to use human behavior generation models for global occlusion-aware human pose estimation with dynamic cameras. The use of deep generative motion and trajectory models allows our method to hallucinate poses for occluded frames and generate consistent global trajectories from estimated body poses.

Generation is the final important piece in human behavior modeling. To improve the physical plausibility of generated behavior, we first apply our simulated-based framework to generate the future motion of a person using egocentric videos. Yet, this approach can only forecast a single future motion, which motivates us to tackle an important aspect of generation, i.e., the uncertainty of future behavior. From a third-person point of view, a person's future behavior is highly uncertain. For instance, we do not know for certain whether a random pedestrian is going to turn left or right at an intersection. To model the uncertainty, we take a deep generative approach to human behavior generation with a special focus on the diversity of the generated behavior. Our deep generative models based on determinantal point processes~\cite{kulesza2012determinantal} and latent normalizing flows~\cite{rezende2015variational} can generate many diverse yet plausible behaviors instead of only the perturbations of the most likely behavior, thus covering as many future scenarios as possible. This is crucial for safety-critical applications such as self-driving vehicles where forecasted human behaviors need to be comprehensive for the safe planning and control of the vehicle. Furthermore, extending from the single-agent setting so far, we also study multi-agent stochastic trajectory generation where multiple agents (e.g., humans, vehicles, robots) interact with each other in complex social scenes. By proposing a new agent-aware Transformer model, we develop an effective multi-agent behavior generation framework that can forecast diverse and socially-aware human trajectories.

\section{Main Contributions and Organization}
\label{sec:intro:main_contribution}

An outline of this thesis is shown in Fig.~\ref{fig:thesis_overview}. The thesis is divided into three parts --- simulation, perception, and generation of human behavior. However, many chapters focus on the combination of the three aspects (Ch.~\ref{chap:rfc}, \ref{chap:egopose18}, \ref{chap:simpoe}, \ref{chap:glamr}, and \ref{chap:egopose19}), which reflects the unified approach of the thesis.

\subsection{Part I: Simulation of Human Behavior}

\paragraph{Robust Simulation of Human Behavior with Residual Force Control.}
In Ch.~\ref{chap:rfc}, we propose an approach called residual force control (RFC) for robustly controlling humanoids in physics simulation, which lays the foundation of our simulation-based framework. Its main idea is to augment a humanoid control policy by adding external residual forces into the action space. During training, the RFC-based policy learns to apply proper residual forces to the humanoid to compensate for the dynamics mismatch and better imitate the human behavior. Experiments demonstrated that RFC significantly outperforms state-of-the-art humanoid control methods in terms of motion quality and learning speed. Equipped with a strong motion imitation framework, RFC, we propose a human motion generation model called dual-policy control which integrates RFC with deep generative human motion models. Specifically, dual-policy control first generates diverse human motions from deep generative models and then uses RFC to track the generated motion by conditioning the control policy on it. Notably, dual-policy control enables synthesizing never-ending diverse human behaviors that are also physically-plausible.

\paragraph{Efficient Automatic Design of Simulated Agents.}
In Ch.~\ref{chap:transform2act}, we propose a new approach for automatic and efficient design of simulated agents by incorporating the design procedure of an agent into its decision-making process. In particular, we learn a transform-and-control policy that first designs an agent and then controls the designed agent. In an episode, we divide the agent's behavior into two consecutive stages: (1) transform stage, where the agent applies a sequence of transform actions to modify its skeletal structure and joint attributes without interacting with the environment; (2) execution stage, where the agent assumes the design resulting from the transform stage and applies motor control actions to interact with the environment and receives rewards. We optimize the transform-and-control policy via a policy gradient method (PPO~\cite{schulman2017proximal}), where the training batch for each iteration includes samples collected under various designs. In contrast to zeroth-order optimization methods such as evolutionary search (ES), we optimize our policy's parameters via policy gradients which use first-order information of the parameterized policy, which improves sample efficiency. Also, unlike ES-based methods which treat optimization for different designs independently, in our actor-critic based policy optimization, both the actor and critic are conditioned on the design, which allows experience sharing and prediction generalization across different designs. We show that our approach outperforms prior art significantly in terms of learning speed and the performance of the designed agents.

\subsection{Part II: Perception of Human Behavior}

\paragraph{Simulation-Based First-Person Human Pose Estimation.}
In Ch.~\ref{chap:egopose18}, we tackle the challenging task of first-person human pose estimation, i.e., using a video from a head-mounted camera to estimate the 3D human motion of the camera wearer. First-person pose estimation has many potential applications such as virtual reality, medical monitoring, and sports training. It is a highly under-constrained problem since the camera has no view of the human body, which motivates us to use physics simulation to better constrain the feasible human motion space. To this end, we leverage our simulation-based behavior modeling framework for first-person pose estimation. Specifically, the camera wearer is modeled as a humanoid agent inside a physics simulator whose state includes its joint angles, velocities, and the egocentric video. The goal is to learn a control policy that maps the agent's current state to control signals (joint torques) which are used by the simulator to generate the agent's next state. The egocentric state features enable better generalization for the policy since it is invariant to the global position and heading of the agent. We apply generative adversarial imitation learning~\cite{ho2016generative} to learn the control policy with synthesized data generated in a virtual environment. Our approach substantially outperforms the kinematics-based baselines and significantly reduces physical artifacts in estimated pose.

\paragraph{Simulation-Based Third-Person Human Pose Estimation.}
In Ch.~\ref{chap:simpoe}, we further tackle third-person human pose estimation using our simulation-based framework. Most third-person pose estimation methods~\cite{pavlakos2019expressive,xiang2019monocular,kolotouros2019learning,kocabas2020vibe,moon2020i2l} only consider human kinematics, which models body motion without physical forces and focuses on the geometric relationships of 3D poses and 2D images. In contrast, few human pose estimation approaches pay attention to human dynamics, which models body motion as the result of physical forces. Kinematics-only methods often suffer from physical artifacts such as jitter, foot sliding, and ground penetration, while dynamics-only approaches typically sacrifice accuracy to ensure physical plausibility. In this work, our goal is to jointly model human kinematics and dynamics to ensure that the estimated pose is both accurate and physically-plausible. To achieve this goal, we extend our simulated-based motion imitation framework by using a structured policy with kinematic reasoning inside. Specifically, the policy contains a kinematic refinement unit that iteratively refines an initial pose estimate based on the matching of 2D keypoints. The refined pose is then used by a control generation unit in the policy to output control signals (joint torques) of the humanoid to control its motion. This design couples the kinematic pose refinement unit with the dynamics-based control generation unit, which are learned jointly with reinforcement learning to achieve accurate and physically-plausible pose estimation. Experiments on large-scale motion datasets demonstrate that our approach outperforms prior methods significantly in terms of pose accuracy and physical plausibility.

\paragraph{Global Occlusion-Aware Human Pose Estimation via Behavior Generation.}
In Ch.~\ref{chap:glamr}, we present an approach for 3D global human mesh recovery from monocular videos recorded with dynamic cameras. Our approach is robust to severe and long-term occlusions and tracks human bodies even when they go outside the camera's field of view. To achieve this, our main idea is to leverage human behavior generation models for perception. Specifically, we first propose a deep generative motion infiller, which autoregressively infills the body motions of occluded humans based on visible motions. Additionally, in contrast to prior work, our approach reconstructs human meshes in consistent global coordinates even with dynamic cameras. Since the joint reconstruction of human motions and camera poses is underconstrained, we propose a global trajectory predictor that generates global human trajectories based on local body movements. Using the predicted trajectories as anchors, we present a global optimization framework that refines the predicted trajectories and optimizes the camera poses to match the video evidence such as 2D keypoints. Experiments on challenging indoor and in-the-wild datasets with dynamic cameras demonstrate that the proposed approach outperforms prior methods significantly in terms of motion infilling and global human pose estimation.

\subsection{Part III: Generation of Human Behavior}

\paragraph{Deterministic Simulation-Based Egocentric Human Motion Generation.}
In Ch.~\ref{chap:egopose19}, we study egocentric human motion generation, i.e., generating a person's future motion from first-person videos, which could enable many applications such as assistive living and sports training. It is even more challenging and under-constrained than first-person pose estimation. A naive application of traditional human pose generation methods would often produce future motions that converge to a static pose or diverge to non-humanlike motions. To address these problems, we again leverage the simulation-based motion imitation framework to constrain the generated motion. To better align with how humans control their joints, we use proportional-derivative (PD) control as the action space of our humanoid agent, i.e., instead of directly producing joint torques, the control policy now produces target joint angles for the agent to reach with PD control. We also address an important challenge in humanoid control, that the agent may lose balance and fall, by designing a falling detection mechanism using the value function in RL. We validate our egocentric generation models trained with motion capture data on both indoor and in-the-wild data, which shows that our method can generate accurate and physically-plausible human behavior and generalize to unseen environments.

\paragraph{Stochastic Human Motion Generation with Determinantal Point Processes.}
In Ch.~\ref{chap:dsf}, we tackle an important aspect of human behavior generation left unaddressed in Ch.~\ref{chap:egopose19}, i.e., the uncertainty of future behavior. In many applications, we need to generate diverse future human behaviors instead of a single likely future behavior. For instance, forecasting diverse future behaviors of people is essential for autonomous driving because many autonomous vehicles (AVs) rely heavily on these behavior forecasts to safely plan their actions. In other words, the diversity of the forecasted trajectories directly impacts the safety of AVs, and more diversity allows AVs to be better informed when making decisions. To improve diversity, recent motion generation methods employ generative models such as GANs and VAEs to capture the multi-modal distribution of future trajectories. However, one aspect that is often overlooked is the sampling method used at test time to produce diverse future behavior samples from a learned generative model. The traditional random sampling approach could generate many similar samples and only cover high-likelihood trajectories while missing other less likely yet possible trajectories, which leads to low sample efficiency and diversity. The system would have to generate a large number of samples to increase sample diversity, which is detrimental to real-time systems like AVs. To address this issue, we propose a novel neural sampler that maps contextual information (e.g., past motion, scene context) to the latent codes of a learned conditional VAE model, which are decoded into diverse behavior samples. We optimize the neural sampler with a novel diversity loss based on determinantal point processes (DPPs)~\cite{kulesza2012determinantal} that strikes a balance between sample diversity and likelihood. Experiments on 2D trajectory data and high-dimensional human motion data show that our approach significantly improves sample diversity.

\paragraph{Stochastic Human Motion Generation with Diversifying Latent Flows.}
In Ch.~\ref{chap:dlow}, we address the limitation of the neural sampling approach described above to further improve the diversity of generated human behaviors. Although the previous approach with DPPs can greatly boost the sample diversity of generative models, it is only able to produce a single set of diverse behaviors, which is undesirable for applications that require more diverse samples from a generative human behavior model. For instance, AVs could use as many future behavior samples as possible to ensure the safety of their planned actions. To tackle this aspect, we propose using normalizing flow models~\cite{rezende2015variational} to dissect the latent space of generative models into multiple regions. We design a diversity loss with an energy-based formulation to optimize the flow models and look for a more diversified latent space partition. Our extensive experiments on human motion generation demonstrate that the proposed approach achieved state-of-the-art performance in terms of both sample diversity and accuracy.

\paragraph{Multi-Agent Stochastic Trajectory Generation with Transformers.}
In Ch.~\ref{chap:aformer}, we tackle multi-agent stochastic trajectory generation, extending the single-agent setting studied previously. Generating accurate future trajectories of multiple agents is essential for autonomous systems, but is challenging due to the complex agent interaction and the uncertainty in each agent's future behavior. Forecasting multi-agent trajectories requires modeling two key dimensions: (1) time dimension, where we model the influence of past agent states over future states; (2) social dimension, where we model how the state of each agent affects others. Most prior methods model these two dimensions separately, e.g., first using a temporal model to summarize features over time for each agent independently and then modeling the interaction of the summarized features with a social model. This approach is suboptimal since independent feature encoding over either the time or social dimension can result in a loss of information. Instead, we would prefer a method that allows an agent's state at one time to directly affect another agent's state at a future time. To this end, we propose a new Transformer, called AgentFormer, that jointly models the time and social dimensions. The model leverages a sequence representation of multi-agent trajectories by flattening trajectory features across time and agents. Since standard attention operations disregard the agent identity of each element in the sequence, AgentFormer uses a novel agent-aware attention mechanism that preserves agent identities by attending to elements of the same agent differently than elements of other agents. Based on AgentFormer, we propose a stochastic multi-agent trajectory prediction model that can attend to features of any agent at any previous timestep when inferring an agent's future position. Our method substantially improves the state of the art on well-established trajectory datasets.

\section{Bibliographical Remarks}
\label{sec:intro:bib}
This thesis only contains works for which the author was a primary contributor. Ch.s~\ref{chap:rfc}, \ref{chap:egopose18}, \ref{chap:egopose19}, \ref{chap:dsf}, and \ref{chap:dlow} are based on joint work with Kris Kitani~\cite{yuan2020residual,yuan20183d,yuan2019ego,yuan2019diverse,yuan2020dlow}. Ch.~\ref{chap:transform2act} is based on joint work with Yuda Song, Zhengyi Luo, Wen Sun, and Kris Kitani~\cite{yuan2022transform}. Ch.~\ref{chap:simpoe} is based on joint work with Shih-En Wei, Tomas Simon, Kris Kitani, and Jason Saragih~\cite{yuan2021simpoe}. Ch.~\ref{chap:glamr} is based on joint work with Umar Iqbal, Pavlo Molchanov, Kris Kitani, and Jan Kautz~\cite{yuan2022glamr}. Ch.~\ref{chap:aformer} is based on joint work with Xinshuo Weng, Yanglan Ou, and Kris Kitani~\cite{yuan2021agent}.

\section{Excluded Research}
\label{sec:intro:excluded}
I excluded a significant portion of research undertaken during my Ph.D. to keep this thesis succint. Below is the excluded research:
\begin{enumerate}
    \item \textbf{Human Behavior Simulation:} Dynamics-regulated kinematic policy for human-scene interaction~\cite{luo2021dynamics}.
    \item \textbf{Human Behavior Perception:} Non-line-of-sight (NLOS) simulation-based human pose estimation~\cite{isogawa2020optical} and imaging~\cite{isogawa2020efficient}; Human pose estimation with chest-mounted cameras~\cite{hwang2020monoeye}; Hand pose estimation with wrist-worn cameras~\cite{wu2020back}.
    \item \textbf{Human Behavior Generation:} Generative human activity and trajecotry forecasting~\cite{guan2020generative}; Joint multi-agent tracking and forecasting~\cite{weng2021ptp}.
    \item \textbf{Reinforcement Learning and Control:}  Online model-based meta RL for personalized navigation~\cite{yuda2022online}.
\end{enumerate}

\part{Simulation of Human Behavior}
\label{part:simulation}
\chapter{Robust Simulation of Human Behavior with Residual Force Control}
\label{chap:rfc}

\section{Introduction}
Understanding human behaviors and creating virtual humans that act like real people has been a mesmerizing yet elusive goal in computer vision and graphics. One important step to achieve this goal is human motion synthesis which has broad applications in animation, gaming and virtual reality. With advances in deep learning, data-driven approaches~\cite{holden2016deep,holden2017phase,peng2018deepmimic,park2019learning,bergamin2019drecon} have achieved remarkable progress in producing realistic motions learned from motion capture data. Among them are physics-based methods~\cite{peng2018deepmimic,park2019learning,bergamin2019drecon} empowered by reinforcement learning (RL), where a humanoid agent in simulation is trained to imitate reference motions. Physics-based methods have many advantages over their kinematics-based counterparts. For instance, the motions generated with physics are typically free from jitter, foot skidding or geometry penetration as they respect physical constraints. Moreover, the humanoid agent inside simulation can interact with the physical environment and adapt to various terrains and perturbations, generating diverse motion variations. 

However, physics-based methods have their own challenges. In many cases, the humanoid agent fails to imitate highly agile motions like ballet dance or long-term motions that involve swift transitions between various locomotions. We attribute such difficulty to the \emph{dynamics mismatch} between the humanoid model and real humans. Humans are very difficult to model because they are very complex creatures with hundreds of bones and muscles.
Although prior work has tried to improve the fidelity of the humanoid model~\cite{lee2019scalable,won2019learning}, it is nonetheless safe to say that these models are not exact replicas of real humans and the dynamics mismatch still exists. The problem is further complicated when motion capture data comprises a variety of individuals with diverse body types.
Due to the dynamics mismatch, motions produced by real humans may not be admissible by the humanoid model, which means no control policy of the humanoid is able to generate those motions.

\begin{figure}
    \centering
    \includegraphics[width=\linewidth]{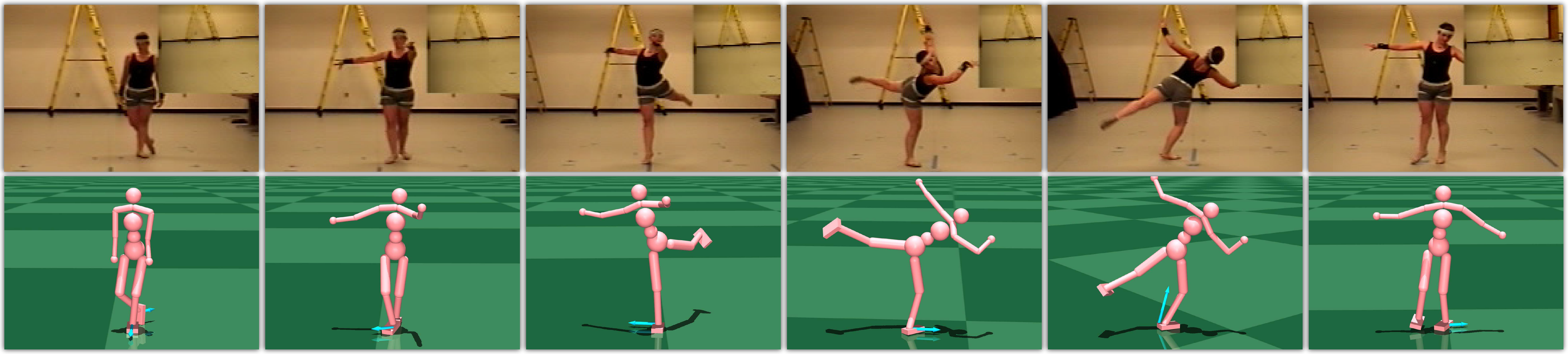}
    \caption{\textbf{Top:} A ballet dancer performing highly agile moves like jeté, arabesque and pirouette. \textbf{Bottom:} A humanoid agent controlled by a policy augmented with the proposed residual forces (blue arrows) is able to dance like the performer. The motion is best viewed in our supplementary \href{https://youtu.be/XuzH1u78o1Y}{video}.}
    \label{rfc:fig:intro}
\end{figure}

To overcome the dynamics mismatch, we propose an approach termed \emph{Residual Force Control (RFC)} which can be seamlessly integrated into existing RL-based humanoid control frameworks. Specifically, RFC augments a control policy by introducing external residual forces into the action space.
During RL training, the RFC-based policy learns to apply residual forces onto the humanoid to compensate for the dynamics mismatch and achieve better motion imitation. 
Intuitively, the residual forces can be interpreted as invisible forces that enhance the humanoid's abilities to go beyond the physical limits imposed by the humanoid model. RFC generates a more expressive dynamics that admits a wider range of human motions since the residual forces serve as a learnable time-varying correction to the dynamics of the humanoid model.
To validate our approach, we perform motion imitation experiments on a wide range of dynamic human motions including ballet dance and acrobatics. The results demonstrate that RFC outperforms state-of-the-art methods with faster convergence and better motion quality. Notably, we are able to showcase humanoid control policies that are capable of highly agile ballet dance moves like pirouette, arabesque and jeté (Fig.~\ref{rfc:fig:intro}).

Another challenge facing physics-based methods is synthesizing multi-modal long-term human motions. Previous work has elicited long-term human motions with hierarchical RL~\cite{merel2018neural,merel2018hierarchical,peng2019mcp} or user interactive control~\cite{bergamin2019drecon,park2019learning}. However, these approaches still need to define high-level tasks of the agent or require human interaction. We argue that removing these requirements could be critical for applications like automated motion generation and large-scale character animation. Thus, we take a different approach to long-term human motion synthesis by leveraging the temporal dependence of human motion. In particular, we propose a \emph{dual-policy control} framework where a kinematic policy learns to predict multi-modal future motions based on the past motion and a latent variable used to model human intent, while an RFC-based control policy learns to imitate the output motions of the kinematic policy to produce physically-plausible motions. Experiments on a large-scale human motion dataset, Human3.6M~\cite{ionescu2013human3}, show that our approach with RFC and dual policy control can synthesize stable long-term human motions without any task guidance or user input.

The main contributions of this work are as follows: (1) We address the dynamics mismatch in motion imitation by introducing the idea of RFC which can be readily integrated into RL-based humanoid control frameworks. (2) We propose a dual-policy control framework to synthesize multi-modal long-term human motions without the need for task guidance or user input. (3) Extensive experiments show that our approach outperforms state-of-the-art methods in terms of learning speed and motion quality. It also enables imitating highly agile motions like ballet dance that evade prior work. With RFC and dual-policy control, we present the first humanoid control method that successfully learns from a large-scale human motion dataset (Human3.6M) and generates diverse long-term motions.

\section{Related Work}
\paragraph{Kinematics-based models\hspace{-0.5em}} for human motion synthesis have been extensively studied by the computer graphics community. Early approaches construct motion graphs from large motion datasets and design controllers to navigate through the graph to generate novel motions~\cite{lee2002interactive,safonova2007construction}. Alternatively, prior work has explored learning a low-dimensional embedding space to synthesize motions continuously~\cite{shin2006motion,lee2010motion,levine2012continuous}. Advances in deep learning have enabled methods that use deep neural networks to design generative models of human motions~\cite{holden2016deep,holden2017phase,starke2019neural}. While the graphics community focuses on user control, computer vision researchers have been increasingly interested in predicting future human motions. A vast body of work has used recurrent neural networks to predict a deterministic future motion from the past motion~\cite{fragkiadaki2015recurrent,jain2016structural,li2017auto,martinez2017human,villegas2017learning,pavllo2018quaternet,aksan2019structured,gopalakrishnan2019neural}. To address the uncertainty of future, stochastic approaches develop deep generative models to predict multi-modal future motions~\cite{yan2018mt,barsoum2018hp,kundu2019bihmp,yuan2019diverse,yuan2020dlow}. The major drawback of kinematics-based approaches is that they are prone to generating physically-invalid motions with artifacts like jitter, foot skidding and geometry (e.g., body, ground) penetration.

\paragraph{Physics-based methods\hspace{-0.5em}} for motion synthesis address the limitation of kinematics-based models by enforcing physical constraints. Early work has adopted model-based methods for tracking reference motions~\cite{yin2007simbicon,muico2009contact,lee2010data,liu2010sampling,liu2016guided}. Recently, deep RL has achieved great success in imitating human motions with manually-designed rewards~\cite{liu2017learning,liu2018learning,peng2018deepmimic}. GAIL~\cite{ho2016generative} based approaches have been proposed to eliminate the need for reward engineering~\cite{merel2017learning,wang2017robust}. RL-based humanoid control has also been applied to estimating physically-plausible human poses from videos~\cite{yuan20183d, yuan2019ego, isogawa2020optical}. To synthesize long-term human motions, prior work has resorted to hierarchical RL with predefined high-level task objectives~\cite{merel2018neural,merel2018hierarchical,peng2019mcp}. Alternatively, recent works use deep RL to learn controllable polices to generate long-term motions with user input~\cite{bergamin2019drecon,park2019learning}. Different from previous work, our dual-policy control framework exploits the temporal dependence of human motion and synthesizes multi-modal long-term motions by forecasting diverse futures, which can be used to replace manual task guidance or user input. Furthermore, our proposed residual force control addresses the dynamics mismatch in humanoid control and enables imitating agile motions like ballet dance that evade prior work.

\paragraph{Inferring external forces\hspace{-0.5em}} from human motion has been an active research area in biomechanics. Researchers have developed models that regress ground reaction forces from human motion using supervised learning~\cite{choi2013ground,oh2013prediction,johnson2018predicting}. These approaches require expensive force data collected in laboratory settings to train the models. On the other hand, machine learning researchers have proposed differentiable physics engines that enable learning forces to control simple simulated systems~\cite{de2018end,degrave2019differentiable}. Trajectory optimization based approaches~\cite{mordatch2012discovery,mordatch2012contact} have also been used to optimize external contact forces to synthesize human motions. A recent work~\cite{ehsani2020use} predicts forces acting on rigid objects in simulation to match image evidence with contact point supervision. Unlike prior work, we use deep RL to learn residual forces that complement contact forces to improve motion imitation without the supervision of forces or contact points.

\section{Preliminaries}
\label{rfc:sec:prelim}

The task of humanoid control-based motion imitation can be formulated as a Markov decision process (MDP), which is defined by a tuple $\mathcal{M} = (\mathcal{S}, \mathcal{A}, \mathcal{T}, R, \gamma)$ of states, actions, transition dynamics, a reward function, and a discount factor. A humanoid agent interacts with a physically-simulated environment according to a policy $\pi(\bs{a}|\bs{s})$, which models the conditional distribution of choosing an action $\bs{a} \in \mathcal{A}$ given the current state $\bs{s} \in \mathcal{S}$. Starting from some initial state $\bs{s}_0$, the agent iteratively samples an action $\bs{a}_t$ from the policy $\pi$ and the simulation environment with transition dynamics $\mathcal{T}(\bs{s}_{t+1}|\bs{s}_t, \bs{a}_t)$ generates the next state $\bs{s}_{t+1}$ and gives the agent a reward $r_t$. The reward is assigned based on how the agent's motion aligns with a given reference motion. The agent's goal is to learn an optimal policy $\pi^\ast$ that maximizes its expected return $J(\pi) = \mathbb{E}_{\pi}\left[\sum_{t}\gamma^t r_t\right]$. To solve for the optimal policy, one can apply one's favorite reinforcement learning algorithm (e.g., PPO~\cite{schulman2017proximal}). In the following, we will give a more detailed description of the states, actions, policy and rewards to show how motion imitation fits in the standard reinforcement learning (RL) framework.

\paragraph{States.} The state $\bs{s}$ is formed by the humanoid state $\bs{x} = (\bs{q}, \dot{\bs{q}})$ which includes all degrees of freedom (DoFs) $\bs{q}$ of the humanoid and their corresponding velocities $\dot{\bs{q}}$. Specifically, the DoFs $\bs{q} = (\bs{q}_\texttt{r}, \bs{q}_\texttt{nr})$ include 6 root DoFs $\bs{q}_\texttt{r}$ (global position and orientation) as well as the angles of other joints~$\bs{q}_\texttt{nr}$. We transform $\bs{q}_\texttt{r}$ to the root's local coordinate to remove dependency on global states.

\paragraph{Actions.} As noticed in previous work~\cite{peng2017learning,yuan2019ego}, using proportional derivative (PD) controllers at each joint yields more robust policies than directly outputting joint torques. Thus, the action $\bs{a}$ consists of the target angles $\bs{u}$ of the PD controllers mounted at non-root joint DoFs $\bs{q}_\texttt{nr}$ (root DoFs $\bs{q}_\texttt{r}$ are not actuated). The joint torques $\bs{\tau}$ can then be computed as
\begin{equation}
\bs{\tau} = \bs{k}_\texttt{p}\circ(\bs{u} - \bs{q}_\texttt{nr}) - \bs{k}_\texttt{d}\circ\dot{\bs{q}}_\texttt{nr},
\end{equation}
where $\bs{k}_\texttt{p}$ and $\bs{k}_\texttt{d}$ are manually-specified gains and $\circ$ denotes element-wise multiplication.

\paragraph{Policy.} As the action $\bs{a}$ is continuous, we use a parametrized Gaussian policy $\pi_\theta(\bs{a}|\bs{s}) = \mathcal{N}(\bs{\mu}_\theta, \bs{\Sigma})$ where the mean $\bs{\mu}_\theta$ is output by a neural network with parameters $\theta$ and $\bs{\Sigma}$ is a fixed diagonal covariance matrix. At test time, instead of sampling we use the mean action to achieve best performance.

\paragraph{Rewards.} Given a reference motion $\widehat{\bs{x}}_{0:T} = (\widehat{\bs{x}}_0, \ldots, \widehat{\bs{x}}_{T-1})$, we need to design a reward function to incentivize the humanoid agent to imitate $\widehat{\bs{x}}_{0:T}$. To this end, the reward $r_t = r^\texttt{im}_t$ is defined by an imitation reward $r^\texttt{im}_t$ that encourages the state  $\bs{x}_t$ of the humanoid agent to match the reference state $\widehat{\bs{x}}_t$.

During RL training, the agent's initial state is intialized to a random frame from the reference motion $\widehat{\bs{x}}_{0:T}$. The episode ends when the agent falls to the ground or the episode horizon $H$ is reached.

\section{Residual Force Control (RFC)}
\label{rfc:sec:rfc}

As demonstrated in prior work~\cite{peng2018deepmimic,yuan2019ego}, we can apply the motion imitation framework described in Sec.~\ref{rfc:sec:prelim} to successfully learn control policies that imitate human locomotions (e.g., walking, running, crouching) or acrobatics (e.g, backflips, cartwheels, jump kicks). However, the motion imitation framework has its limit on the range of motions that the agent is able to imitate. In our experiments, we often find the framework unable to learn more complex motions that require sophisticated foot interaction with the ground (e.g., ballet dance) or long-term motions that involve swift transitions between different modes of locomotion. We posit that the difficulty in learning such highly agile motions can be attributed to the \emph{dynamics mismatch} between the humanoid model and real humans, i.e., the humanoid transition dynamics $\mathcal{T}(\bs{s}_{t+1}|\bs{s}_t, \bs{a}_t)$ is different from the real human dynamics. Thus, due to the dynamics mismatch, a reference motion $\widehat{\bs{x}}_{0:T}$ generated by a real human may not be admissible by the transition dynamics $\mathcal{T}$, which means no policy under $\mathcal{T}$ can generate $\widehat{\bs{x}}_{0:T}$. 
 
To overcome the dynamics mismatch, our goal is to come up with a new transition dynamics $\mathcal{T}'$ that admits a wider range of motions. The new transition dynamics $\mathcal{T}'$ should ideally satisfy two properties: (1)~$\mathcal{T}'$ needs to be expressive and overcome the limitations of the current dynamics~$\mathcal{T}$; (2)~$\mathcal{T}'$ needs to be physically-valid and respect physical constraints (e.g., contacts), which implies that kinematics-based approaches such as directly manipulating the resulting state $\bs{s}_{t+1}$ by adding some residual $\delta \bs{s}$ are not viable as they may violate physical constraints.

Based on the above considerations, we propose \emph{residual force control (RFC)}, that considers a more general form of dynamics $\widetilde{\mathcal{T}}(\bs{s}_{t+1}|\bs{s}_t, \bs{a}_t, \widetilde{\bs{a}}_t)$ where we introduce a corrective control action $\widetilde{\bs{a}}_t$ (i.e., external residual forces acting on the humanoid) alongside the original humanoid control action~$\bs{a}_t$. We also introduce a corresponding RFC-based composite policy $\widetilde{\pi}_\theta(\bs{a}_t, \widetilde{\bs{a}}_t| \bs{s}_t)$ which can be decomposed into two policies: (1) the original policy $\widetilde{\pi}_{\theta_1}(\bs{a}_t | \bs{s}_t)$ with parameters $\theta_1$ for humanoid control and (2) a residual force policy $\widetilde{\pi}_{\theta_2}(\widetilde{\bs{a}}_t| \bs{s}_t)$ with parameters $\theta_2$ for corrective control. The RFC-based dynamics and policy are more general as the original policy $\widetilde{\pi}_{\theta_1}(\bs{a}_t|\bs{s}_t) \equiv \widetilde{\pi}_\theta(\bs{a}_t, \bs{0}| \bs{s}_t)$ corresponds to a policy $\widetilde{\pi}_\theta$ that always outputs zero residual forces. Similarly, the original dynamics $\mathcal{T}(\bs{s}_{t+1}|\bs{s}_t, \bs{a}_t)\equiv \widetilde{\mathcal{T}}(\bs{s}_{t+1}|\bs{s}_t, \bs{a}_t, \mathbf{0})$ corresponds to the dynamics $\widetilde{\mathcal{T}}$ with zero residual forces. During RL training, the RFC-based policy $\widetilde{\pi}_\theta(\bs{a}_t, \widetilde{\bs{a}}_t| \bs{s}_t)$ learns to apply proper residual forces $\widetilde{\bs{a}}_t$ to the humanoid to compensate for the dynamics mismatch and better imitate the reference motion. 
Since $\widetilde{\bs{a}}_t$ is sampled from $\widetilde{\pi}_{\theta_2}(\widetilde{\bs{a}}_t| \bs{s}_t)$, the dynamics of the original policy $\widetilde{\pi}_{\theta_1}(\bs{a}_t | \bs{s}_t)$ is parametrized by~$\theta_2$ as $\mathcal{T}'_{\theta_2}(\bs{s}_{t+1}|\bs{s}_t, \bs{a}_t)\equiv\widetilde{\mathcal{T}}(\bs{s}_{t+1}|\bs{s}_t, \bs{a}_t, \widetilde{\bs{a}}_t)$. From this perspective, $\widetilde{\bs{a}}_t$ are learnable time-varying dynamics correction forces governed by $\widetilde{\pi}_{\theta_2}$.
Thus, by optimizing the composite policy $\widetilde{\pi}_\theta(\bs{a}_t, \widetilde{\bs{a}}_t| \bs{s}_t)$, we are in fact jointly optimizing the original humanoid control action $\bs{a}_t$ and the dynamics correction (residual forces)~$\widetilde{\bs{a}}_t$. In the following, we propose two types of RFC, each with its own advantages.

\subsection{RFC-Explicit}

One way to implement RFC is to explicitly model the corrective action $\widetilde{\bs{a}}_t$ as a set of residual force vectors $\{ \bs{\xi}_1, \ldots, \bs{\xi}_M \}$ and their respective contact points $\{ \bs{e}_1, \ldots, \bs{e}_M\}$. As the humanoid model is formed by a set of rigid bodies, the residual forces are applied to $M$ bodies of the humanoid, where $\bs{\xi}_j$ and $\bs{e}_j$ are represented in the local body frame. To reduce the size of the corrective action space, one can apply residual forces to a limited number of bodies such as the hip or feet. In RFC-Explicit, the corrective action of the policy $\widetilde{\pi}_\theta(\bs{a}, \widetilde{\bs{a}}| \bs{s})$ is defined as $\widetilde{\bs{a}} = (\bs{\xi}_1, \ldots, \bs{\xi}_M, \bs{e}_1, \ldots, \bs{e}_M)$ and the humanoid control action is $\bs{a} = \bs{u}$ as before (Sec.~\ref{rfc:sec:prelim}). We can describe the humanoid motion using the equation of motion for multibody systems~\cite{siciliano2010robotics} augmented with the proposed residual forces:
\begin{equation}
\label{rfc:eq:eom}
    \bs{B}(\bs{q}) \ddot{\bs{q}}+\bs{C}(\bs{q}, \dot{\bs{q}}) \dot{\bs{q}}\ + \bs{g}(\bs{q})= \begin{bmatrix}
    \bs{0} \\ \bs{\tau}
    \end{bmatrix}
    + \underbrace{\sum_i \bs{J}^{T}_{\bs{v}_i}\bs{h}_i\vphantom{\sum_{j=1}^M \bs{J}^{T}_{\bs{e}_j} \bs{\xi}_j}}_\text{Contact Forces} + \underbrace{\sum_{j=1}^M \bs{J}^{T}_{\bs{e}_j} \bs{\xi}_j}_\textbf{Residual Forces}\,,
\end{equation}
where we have made the residual forces term explicit. Eq.~\eqref{rfc:eq:eom} is an ordinary differential equation (ODE), and by solving it with an ODE solver we obtain the aforementioned RFC-based dynamics $\widetilde{\mathcal{T}}(\bs{s}_{t+1}|\bs{s}_t, \bs{a}_t, \widetilde{\bs{a}}_t)$. On the left hand side $\ddot{\bs{q}}, \bs{B},\bs{C},\bs{g}$ are the joint accelerations, the inertial matrix, the matrix of Coriolis and centrifugal terms, and the gravity vector, respectively. On the right hand side, the first term contains the torques $\bs{\tau}$ computed from $\bs{a}$ (Sec.~\ref{rfc:sec:prelim}) applied to the non-root joint DoFs $\bs{q}_\texttt{nr}$ and $\bs{0}$ corresponds to the 6 non-actuated root DoFs $\bs{q}_\texttt{r}$. The second term involves existing contact forces $\bs{h}_i$ on the humanoid (usually exerted by the ground plane) and the contact points $\bs{v}_i$ of $\bs{h}_i$, which are determined by the simulation environment. Here, $\bs{J}_{\bs{v}_i} = d\bs{v}_i/d\bs{q}$ is the Jacobian matrix that describes how the contact point $\bs{v}_i$ changes with the joint DoFs $\bs{q}$. By multiplying $\bs{J}_{\bs{v}_i}^T$, the contact force $\bs{h}_i$ is transformed from the world space to the joint space, which can be understood using the principle of virtual work, i.e., the virtual work in the joint space equals that in the world space or $ (\bs{J}_{\bs{v}_i}^T \bs{h}_i )^Td\bs{q} = \bs{h}_i^T d\bs{v}_i$. Unlike the contact forces $\bs{h}_i$ which are determined by the environment, the policy can control the corrective action $\widetilde{\bs{a}}$ which includes the residual forces $\bs{\xi}_j$ and their contact points $\bs{e}_j$ in the proposed third term. The Jacobian matrix $\bs{J}_{\bs{e}_j} = d\bs{e}_j/d\bs{q}$ is similarly defined as $\bs{J}_{\bs{v}_i}$. During RL training, the policy will learn to adjust $\bs{\xi}_j$ and $\bs{e}_j$ to better imitate the reference motion. Most popular physics engines (e.g., MuJoCo~\cite{todorov2012mujoco}, Bullet~\cite{coumans2010bullet}) use a similar equation of motion to Eq.~\eqref{rfc:eq:eom} (without residual forces), which makes our approach easy to integrate.

As the residual forces are designed to be a correction mechanism to the original humanoid dynamics $\mathcal{T}$, we need to regularize the residual forces so that the policy only invokes the residual forces when necessary. Consequently, the regularization keeps the new dynamics $\mathcal{T}'$ close to the original dynamics~$\mathcal{T}$. Formally, we change the RL reward function by adding a regularizing reward $r_t^\texttt{reg}$:
\begin{equation}
\label{rfc:eq:rfce-reward}
r_t = r_t^\texttt{im} + w_\texttt{reg}r_t^\texttt{reg}\,, \quad r_t^\texttt{reg} = \exp\left(-\sum_{j=1}^M \left(k_\texttt{f}\left\|\bs{\xi}_j\right\|^2 +  k_\texttt{cp}\left\|\bs{e}_j\right\|^2  \right) \right),
\end{equation}
where $w_\texttt{reg}$, $k_\texttt{f}$ and $k_\texttt{cp}$ are weighting factors. The regularization constrains the residual force $\bs{\xi}_j$ to be as small as possible and pushes the contact point $\bs{e}_j$ to be close to the local body origin.
 
\subsection{RFC-Implicit}
One drawback of RFC-explicit is that one must specify the number of residual forces and the contact points. To address this issue, we also propose an implicit version of RFC where we directly model the total joint torques $\bs{\eta} = \sum \bs{J}^{T}_{\bs{e}_j} \bs{\xi}_j$ of the residual forces. In this way, we do not need to specify the number of residual forces or the contact points. We can decompose $\bs{\eta}$ into two parts $(\bs{\eta}_\texttt{r}, \bs{\eta}_\texttt{nr})$ that correspond to the root and non-root DoFs respectively. We can merge $\bs{\eta}$ with the first term on the right of Eq.~\eqref{rfc:eq:eom} as they are both controlled by the policy, which yields the new equation of motion:
\begin{equation}
\label{rfc:eq:eom_new}
    \bs{B}(\bs{q}) \ddot{\bs{q}}+\bs{C}(\bs{q}, \dot{\bs{q}}) \dot{\bs{q}}\ + \bs{g}(\bs{q})= \begin{bmatrix}
    \bs{\eta}_\texttt{r} \\ \bs{\tau} \; \cancel{+ \; \bs{\eta}_\texttt{nr}}
    \end{bmatrix}
    + \sum_i \bs{J}^{T}_{\bs{v}_i}\bs{h}_i\,,
\end{equation}
where we further remove $\bs{\eta}_\texttt{nr}$ (crossed out) because the torques applied at non-root DoFs are already modeled by the policy $\widetilde{\pi}_\theta(\bs{a}, \widetilde{\bs{a}}| \bs{s})$ through $\bs{\tau}$ which can absorb $\bs{\eta}_\texttt{nr}$. In RFC-Implicit, the corrective action of the policy is defined as $\widetilde{\bs{a}} = \bs{\eta}_\texttt{r}$.
To regularize $\bs{\eta}_\texttt{r}$, we use a similar reward to Eq.~\eqref{rfc:eq:rfce-reward}:
\begin{equation}
\label{rfc:eq:rfci-reward}
r_t = r_t^\texttt{im} + w_\texttt{reg}r_t^\texttt{reg}\,, \quad r_t^\texttt{reg} = \exp\left(-k_\texttt{r}\left\|\bs{\eta}_\texttt{r}\right\|^2  \right),
\end{equation}
where $k_\texttt{r}$ is a weighting factor. While RFC-Explicit provides more interpretable results by exposing the residual forces and their contact points, RFC-Implicit is computationally more efficient as it only increases the action dimensions by 6 which is far less than that of RFC-Explicit and it does not require Jacobian computation. Furthermore, RFC-Implicit does not make any underlying assumptions about the number of residual forces or their contact points.

\subsection{Discussion}
As shown in Eq.~\eqref{rfc:eq:eom_new}, RFC-explicit can generate torques $\bs{\eta}_\texttt{r}$ and $\bs{\eta}_\texttt{nr}$ for both the root and non-root DoFs. So another variant of RFC-explicit is to remove the original torque action $\bs{\tau}$, which can be absorbed into $\bs{\eta}_\texttt{nr}$. However, this approach is less desirable than the original RFC-Explicit and RFC-Implicit. This is because it would need many residual forces acting on different body parts to actuate every joint DoFs, leading to larger action space and less computational efficiency. In particular, each residual force adds 9 DoFs (force, torque, and contact point), which are much more than the 3 DoFs typically required for each joint. Additionally, it is also harder to regularize the residual forces since it is now also responsible for the normal joint actuation of the humanoid instead of just compensating for the dynamics mismatch.

\section{Dual-Policy Control for Extended Motion Generation}
So far our focus has been on imitating a given reference motion, which in practice is typically a short and segmented motion capture sequence (e.g., within 10 seconds). In some applications (e.g., behavior simulation, large-scale animation), we want the humanoid agent to autonomously exhibit long-term behaviors that consist of a sequence of diverse agile motions. Instead of guiding the humanoid using manually-designed tasks or direct user input, our goal is to let the humanoid learn long-term behaviors directly from data. To achieve this, we need to develop an approach that (i)~infers future motions from the past and (ii) captures the multi-modal distribution of the future.

\begin{center}
\begin{algorithm}
\caption{Learning RFC-based policy $\widetilde{\pi}_\theta$ in dual-policy control}
\label{alg:dpc}
\begin{algorithmic}[1]
\State \textbf{Input:} motion data $\mathcal{X}$, pretrained kinematic policy $\kappa_\psi$
\State $\theta \leftarrow$ random weights
\While{not converged}
    \State $\mathcal{D} \leftarrow \emptyset$ \Comment{initialize sample memory}
    \While{$\mathcal{D}$ is not full}
    \State $\widehat{\bs{x}}_{0:p} \leftarrow$ random motion from $\mathcal{X}$
    \State $\bs{x}_{p-1} \leftarrow \widehat{\bs{x}}_{p-1}$ \Comment{initialize humanoid state}
    \For{$t \leftarrow p, \ldots, p + nf - 1$}
    \If{$(t - p)\bmod f = 0$} \Comment{if reaching end of reference motion segment}
        \State $\bs{z} \sim p(\bs{z})$
        \State $\widehat{\bs{x}}_{t:t+f} \leftarrow \kappa_\psi(\widehat{\bs{x}}_{t:t+f}|\widehat{\bs{x}}_{t-p:t}, \bs{z})$ \Comment{generate next reference motion segment}
    \EndIf
    \State $\bs{s}_t \leftarrow (\bs{x}_{t-1},\widehat{\bs{x}}_{t-1}, \bs{z})$; \, $\bs{a}_t, \widetilde{\bs{a}}_t \leftarrow \widetilde{\pi}_\theta(\bs{a}_t, \widetilde{\bs{a}}_t|\bs{s}_t)$
    \State $\bs{x}_{t} \leftarrow$ next state from simulation with $\bs{a}_t$ and $\widetilde{\bs{a}}_t$
    \State $r_{t} \leftarrow$ reward from Eq.~\eqref{rfc:eq:rfce-reward} or~\eqref{rfc:eq:rfci-reward}
    \State $\bs{s}_{t+1} \leftarrow (\bs{x}_t,\widehat{\bs{x}}_t, \bs{z})$
    \State store $(\bs{s}_t, \bs{a}_{t}, \widetilde{\bs{a}}_t, r_t, \bs{s}_{t+1})$ into memory $\mathcal{D}$
    \EndFor
    \EndWhile
    \State $\theta \leftarrow$ PPO~\cite{schulman2017proximal} update using trajectory samples in $\mathcal{D}$ \Comment{update control policy $\widetilde{\pi}_\theta$}
\EndWhile
\end{algorithmic}
\end{algorithm}
\end{center}

As multi-modal behaviors are usually difficult to model in the control space due to non-differentiable dynamics, we first model human behaviors in the kinematic space. We propose a \emph{dual-policy control} framework that consists of a kinematic policy $\kappa_\psi$ and an RFC-based control policy $\widetilde{\pi}_\theta$. The $\psi$-parametrized kinematic policy $\kappa_\psi(\bs{x}_{t:t+f}|\bs{x}_{t-p:t}, \bs{z})$ models the conditional distribution over a $f$-step future motion $\bs{x}_{t:t+f}$, given a $p$-step past motion $\bs{x}_{t-p:t}$ and a latent variable $\bs{z}$ used to model human intent. We learn the kinematic policy $\kappa_\psi$ with a conditional variational autoencoder (CVAE~\cite{kingma2013auto}), where we optimize the evidence lower bound (ELBO):
\begin{equation}
\label{rfc:eq:vae}
\mathcal{L}=\mathbb{E}_{q_{\phi}(\bs{z} | \bs{x}_{t-p:t}, \bs{x}_{t:t+f})}\left[\log \kappa_\psi(\bs{x}_{t:t+f}|\bs{x}_{t-p:t}, \bs{z})\right]-\mathrm{KL}\left(q_{\phi}(\bs{z} | \bs{x}_{t-p:t}, \bs{x}_{t:t+f}) \| p(\bs{z})\right),
\end{equation}
where $q_{\phi}(\bs{z} | \bs{x}_{t-p:t}, \bs{x}_{t:t+f})$ is a $\phi$-parametrized approximate posterior (encoder) distribution and $p(\bs{z})$ is a Gaussian prior. The kinematic policy $\kappa_\psi$ and encoder $q_\phi$ are instantiated as Gaussian distributions whose parameters are generated by two recurrent neural networks (RNNs) respectively.

Once the kinematic policy $\kappa_\psi$ is learned, we can generate multi-modal future motions $\widehat{\bs{x}}_{t:t+f}$ from the past motion $\bs{x}_{t-p:t}$ by sampling $\bs{z}\sim p(\bs{z})$ and decoding $\bs{z}$ with $\kappa_\psi$. To produce physically-plausible motions, we use an RFC-based control policy $\widetilde{\pi}_\theta(\bs{a}, \widetilde{\bs{a}}|\bs{x},\widehat{\bs{x}}, \bs{z})$ to imitate the output motion $\widehat{\bs{x}}_{t:t+f}$ of $\kappa_\psi$ by treating $\widehat{\bs{x}}_{t:t+f}$ as the reference motion in the motion imitation framework (Sec.~\ref{rfc:sec:prelim} and \ref{rfc:sec:rfc}). The state $\bs{s}$ of the policy now includes the state $\bs{x}$ of the humanoid, the reference state $\widehat{\bs{x}}$ from $\kappa_\psi$, and the latent code $\bs{z}$. 
To fully leverage the reference state $\widehat{\bs{x}}$, we use the non-root joint angles $\widehat{\bs{q}}_\texttt{nr}$ inside $\widehat{\bs{x}}$ to serve as bases for the target joint angles $\bs{u}$ of the PD controllers. For this purpose, we change the humanoid control action $\bs{a}_t$ from $\bs{u}$ to residual angles $\delta\bs{u}$, and $\bs{u}$ can be computed as $\bs{u} = \widehat{\bs{q}}_\texttt{nr} + \delta\bs{u}$. This additive action will improve policy learning because $\widehat{\bs{q}}_\texttt{nr}$ provides a good guess for $\bs{u}$.

The learning procedure for the control policy $\widetilde{\pi}_\theta$ is outlined in Alg.~\ref{alg:dpc}. In each RL episode, we autoregressively apply the kinematic policy $n$ times to generate reference motions $\widehat{\bs{x}}_{p:p+nf}$ of $nf$ steps, and the agent with policy $\widetilde{\pi}_\theta$ is rewarded for imitating $\widehat{\bs{x}}_{p:p+nf}$. The reason for autoregressively generating $n$ segments of future motions is to let the policy $\widetilde{\pi}_\theta$ learn stable transitions through adjacent motion segments (e.g., $\widehat{\bs{x}}_{p:p+f}$ and $\widehat{\bs{x}}_{p+f:p+2f}$). At test time, we use the kinematic policy $\kappa_\psi$ and control policy $\widetilde{\pi}_\theta$ jointly to synthesize infinite-horizon human motions by continuously forecasting futures with $\kappa_\psi$ and physically tracking the forecasted motions with~$\widetilde{\pi}_\theta$.

\section{Experiments}
Our experiments consist of two parts: (1) Motion imitation, where we examine whether the proposed RFC can help overcome the dynamics mismatch and enable the humanoid to learn more agile behaviors from reference motions; (2) Extended motion synthesis, where we evaluate the effectiveness of the proposed dual-policy control along with RFC in synthesizing long-term human motions. 

\subsection{Motion Imitation}

\paragraph{Reference Motions.}
We use the CMU motion capture (MoCap) database (\href{http://mocap.cs.cmu.edu/}{link}) to provide reference motions for imitation. Specifically, we deliberately select eight clips of highly agile motions to increase the difficulty. We use clips of ballet dance with signature moves like pirouette, arabesque and jeté, which have sophisticated foot-ground interaction. We also include clips of acrobatics such as handsprings, backflips, cartwheels, jump kicks and side flips, which involve dynamic body rotations.

\paragraph{Implementation Details.}
We use MuJoCo~\cite{todorov2012mujoco} as the physics engine. We construct the humanoid model from the skeleton of subject 8 in the CMU Mocap database while the reference motions we use are from various subjects. The humanoid model has 38 DoFs and 20 rigid bodies with properly assigned geometries. Following prior work~\cite{peng2018deepmimic}, we add the motion phase to the state of the humanoid agent. We also use the stable PD controller~\cite{tan2011stable} to compute joint torques. The simulation runs at 450Hz and the policy operates at 30Hz. We use PPO~\cite{schulman2017proximal} to train the policy for 2000 epochs, each with 50,000 policy steps. Each policy takes about 1 day to train on a 20-core machine with an NVIDIA RTX 2080 Ti.

\begin{figure}
    \centering
    \includegraphics[width=\linewidth]{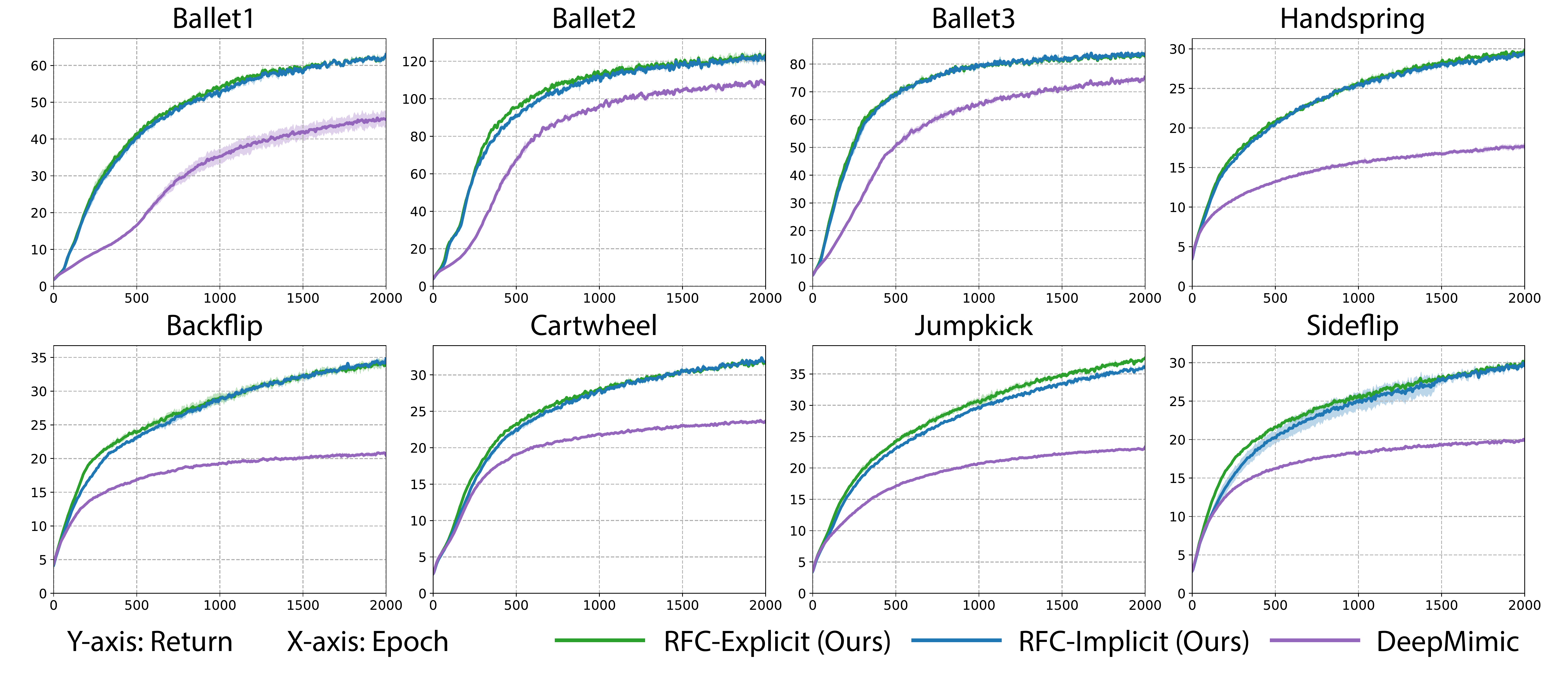}
    \caption{Learning curves of our RFC models and DeepMimic for imitating various agile motions.}
    \label{rfc:fig:plot}
\end{figure}

\paragraph{Comparisons.}
We compare the two variants -- RFC-Explicit and RFC-Implicit -- of our approach against the state-of-the-art method for motion imitation, DeepMimic~\cite{peng2018deepmimic}. For fair comparison, the only differences between our RFC models and the DeepMimic baseline are the residual forces and the regularizing reward.
Fig.~\ref{rfc:fig:plot} shows the learning curves of our models and DeepMimic, where we plot the average return per episode against the training epochs for all eight reference motions. We train three models with different initial seeds for each method and each reference motion. The return is computed using only the motion imitation reward and excludes the regularizing reward. We can see that both variants of RFC converge faster than DeepMimic consistently. Moreover, our RFC models always converge to better motion policies as indicated by the higher final returns. One can also observe that RFC-Explicit and RFC-Implicit perform similarly, suggesting that they are equally capable of imitating agile motions. Since the motion quality of learned policies is best seen in videos, we encourage the reader to refer to the supplementary \href{https://youtu.be/XuzH1u78o1Y}{video}\footnote{Video: \url{https://youtu.be/XuzH1u78o1Y}.} for qualitative comparisons. One will observe that RFC can successfully imitate the sophisticated ballet dance skills while DeepMimic fails to reproduce them. We believe the failure of DeepMimic is due to the dynamics mismatch between the humanoid model and real humans, which results in the humanoid unable to generate the external forces needed to produce the motions. On the other hand, RFC overcomes the dynamics mismatch by augmenting the original humanoid dynamics with learnable residual forces, which enables a more flexible new dynamics that admits a wider range of agile motions. We note that the comparisons presented here are only for simulation domains (e.g., animation and motion synthesis) since external residual forces are not directly applicable to real robots. However, we do believe that RFC could be extended to a warm-up technique to accelerate the learning of complex policies for real robots, and the residual forces needed to overcome the dynamics mismatch could be used to guide agent design.

\begin{table}[t]
\footnotesize
\centering
\begin{tabular}{@{\hskip 1mm}lcccccccc@{\hskip 0mm}}
\toprule
\multirow{3}{*}[2pt]{Method} & \multirow{3}{12mm}[2pt]{\centering Phsyics-\\based} & \multicolumn{2}{c}{Human3.6M (Mix)} & \multicolumn{2}{c}{Human3.6M (Cross)} & \multicolumn{2}{c}{EgoMocap}\\ \cmidrule(l{0.5mm}r{1mm}){3-4} \cmidrule(l{1mm}r{1mm}){5-6} \cmidrule(l{1mm}r{0mm}){7-8}
 &  & \hspace{1mm}MAE $\downarrow$ & \hspace{1mm}FAE $\downarrow$ & \hspace{1mm}MAE $\downarrow$ & \hspace{1mm}FAE $\downarrow$ & \hspace{1mm}MAE $\downarrow$ & \hspace{1mm}FAE $\downarrow$ \\ \midrule
RFC-Explicit (Ours) & \cmark & \textbf{2.498} & \textbf{2.893} & 2.379 & \textbf{2.802} & 0.557 & 0.710 \\
RFC-Implicit (Ours) & \cmark & \textbf{2.498} & 2.905 & \textbf{2.377} & \textbf{2.802} & \textbf{0.556} & \textbf{0.701} \\ \midrule
EgoPose~\cite{yuan2019ego}      & \cmark & 2.784 & 3.732 & 2.804 & 3.893 & 0.922 & 1.164 \\
ERD~\cite{fragkiadaki2015recurrent}          & \xmark & 2.770 & 3.223 & 3.066 & 3.578 & 0.682 & 1.092 \\
acLSTM~\cite{li2017auto}       & \xmark & 2.909 & 3.315 & 3.386 & 3.860 & 0.715 & 1.130 \\
\bottomrule
\end{tabular}
\vspace{5mm}
\caption{Quantitative results for human motions synthesis.}
\label{rfc:table:quan}
\end{table} 
\subsection{Extended Motion Synthesis}
\paragraph{Datasets.}
Our experiments are performed with two motion capture datasets: Human3.6M~\cite{ionescu2013human3} and EgoMocap~\cite{yuan2019ego}. Human3.6M is a large-scale dataset with 11 subjects (7 labeled) and 3.6 million total video frames. Each subject performs 15 actions in 30 takes where each take lasts from 1 to 5 minutes. We consider two evaluation protocols: (1) Mix, where we train and test on all 7 labeled subjects but using different takes; (2) Cross, where we train on 5 subjects (S1, S5, S6, S7, S8) and test on 2 subjects (S9 and S11). We train a model for each action for all methods. The other dataset, EgoMocap, is a relatively small dataset including 5 subjects and around 1 hour of motions. We train the models using the default train/test split in the mixed subject setting. Both datasets are resampled to 30Hz to conform to the policy.

\paragraph{Implementation Details.}
The simulation setup is the same as the motion imitation task. We build two humanoids, one with 52 DoFs and 18 rigid bodies for Human3.6M and the other one with 59 DoFs and 20 rigid bodies for EgoMocap. For both datasets, the kinematic policy $\kappa_\psi$ observes motions of $p=30$ steps (1s) to forecast motions of $f=60$ steps (2s). When training the control policy $\widetilde{\pi}_\theta$, we generate $n=5$ segments of future motions with $\kappa_\psi$.

\paragraph{Baselines and Metrics.}
We compare our approach against two well-known kinematics-based motion synthesis methods, ERD~\cite{fragkiadaki2015recurrent} and acLSTM~\cite{li2017auto}, as well as a physics-based motion synthesis method that does not require task guidance or user input, EgoPose~\cite{yuan2019ego}. We use two metrics, mean angle error (MAE) and final angle error (FAE). MAE computes the average Euclidean distance between predicted poses and ground truth poses in angle space, while FAE computes the distance for the final frame. Both metrics are computed with a forecasting horizon of 2s. For stochastic methods, we generate 10 future motion samples to compute the mean of the metrics.

\begin{wraptable}{r}{0.42\textwidth}
\footnotesize
\centering
\begin{tabular}{@{\hskip 1mm}cccc@{\hskip 1mm}}
\toprule
 \multicolumn{2}{c}{Component} & \multicolumn{2}{c}{Metric}\\ \cmidrule(l{0mm}r{1mm}){1-2}\cmidrule(l{1mm}r{0mm}){3-4}
AddAct & ResForce & \hspace{1mm}MAE $\downarrow$ & \hspace{1mm}FAE $\downarrow$\\ \midrule
\cmark & \cmark & \textbf{2.498} & \textbf{2.893} \\
\cmark & \xmark & 2.610 & 3.150 \\
\xmark & \cmark & 3.099 & 3.634 \\
\bottomrule
\end{tabular}
\vspace{5mm}
\caption{Ablation Study.}
\label{rfc:table:abl}
\end{wraptable}
\paragraph{Results.} 
In Table~\ref{rfc:table:quan}, we show quantitative results of all models for motion forecasting over the 2s horizon, which evaluates the methods' ability to infer future motions from the past. For all datasets and evaluation protocols, our RFC models with dual-policy control outperform the baselines consistently in both metrics. We hypothesize that the performance gain over the other physics-based method, EgoPose, can be attributed to the use of kinematic policy and the residual forces. To verify this hypothesis, we conduct an ablation study in the Human3.6M (Mix) setting. We train model variants of RFC-Implicit by removing the residual forces (ResForce) or the additive action (AddAct) that uses the kinematic policy's output. Table~\ref{rfc:table:abl} demonstrates that, in either case, the performance decreases for both metrics, which supports our previous hypothesis. Unlike prior physics-based methods, our approach also enables synthesizing sitting motions even when the chair is not modeled in the physics environment, because the learned residual forces can provide the contact forces need to support the humanoid. Furthermore, our model allows infinite-horizon stable motion synthesis by autoregressively applying dual policy control. As motions are best seen in videos, please refer to the supplementary \href{https://youtu.be/XuzH1u78o1Y}{video} for qualitative results.

\section{Conclusion}
In this work, we proposed residual force control (RFC), a novel and simple method to address the dynamics mismatch between the humanoid model and real humans. RFC uses external residual forces to provide a learnable time-varying correction to the dynamics of the humanoid model, which results in a more flexible new dynamics that admits a wider range of agile motions. Experiments showed that RFC outperforms state-of-the-art motion imitation methods in terms of convergence speed and motion quality. RFC also enabled the humanoid to learn sophisticated skills like ballet dance which have eluded prior work. Furthermore, we proposed a dual-policy control framework to synthesize multi-modal infinite-horizon human motions without any task guidance or user input, which opened up new avenues for automated motion generation and large-scale character animation. We hope our exploration of the two aspects of human motion, dynamics and kinematics, can encourage more work to view the two from a unified perspective. One limitation of the RFC framework is that it can only be applied to simulation domains (e.g., animation, motion synthesis, pose estimation) in its current form, as real robots cannot generate external residual forces. However, we do believe that RFC could be applied as a warm-up technique to accelerate the learning of complex policies for real robots. Further, the residual forces needed to overcome the dynamics mismatch could also be used to inform and optimize agent design. These are all interesting avenues for future work.

\chapter{Efficient Automatic Design of Simulated Agents}
\label{chap:transform2act}

\vspace{-5mm}
\section{Introduction}

\begin{figure}[t]
    \centering
    \includegraphics[width=\linewidth]{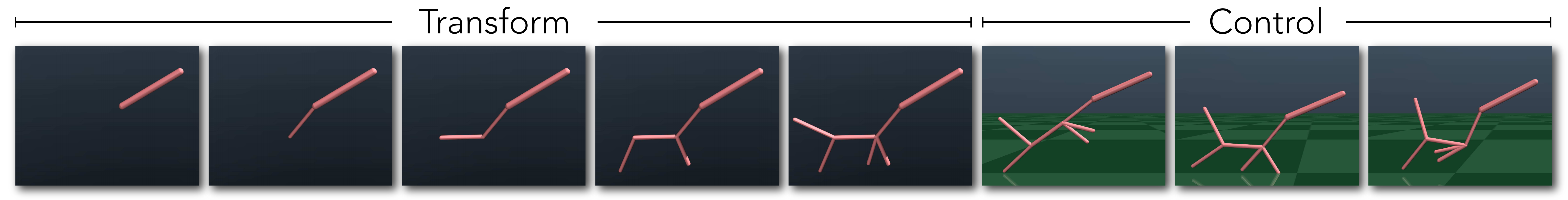}
    \vspace{-2mm}
    \caption{Transform2Act learns a transform-and-control policy that first applies transform actions to design an agent and then controls the designed agent to interact with the environment. The giraffe-like agent obtained by Transform2Act can run extremely fast (see \href{https://sites.google.com/view/transform2act}{video}).}
    \label{fig:teaser}
\end{figure}

Automatic and efficient design of robotic agents in simulation holds great promise in complementing and guiding the traditional physical robot design process~\cite{ha2017joint,ha2018computational}, which can be laborious and time-consuming. In this paper, we consider a setting where we optimize both the skeletal structure and joint attributes (e.g., bone length, size, and motor strength) of an agent to maximize its performance on a given task. This is a very challenging problem for two main reasons. First, the design space, i.e., all possible skeletal structures and their joint attributes, is prohibitively vast and combinatorial, which also makes applying gradient-based continuous optimization methods difficult. Second, the problem is inherently bi-level: (1) we need to search an immensely large design space and (2) the evaluation of each candidate design entails solving a computationally expensive inner optimization to find its optimal controller. Prior work typically uses evolutionary search (ES) algorithms for combinatorial design optimization~\cite{sims1994evolving, wang2019neural}. During each iteration, ES-based methods maintain a large population of agents with various designs where each agent learns to perform the task independently. When the learning ends, agents with the worst performances are eliminated while surviving agents produce child agents with random mutation to maintain the size of the population. ES-based methods have low sample efficiency since agents in the population do not share their experiences and many samples are wasted on eliminated agents. Furthermore, zeroth-order optimization methods such as ES are known to be sample-inefficient when the optimization search space (i.e., design space) is high-dimensional \cite{vemula2019contrasting}.

In light of the above challenges, we take a new approach to agent design optimization by incorporating the design procedure of an agent into its decision-making process. Specifically, we learn a transform-and-control policy, called Transform2Act, that first designs an agent and then controls the designed agent. In an episode, we divide the agent's behavior into two consecutive stages, a transform stage and an execution stage, on which the policy is conditioned. In the transform stage, the agent applies a sequence of transform actions to modify its skeletal structure and joint attributes without interacting with the environment. In the execution stage, the agent assumes the design resulting from the transform stage and applies motor control actions to interact with the environment and receives rewards. Since the policy needs to be used across designs with a variable number of joints, we adopt graph neural networks (GNNs) as the policy's main network architecture. Each graph node in the GNNs represents a joint and uses message passing with its neighbors to output joint-specific actions. While GNNs can improve the generalizability of learned control policies across different skeletons through weight sharing, they can also limit the specialization of each joint's design since similar joints tend to output similar transform actions due to the weight sharing in GNNs. To tackle this problem, we propose to attach a joint-specialized multilayer perceptron (JSMLP) on top of the GNNs in the policy. The JSMLP uses different sets of weights for each joint, which allows more flexible transform actions to enable asymmetric designs with more specialized joint functions.

The proposed Transform2Act policy jointly optimizes its transform and control actions via policy gradient methods, where the training batch for each iteration includes samples collected under various designs. Unlike zeroth-order optimization methods (e.g., ES) that do not use a policy to change designs but instead mutate designs randomly, our approach stores information about the goodness of a design into our transform policy and uses it to select designs to be tested in the execution stage. The use of GNNs further allows the information stored in the policy to be shared across joints, enabling better experience sharing and prediction generalization for both design and control. Furthermore, in contrast to ES-based methods that do not share training samples across designs in a generation, our approach uses all the samples from all designs to train our policy, which improves sample efficiency.

The main contributions of this paper are: (1) We propose a transform-and-control paradigm that formulates design optimization as learning a conditional policy, which can be solved using the rich tools of RL. (2) Our GNN-based conditional policy enables joint optimization of design and control as well as experience sharing across all designs, which improves sample efficiency substantially. (3)~We further enhance the GNNs by proposing a joint-specialized MLP to balance the generalization and specialization abilities of our policy. (4) Experiments show that our approach outperforms previous methods significantly in terms of convergence speed and final performance, and is also able to discover familiar designs similar to giraffes, squids, and spiders. 

\section{Related Work}
\label{transform2act:sec:related}
\paragraph{Continuous Design Optimization.} Considerable research has examined optimizing an agent's continuous design parameters without changing its skeletal structure. For instance, \cite{baykal2017asymptotically} introduce a simulated annealing-based optimization framework for designing piecewise
cylindrical robots. Alternatively, trajectory optimization and the implicit function theorem have been used to adapt the design of legged robots~\cite{ha2017joint,ha2018computational,desai2018interactive}. Recently, deep RL has become a popular approach for design optimization. \cite{chen2020hardware} model robot hardware as part of the policy using computational graphs. \cite{luck2020data} learn a design-conditioned value function and optimize design via CMA-ES. \cite{ha2019reinforcement} uses a population-based policy gradient method for design optimization. \cite{schaff2019jointly} employs RL and evolutionary strategies to maintain a distribution over design parameters. Another line of work \cite{yu2018policy,exarchos2020policy,jiang2021simgan} uses RL to find the robot parameters that best fit an incoming domain. Unlike the above works, our approach can optimize the skeletal structure of an agent in addition to its continuous design parameters.

\paragraph{Combinatorial Design Optimization.} Towards jointly optimizing the skeletal structure and design parameters, the seminal work by \cite{sims1994evolving} uses evolutionary search (ES) to optimize the design and control of 3D blocks. Extending this method, \cite{cheney2014unshackling, cheney2018scalable} adopt oscillating 3D voxels as building blocks to reduce the search space. \cite{desai2017computational} use human-in-the-loop tree search to optimize the design of modular robots. \cite{wang2019neural} propose an evolutionary graph search method that uses GNNs to enable weight sharing between an agent and its offspring. Recently, \cite{hejna2021task} employ an information-theoretic objective to evolve task-agnostic agent designs. Due to the combinatorial nature of skeletal structure optimization, most prior works use ES-based optimization frameworks which can be sample-inefficient since agents with various designs in the population learn independently. In contrast, we learn a transform-and-control policy using samples collected from all designs, which improves sample efficiency significantly.

\paragraph{GNN-based Control.}
Graph neural networks (GNNs) \cite{scarselli2008graph, bruna2013spectral, kipf2016semi} are a class of models that use message passing to aggregate and extract features from a graph. GNN-based control policies have been shown to greatly improve the generalizability of learned controllers across agents with different skeletons~\cite{wang2018nervenet,wang2019neural,huang2020one}. Along this line, \cite{pathak2019learning} use a GNN-based policy to control self-assembling modular robots to perform tasks. Recently, \cite{kurin2021my} show that GNNs can hinder learning in incompatible multitask RL and propose to use attention mechanisms instead. In this paper, we also study the lack of per-joint specialization caused by GNNs due to weight sharing, and we propose a remedy, joint-specialized MLP, to improve the specialization of GNN-based policies.

\section{Background}
\label{transform2act:sec:background}

\paragraph{Reinforcement Learning.}
Given an agent interacting with an episodic environment, reinforcement learning (RL) formulates the problem as a Markov Decision Process (MDP) defined by a tuple $\mathcal{M} = (\mathcal{S}, \mathcal{A}, \mathcal{T}, R, \gamma)$ of state space, action space, transition dynamics, a reward function, and a discount factor. The agent's behavior is controlled by a policy $\pi(a_t|s_t)$, which models the probability of choosing an action $a_t \in \mathcal{A}$ given the current state $s_t \in \mathcal{S}$. Starting from some initial state $s_0$, the agent iteratively samples an action $a_t$ from the policy $\pi$ and the environment generates the next state $s_{t+1}$ based on the transition dynamics $\mathcal{T}(s_{t+1}|s_t, a_t)$ and also assigns a reward $r_t$ to the agent. The goal of RL is to learn an optimal policy $\pi^\ast$ that maximizes the expected total discounted reward received by the agent: $ J(\pi) = \mathbb{E}_{\pi}\left[\sum_{t=0}^H\gamma^t r_t\right]$, where $H$ is the variable time horizon. In this paper, we use a standard policy gradient method, PPO~\cite{schulman2017proximal}, to optimize our policy with both transform and control actions. PPO is particularly suitable for our approach since it has a KL regularization between current and old policies, which can prevent large changes to the transform actions and the resulting design in each optimization step, thus avoiding catastrophic failure.

\paragraph{Design Optimization.}
An agent's design $D \in \mathcal{D}$ plays an important role in its functionality. In our setting, the design $D$ includes both the agent's skeletal structure and joint-specific attributes (e.g., bone length, size, and motor strength). To account for changes in design, we now consider a more general transition dynamics $\mathcal{T}(s_{t+1}|s_t, a_t, D)$ conditioned on design $D$. The total expected reward is also now a function of design $D$: $ J(\pi, D) = \mathbb{E}_{\pi, D}\left[\sum_{t=0}^H\gamma^t r_t\right]$. One main difficulty of design optimization arises from its bi-level nature, i.e., we need to search over a large design space and solve for the optimal policy under each candidate design for evaluation. Formally, the bi-level optimization is defined as:
\begin{align}
    D^\ast & = \argmax_{D} J(\pi_D, D) \\
    \hspace{-10mm}\text{subject to} \quad \pi_D & = \argmax_{\pi} J(\pi, D)
    \label{transform2act:eq:biopt_inner}
\end{align}
The inner optimization described by Equation~(\ref{transform2act:eq:biopt_inner}) typically requires RL, which is computationally expensive and may take up to several days depending on the task. Additionally, the design space $\mathcal{D}$ is extremely large and combinatorial due to a vast number of possible skeletal structures. To tackle these problems, in Sec.~\ref{transform2act:sec:main} we will introduce a new transform-and-control paradigm that formulates design optimization as learning a conditional policy to both design and control the agent. It uses first-order policy optimization via policy gradient methods and enables experience sharing across designs, which improves sample efficiency significantly.

\paragraph{Graph Neural Networks.}
Since our goal is to learn a policy to dynamically change an agent's design, we need a network architecture that can deal with variable input sizes across different skeletal structures. As skeletons can naturally be represented as graphs, graph neural networks (GNNs)~\cite{scarselli2008graph,bruna2013spectral,kipf2016semi} are ideal for our use case.

We denote a graph as $G = (V, E, A)$ where $u \in V$ and $e \in E$ are nodes and edge respectively, and each node $u$ also includes an input feature $x_u \in A$. A GNN uses multiple GNN layers to extract and aggregate features from the graph $G$ through message passing. For the $i$-th of $N$ GNN layers, the message passing process can be written as:
\begin{align}
    m_u^i &= M(h_u^{i-1})\,, \\
    c_{u}^{i} &=C(\left\{m_{v}^i \mid \forall v \in \mathcal{N}(u)\right\}), \\
    h_{u}^t &=U(h_u^{i-1}, c_{u}^i),
\end{align}
where a message sending module $M$ first computes a message $m_u^i$ for each node $u$ from the hidden features $h_u^{i-1}$ ($h_u^0 = x_u$) of the previous layer. Each node's message is then sent to neighboring nodes $\mathcal{N}(u)$, and an aggregation module $C$ summarizes the messages received by every node and outputs a combined message $c_u^i$. Finally, a state update module $U$ updates the hidden state $h_u^i$ of each node using $c_u^i$ and previous hidden states $h_u^{i-1}$.
After $N$ GNN layers, an output module $P$ is often used to regress the final hidden features $h_u^N$ to desired outputs $y_u = P(h_u^N)$. Different designs of modules $M, C, U, P$ lead to many variants of GNNs~\cite{wu2020comprehensive}. We use the following equation to summarize GNNs' operations:
\begin{equation}
y_u = \mathrm{GNN}(u, A; V, E)\,
\end{equation}
where $\mathrm{GNN}(u, \cdot)$ is used to denote the output for node $u$.

\section{Transform2Act: a Transform-and-Control Policy}
\label{transform2act:sec:main}

\begin{figure}[t]
    \centering
    \includegraphics[width=\linewidth]{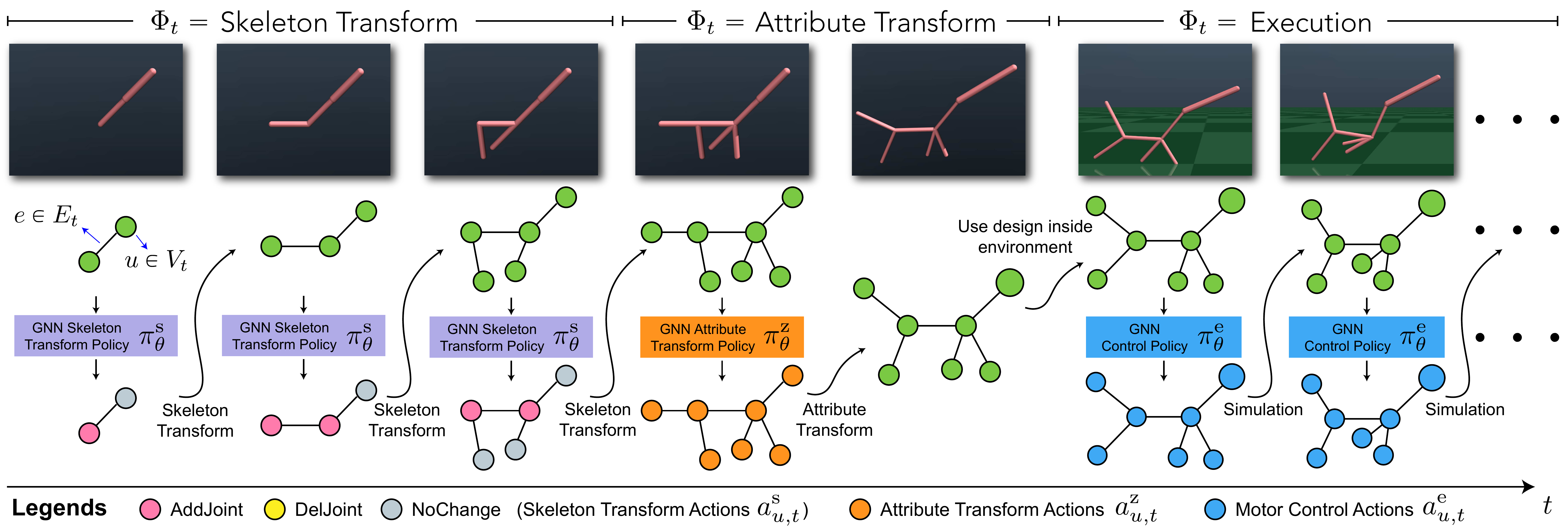}
    \caption{Transform2Act divides an episode into three stages: (1) Skeleton transform stage, where sub-policy $\pi_\theta^\mathrm{s}$ changes the agent's skeleton by adding or removing joints; (2) Attribute transform stage, where sub-policy $\pi_\theta^\mathrm{z}$ changes joint attributes (e.g., length, size); (3) Execution stage, sub-policy $\pi_\theta^\mathrm{e}$ selects control actions for the newly-designed agent to interact with the environment.}
    \label{transform2act:fig:overview}
\end{figure}

To tackle the challenges in design optimization, our key approach is to incorporate the design procedure of an agent into its decision-making process. In each episode,  the agent's behavior is separated into two consecutive stages: (1) \textbf{{Transform Stage}}, where the agent applies transform actions to modify its design, including skeletal structure and joint attributes, \emph{without} interacting with the environment; (2) \textbf{{Execution Stage}}, where the agent assumes the new transformed design and applies motor control actions to interact with the environment. In both stages, the agent is governed by a conditional policy, called \emph{Transform2Act}, that selects transform or control actions depending on which stage the agent is in. In the transform stage, no environment reward is assigned to the agent, but the agent will see future rewards accrued in the execution stage under the transformed design, which provide learning signals for the transform actions.
By training the policy with PPO~\cite{schulman2017proximal}, the transform and control actions are optimized jointly to improve the agent's performance for the given task. An overview of our method is provided in Figure~\ref{transform2act:fig:overview}. We also outline our approach in Algorithm~\ref{alg:design_opt}. In the following, we first introduce the preliminaries before describing the details of the proposed Transform2Act policy and the two stages.

\paragraph{Design Representation.}
To represent various skeletal structures, we denote an agent design as a graph $D_t = (V_t, E_t, A_t)$ where each node $u \in V_t$ represents a joint $u$ in the skeleton and each edge $e \in E_t$ represents a bone connecting two joints, and $z_{u,t} \in A_t$ is a vector representing the attributes of joint $u$ including bone length, size, motor strength, etc. Here, the design $D_t$ is indexed by $t$ since it can be changed by transform actions during the transform stage.

\paragraph{MDP with Design.}
To accommodate the transform actions and changes in agent design $D_t$, we redefine the agent's MDP by modifying the state and action space as well as the transition dynamics. Specifically, the new state $s_t = (s_t^\mathrm{e}, D_t, \Phi_t)$ includes the agent's state $s_t^\mathrm{e}$ in the environment and the agent design $D_t = (V_t, E_t, A_t)$, as well as a stage flag $\Phi_t$. The new action $a_t \in \{a_t^\mathrm{d}, a_t^\mathrm{e} \}$ consists of both the transform action $a_t^\mathrm{d}$ and the motor control action $a_t^\mathrm{e}$. The new transition dynamics $\mathcal{T}(s_{t+1}^\mathrm{e}, D_{t+1}, \Phi_{t+1}|s_t^\mathrm{e}, D_t, \Phi_t, a_t)$ reflects the changes in the state and action.

\paragraph{Transform2Act Policy.}
The policy $\pi_\theta$ with parameters $\theta$ is a conditional policy that selects the type of actions based on which stage the agent is in:
\begin{equation}
\pi_\theta(a_t|s_t^\mathrm{e}, D_t, \Phi_t) =  
\begin{cases}
    \pi^\mathrm{d}_\theta(a_t^\mathrm{d}| D_t, \Phi_t) ,& \text{if } \Phi_t = \mathrm{Transform}\\
    \pi^\mathrm{e}_\theta(a_t^\mathrm{e}| s_t^\mathrm{e}, D_t, \Phi_t), & \text{if } \Phi_t = \mathrm{Execution}
\end{cases}
\end{equation}
where two sub-policies $\pi_\theta^\mathrm{d}$ and $\pi_\theta^\mathrm{e}$ are used in the transform and execution stages respectively. As the agent in the transform stage does not interact with the environment, the transform sub-policy $\pi^\mathrm{d}_\theta(a_t^\mathrm{d}| D_t, \Phi_t)$ is not conditioned on the environment state $s_t^\mathrm{e}$ and only outputs transform actions $a_t^\mathrm{d}$. The execution sub-policy $\pi^\mathrm{e}_\theta(a_t^\mathrm{e}| s_t^\mathrm{e}, D_t, \Phi_t)$ is conditioned on both $s_t^\mathrm{e}$ and the design $D_t$ to output control actions $a_t^\mathrm{e}$. $D_t$ is needed since the transformed design will affect the dynamics of the environment in the execution stage. As a notation convention, policies with different superscripts (e.g., $\pi^\mathrm{d}_\theta$ and $\pi^\mathrm{e}_\theta$) do not share the same set of parameters.

\subsection{Transform Stage}
\label{transform2act:sec:transform}
In the transform stage, starting from an initial design $D_0$, the agent follows the transform sub-policy $\pi^\mathrm{d}_\theta(a_t^\mathrm{d}| D_t, \Phi_t)$ which outputs transform actions to modify the design. Since the design $D_t = (V_t, E_t, A_t)$ includes both the skeletal graph $(V_t, E_t)$ and joint attributes $A_t$, the transform action $a_t^\mathrm{d} \in \{a_t^\mathrm{s}, a_t^\mathrm{z}\}$ consists of two types: (1) \textbf{{Skeleton Transform Action $a_t^\mathrm{s}$}}, which is discrete and changes the skeletal graph $(V_t, E_t)$ by adding or deleting joints;
(2) \textbf{{Attribute Transform Action $a_t^\mathrm{z}$}}, which modifies the attributes of each joint and can be either continuous or discrete. Based on the two types of transform actions, we further divide the transform stage into two sub-stages -- \textbf{{Skeleton Transform Stage}} and \textbf{{Attribute Transform Stage}} -- where $a_t^\mathrm{s}$ and $a_t^\mathrm{z}$ are taken by the agent respectively. We can then write the transform sub-policy as conditioned on the agent's stage:
\begin{equation}
\pi^\mathrm{d}_\theta(a_t^\mathrm{d}| D_t, \Phi_t) =  
\begin{cases}
    \pi^\mathrm{s}_\theta(a_t^\mathrm{s}| D_t, \Phi_t) ,& \text{if } \Phi_t = \mathrm{Skeleton\text{ }Transform}\\
    \pi^\mathrm{z}_\theta(a_t^\mathrm{z}| D_t, \Phi_t), & \text{if } \Phi_t = \mathrm{Attribute\text{ }Transform}
\end{cases}
\end{equation}
where the agent follows sub-policies $\pi_\theta^\mathrm{s}$ for $N_\mathrm{s}$ timesteps in the skeleton transform stage and then follows $\pi_\theta^\mathrm{z}$ for $N_\mathrm{z}$ timesteps in the attribute transform stage. It may be tempting to merge the two stages and apply skeleton and attribute transform actions together. However, we separate the two stages since the skeleton action can increase the number of joints in the graph, while attribute transform actions can only output attribute changes for existing joints.

\paragraph{Skeleton Transform.}
The skeleton transform sub-policy $\pi^\mathrm{s}_\theta(a_t^\mathrm{s}| D_t, \Phi_t)$ adopts a GNN-based network, where the action $a_t^\mathrm{s}=\{a_{u,t}^\mathrm{s}|u\in V_t\}$ is factored across joints and each joint $u$ outputs its categorical action distribution $\pi^\mathrm{s}_\theta(a_{u,t}^\mathrm{s}| D_t, \Phi_t)$. The policy distribution is the product of all joints' action distributions: 
\begin{align}
     \pi^\mathrm{s}_\theta(a_{u,t}^\mathrm{s}| D_t, \Phi_t) = \mathcal{C}(a_{u,t}^\mathrm{s}; l_{u,t}), \quad l_{u,t} &= \mathrm{GNN_s}(u, A_t; V_t, E_t), \quad  \forall u \in V_t, \\
    \pi^\mathrm{s}_\theta(a_t^\mathrm{s}| D_t, \Phi_t) = & \prod_{u \in V_t} \pi^\mathrm{s}_\theta(a_{u,t}^\mathrm{s}| D_t, \Phi_t)\,,
\end{align}
where the GNN uses the joint attributes $A_t$ as input node features to output the logits $l_{u,t}$ of each joint's categorical action distribution $\mathcal{C}$. The skeleton transform action $a_{u,t}^\mathrm{s}$ has three choices:
\begin{itemize}[leftmargin=7mm]
    \item \textbf{AddJoint}: joint $u$ will add a child joint $v$ to the skeletal graph, which inherits its attribute $z_{u,t}$.
    \item \textbf{DelJoint}: joint $u$ will remove itself from the skeletal graph. The action is only performed when the joint $u$ has no child joints, which is to prevent aggressive design changes.
    \item \textbf{NoChange}: no changes will be made to joint $u$.
\end{itemize}
After the agent applies the skeleton transform action $a_{u,t}^\mathrm{s}$ at each joint $u$, we obtain the design $D_{t+1} = (V_{t+1}, E_{t+1}, A_{t+1})$ for the next timestep with a new skeletal structure.

The GNN-based skeleton transform policy enables rapid growing of skeleton structures since every joint can add a child joint at each timestep. Additionally, it also encourages symmetric structures due to weight sharing in GNNs where mirroring joints can choose the same action.

\begin{algorithm}[t]
\caption{Agent Design Optimization with Transform2Act Policy}\label{alg:design_opt}
\begin{algorithmic}[1]
\State initialize Transform2Act policy $\pi_\theta$
\While{not reaching max iterations}
    \State memory $\mathcal{M} \leftarrow \emptyset$
    \While{$\mathcal{M}$ not reaching batch size}
        \State $D_0 \gets$ initial agent design
        \State // Skeleton Transform Stage
        \For{$t = 0,1, \ldots, N_\mathrm{s}-1$}
            \State sample skeleton transform action $a_t^\mathrm{s} \sim \pi^\mathrm{s}_\theta$; $\Phi_t \gets \mathrm{Skeleton\text{ }Transform}$
            \State $D_{t+1} \gets$ apply $a_t^\mathrm{s}$ to modify skeleton $(V_t, E_t)$ in $D_t$
            \State $r_t \gets 0$; store $(r_t, a_t^\mathrm{s}, D_t, \Phi_t)$ into $\mathcal{M}$
        \EndFor
        \State // Attribute Transform Stage
        \For{$t = N_\mathrm{s}, \ldots, N_\mathrm{s} + N_\mathrm{z} - 1$}
            \State sample attribute transform action $a_t^\mathrm{z} \sim \pi^\mathrm{z}_\theta$; $\Phi_t \gets \mathrm{Attribute\text{ }Transform}$
            \State $D_{t+1} \gets$ apply $a_t^\mathrm{z}$ to modify attributes $A_t$ in $D_t$
            \State $r_t \gets 0$; store $(r_t, a_t^\mathrm{z}, D_t, \Phi_t)$ into $\mathcal{M}$
        \EndFor
        \State // Execution Stage
        \State $s_{N_\mathrm{s} + N_\mathrm{z}}^\mathrm{e} \gets$ initial environment state 
        \For{$t = N_\mathrm{s} + N_\mathrm{z}, \ldots, H$}
            \State sample motor control action $a_t^\mathrm{e} \sim \pi^\mathrm{e}_\theta$; $\Phi_t \gets \mathrm{Execution}$
            \State $s_{t+1}^\mathrm{e} \gets$ environment dynamics $\mathcal{T}^\mathrm{e}(s_{t+1}^\mathrm{e}|s_t^\mathrm{e},a_t^\mathrm{e})$; $D_{t+1} \gets D_t$
            \State $r_t \gets$ environment reward; store $(r_t, a_t^\mathrm{s}, s_t^\mathrm{e}, D_t, \Phi_t)$ into $\mathcal{M}$
        \EndFor
    \EndWhile
    \State update $\pi_\theta$ with PPO using samples in $\mathcal{M}$
\EndWhile\\
\Return{$\pi_\theta$}
\end{algorithmic}
\end{algorithm}

\paragraph{Attribute Transform.}
The attribute transform sub-policy $\pi^\mathrm{z}_\theta(a_t^\mathrm{z}| D_t, \Phi_t)$ adopts the same GNN-based network as the skeleton transform sub-policy $\pi^\mathrm{s}_\theta$. The main difference is that the output action distribution can be either continuous or discrete. In this paper, we only consider continuous attributes including bone length, size, and motor strength, but our method by design can generalize to discrete attributes such as geometry types. The policy distribution is defined as:
\begin{align}
     \pi^\mathrm{z}_\theta(a_{u,t}^\mathrm{z}| D_t, \Phi_t) = \mathcal{N}(a_{u,t}^\mathrm{z};\mu_{u,t}^\mathrm{z}, \Sigma^\mathrm{z}), \quad \mu_{u,t}^\mathrm{z} &= \mathrm{GNN_z}(u, A_t; V_t, E_t), \quad  \forall u \in V_t, \\
    \pi^\mathrm{z}_\theta(a_t^\mathrm{z}| D_t, \Phi_t) = & \prod_{u \in V_t} \pi^\mathrm{z}_\theta(a_{u,t}^\mathrm{z}| D_t, \Phi_t)\,,
\end{align}
where the GNN outputs the mean $\mu_{u,t}^\mathrm{z}$ of joint $u$'s Gaussian action distribution and $\Sigma^\mathrm{z}$ is a learnable diagonal covariance matrix independent of $D_t,\Phi_t$ and shared by all joints.
Each joint's action $a_{u,t}^\mathrm{z}$ is used to modify its attribute feature: $z_{u,t+1} = z_{u,t} + a_{u,t}^\mathrm{z}$, and the new design becomes $D_{t+1} = (V_t, E_t, A_{t+1})$ where the skeleton $(V_t, E_t)$ remains unchanged.

\paragraph{Reward.}
During the transform stage, the agent does not interact with the environment, because changing the agent's design such as adding or removing joints while interacting with the environment does not obey the laws of physics and may be exploited by the agent. Since there is no interaction, we do not assign any environment rewards to the agent. While it is possible to add rewards based on the current design to guide the transform actions, in this paper, we do not use any design-related rewards for fair comparison with the baselines. Thus, no reward is assigned to the agent in the transform stage, and the transform sub-policies are only trained using future rewards from the execution stage.

\paragraph{Inference.}
At test time, the most likely action will be chosen by both the skeleton and attribute transform policies. The design $D_t$ after the transform stage is the final design output.

\subsection{Execution Stage}
\label{transform2act:sec:execute}
After the agent performs $N_\mathrm{s}$ skeleton transform and $N_\mathrm{z}$ attribute transform actions, it enters the execution stage where the agent assumes the transformed design and interacts with the environment. A GNN-based execution policy $\pi^\mathrm{e}_\theta(a_t^\mathrm{e}| s_t^\mathrm{e}, D_t, \Phi_t)$ is used in this stage to output motor control actions $a_t^\mathrm{e}$ for each joint. Since the agent now interacts with the environment, the policy $\pi^\mathrm{e}_\theta$ is conditioned on the environment state $s_t^\mathrm{e}$ as well as the transformed design $D_t$, which affects the dynamics of the environment. Without loss of generality, we assume the control actions are continuous. The execution policy distribution is defined as:
\begin{align}
     \pi^\mathrm{e}_\theta(a_{u,t}^\mathrm{e}| s_t^\mathrm{e}, D_t, \Phi_t) = \mathcal{N}(a_{u,t}^\mathrm{e};\mu_{u,t}^\mathrm{e}, \Sigma^\mathrm{e}), \quad \mu_{u,t}^\mathrm{e} &= \mathrm{GNN_e}(u, s_t^\mathrm{e}, A_t; V_t, E_t), \quad  \forall u \in V_t, \\
    \pi^\mathrm{e}_\theta(a_t^\mathrm{e}|s_t^\mathrm{e}, D_t, \Phi_t) = & \prod_{u \in V_t} \pi^\mathrm{e}_\theta(a_{u,t}^\mathrm{e}|s_t^\mathrm{e}, D_t, \Phi_t)\,,
\end{align}
where the environment state $s_t^\mathrm{e}=\{s_{u,t}^\mathrm{e}|u\in V_t\}$ includes the state of each node $u$ (e.g., joint angle and velocity). The GNN uses the environment state $s_t^\mathrm{e}$ and joint attributes $A_t$ as input node features to output the mean $\mu_{u,t}^\mathrm{e}$ of each joint's Gaussian action distribution. $\Sigma^\mathrm{e}$ is a state-independent learnable diagonal covariance matrix shared by all joints. The agent applies the motor control actions $a_t^\mathrm{e}$ to all joints and the environment transitions the agent to the next environment state $s_{t+1}^\mathrm{e}$ according to the environment's transition dynamics $\mathcal{T}^\mathrm{e}(s_{t+1}^\mathrm{e}|s_t^\mathrm{e},a_t^\mathrm{e})$. The design $D_t$ remains unchanged throughout the execution stage.

\subsection{Value Estimation}
\label{transform2act:sec:value_est}
As we use an actor-critic method (PPO) for policy optimization, we need to approximate the value function $\mathcal{V}$, i.e., the expected total discounted rewards starting from state $s_t = (s_t^\mathrm{e}, D_t, \Phi_t)$:
\begin{equation}
    \mathcal{V} (s_t^\mathrm{e}, D_t, \Phi_t) \triangleq \mathbb{E}_{\pi_\theta} \left[\sum_{t=0}^{H}\gamma^t r_t\right].
\end{equation}
We learn a GNN-based value network $\hat{\mathcal{V}}_\phi$ with parameters $\phi$ to approximate the true value function:
\begin{equation}
    \hat{\mathcal{V}}_\phi (s_t^\mathrm{e}, D_t, \Phi_t) = \mathrm{GNN_v}(\mathrm{root}, s_t^\mathrm{e}, A_t; V_t, E_t)\,
\end{equation}
where the GNN takes the environment state $s_t^\mathrm{e}$, joint attributes $A_t$ and stage flag $\Phi_t$ (one-hot vector) as input node features to output a scalar at each joint. We use the output of the root joint as the predicted value. The value network $\hat{\mathcal{V}}_\phi$ is used in all stages. In the transform stage, the environment state $s_t^\mathrm{e}$ is unavailable so we set it to 0.

\subsection{Improve Specialization with Joint-Specialized MLP}
\label{transform2act:sec:jsmlp}
As demonstrated in prior work~\cite{wang2018nervenet,huang2020one}, GNN-based control policies enjoy superb generalizability across different designs. The generalizability can be attributed to GNNs' weight sharing across joints, which allows new joints to share the knowledge learned by existing joints. However, weight sharing also means that joints in similar states will choose similar actions, which can seriously limit the transform policies and the per-joint specialization of the resulting design. Specifically, as both skeleton and attribute transform policies are GNN-based, due to weight sharing, joints in similar positions in the graph will choose similar or the same transform actions. While this does encourage the emergence of symmetric structures, it also limits the possibility of asymmetric designs such as different lengths, sizes, or strengths of the front and back legs.

To improve the per-joint specialization of GNN-based policies, we propose to add a joint-specialized multi-layer perceptron (JSMLP) after the GNNs. Concretely, the JSMLP uses different MLP weights for each joint, which allows the policy to output more specialized joint actions. To achieve this, we design a joint indexing scheme that is consistent across all designs and maintain a joint-indexed weight memory. The joint indexing is used to identify joint correspondence across designs where two joints with the same index are deemed the same. This allows the same joint in different designs to use the same MLP weights. However, we cannot simply index joints using a breadth-first search (BFS) for each design since some joints may appear or disappear across designs, and completely different joints can be assigned the same index in BFS. Instead, we index a joint based on the path from the root to the joint. As we will show in the ablation studies, JSMLPs can improve the performance of both transform and control policies.

\vspace{-2mm}
\section{Experiments}
\label{transform2act:sec:exp}
We design our experiments to answer the following questions: (1) Does our method, Transform2Act, outperform previous methods in terms of convergence speed and final performance? (2) Does \mbox{Transform2Act} create agents that look plausible? (3) How do critical components -- GNNs and JSMLPs -- affect the performance of Transform2Act? (4) Can Transform2Act design humanoids with better motion imitation ability?

\vspace{-1mm}
\paragraph{Environments.}
We evaluate Transform2Act on four distinct environments using the MuJoCo simulator~\cite{todorov2012mujoco}: (1) \textbf{2D Locomotion}, where a 2D agent living in an $xz$-plane is tasked with moving forward as fast as possible, and the reward is its forward speed. (2) \textbf{3D Locomotion}, where a 3D agent's goal is to move as fast as possible along $x$-axis and is rewarded by its forward speed along $x$-axis. (3) \textbf{Swimmer}, where a 2D agent living in water with 0.1 viscosity and confined in an $xy$-plane is rewarded by its moving speed along $x$-axis. (4) \textbf{Gap Crosser}, where a 2D agent living in an $xz$-plane needs to cross periodic gaps and is rewarded by its forward speed.

\vspace{-1mm}
\paragraph{Baselines.}
We compare Transform2Act with the following baselines that also optimize both the skeletal structure and joint attributes of an agent: (1) \textbf{Neural Graph Evolution (NGE)}~\cite{wang2019neural}, which is an ES-based method that uses GNNs to enable weight sharing between an agent and its offspring. (2) \textbf{Evolutionary Structure Search (ESS)}~\cite{sims1994evolving}, which is a classic ES-based method that has been used in recent works~\cite{cheney2014unshackling,cheney2018scalable}. (3) \textbf{Random Graph Search (RGS)}, which is a baseline employed in~\cite{wang2019neural} that trains a population of agents with randomly generated skeletal structures and joint attributes.

\begin{figure}[t]
    \centering
    \includegraphics[width=\linewidth]{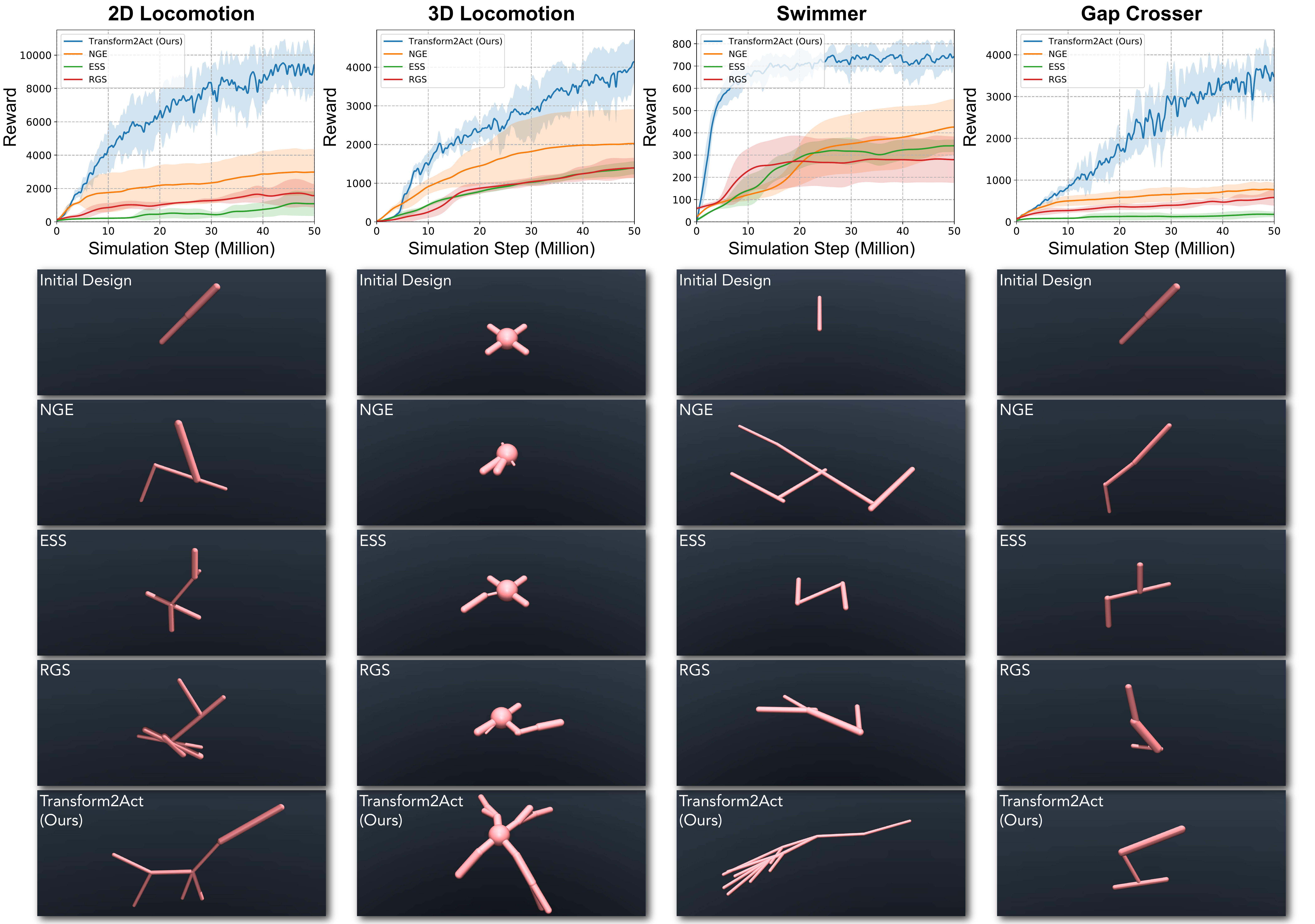}
    \caption{\textbf{Baseline comparison.} For each environment, we plot the mean and standard deviation of total rewards against the number of simulation steps for all methods, and show their final designs.}
    \label{transform2act:fig:plot_baseline}
\end{figure}

\vspace{-1mm}
\subsection{Comparison with Baselines}
In Figure~\ref{transform2act:fig:plot_baseline} we show the learning curves of each method and their final agent designs for all four environments. For each method, the learning curve plots use six seeds per environment and plot the agent's total rewards against the total number of simulation steps used by the method. For ES-based baselines with a population of agents, we plot the performance of the best agent at each iteration. We can clearly see that our method, Transform2Act, consistently and significantly outperforms the baselines in terms of convergence speed and final performance.

Next, let us compare the final designs generated by each method in Figure~\ref{transform2act:fig:plot_baseline}. For better comparison, we encourage the reader to see these designs in \href{https://sites.google.com/view/transform2act}{video} on the project website. For 2D locomotion, Transform2Act is able to discover a giraffe-like agent that can run extremely fast and remain stable. The design is suitable for optimizing the given reward function  since it has a long neck to increase its forward momentum and balance itself when jumping forward. For 3D Locomotion, Transform2Act creates a spider-like agent with long legs. The design is roughly symmetric due to the GNN-based transform policy, but it also contains joint-specific features thanks to the JSMLPs which help the agent attain better performance. For Swimmer, Transform2Act produces a squid-like agent with long bodies and numerous tentacles. As shown in the \href{https://sites.google.com/view/transform2act}{video}, the movement of these tentacles propels the agent forward swiftly in the water. Finally, for Gap Crosser, Transform2Act designs a Hopper-like agent that can jump across gaps. Overall, we can see that Transform2Act is able to find plausible designs similar to giraffes, squids, and spiders, while the baselines fail to discover such designs and have much lower performance in all four environments.

\begin{figure}[t]
    \centering
    \includegraphics[width=\linewidth]{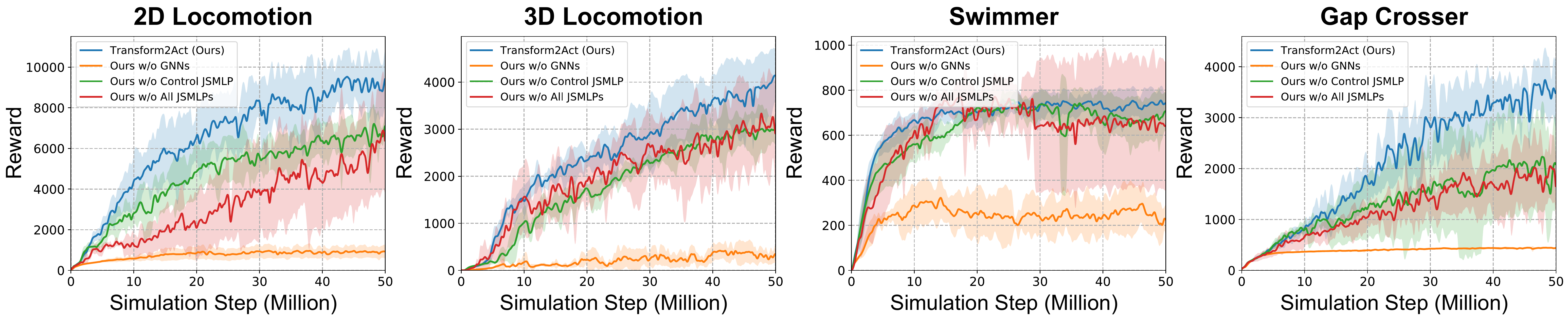}
    \caption{\textbf{Ablation studies.} The plots indicate that GNNs and JSMLPs both contribute to the performance and stability of our approach greatly.}
    \label{transform2act:fig:plot_ablation}
\end{figure}

\begin{figure}[t]
    \centering
    \includegraphics[width=\linewidth]{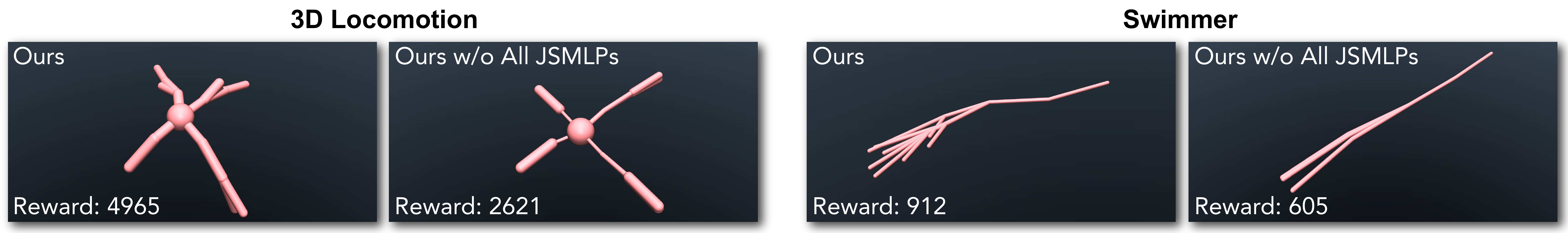}
    \caption{\textbf{Effect of JSMLPs.} Designs without JSMLPs are overly symmetric with little per-joint specialization, which leads to worse performance.}
    \label{transform2act:fig:plot_smlp}
\end{figure}

\vspace{-2mm}
\subsection{Ablation Studies}
We aim to investigate the importance of two critical components in our approach -- GNNs and JSMLPs. We design three variants of our approach: (1) \textbf{Ours w/o GNNs}, where we remove all the GNNs from our Transform2Act policy and uses JSMLP only; (2) \textbf{Ours w/o Control JSMLP}, where we remove the JSMLP from our execution sub-policy $\pi_\theta^\mathrm{e}$; (3) \textbf{Ours w/o All JSMLPs}, where we remove all the JSMLPs from the execution sub-policy $\pi_\theta^\mathrm{e}$ and transform sub-policies $\pi_\theta^\mathrm{s}$ and $\pi_\theta^\mathrm{z}$. The learning curves of all variants are shown in Figure~\ref{transform2act:fig:plot_ablation}. It is evident that GNNs are a crucial part of our approach as indicated by the large decrease in performance of the corresponding variant. Additionally, JSMLPs are very instrumental for both design and control in our method, which not only increase the agent's performance but also make the learning more stable. For Swimmer, the variant without JSMLPs has a very large performance variance. To further study the effect of JSMLPs, we also show the design with and without JSMLPs for 3D Locomotion and Swimmer in Figure~\ref{transform2act:fig:plot_smlp}. We can observe that the designs without JSMLPs are overly symmetric and uniform while the designs with JSMLPs contain joint-specialized features that help the agent achieve better performance.

\subsection{Discussion}
In this section, we will discuss the key reasons why our method outperforms the strongest baseline, NGE~\cite{wang2019neural}:
\begin{enumerate}[leftmargin=7mm]
    \item \textbf{NGE does not allow experience sharing among species in a generation.} As shown in Algorithm 1 of NGE, each species inside a generation $j$ has its own set of weights $\theta_i^j$ and is trained independently without sharing experiences among different species. The experience sharing in NGE is only enabled through weighting sharing between a species and its parent species from the previous generation. This means that if there are $N$ species in a generation, in every epoch, each species is only trained with $M/N$ number of experience samples where $M$ is the sample budget for each epoch. At the end of the training, each species has only used $EM/N$ samples for training where $E$ is the number of epochs. In contrast, in our method, every design shares the same control policy, so the policy can use all $EM$ samples. Therefore, our method allows better experience sharing across different designs, which improves sample efficiency.
    \item \textbf{Our method uses a transform policy to change designs instead of random mutation.} Our transform policy takes the current design as input to output the transform actions (design changes). Through training, the policy learns to store information about which design to prefer and which design to avoid. This information is also shared among different joints via the use of GNNs, where joints in similar states choose similar transform actions (which is also balanced by JSMLPs for joint specialization). Additionally, the policy also allows every joint to simultaneously change its design (e.g., add a child joint, change joint attributes). For example, in 3D Locomotion, the agent can simultaneously grow its four feet in a single timestep, while ES-based methods such as NGE will take four different mutations to obtain four feet. Therefore, our method with the transform policy allows better generalization and experience sharing among joints, compared to ES-based methods that perform random mutation.
    \item \textbf{Our method allows more exploration.} Our transform-and-control policy tries out a new design every episode, which means our policy can try $M/H_\text{avg}$ designs every epoch, where $M$ is the total number of sample timesteps and $H_\text{avg}$ is the average episode length. There is also more exploration for our approach at the start of the training when $H_\text{avg}$ is small. On the other hand, ES-based methods such as NGE only try $N$ (num of species) different designs every epoch. If NGE uses too many species (large $N$), each species will have few samples to train as mentioned in point 1. Therefore, $N$ is typically set to be $\ll M/H_\text{avg}$. For example, $N$ is set from 16 to 100 in NGE, while $M/H_\text{avg}$ in our method can be more than 2000.
\end{enumerate}
All the factors above contribute to the sample efficiency of our approach, allowing it to discover interesting and performant designs within a reasonable computing budget.

\subsection{Finetuning Humanoid Design}
To further evaluate our approach, Transform2Act, we use it for the task of finetuning the design of a humanoid to improve its motion imitation ability. The design search space of the humanoid include the geometries, density, motor gears, and friction coefficients of each joint. The humanoid is tasked to imitate challenging motion sequences in a large motion database, AMASS~\cite{AMASS:ICCV:2019}, such as belly dance, parkour, karate, cartwheeling, etc. Such sequences are highly dynamic and often require the human performer to have a special physique to properly carry out these motions. For instance, an expert dancer needs to be well-balanced while a crawler may have more prominent elbows. As a result, a ``vanilla'' humanoid may fail at imitating such sequences, and finetuning the humanoid's design can improve its motion imitation ability. To achieve this, we use a motion imitation reward in the execution stage of Transform2Act, which is based on how the motion performed by the humanoid aligns with the GT. We consider two training settings: single-sequence finetuning and multiple-sequence finetuning, where the former finetunes the humanoid design to best perform one sequence while the latter finetunes the design for a category of motion sequences.

\paragraph{Metrics.} We use the following metrics to assess the motion imitation performance of the designed humanoid: (1)~Mean per joint position error $ \bf E_{\text{mpjpe}}$ (mm), which is a popular metric for 3D human pose estimation and is computed after setting the root translation to zero; (2)~Global mean per joint position error $ \bf E_{\text{mpjpe-g}}$ (mm), which computes the joint position error in global space without zeroing out the root translation, thus better reflecting the overall motion tracking quality; (3)~Acceleration error $\text{E}_{\text{acc}}$ (mm/frame$^2$), which measures the difference between the estimated joint position acceleration and the GT;
(4)~Success rate $\text{S}_{\text{succ}}$, which measures whether the humanoid has fallen or deviates too far away from the reference motion during the sequence.

\begin{table}[t]
\footnotesize
\centering
\begin{tabular}{lrrrr} 
\toprule
Sequences  & $\text{S}_{\text{succ}} \uparrow$ & $\text{E}_{\text{mpjpe}}\downarrow$ & $\text{E}_{\text{mpjpe-g}} \downarrow$   & $\text{E}_{\text{acc}} \downarrow$     \\ \midrule
Belly Dance-1  &  ${0\%} \rightarrow \textbf{100\%}$ &  ${183.3} \rightarrow \textbf{36.9}$ &  ${347.1} \rightarrow \textbf{55.0}$ &  ${7.6} \rightarrow {7.6}$ \\  
Parkour-1  & ${0\%} \rightarrow \textbf{100\%}$ &  ${175.6} \rightarrow \textbf{87.1}$ &  ${324.2} \rightarrow \textbf{146.4}$ &  ${24.8} \rightarrow \textbf{15.0}$ \\  
Karate-1  &  ${100\%} \rightarrow {100\%}$ &  ${35.9} \rightarrow \textbf{30.1}$ &  ${45.8} \rightarrow \textbf{39.9}$ &  ${7.1} \rightarrow {7.1}$ \\  
Crawl-1  &  ${100\%} \rightarrow {100\%}$ &  ${66.4} \rightarrow \textbf{40.6}$ &  ${307.6} \rightarrow \textbf{67.2}$ &  $\textbf{3.8} \rightarrow {4.3}$ \\  
Cartwheel-1  & ${0\%} \rightarrow \textbf{100\%}$ &  ${160.9} \rightarrow \textbf{37.4}$ &  ${284.8} \rightarrow \textbf{66.2}$ &  ${8.1} \rightarrow \textbf{4.9}$ \\  
\hline
Dance-200  & ${57.0\%} \rightarrow \textbf{72.0\%}$ &  ${84.1} \rightarrow \textbf{58.0}$ &  ${146.7} \rightarrow \textbf{98.7}$ &  ${13.3} \rightarrow {13.3}$ \\  
Tennis-60 &  ${96.7\%} \rightarrow \textbf{100\%}$ &  ${27.8} \rightarrow \textbf{21.8}$ &  ${40.5} \rightarrow \textbf{30.7}$ &  ${4.2} \rightarrow \textbf{4.1}$ \\
Crawl-37  &  ${94.6\%} \rightarrow \textbf{94.6\%}$ &  ${62.0} \rightarrow \textbf{40.6}$ &  ${163.6} \rightarrow \textbf{87.9}$ &  $\textbf{5.6} \rightarrow {9.1}$ \\
Cartwheel-4  &  ${25\%} \rightarrow \textbf{75\%}$ &  ${219.6} \rightarrow \textbf{89.4}$ &  ${393.6} \rightarrow \textbf{166.1}$ &  ${19.4} \rightarrow \textbf{12.5}$ \\
Kick-302 &  ${98.3\%} \rightarrow {98.3\%}$ &  ${45.5} \rightarrow \textbf{38.8}$ &  ${75.4} \rightarrow \textbf{62.4}$ &  $\textbf{7.5} \rightarrow {9.1}$ \\
\bottomrule 
\end{tabular}
\vspace{5mm}
\caption{Performance improvement of the humanoid after finetuning for different sequences. Here, the suffix indicates the number of motion sequences used for imitation.}
\label{transform2act:table:design}
\end{table}

\begin{table}[t]
\footnotesize
\centering
\begin{tabular}[b]{lrrrr} 
\toprule
\multicolumn{5}{c}{AMASS Test Set}\\ \midrule
Training Sequences  & $\text{S}_{\text{succ}} \uparrow$ & $\text{E}_{\text{mpjpe}}\downarrow$ & $\text{E}_{\text{mpjpe-g}} \downarrow$   & $\text{E}_{\text{acc}} \downarrow$     \\ \midrule
Dance-200  &  \textbf{92.8\%} &  {40.4} &  {70.4} &  {12.1}\\  
Tennis-60  &  {89.9\%} &  {33.9} &  {55.7} &  {11.9}\\  
Crawl-37  &  {86.3\%} &  {36.7} &  {54.4} &  {13.0}\\  
Cartwheel-4 &  {89.9\%} &  {36.8} &  {64.9} &  {11.8}\\  
Kick-203  &  {90.6\%} &  {52.9} &  {92.9} &  {14.6}\\  
\midrule
RFC & {91.4\%} &  \textbf{35.3} &  \textbf{60.1} &  \textbf{10.5} \\
\bottomrule 
\end{tabular}
\vspace{3mm}
\caption{Performance of the humanoid finetuned with different training sequences on the AMASS test split.}
\label{transform2act:table:transfer}
\end{table}

\paragraph{Single-Sequence Finetuning.}
The top half of Table \ref{transform2act:table:design} shows the results of finding the humanoid design that performs a single sequence the best. We can see that for each individual sequence, optimizing the design parameters can significantly improve the motion imitation performance, and often enables the humanoid to successfully imitate a sequence without falling. This demonstrates our approach's ability to finetune the humanoid design for a single sequence.

\paragraph{Category-Level Finetuning.}
The bottom half of Table \ref{transform2act:table:design} shows the category-level motion imitation results. Similar to the single-sequence case, our approach is able to find a suitable humanoid design for a whole category of motion sequences, demonstrating its ability to generalize to a suite of diverse motions that share similar traits.

\paragraph{Test Set Transfer.}
To further evaluate the generalization of the finetuned design and test its ability to perform general motions, we directly use the humanoid design obtained by category-level finetuning for the test set of AMASS, where the design is kept fixed and only the humanoid control policy is trained. As shown in Table \ref{transform2act:table:transfer}, the humanoid designs finetuned with different training sequences all maintain a similar level of motion imitation ability as a ``vanilla humanoid'' using an RFC-based control policy. This shows that, unlike RFC, our approach can design humanoids with better motion imitation ability without sacrificing physical plausibility.

\paragraph{Discussion.}
Although our approach can find a humanoid design with better motion imitation ability, the found design does not necessarily correspond to the actual physique of the actor who performed the training motion. The main reasons are twofold. First, there are redundant degrees of freedom (DoFs) and local minima in the design optimization problem since both the humanoid design and the control policy can change during optimization, and the reward function only measures the motion imitation performance. So two different designs can have similar motion imitation performance as long as their control policies can accommodate the designs. Second, the underlying simulation model of the humanoid is oversimplified and cannot simulate many aspects of human dynamics. For example, the physics simulator cannot simulate muscles, tendons, and soft tissues. Therefore, many agile human motions cannot be modeled using these aspects, and the design optimization may change other aspects of the humanoid such as body proportions and motor strengths to force the humanoid to imitate the motion.

In order to produce more biologically-plausible humanoid designs, one possible way is to develop more expressive and biologically-sound simulation models of the humanoid. However, these models are often computationally-expensive, which may not be compatible with RL and design optimization. Another direction is to add constraints to the design search space using visual data. For example, we can recover the 3D shape of the human actor using human mesh recovery methods (e.g., \cite{iqbal2021kama}), and we can then constrain the humanoid design to have roughly the same shape as the actor. The added constraints alleviate the problem of redundant DoFs mentioned previously. Finally, we can also leverage biomechanical models to other aspects of human dynamics besides pose. For example, we can use the 3CC model~\cite{xia2008theoretical} to model muscle fatigue which could further constrain the humanoid design to be biologically-plausible.

\section{Conclusion}
In this paper, we proposed a new transform-and-control paradigm that formulates design optimization as conditional policy learning with policy gradients. Compared to prior ES-based methods that randomly mutate designs, our approach is more sample-efficient as it leverages a parameterized policy to select designs based on past experience and also allows experience sharing across different designs. Experiments show that our approach outperforms prior methods in both convergence speed and final performance by an order of magnitude. Our approach can also automatically discover plausible designs similar to giraffes, squids, and spiders. We further showed that our approach can be used to finetune a humanoid design to improve its motion imitation ability. For future work, we are interested in leveraging RL exploration methods to further improve the sample efficiency of our approach.

\part{Perception of Human Behavior}
\label{part:perception}
\chapter{Simulation-Based First-Person Human Pose Estimation}
\label{chap:egopose18}

\section{Introduction}
Our task is to use a single head-mounted wearable camera to estimate the 3D body pose sequence of the camera wearer, such that the estimated motion sequence obeys basic rules of physics (e.g., joint limits are observed, feet contact the ground, motion conserves momentum). Employing a single wearable camera to estimate the pose of the camera wearer is useful for many applications. In medical monitoring, the inferred pose sequence can help doctors diagnose patients' conditions during motor rehabilitation or general activity monitoring. For athletes, egocentric pose estimation provides motion feedback without instrumenting the environment with cameras, which may be impractical for sports like marathon running or cross-country skiing. In virtual reality games, the headset wearer's poses can be reproduced in the virtual environment to create a better multi-player gaming experience without additional sensors. In many applications, accurate and \emph{physically-valid} pose sequences are desired.

However, estimating physically-valid 3D body poses from egocentric videos is challenging. First, egocentric cameras typically face forward and have almost no view of the camera wearer's body. The task of estimating 3D pose is under-constrained as the video only encodes information about the position and orientation of the camera's viewpoint. Second, with a single wearable camera, we have no access to the forces being applied to the body, such as joint torques or ground contact forces. Without observations of these forces, it is very difficult to learn the relationship between camera-based motion features and body pose using physics simulation in a data-driven way. Most traditional approaches to human pose estimation in computer vision side-step the issue of physics completely by focusing primarily on the kinematics of human motion. Unfortunately, this can sometimes result in awkward pose estimates that allow the body to float in the air or joints to flex beyond what is physically possible, which makes it difficult to use for motion analysis applications. New technical approaches are needed to tackle these challenges of generating physically-valid 3D body poses from egocentric video.

In light of these challenges, we take a radical departure from the kinematics-based representation traditionally used in computer vision towards a control-based representation of humanoid motion commonly used in robotics. In the traditional kinematics-based representation used for pose estimation from videos, a human pose sequence is typically modeled as a sequence of poses $\{ p_1,\ldots, p_T \}$. It is common to use a temporal sequence model (e.g., hidden Markov model, linear chain CRF, recurrent neural network) where the estimate of each pose $p_t$ is conditioned on image evidence $I_t$ and a prior pose $p_{t-1}$ (or some sufficient statistics of the past, e.g., hidden layer in the case of RNN). While it is often sufficient to reason only about the kinematics of the pose sequence for pose estimation, when one would like to evaluate the physical validity of the sequence, it becomes necessary to understand the \textit{control input} that has generated each pose transition. In other words, we must make explicit the torque (control input) that is applied to every joint to move a person from pose $p_t$ to $p_{t+1}$. Under a control-based method, a human pose sequence needs to be described by a sequence of states and actions (control inputs) $\{ s_1,a_1,s_2,a_2, \ldots, s_T\}$ where state $s_t$ contains both the pose $p_t$ and velocity $v_t$ of the human. A control-based model explicitly takes into account the control input sequence and learns a control policy $\pi(a|s)$, that maps states to actions for optimal control. Making explicit the control input is essential for generating a state sequence based on the laws of physics.

\begin{figure}[t]
	\centering
	\includegraphics[width=\textwidth]{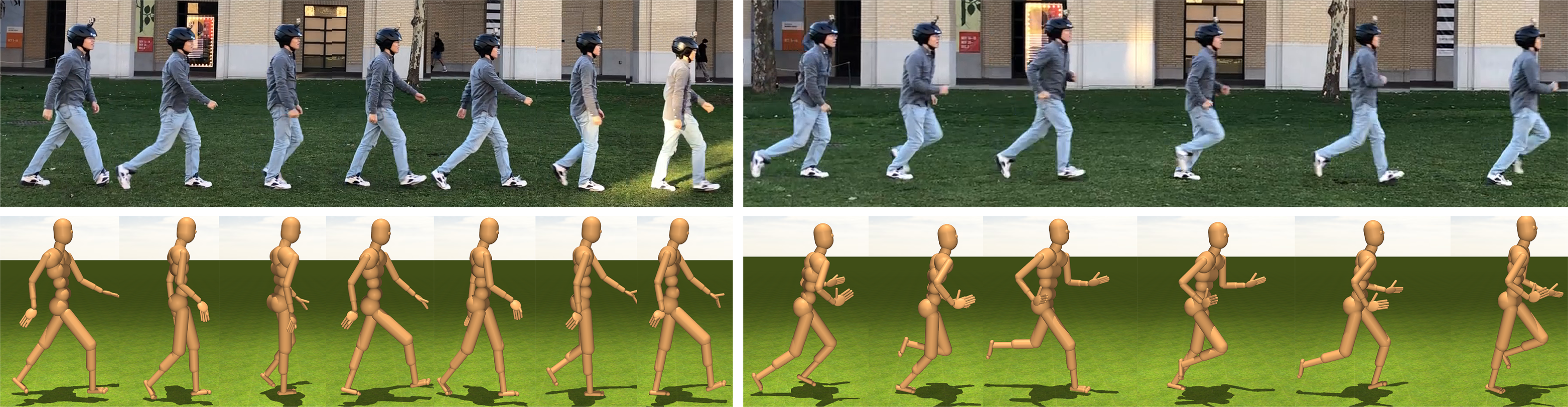}
	\caption{Our 3D ego-pose estimation results using egocentric videos.}
	\label{egopose18:fig:teaser}
\end{figure}

The use of a control-based method requires access to interaction with real-world physics or in our scenario, a physics simulator. The use of a physics simulator for learning a control policy provides two major advantages. First, the physical properties of the virtual humanoid, such as joint actuation limits and range limits, used in the simulator serve as a gating mechanism to constrain the learning process to generate actions that are humanly possible. Second, the physical constraints of the simulation environment ensure that only physically-valid pose sequences are estimated such that the feet will not penetrate the ground or slip during contact. Within the confines of the physic simulator, the goal of control policy learning is to learn a virtual humanoid policy that maps the current state (pose and velocity, optionally egocentric video) to an action (joint torques). Formally, we frame first-person pose estimation as a sequential decision process using a Markov decision process (MDP). The state of the MDP is the state of our humanoid model defined in terms of joint positions, joint velocities and the observed first-person POV video. The action is the joint torques exerted by joint actuators. The transition probability is the humanoid dynamics provided by the physics simulation environment. In our imitation learning framework, the reward function is based on the similarity between the generated pose and its corresponding training pose. Based on this MDP, we perform imitation learning (IL) to obtain a humanoid control policy that is conditioned on the egocentric video. Once an optimal policy is learned, it can be used to generate a physically-grounded pose sequence given an egocentric video sequence.

The use of imitation learning to estimate pose from egocentric video requires a set of demonstrated `expert behaviors' which in our scenario would be a set of egocentric videos labeled with 3D joint positions and joint torques. However, it is not easy to obtain such data without instrumenting the body with other sensors such as an exoskeleton \cite{hwang2015method}. Instead, we propose a two-step imitation learning process to learn a video-conditioned humanoid control policy for ego-pose estimation. In the first step, following Merel et al. \cite{merel2017learning}, we learn a set of humanoid control policies imitating different human behaviors in motion capture data to generate virtual humanoid pose sequences, from which we can render first-person POV videos. In the second step, imitation learning is again used to learn a video-conditioned policy which maps video features to optimal joint torques, to yield a physically valid 3D pose sequence. In this way, we are able to learn a video-conditioned control policy without the need for direct measurements of joint torques from the camera wearer.

We note that the two-stage imitation learning process described thus far relies only on simulations in a virtual environment and overlooks the problem of the domain gap between the virtual and real data. Thus, we further propose to fine-tune the video-conditioned policy at test time using real data to perform domain adaptation. We use regression to estimate the best initial state that maximizes the policy's expected return and fine-tune the policy with policy gradient methods. We evaluate our approach on both virtual world data and real-world data and show that our pose estimation technique can generalize well to real first-person POV video data despite being trained on virtual data.

In this work, we aim to show that a decision-theoretic approach to human motion estimation offers a powerful representation that can naturally map the visual input of the human visual system (i.e., egocentric video) to body dynamics while taking into account the role of physics. Towards this aim, we focus on estimating the pose of human locomotion using a head-mounted camera. To the best of our knowledge, this is the first work to utilize physically grounded imitation learning to generate ego-pose estimates using a wearable camera.

\section{Related work}
\textbf{Third-person pose estimation.}
Pose estimation from third-person images or videos has been studied for decades \cite{sarafianos20163d,liu2015survey}. Existing work leverages the fact that full human body can be seen by the third-person camera. In contrast, we consider the case where the person is entirely out of sight. Thus, existing pose estimation methods are not immediately applicable to our problem setting. Some of these methods use regression to map from images to pose parameters \cite{agarwal20043d,shakhnarovich2003fast,toshev2014deeppose,sminchisescu2007bm3e}, including the recent work DeepPose \cite{toshev2014deeppose} that uses convolutional neural networks. It is tempting to directly apply regression-based methods to egocentric pose estimation. However, such approaches are inadequate since the egocentric images only contain information about the position and orientation of the camera. Even if the method can perfectly reconstruct the motion of the camera, the underlying human poses are still under-constrained. Without prior information as regularization, unnatural human poses will emerge. This  motivates us to physically model and simulate the human, and use the human dynamics as a natural regularization.

\paragraph{Egocentric pose estimation.} 
A limited amount of research has looked into the problem of inferring human poses from egocentric images or videos. Most of existing methods still assume the estimated human body or part of the body is visible \cite{li2013model,li2013pixel,ren2010figure,arikan2003motion,rogez2015first}. The ``inside-out'' mocap approach of \cite{shiratori2011motion} gets rid of the visibility assumption and infer the 3D locations of 16 or more body-mounted cameras via structure from motion. Recently, \cite{jiang2017seeing} show that it is possible to estimate human pose using a single wearable camera. They construct a motion graph from the training data and recover the pose sequence by solving for the optimal pose path. In contrast, we explicitly model and simulate human dynamics, and learns a video-conditioned control policy.

\paragraph{Adversarial imitation learning.}
Our problem suits a specific setting of imitation learning in which the learner only has access to samples of expert trajectories and is not allowed to query the expert during training. Behavior cloning \cite{pomerleau1991efficient}, which treats the problem as supervised learning and directly learns the mapping from state to action for each timestep, suffers from compounding error caused by covariate shift \cite{ross2010efficient,ross2011reduction}. Another approach, inverse reinforcement learning (IRL) \cite{russell1998learning,ng2000algorithms}, learns a cost function by prioritizing expert trajectories over others and thus avoids the compounding error problem common in methods that fit single-timestep decisions. However, IRL algorithms are very expensive to run because they need to solve a reinforcement learning problem in the inner loop. Generative adversarial imitation learning (GAIL; \cite{ho2016generative}) extends the GAN framework to solve this problem. A policy acts as a generator to produce sample trajectories and a discriminator is used to distinguish between expert trajectories and generated ones. It uses reinforcement learning algorithms to optimize the policy and the policy is rewarded for fooling the discriminator. The key benefit of GAIL is that no explicit hand-designed metric is needed for measuring the similarity between imitation and demonstration data.

\paragraph{Learning human behaviors.} 
There have been two types of approaches for modeling human movements: one is purely kinematic, and the other is physical control-based. For the former, a good amount of research models the kinematic trajectories of humans from motion capture data in the absence of physics \cite{taylor2007modeling,sussillo2009generating,holden2017phase}. The latter has long been studied in the graphics community and is more relevant to our scenario. Many of these methods are model-based and require significant domain expertise. With rapid development in deep reinforcement learning (Deep RL), exciting recent work has used Deep RL for the locomotion of 2D creatures \cite{peng2016terrain} and 3D humanoid \cite{peng2017deeploco}. More recently, adversarial imitation learning from motion capture data \cite{merel2017learning} has shown beautiful results. They use context variables to learn a single policy for different behaviors such as walking and running. As a follow-up work, \cite{wang2017robust} propose to learn the context variables by a variational autoencoder (VAE).

\begin{figure}[t]
	\centering
	\includegraphics[width=\textwidth]{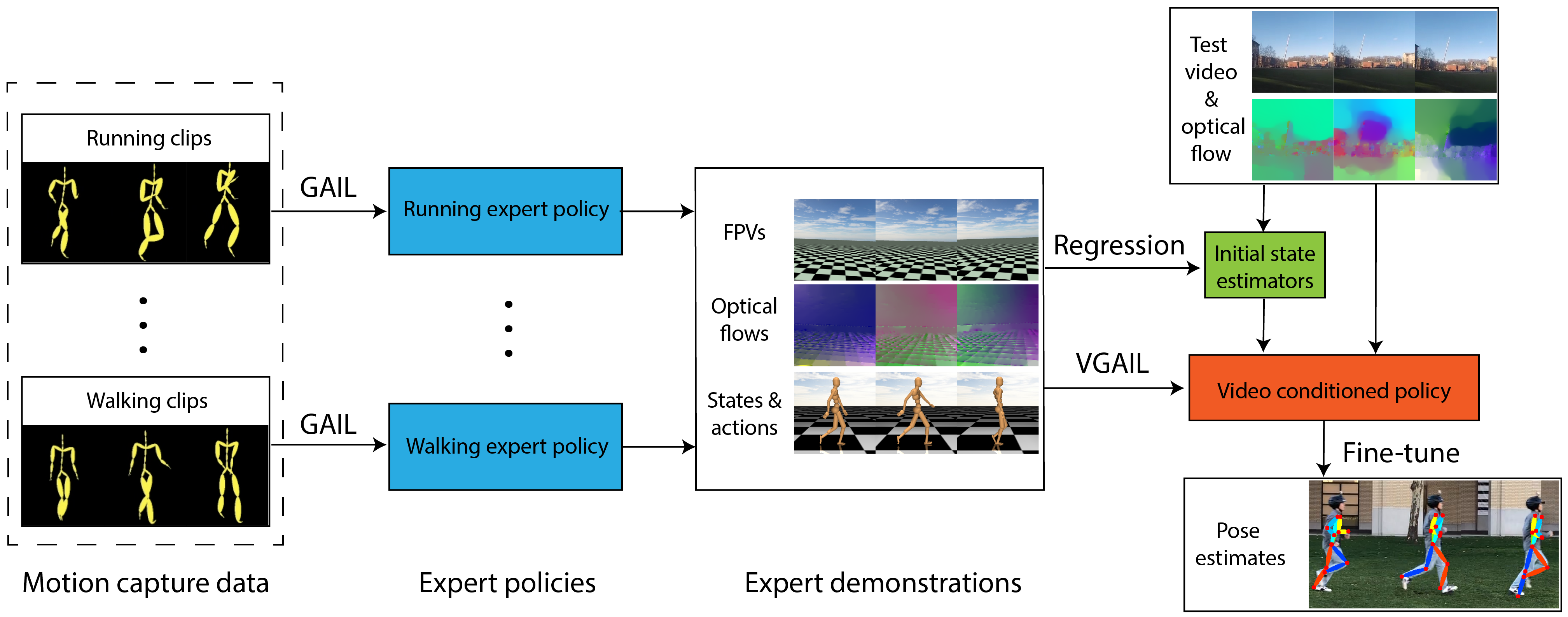}
	\caption{Overview of our proposed pose estimation pipeline.}
	\label{egopose18:fig:overview}
\end{figure}

\section{Approach}
Towards our goal of estimating a physically valid 3D body pose sequence of a person using video acquired with a head-mounted camera, we propose a two-step imitation learning technique that leverages motion capture data, a humanoid model and a physics simulator. As shown in Figure~\ref{egopose18:fig:overview}, in our first phase, our proposed method starts with learning an initial set of $C$ expert policies $\{\hat{\pi}_c\}_{c=1}^C$, each of which represents a specific type of human behavior, e.g.,  walking or running. In the second phase, virtual demonstrations of the humanoid are generated from each of the $C$ policies, including state and action sequences of the humanoid, along with virtual egocentric video sequences  captured by the humanoid's head-mounted camera. With this virtual expert demonstration data, we again use imitation learning to learn a video-conditioned policy that can map egocentric video features directly to joint torques which generate the pose sequence.

\begin{figure}[t]
	\centering
	\includegraphics[width=\textwidth]{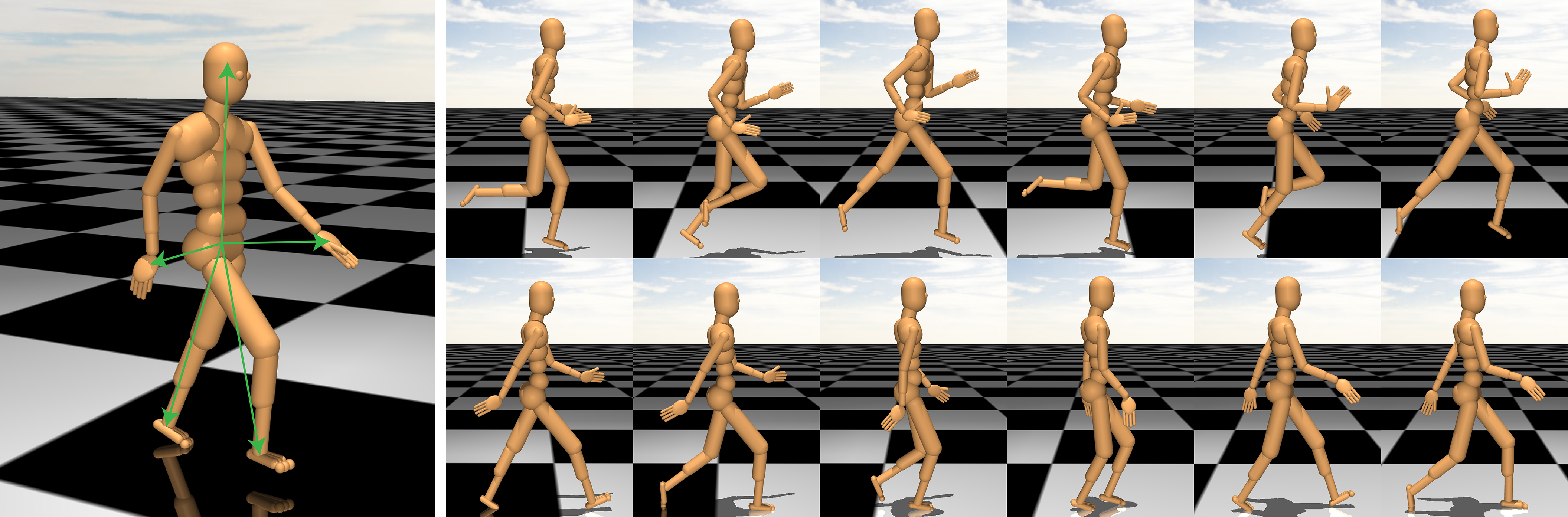}
	\caption{\textbf{Left:} The humanoid model. Green arrows illustrate 3D vectors from the root to the feet, head and hands provided to the policy and discriminator. \textbf{Right:} Selected key frames of running and walking clips in motion capture data animated using the humanoid model.}
	\label{egopose18:fig:humanoid}
\end{figure}

\paragraph{Humanoid Model.} By design, our underlying control policy assumes a pre-defined humanoid model which can be actuated in a virtual environment (see Figure~\ref{egopose18:fig:humanoid}). The humanoid model we use consists of 31 rigid bodies, 56 hinge joints and 63 degrees of freedom (DoFs). All the hinge joints can be actuated and have torque limits and range limits. The joints also have physical properties such as stiffness and damping. It is important to note here that the careful design of the humanoid is critical for solving the under-constrained problem of pose estimation from egocentric videos because the model must be similar enough to the human body for the physics simulation to match real human motion.

\paragraph{Humanoid Control Policy.} It is common to use a Markov decision process (MDP) to model the effect of control on the dynamics of a system. In our scenario, given the humanoid model, we can formulate human(oid) motion as the output of an MDP, where any given 3D body pose sequence is assumed to be generated by an optimal policy derived from the MDP. The MDP is defined by a tuple $\mathcal{M} = \langle S,A,P,R, \gamma \rangle $, where $S$ is the state space, $A$ is the action (or control) space, $T$ is the state transition dynamics, $\gamma$ is the discount factor and $R$ is the reward or cost function typically defined over the state and action space. In our formulation, the state $s$ represents the state of the humanoid and optionally the egocentric video (second step of our learning task). The state $z$ of the humanoid consists of the pose $p$ and velocity $v$. The pose $p$ contains the position and orientation of the root, as well as the 56 joint angles. The velocity $v$ consists of the linear and angular velocities of the root as well as the joint velocities. The action is composed of the joint torques of all actuated hinge joints. The dynamics of the humanoid is denoted by $P(s_{t+1}|s_t, a_t)$ (i.e., how the control or action $a$ affects the pose transition) which is determined by the simulation environment (we use the MuJoCo simulator \cite{todorov2012mujoco}).

The solution of a given MDP is an optimal policy $\pi$ that maximizes the expected return. We use $\pi(a|s)$ to denote the policy, which outputs the probability of choosing action $a \in A$ when the agent is in state $s \in S$. We use a multivariate normal distribution to model the policy $\pi$ where the mean and log standard deviation are parameterized by neural networks. In our final task, we want to learn a video-conditioned policy that maps humanoid state $z$ and egocentric video $V_{1:T}$ to joint torques, to estimate a physically valid 3D pose sequence. In what follows, we describe a two-step imitation learning method for learning this video-conditioned policy.

\subsection{Stage 1: Imitation Learning for Data Generation}
\label{egopose18:sec:stage1}
Instead of directly generating virtual egocentric POV videos using motion capture data, we propose to first learn a set of expert control policies imitating the human behaviors from the motion capture data and then use the expert policy for egocentric video generation. This provides two advantages. First, motion capture data is often noisy and our humanoid model cannot perfectly match the real human motion sequence. In contrast, an expert policy successfully learned from motion capture data can generate pose sequences that are noise-free and realizable by our humanoid model. Second, the imitation learning procedure solves the inverse dynamics problem (i.e., the control policy $\pi(a|s)$ is learned from observed state transition dynamics $p(s'|s)$) and the policy provides the joint torques for generating novel pose sequences and egocentric videos, which we show later is needed for learning the video-conditioned policy.

Our method first learns a set of expert policies $\{\hat{\pi}_c\}_{c=1}^C$ from motion capture data using generative adversarial imitation learning (GAIL) following Merel et al. \cite{merel2017learning}. Each of the expert policies represents a specific type of human behavior. In this stage, the state $s$ of the MDP is just the state of the humanoid $z$ as no video input is involved. Similar to GAN, the loss function of GAIL takes the form:
\begin{align}
	\label{egopose18:eq:GAIL}
	\ell(\theta, \phi) = \mathbb{E}_{z\sim \pi_\theta}[\log\left(1-D_\phi(z)\right)] + \mathbb{E}_{\hat{z}\sim \hat{\pi}}[\log \left(D_\phi\left(\hat{z}\right)\right)]\,,
\end{align}
where $\pi_\theta$ is the policy we want to learn and $\hat{\pi}$ is the expert policy implicitly represented by expert demonstrations $\left\{\hat{z}_i\right\}_{i=1}^N$. At each iteration, the policy acts as a generator to collect samples $\{z_i\}_{i=1}^M$ and rewards $\{r_i\}_{i=1}^M$. Using these samples and rewards, policy gradient methods (e.g., TRPO \cite{schulman2015trust}, PPO \cite{schulman2017proximal}) are employed to update the policy and thus decrease the loss $\ell$ w.r.t $\theta$. Once the generator update is done, we also need to update the discriminator to distinguish between generated samples and expert demonstrations. As argued by Merel et al. \cite{merel2017learning}, using the full state $z$ of the humanoid performs poorly because our simplified humanoid model cannot perfectly match the real human. Thus, we only use a partial state representation  of the humanoid as state input $z$ to both the policy and discriminator. Our partial state includes the root's linear and rotational velocities axis-aligned to the root orientation frame, upward direction of the root, as well as 3D displacement vectors from the root to each foot, each hand, and head, also in the root coordinate frame (see Figure~\ref{egopose18:fig:humanoid}~(Left)). We also added the orientation of the head in the root coordinate to the partial state for GAIL to learn natural head motions. After we train expert policies $\{\hat{\pi}_c\}_{c=1}^C$ using GAIL, we can generate a large amount of expert trajectories $\left\{\hat{\tau}_i\right\}_{i=1}^N$ from different human behaviors, where each expert trajectory $\hat{\tau}_i$ contains a state sequence $\hat{z}_{1:\hat{T}_i}^i$, an action sequence $\hat{a}_{1:\hat{T}_i}^i$ and a virtual egocentric video sequence $\hat{V}_{1:\hat{T}_i}^i$.

\begin{algorithm}[tb]
	\caption{Video-conditioned generative adversarial imitation learning}
	\label{alg:vgail}
	\begin{algorithmic}
		\State \textbf{Input:} Set of expert demonstrations $\left\{\hat{\tau}_i\right\}_{i=1}^N$
		\State \textbf{Output:} Learned policy $\pi_\theta\left(a | z, V_{1:T}\right)$
		\State Randomly Initialize policy $\pi_\theta$ and discriminator $D_\phi$
		\Repeat
		\State // Perform generator updates
		\For{$k$ in $1\ldots N$}
		\State Sample an expert trajectory $\hat{\tau}_k$ from $\left\{\hat{\tau}_i\right\}_{i=1}^N$
		\State Conditioned on $\hat{V}^k_{1:\hat{T}_k}$, execute policy $\pi_\theta$ to collect learner's trajectory $\tau_k$
		\State Compute rewards $r_t^k = -\log \left(1-D_\phi(z_t^k,  \hat{V}_{1:\hat{T}_k}^k)\right) -\alpha||p_t^k - \hat{p}_t^k ||_2 + \beta$
		\EndFor
		\State Update $\theta$ by policy gradient methods (e.g. TRPO, PPO) using rewards $\{r_t^k\}$
		\State // Perform discriminator updates
		\For{$j$ in $1\ldots J$}
		\State $\ell(\phi) = \frac{1}{N}\sum_{k=1}^N\left[ \frac{1}{T_k} \sum_{t=1}^{T_k}\log \left(1-D_\phi(z_t^k , \hat{V}_{1:\hat{T}_k}^k)\right) +  \frac{1}{\hat{T}_k} \sum_{t=1}^{\hat{T}_k}\log \left(D_\phi(\hat{z}_t^k, \hat{V}_{1:\hat{T}_k}^k)\right)\right]$
		\State Update $\phi$ by a gradient method w.r.t. $\ell(\phi)$
		\EndFor
		\Until{Max iteration reached}
	\end{algorithmic}
\end{algorithm}

\subsection{Stage 2: Imitation Learning for Ego-Pose Estimation}
\label{egopose18:sec:stage2}

Using the expert trajectories $\left\{\hat{\tau}_i\right\}_{i=1}^N$ generated in the first stage, we can now learn a video-conditioned policy $\pi_\theta\left(a | z, V_{1:T}\right)$ with our video-conditioned GAIL (VGAIL) algorithm outlined in Algorithm~\ref{alg:vgail}. As we only care about the motion of the camera, we extract optical flow from egocentric videos and overload the notation to use optical flow as the video motion features $V_{1:T}$. In this stage, the state $s$ of the MDP is the combination of the state $z$ of the humanoid and the egocentric optical flow $V_{1:T}$. The VGAIL loss becomes
\begin{align}
	\label{egopose18:eq:VGAIL}
	\ell(\theta, \phi) = \mathbb{E}_{z\sim \pi_\theta}\left[\log\left(1-D_\phi(z, \hat{V}_{1:T})\right)\right] + \mathbb{E}_{\hat{z}\sim \hat{\pi}}\left[\log \left(D_\phi\left(\hat{z}, \hat{V}_{1:T}\right)\right)\right]\,.
\end{align}
We use $\hat{V}_{1:T}$ in the above equation since the policy is trained on egocentric videos in expert demonstrations. In GAIL, expert demonstrations are a set of expert states $\{\hat{z}_i\}$ of the humanoid and their temporal correlation is dismissed. In VGAIL, expert demonstrations become a set of expert trajectories $\{\hat{\tau}_i\}$ with sampled expert trajectory $\hat{\tau}_k$ containing a state sequence $\hat{s}^k_{1:\hat{T}_k}$ (poses $\hat{p}^k_{1:\hat{T}_k}$ and velocities $\hat{v}^k_{1:\hat{T}_k}$), an action sequence $\hat{a}^k_{1:\hat{T}_k}$ and a video sequence $\hat{V}^k_{1:\hat{T}_k}$. This provides two benefits. First, as we want our policy-generated pose sequence $p_{1:T_k}^k$ to match with the expert pose sequence, we use the expert pose sequence $\hat{p}_{1:\hat{T}_k}^k$ to augment the reward with an additional pose alignment term $-||p_t^k - \hat{p}_t^k ||_2$, which uses L2-norm to penalize pose difference. Second, we can use the action sequence $\hat{a}_{1:\hat{T}_k}^k$ to pre-train the policy with behavior cloning \cite{pomerleau1991efficient}, which accelerates the training significantly. The reward for VGAIL is
\begin{align}
	\label{egopose18:eq:reward}
	r_t^k = -\log \left(1-D_\phi(z_t^k , \hat{V}_{1:T_k}^k)\right) -\alpha||p_t^k - \hat{p}_t^k ||_2 + \beta\,,
\end{align}
where $\alpha$ is a weighting coefficient and $\beta$ is a `living' bonus to encourage longer episode (the simulation episode will end if the humanoid falls down). $\alpha$ and $\beta$ are set to 3.0 and 5.0 respectively in our implementation.

Again, we use the partial state of the humanoid discussed in Sec.~\ref{egopose18:sec:stage1} as humanoid state $z$ to both the policy and discriminator. As shown in Figure~\ref{egopose18:fig:net_ft}(Bottom), for both the policy and discriminator networks, we use a CNN to extract visual motion features and pass them to a bidirectional LSTM to process temporal information, and a multilayer perceptron (MLP) following the LSTM outputs the action distribution (policy) or the classification probability (discriminator). Once the video-conditioned policy $\pi_\theta\left(a | z, V_{1:T}\right)$ is learned, given an egocentric video with its optical flow $V_{1:T}$ and the initial state of the humanoid, we execute the policy $\pi_\theta$ inside the physics simulator and always choose mean actions to generate the corresponding pose sequence of the video.

\subsection{Initial State Estimation and Domain Adaptation}

The straightforward use of the video-conditioned policy on real egocentric video data will lead to failure for two reasons. First, without a mechanism for reliably estimating the initial state $z_1$ of the humanoid, the actions generated by the policy cause the humanoid to fall down in the physics simulator because it cannot reconcile extreme offset between the phase of the body motion and the video motion. Second, the visual features learned from the optical flow in the virtual world (checkered floor and skybox) are usually very different from the environment in real egocentric videos, and therefore the policy is not able to accurately interpret the optical flow. We propose two important techniques to enable pose estimation with real-world video data.

\begin{figure}[t]
	\centering
	\includegraphics[width=\textwidth]{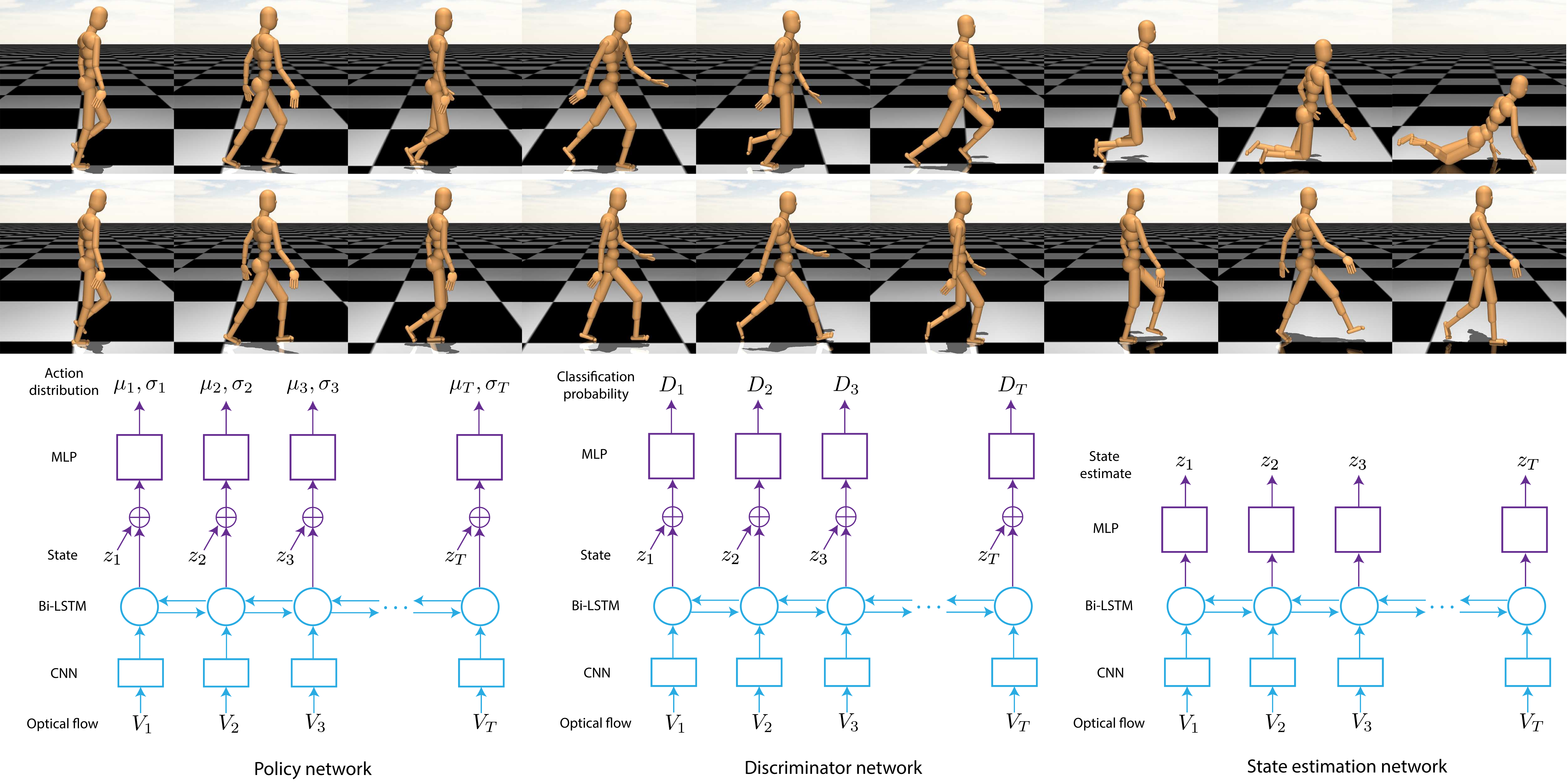}
	\caption{\textbf{Top:} Humanoid falling down due to the error in initial state estimate. \textbf{Mid:} After fine-tuning for 20 iterations, the policy can generate correct walking estimates. \textbf{Bottom:} Network architecture for the policy, discriminator, and state estimator. All three networks employ the same architecture for processing the optical flow: a CNN with three convolutional layers of kernel size 4 and stride 4 is used and the size of its hidden channels are (32, 32, 8), and a bidirectional LSTM is used to distill temporal information from the CNN features. For the policy and discriminator, we concatenate the LSTM output with the humanoid state $z$ and pass it to an MLP with hidden size (300, 300, 200, 100), which outputs the action distribution (policy) or classification probability (discriminator). For the state estimation network, the LSTM output is passed to an MLP with hidden size (300, 300, 200) which outputs the state estimate.}
	\label{egopose18:fig:net_ft}
\end{figure}

\paragraph{Initial state estimation.}
\label{egopose18:sec:init}
We propose to learn a set of state estimators $\{f_c\}_{c=1}^C$ where $f_c$ maps an optical flow $V_{1:T}$ to its corresponding state sequence $z_{1:T}$ and is learned using expert trajectories generated by expert policy $\hat{\pi}_c$. The state at time $t$ can be extracted by $f_c(V_{1:T})_t$. $f_c$ is implemented as the state estimation network in Figure~\ref{egopose18:fig:net_ft}(Bottom). Visual motion features from the optical flow are extracted by a CNN and passed to a bidirectional LSTM before going into a multilayer perceptron (MLP) which makes the state predictions. We use the mean square error as loss which takes the form $l_c(\psi) = \frac{1}{T}\sum_{t=1}^T||f_c(V_{1:T})_t - z_t||^2$, where $\psi$ is the parameters of $f_c$. We can get an optimal $f_c$ by a SGD-based method. The state estimators are used for initial state estimation in the policy fine-tuning step described below.

\paragraph{Policy fine-tuning.}
\label{egopose18:sec:finetune}
Our imitation learning framework allows us to fine-tune the policy on test data (of course without requiring any ground truth pose data). This fine-tuning step is essentially a reinforcement learning step that adapts the policy network to the video input $V_{1:T}$ while maximizing the reward for matching the training data distribution. In order to utilize a policy gradient method to improve and adapt the policy, we need a reward function and an initial state estimate. We define a reward function that will help to ensure that the fine-tuned policy generates pose sequences that are similar to the training data. Given the test video's optical flow $V_{1:T}$, the fine-tuning reward is defined as
\begin{align}
	\label{egopose18:eq:finetune}
	r_t = -\log \left(1-D_\phi(z_t, V_{1:T})\right) + \xi\,,
\end{align}
where $\xi$ is a `living' bonus (set to $0.5$ in our implementation). The initial state estimate can be obtained using the state estimators described above by solving the following optimization problem:
\begin{equation}
	\label{egopose18:eq:model_select}
	c^\ast, b^\ast = \argmax_{c=1...C,\,b=1...10} \quad \mathbb{E}_{z_1 = f_c(V_{1:T})_b,\,a_t\sim\pi_\theta} \left[\sum_{t=1}^T\gamma^t r_t\right]\,,
\end{equation}
where $c^*$ is the index of the optimal estimator and $b^*$ is the optimal start frame offset. This step enables our method to find the best initial state estimator $f_{c^\ast}$ and the best start frame $b^\ast$ by maximizing the expected return, where the expected return can be estimated by sampling trajectories from the video-conditioned policy. We then perform fine-tuning by sampling trajectories of the policy starting from the initial state $f_{c^\ast}(V_{1:T})_{b^\ast}$ and computing rewards using Equation~\ref{egopose18:eq:finetune}. We employ policy gradient methods (e.g., PPO \cite{schulman2017proximal}) to update the policy using the sampled trajectories and rewards.

\section{Experimental Setup}

To evaluate our proposed method's ability to estimate both accurate and physically valid pose sequences from an egocentric video, we tested our method on two datasets. The first one is a synthetic dataset using the same expert policies we learned in Sec.~\ref{egopose18:sec:stage1}. The synthetic dataset will allow us to evaluate the accuracy of the 3D pose estimates and control actions since we have access to the ground truth through the simulator. The second dataset is composed of real-world first-person videos of different people walking and running. Our aim is to show the robustness of our technique through domain adaptation and initial pose estimation for real-world videos. Evaluations, however, are based on noisy ground truth estimates using a 2D projection of joint positions using a second static camera.

\paragraph{Baselines.} We compare our methods against two baselines:
\begin{itemize}[leftmargin=*]
	\item \textbf{Pose regression:} direct regression from video motion features to poses.
	Similar to the initial state estimation in Sec.~\ref{egopose18:sec:init}, pose regression learns a mapping from egocentric optical flow $V_{1:T}$ to its corresponding pose sequence $p_{1:T}$. The regression network is the same as the state estimation network in Figure~\ref{egopose18:fig:net_ft}(Bottom), except the final outputs are poses instead of states.
	\item \textbf{Path pose:} 
	an adaptation of the method proposed by Jiang and Grauman \cite{jiang2017seeing}. This method maps a sequence of planar homographies to poses along with temporal conditional random field (CRF) smoothing to estimate the pose sequence. We do not use static scene cues as the original work since our training data is synthetic.
\end{itemize}

Both of these baselines do not impose any physical constraints on their solutions but rather attempt to directly estimate body poses.

\paragraph{Evaluation Metrics.}
To evaluate the accuracy and physical soundness of all methods, we use both pose-based and physics-based metrics:

\begin{itemize}[leftmargin=*]
	\item \textbf{Pose error:} Pose-based metric that measures the euclidean distance between the generated pose sequence $p_{1:T}$ and the true pose sequence $\hat{p}_{1:T}$. It can be calculated as $\frac{1}{T}\sum_{t=1}^T||p_t - \hat{p}_t||_2$.
	\item \textbf{2D projection error:} Pose-based metric used for real-world datasets where the ground-truth 3D pose sequence of the person is unknown. We project the 3D joint locations of our estimated pose into a 2D image plane using a side-view virtual camera. The 2D projection error can be calculated as $\frac{1}{TJ}\sum_{t=1}^T\sum_{j=1}^J||q_t^j - \hat{q}_t^j||_2$ where $q_t^j$ is the $j$-th joint's 2D position of our estimated pose and $\hat{q}_t^j$ is the ground-truth.  We use OpenPose \cite{cao2017realtime} to extract the ground-truth 2D joint positions from the side-view video. To comply with OpenPose, we only evaluate 12 joints (hips, knees, ankles, shoulders, elbows, and wrists). For the 2D poses from our method and OpenPose, we align their positions of the center of the hips and scale the 2D coordinates to make the distance between the shoulder and hip equals 0.5.
	\item \textbf{Velocity error:} Physics-based metric that measures the euclidean distance between generated velocity sequence $v_{1:T}$ and true velocity sequence $\hat{v}_{1:T}$. It can be calculated as $\frac{1}{T}\sum_{t=1}^T||v_t - \hat{v}_t||_2$. $v_t$ can be approximated by $(p_{t+1} - p_t) / h$ using finite difference method where $h$ is the time step and $\hat{v}_t$ is computed in the same fashion. 
	\item \textbf{Smoothness:} Physics-based metric that uses average magnitude of joint accelerations to measure the smoothness of the generated pose sequence. It can be calculated as $\frac{1}{TG}\sum_{t=1}^T||a_t||_1$ where $G$ is the number of actuated DoFs and $a_t$ can be approximated by $(v_{t+1} - v_t) / h$. 
\end{itemize}

\subsection{Implementation Details}
\textbf{Motion capture data and simulation.}
We use CMU graphics lab motion capture database to learn expert policies as described in Sec.~\ref{egopose18:sec:stage1}. The humanoid is similarly constructed as the CMU humanoid model in DeepMind control suite \cite{tassa2018deepmind} with tweaks on joint stiffness, damping and torque limits. Please refer to the \href{https://github.com/Khrylx/EgoPose}{code} for further details. We learn 4 expert policies from 4 clips (0801, 0804, 0807, 0901) of the motion capture data corresponding to three styles of walking (slow, normal, fast) and one style of running. The physics simulation environment has a simulation timestep of 6.67ms and a control timestep of 33.3ms (control changes after 5 simulation steps).

\paragraph{Imitation learning parameters.}
The video-conditioned policy is pre-trained using behavior cloning for 100 iterations. In VGAIL, at every iteration, the policy generates sample trajectories with a total batch size of 50k timesteps. We perform online z-filtering of state inputs for normalization. The standard deviation for each action dimension is initialized to 0.1. The reward is clipped with a max value of 10 and advantages are normalized. For policy optimization, we use proximal policy optimization (PPO \cite{schulman2017proximal}) with a 0.2 clipping threshold. The discount factor $\gamma$ is 1. The learning rate for the policy and discriminator is 5e-5 and 1e-5 respectively with the discriminator updated 5 times in the inner loop. We terminate the training after 6000 iterations to prevent over-fitting. When fine-tuning the policy, we reduce the batch size to 5k and it takes about 2s per iteration on a GTX 1080Ti.

\begin{table}[ht]
	\centering
	\begin{tabular}{cccc}
		\hline
		& Smoothness & ~~Velocity error~~  & Pose error \\\hline
		Ours  & \textbf{11.9876} &  \textbf{6.5143} & 0.9779   \\
		Pose regression & 36.1628 & 9.0611  & \bf 0.8310  \\
		Path pose \cite{jiang2017seeing} & 198.6509 & 45.0189 & 1.7643    \\\hline\hline
		& Smoothness & ~~Velocity error~~  & Pose error \\\hline
		Ours  & \bf 11.9876 &  6.5143 & 0.9779   \\
		Ours-IE & 12.2472 & 7.5337 & 1.2219   \\
		Ours-GTI & 12.2968 & \bf 6.0761 & \bf 0.6688   \\\hline\hline
		& Smoothness  & ~~~ 2D projection error ~~~ \\\hline
		Ours  & \textbf{11.54} & \textbf{0.1325}   \\
		Pose regression & 44.11 & 0.1621   \\
		Path pose \cite{jiang2017seeing} & 214.21 & 0.1738 \\\hline
	\end{tabular}
	\bigskip
	\vspace{3mm}
	\caption{\textbf{Top:} Results for pose-based and physics-based metrics on the virtual test dataset. \textbf{Mid:} Ablation Study. (Ours-GTI) our method with ground-truth initial states. (Ours-IE) Our method with estimated initial states before fine-tuning. \textbf{Bottom:} Results for physics-based and pose-based metrics on real-world data.}
	\label{egopose18:table:results}
\end{table}

\section{Virtual World Validation}

We first evaluate our method on a test dataset generated using expert policies learned in Sec.~\ref{egopose18:sec:stage1}. The dataset consists of 20 trajectories, each of which is 100 timesteps long. The policy is fine-tuned for 20 iterations for each sequence.

\begin{figure}[t]
	\centering
	\includegraphics[width=\textwidth]{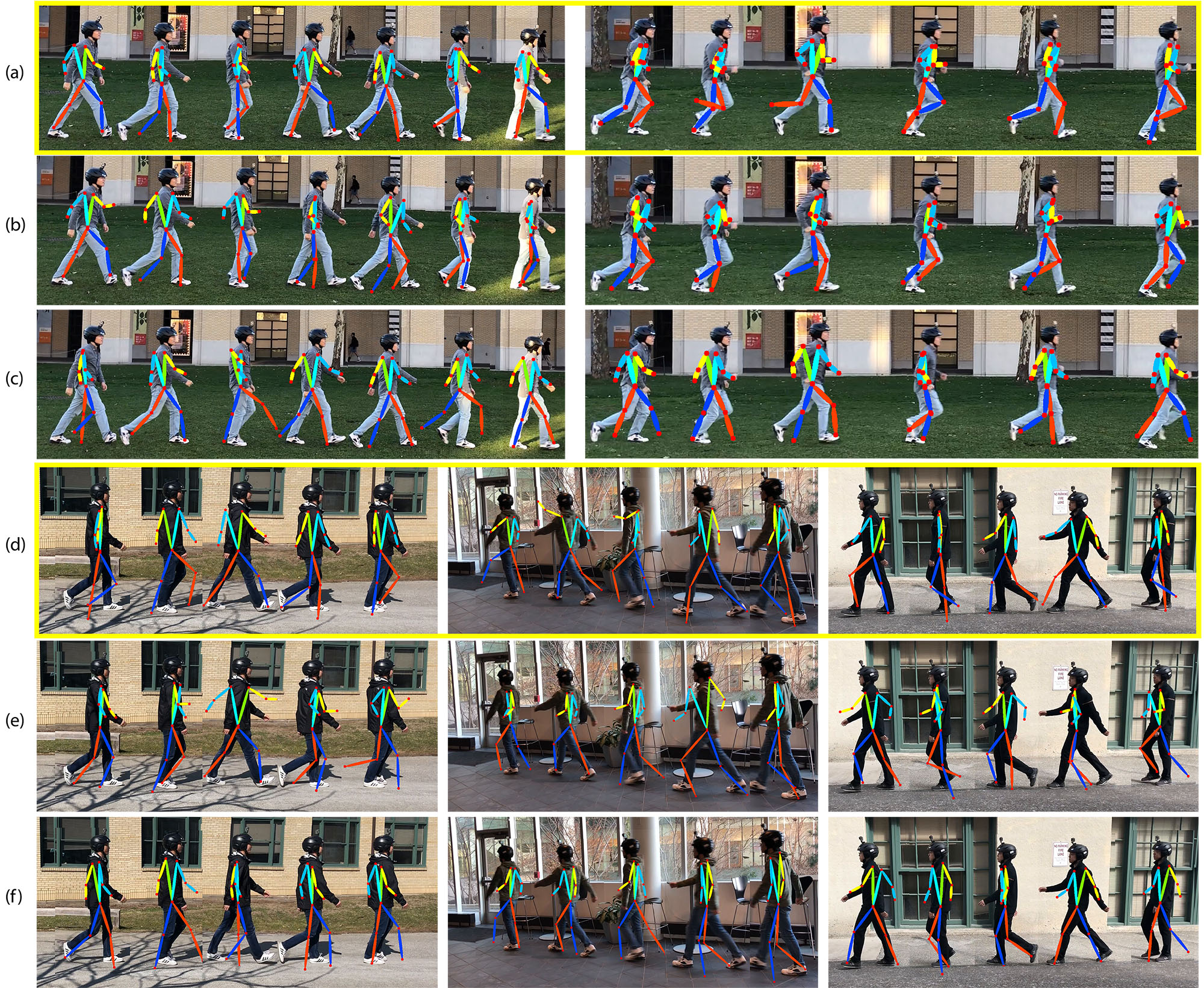}
	\caption{Qualitative results on real world dataset. (a)(d) Our method (yellow box); (b)(e) Pose regression; (c)(f) Path pose \cite{jiang2017seeing}. Yellow and orange bones correspond to the left arm and leg respectively.} 
	\label{egopose18:fig:real}
\end{figure}

Table~\ref{egopose18:table:results}(Top) shows a comparison of our method against the two baselines (pose regression and path pose). We observe that our method outperforms the baselines in terms of physics-based metrics (acceleration and velocity error), and the pose estimation is reasonably accurate.

\paragraph{Ablation Study.} As shown in Table~\ref{egopose18:table:results}(Mid), the accuracy of the initial state plays an important role in our method. As expected, our method with ground-truth initial states is much more accurate than with estimated initial states. This is because sometimes the humanoid falls down due to the error in the initial state estimate as shown in Figure~\ref{egopose18:fig:net_ft}(Top). Our fine-tuning approach can adapt the policy to recover from the error in the initial state and generate a more accurate pose sequence (see Figure~\ref{egopose18:fig:net_ft}(Mid)).

\section{Real World Validation}

To understand the true utility of our approach, we must evaluate its performance on real-world first-person videos. In this experiment, we apply our virtually trained video-conditioned policy on real video data and show that our approach is able to estimate both accurate and physically-valid pose sequences. Since we do not have access to the true 3D poses of the person recording the egocentric video, we use a secondary static camera (third-person POV) to measure the error of our pose estimation based on 2D projections of joint positions.

We evaluate our proposed method on 12 video sequences composed of 3 different people performing the walking activity and running activity, in both outdoor and indoor scenes. Each egocentric video is 3-7 seconds long and obtained by a head-mounted GoPro camera. For each sequence, the policy is fine-tuned for 50 iterations. As indicated in Table~\ref{egopose18:table:results}(Bottom), our method estimates much smoother (3.8x, 18.5x) pose sequences and is also more accurate in terms of 2D projection error (18\%, 24\%). Figure~\ref{egopose18:fig:real} shows a qualitative comparison of our approach against the two baselines.

\section{Conclusion}
We proposed a physically-grounded ego-pose estimation method that learns a video-conditioned control policy to generate the pose estimates in physics simulation. We evaluated our method on both simulation data and real-world data and show that our approach significantly outperforms baseline methods in terms of physics-based metrics and is also accurate. Our experiments also demonstrated the effectiveness of our proposed fine-tuning approach for domain adaptation from synthetic to real data. We believe our work is one of the first to open new research directions that consider the role of physics in understanding human motion in computer vision.
\chapter{Simulation-Based Third-Person Human Pose Estimation}
\label{chap:simpoe}

\section{Introduction}
\label{simpoe:sec:intro}
We aim to show that accurate 3D human pose estimation from monocular video requires modeling both kinematics and dynamics. Human dynamics, \emph{i.e.,} body motion modeling with physical forces, has gained relatively little attention in 3D human pose estimation compared to its counterpart, kinematics, which models motion without physical forces. There are two main reasons for the disparity between these two equally important approaches. First, kinematics is a more direct approach that focuses on the geometric relationships of 3D poses and 2D images; it sidesteps the challenging problem of modeling the physical forces underlying human motion, which requires significant domain knowledge about physics and control. Second, compared to kinematic measurements such as 3D joint positions, physical forces present unique challenges in their measurement and annotation, which renders standard supervised learning paradigms unsuitable. Thus, almost all state-of-the-art methods~\cite{pavlakos2019expressive,xiang2019monocular,kolotouros2019learning,kocabas2020vibe,moon2020i2l} for 3D human pose estimation from monocular video are based only on kinematics.
Although these kinematic methods can estimate human motion with high pose accuracy, they often fail to produce physically-plausible motion. Without modeling the physics of human dynamics, kinematic methods have no notion of force, mass or contact; they also do not have the ability to impose physical constraints such as joint torque limits or friction. As a result, kinematic methods often generate physically-implausible motions with pronounced artifacts: body parts (\emph{e.g.,} feet) penetrate the ground; the estimated poses are jittery and vibrate excessively; the feet slide back and forth when they should be in static contact with the ground. All these physical artifacts significantly limit the application of kinematic pose estimation methods. For instance, jittery motions can be misleading for medical monitoring and sports training; physical artifacts also prevent applications in computer animation and virtual/augmented reality since people are exceptionally good at discerning even the slightest clue of physical inaccuracy~\cite{reitsma2003perceptual,hoyet2012push}.

\begin{figure}[t]
    \centering
    \includegraphics[width=\textwidth]{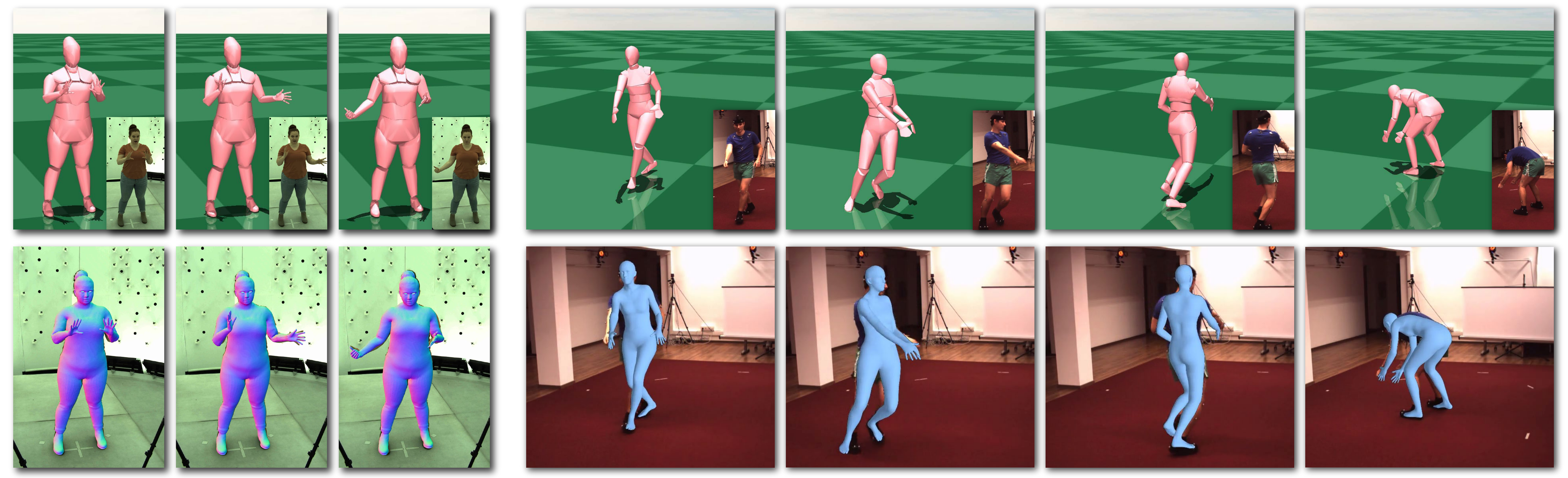}
    \caption{Our SimPoE framework learns a kinematics-aware video-conditioned policy that controls a character in a physics simulator (\textbf{Top}) and estimates accurate and physically-plausible human motion (\textbf{Bottom}).}
    \label{simpoe:fig:teaser}
\end{figure}

To improve the physical plausibility of estimated human motion from video, recent work~\cite{li2019estimating,rempe2020,shimada2020physcap} has started to adopt the use of dynamics in their formulation. These methods first estimate kinematic motion and then use physics-based trajectory optimization to optimize the forces to induce the kinematic motion. Although they can generate physically-grounded motion, there are several drawbacks of trajectory optimization-based approaches. First, trajectory optimization entails solving a highly-complex optimization problem at test time. This can be computationally intensive and requires the batch processing of a temporal window or even the entire motion sequence, causing high latency in pose predictions and making it unsuitable for interactive real-time applications. Second, trajectory optimization requires simple and differentiable physics models to make optimization tractable, which can lead to high approximation errors compared to advanced and non-differentiable physics simulators (\emph{e.g.,} MuJoCo~\cite{todorov2012mujoco}, Bullet~\cite{coumans2010bullet}). Finally and most importantly, the application of physics in trajectory optimization-based methods is implemented as a post-processing step that projects a given kinematic motion to a physically-plausible one. Since it is optimization-based, there is no learning mechanism in place that tries to match the optimized motion to the ground truth. As such, the resulting motion from trajectory optimization can be physically-plausible but still far from the ground-truth, especially when the input kinematic motion is inaccurate.

To address these limitations, we present a new approach, SimPoE (\emph{\textbf{Sim}ulated Character Control for Human \textbf{Po}se \textbf{E}stimation}), that tightly integrates image-based kinematic inference and physics-based dynamics modeling into a joint learning framework. Unlike trajectory optimization, SimPoE is a causal temporal model with an integrated physics simulator. Specifically, SimPoE learns a policy that takes the current pose and the next image frame as input, and produces controls for a proxy character inside the simulator that outputs the pose estimate for the next frame. To perform kinematic inference, the policy contains a learnable kinematic pose refinement unit that uses image evidence (2D keypoints) to iteratively refine a kinematic pose estimate. Concretely, the refinement unit takes as input the gradient of keypoint reprojection loss, which encodes rich information about the geometry of pose and keypoints, and outputs the kinematic pose update. Based on this refined kinematic pose, the policy then computes a character control action, \emph{e.g.,} target joint angles for the character's proportional-derivative (PD) controllers, to advance the character state and obtain the next-frame pose estimate. This policy design couples the kinematic pose refinement unit with the dynamics-based control generation unit, which are learned jointly with reinforcement learning (RL) to ensure both accurate and physically-plausible pose estimation. At each time step, a reward is assigned based on the similarity between the estimated motion and the ground truth. To further improve pose estimation accuracy, SimPoE also includes a new control mechanism called meta-PD control. PD controllers are widely used in prior work~\cite{peng2017learning,peng2018deepmimic,yuan2019ego} to convert the action produced by the policy into the joint torques that control the character. However, the PD controller parameters typically have fixed values that require manual tuning, which can produce sub-optimal results. Instead, in meta-PD control, SimPoE's policy is also trained to dynamically adjust the PD controller parameters across simulation steps based on the state of the character to achieve a finer level of control over the character's motion. 

We validate our approach, SimPoE, on two large-scale datasets, Human3.6M~\cite{ionescu2013human3} and an in-house human motion dataset that also contains \emph{detailed finger motion}. We compare \mbox{SimPoE} against state-of-the-art monocular 3D human pose estimation methods including both kinematic and physics-based approaches. On both datasets, SimPoE outperforms previous art in both pose-based and physics-based metrics, with significant pose accuracy improvement over prior physics-based methods. We further conduct extensive ablation studies to investigate the contribution of our proposed components including the kinematic refinement unit, meta-PD control, as well as other design choices.

The main contributions of this paper are as follows: (1)~We present a joint learning framework that tightly integrates image-based kinematic inference and physics-based dynamics modeling to achieve accurate and physically-plausible 3D human pose estimation from monocular video. (2)~Our approach is causal, runs in real-time without batch trajectory optimization, and addresses several drawbacks of prior physics-based methods. (3)~Our proposed meta-PD control mechanism eliminates manual dynamics parameter tuning and enables finer character control to improve pose accuracy. (4) Our approach outperforms previous art in both pose accuracy and physical plausibility. (5) We perform extensive ablations to validate the proposed components to establish good practices for RL-based human pose estimation.

\section{Related Work}
\paragraph{Kinematic 3D Human Pose Estimation.}
Numerous prior works estimate 3D human joint locations from monocular video using either two-stage~\cite{dabral2018learning,rayat2018exploiting,pavllo20193d} or end-to-end~\cite{mehta2017vnect,mehta2018single} frameworks. On the other hand, parametric human body models~\cite{anguelov2005scape,loper2015smpl,pavlakos2019expressive} are widely used as the human pose representation since they additionally provide skeletal joint angles and a 3D body mesh. Optimization-based methods have been used to fit the SMPL body model~\cite{loper2015smpl} to 2D keypoints extracted from an image ~\cite{bogo2016keep,lassner2017unite}. Alternatively, regression-based approaches use deep neural networks to directly regress the parameters of the SMPL model from an image~\cite{tung2017self,tan2017indirect,pavlakos2018learning,omran2018neural,kanazawa2018end,guler2019holopose}, using weak supervision from 2D keypoints~\cite{tung2017self,tan2017indirect,kanazawa2018end} or body part segmentation~\cite{omran2018neural,pavlakos2018learning}. Song \emph{et al.}~\cite{song2020human} propose neural gradient descent to fit the SMPL model using 2D keypoints. Regression-based~\cite{kanazawa2018end} and optimization-based~\cite{bogo2016keep} methods have also been combined to produce pseudo ground truth from weakly-labeled images~\cite{kolotouros2019learning} to facilitate learning. Recent work~\cite{arnab2019exploiting,huang2017towards,kanazawa2018learning,sun2019human,kocabas2020vibe,luo20203d} starts to exploit the temporal structure of human motion to estimate smooth motion. Kanazawa \emph{et al.}~\cite{kanazawa2018learning} model human kinematics by predicting past and future poses. Transformers~\cite{vaswani2017attention} have also been used to improve the temporal modeling of human motion~\cite{sun2019human}. All the aforementioned methods disregard human dynamics, \emph{i.e.,} the physical forces that generate human motion. As a result, these methods often produce physically-implausible motions with pronounced physical artifacts such as jitter, foot sliding, and ground penetration.

\paragraph{Physics-Based Human Pose Estimation.} A number of works have addressed human dynamics for 3D human pose estimation. Most prior works~\cite{brubaker2009estimating,wei2010videomocap,vondrak2012video,zell2017joint,yuan2019ego,rempe2020,shimada2020physcap} use trajectory optimization to optimize the physical forces to induce the human motion in a video. As discussed in Sec.~\ref{simpoe:sec:intro}, trajectory optimization is a batch procedure which has high latency and is typically computationally expensive, making it unsuitable for real-time applications. Furthermore, these methods cannot utilize advanced physics simulators with non-differentiable dynamics. Most importantly, there is no learning mechanism in trajectory optimization-based methods that tries to match the optimized motion to the ground truth. Our approach addresses these drawbacks with a framework that integrates kinematic inference with RL-based character control, which runs in real-time, is compatible with advanced physics simulators, and has learning mechanisms that aim to match the output motion to the ground truth. Although prior work~\cite{yuan20183d,yuan2019ego,isogawa2020optical} has used RL to produce simple human locomotions from videos, these methods only learn policies that coarsely mimic limited types of motion instead of precisely tracking the motion presented in the video. In contrast, our approach can achieve accurate pose estimation by integrating images-based kinematic inference and RL-based character control with the proposed policy design and meta-PD control.

\paragraph{Reinforcement Learning for Character Control.}
Deep RL has become the preferred approach for learning character control policies with manually-designed rewards~\cite{liu2017learning,liu2018learning,peng2018deepmimic,peng2018sfv}. GAIL \cite{ho2016generative} based methods are proposed to learn character control without reward engineering~\cite{merel2017learning,wang2017robust}. To produce long-term behaviors, prior work has used hierarchical RL to control characters to achieve high-level tasks~\cite{merel2018neural,merel2018hierarchical,peng2019mcp,merel2020catch}. Recent work also uses deep RL to learn user-controllable policies from motion capture data for character animation~\cite{bergamin2019drecon,park2019learning,won2020scalable}. Prior work in this domain learns control policies that reproduce training motions, but the policies do not transfer to unseen test motions, nor do they estimate motion from video as our method does.

\section{Approach}

\begin{figure*}
    \centering
    \includegraphics[width=\textwidth]{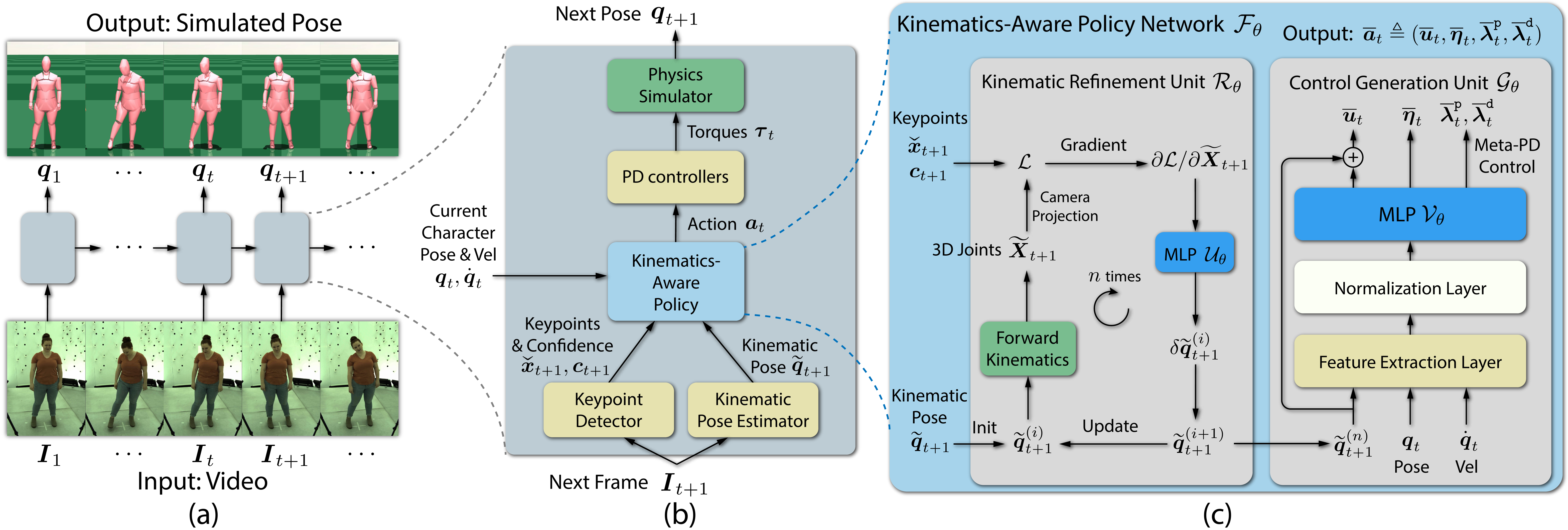}
    \caption{\textbf{Overview of our SimPoE framework.} (a) SimPoE is a physics-based causal temporal model. (b) At each frame (30Hz), the policy network $\mathcal{F}_\theta$ use the current pose $\bs{q}_t$, velocities $\dot{\bs{q}}_t$, and the next frame's estimated kinematic pose $\widetilde{\bs{q}}_{t+1}$ and keypoints $(\widecheck{\bs{x}}_{t+1}, \bs{c}_{t+1})$ to generate an action $\bs{a}_t$, which controls the character in the physics simulator (450Hz) via PD controllers to produce the next pose $\bs{q}_{t+1}$. (c) The policy network $\mathcal{F}_\theta$ outputs the mean action $\overline{\bs{a}}_t \triangleq (\overline{\bs{u}}_t, \overline{\bs{\eta}}_t,\overline{\bs{\lambda}}_t^\texttt{p},\overline{\bs{\lambda}}_t^\texttt{d})$. The kinematic refinement unit iteratively refines a kinematic pose estimate by learning pose updates. The refined pose $\widetilde{\bs{q}}^{(n)}_{t+1}$ is used by the control generation unit to produce the mean action $\overline{\bs{a}}_t$.}
    \label{simpoe:fig:overview}
\end{figure*}

The overview of our SimPoE (\emph{\textbf{Sim}ulated Character Control for Human \textbf{Po}se \textbf{E}stimation}) framework is illustrated in Fig.~\ref{simpoe:fig:overview}.
The input to SimPoE is a video $\bs{I}_{1:T} = (\bs{I}_1, \ldots, \bs{I}_T)$ of a person with $T$ frames. For each frame $\bs{I}_t$, we first use an off-the-shelf kinematic pose estimator to estimate an initial kinematic pose $\widetilde{\bs{q}}_t$, which consists of the joint angles and root translation of the person; we also extract 2D keypoints $\widecheck{\bs{x}}_t$ and their confidence $\bs{c}_t$ from $\bs{I}_t$ using a given pose detector (\emph{e.g.,} OpenPose~\cite{cao2017realtime})). As the estimated kinematic motion $\widetilde{\bs{q}}_{1:T} = (\widetilde{\bs{q}}_1, \ldots, \widetilde{\bs{q}}_T)$ is obtained without modeling human dynamics, it often contains physically-implausible poses with artifacts like jitter, foot sliding, and ground penetration. This motivates the main stage of our method, \emph{simulated character control}, where we model human dynamics with a proxy character inside a physics simulator. The character's initial pose $\bs{q}_1$ is set to $\widetilde{\bs{q}}_1$. At each time step $t$ shown in Fig.~\ref{simpoe:fig:overview}\,(b), SimPoE learns a policy that takes as input the current character pose $\bs{q}_t$, velocities $\dot{\bs{q}}_t$, as well as the next frame's kinematic pose $\widetilde{\bs{q}}_{t+1}$ and keypoints $(\widecheck{\bs{x}}_{t+1},\bs{c}_{t+1})$ to produce an action that controls the character in the simulator to output the next pose $\bs{q}_{t+1}$. By repeating this causal process, we obtain the physically-grounded estimated motion $\bs{q}_{1:T} = (\bs{q}_1, \ldots, \bs{q}_T)$ of SimPoE.

\subsection{Automated Character Creation}
\label{simpoe:sec:character_creation}
The character we use as a proxy to simulate human motion is created from skinned human mesh models, \emph{e.g.,} the SMPL model~\cite{loper2015smpl}, which can be recovered via SMPL-based pose estimation methods such as VIBE~\cite{kocabas2020vibe}. These skinned mesh models provide a skeleton of $B$ bones, a mesh of $V$ vertices, and a skinning weight matrix $\bs{W} \in \mathbb{R}^{V \times B}$ where each element $W_{ij}$ specifies the influence of the $j$-th bone's transformation on the $i$-th vertex's position. We can obtain a rigid vertex-to-bone association $\bs{A} \in \mathbb{R}^{V}$ by assigning each vertex~$i$ to the bone with the largest skinning weight for it: $A_i = \argmax_j W_{ij}$. With the vertex-to-bone association $\bs{A}$, we can then create the geometry of each bone by computing the 3D convex hull of all the vertices assigned to the bone. Assuming constant density, the mass of each bone is determined by the volume of its geometry. Our character creation process is fully automatic, is compatible with popular body mesh models (\emph{e.g.,} SMPL), and ensures proper body geometry and mass assignment.

\subsection{Simulated Character Control}

The task of controlling a character agent in physics simulation to generate desired human motions can be formulated as a Markov decision process (MDP), which is defined by a tuple $\mathcal{M} = (\mathcal{S}, \mathcal{A}, \mathcal{T}, R, \gamma)$ of states, actions, transition dynamics, a reward function, and a discount factor. The character agent interacts with the physics simulator according to a policy $\pi(\bs{a}_t|\bs{s}_t)$, which models the conditional distribution of choosing an action $\bs{a}_t \in \mathcal{A}$ given the current state $\bs{s}_t \in \mathcal{S}$ of the agent. Starting from some initial state $\bs{s}_1$, the character agent iteratively samples an action $\bs{a}_t$ from the policy $\pi$ and the simulation environment with transition dynamics $\mathcal{T}(\bs{s}_{t+1}|\bs{s}_t, \bs{a}_t)$ generates the next state $\bs{s}_{t+1}$ and gives the agent a reward $r_t$. The reward is assigned based on how well the character's motion aligns with the ground-truth human motion. The goal of our character control learning process is to learn an optimal policy $\pi^\ast$ that maximizes the expected return $J(\pi) = \mathbb{E}_{\pi}\left[\sum_{t}\gamma^t r_t\right]$ which translates to imitating the ground-truth motion as closely as possible. We apply a standard reinforcement learning algorithm (PPO~\cite{schulman2017proximal}) to solve for the optimal policy. In the following, we provide a detailed description of the states, actions and rewards of our control learning process. We then use a dedicated Sec.~\ref{simpoe:sec:policy} to elaborate on our policy design.

\paragraph{States.} The character state $\bs{s}_t \triangleq (\bs{q}_t, \dot{\bs{q}}_t, \widetilde{\bs{q}}_{t+1}, \widecheck{\bs{x}}_{t+1}, \bs{c}_{t+1})$ consists of the character's current pose $\bs{q}_t$, joint velocities (time derivative of the pose) $\dot{\bs{q}}_t$, as well as the estimated kinematic pose $\widetilde{\bs{q}}_{t+1}$, 2D keypoints $\widecheck{\bs{x}}_{t+1}$ and keypoint confidence $\bs{c}_{t+1}$ of the next frame. The state includes information of both the current frame ($\bs{q}_t,\dot{\bs{q}}_t$) and next frame ($\widetilde{\bs{q}}_{t+1}$, $\widecheck{\bs{x}}_{t+1}$,$\bs{c}_{t+1}$), so that the agent learns to take the right action $\bs{a}_{t}$ to transition from the current pose $\bs{q}_t$ to a desired next pose $\bs{q}_{t+1}$, \emph{i.e.,} pose close to the ground truth.

\paragraph{Actions.} The policy $\pi(\bs{a}_t|\bs{s}_t)$ runs at 30Hz, the input video's frame rate, while our physics simulator runs at 450Hz to ensure stable simulation. This means one policy step corresponds to 15 simulation steps. One common design of the policy's action $\bs{a}_t$ is to directly output the torques $\bs{\tau}_t$ to be applied at each joint (except the root), which are used repeatedly by the simulator during the 15 simulation steps. However, finer control can be achieved by adjusting the torques at each step based on the state of the character. Thus, we follow prior work~\cite{peng2017learning,yuan2019ego} and use proportional-derivative (PD) controllers at each non-root joint to produce torques. With this design, the action $\bs{a}_t$ includes the target joint angles $\bs{u}_t$ of the PD controllers. At the $j$-th of the 15 simulation (PD controller) steps, the joint torques $\bs{\tau}_t$ are computed as 
\begin{equation}
\label{simpoe:eq:pd}
\bs{\tau}_t = \bs{k}_\texttt{p}\circ(\bs{u}_t - \bs{q}_t^\texttt{nr}) - \bs{k}_\texttt{d}\circ\dot{\bs{q}}_t^\texttt{nr},
\end{equation}
where $\bs{k}_\texttt{p}$ and $\bs{k}_\texttt{d}$ are the parameters of the PD controllers, $\bs{q}_t^\texttt{nr}$ and $\dot{\bs{q}}_t^\texttt{nr}$ denote the joint angles and velocities of non-root joints at the start of the simulation step, and $\circ$ denotes element-wise multiplication. The PD controllers act like damped springs that drive joints to target angles $\bs{u}_t$, where $\bs{k}_\texttt{p}$ and $\bs{k}_\texttt{d}$ are the stiffness and damping of the springs. In Sec.~\ref{simpoe:sec:meta_pd}, we will introduce a new control mechanism, meta-PD control, that allows $\bs{k}_\texttt{p}$ and $\bs{k}_\texttt{d}$ to be dynamically adjusted by the policy to achieve an even finer level of character control. With Meta-PD control, the action $\bs{a}_t$ includes elements $\bs{\lambda}_t^\texttt{p}$ and $\bs{\lambda}_t^\texttt{d}$ for adjusting $\bs{k}_\texttt{p}$ and $\bs{k}_\texttt{d}$ respectively. As observed in prior work~\cite{yuan2020residual}, allowing the policy to apply external residual forces to the root greatly improves the robustness of character control. Thus, we also add the residual forces and torques $\bs{\eta}_t$ of the root into the action $\bs{a}_t$. Overall, the action is defined as $\bs{a}_t \triangleq (\bs{u}_t, \bs{\eta}_t, \bs{\lambda}_t^\texttt{p}, \bs{\lambda}_t^\texttt{d})$.

\paragraph{Rewards.} In order to learn the policy, we need to define a reward function that encourages the motion $\bs{q}_{1:T}$ generated by the policy to match the ground-truth motion $\widehat{\bs{q}}_{1:T}$. Note that we use $\;\widehat{\cdot}\;$ to denote ground-truth quantities. The reward $r_t$ at each time step is defined as the multiplication of four sub-rewards:
\begin{align}
    r_t = r^\texttt{p}_t \cdot r^\texttt{v}_t \cdot r^\texttt{j}_t \cdot r^\texttt{k}_t\,.
\end{align}
The pose reward $r^\texttt{p}_t$ measures the difference between the local joint orientations $\bs{o}^j_t$ and the ground truth~$\widehat{\bs{o}}^j_t$:
\begin{equation}
\label{simpoe:eq:r_p}
    r^\texttt{p}_t = \exp\left[-\alpha_\texttt{p}\left(\sum_{j=1}^J\|\bs{o}_t^j\ominus \widehat{\bs{o}}_t^j\|^2\right)\right],
\end{equation}
where $J$ is the total number of joints, $\ominus$ denotes the relative rotation between two rotations, and $\|\cdot\|$ computes the rotation angle. The velocity reward $r^\texttt{v}_t$ measures the mismatch between joint velocities $\dot{\bs{q}}_t$ and the ground truth $\widehat{\dot{\bs{q}}}_t$:
\begin{equation}
\label{simpoe:eq:r_v}
    r^\texttt{v}_t = \exp\left[-\alpha_\texttt{v}\|\dot{\bs{q}}_t -  \widehat{\dot{\bs{q}}}_t\|^2\right].
\end{equation}
The joint position reward $r^\texttt{j}_t$ encourages the 3D world joint positions $\bs{X}_t^j$ to match the ground truth  $\widehat{\bs{X}}_t^j$:
\begin{equation}
\label{simpoe:eq:r_j}
    r^\texttt{j}_t = \exp\left[-\alpha_\texttt{j}\left(\sum_{j=1}^J\|\bs{X}_t^j - \widehat{\bs{X}}_t^j\|^2\right)\right].
\end{equation}
Finally, the keypoint reward $r^\texttt{k}_t$ pushes the 2D image projection $\bs{x}_t^j$ of the joints to match the ground truth $\widehat{\bs{x}}_t^j$:
\begin{equation}
\label{simpoe:eq:r_k}
    r^\texttt{k}_t = \exp\left[-\alpha_\texttt{k}\left(\sum_{j=1}^J\|\bs{x}_t^j - \widehat{\bs{x}}_t^j\|^2\right)\right].
\end{equation}
Note that the orientations $\bs{o}^j_t$, 3D joint positions $\bs{X}_t^j$ and 2D image projections $\bs{x}_t^j$ are functions of the pose $\bs{q}_t$. The joint velocities $\dot{\bs{q}}_t$ are computed via finite difference. There are also weighting factors $\alpha_\texttt{p}, \alpha_\texttt{v}, \alpha_\texttt{j}, \alpha_\texttt{k}$ inside each reward.
These sub-rewards complement each other by matching different features of the generated motion to the ground-truth: joint angles, velocities, as well as 3D and 2D joint positions. Our reward design is multiplicative, which eases policy learning as noticed by prior work~\cite{won2020scalable}. The multiplication of the sub-rewards ensures that none of them can be overlooked in order to achieve a high reward.

\subsection{Kinematics-Aware Policy}
\label{simpoe:sec:policy}

As the action $\bs{a}_t$ is continuous, we adopt a parametrized Gaussian policy $\pi_\theta(\bs{a}_t|\bs{s}_t) = \mathcal{N}(\overline{\bs{a}}_t, \bs{\Sigma})$ where the mean $\overline{\bs{a}}_t \triangleq (\overline{\bs{u}}_t, \overline{\bs{\eta}}_t, \overline{\bs{\lambda}}_t^\texttt{p}, \overline{\bs{\lambda}}_t^\texttt{d})$ is output by a neural network $\mathcal{F}_\theta$ with parameters $\theta$, and $\bs{\Sigma}$ is a fixed diagonal covariance matrix whose elements are treated as hyperparameters. The noise inside the Gaussian policy governed by $\bs{\Sigma}$ allows the agent to explore different actions around the mean action~$\overline{\bs{a}}_t$ and use these explorations to improve the policy during training. At test time, the noise is removed and the character agent always takes the mean action $\overline{\bs{a}}_t$ to improve performance. 

Now let us focus on the design of the policy network $\mathcal{F}_\theta$ that maps the state $\bs{s}_t$ to the mean action~$\overline{\bs{a}}_t$. Based on the design of $\bs{s}_t$, the mapping can be written as 
\begin{equation}
    \overline{\bs{a}}_t = \mathcal{F}_\theta\left(\bs{q}_t, \dot{\bs{q}}_t, \widetilde{\bs{q}}_{t+1}, \widecheck{\bs{x}}_{t+1}, \bs{c}_{t+1}\right).
\end{equation}
Recall that $\widetilde{\bs{q}}_{t+1}$ is the kinematic pose, $\widecheck{\bs{x}}_{t+1}$ and $\bs{c}_{t+1}$ are the detected 2D keypoints and their confidence, and that they are all information about the next frame. 
The overall architecture of our policy network $\mathcal{F}_\theta$ is illustrated in Fig.~\ref{simpoe:fig:overview}\,(c).
The components $(\overline{\bs{u}}_t, \overline{\bs{\eta}}_t, \overline{\bs{\lambda}}_t^\texttt{p}, \overline{\bs{\lambda}}_t^\texttt{d})$ of the mean action $\overline{\bs{a}}_t$ are computed as follows:
\begin{align}
\label{simpoe:eq:ref}
    \widetilde{\bs{q}}_{t+1}^{(n)} &= \mathcal{R}_\theta\left(\widetilde{\bs{q}}_{t+1}, \widecheck{\bs{x}}_{t+1}, \bs{c}_{t+1}\right),\\
\label{simpoe:eq:action}
    (\delta\overline{\bs{u}}_t, \overline{\bs{\eta}}_t, \overline{\bs{\lambda}}_t^\texttt{p}, \overline{\bs{\lambda}}_t^\texttt{d}) &=  \mathcal{G}_\theta\left(\widetilde{\bs{q}}_{t+1}^{(n)}, \bs{q}_t, \dot{\bs{q}}_t\right), \\
\label{simpoe:eq:residual}
    \overline{\bs{u}}_t &= \widetilde{\bs{q}}_{t+1}^{(n)} + \delta\overline{\bs{u}}_t\,.
\end{align}
In Eq.~\eqref{simpoe:eq:ref}, $\mathcal{R}_\theta$ is a kinematic refinement unit that iteratively refines the kinematic pose $\widetilde{\bs{q}}_{t+1}$ using the 2D keypoints $\widecheck{\bs{x}}_{t+1}$ and confidence $\bs{c}_{t+1}$, and $\widetilde{\bs{q}}_{t+1}^{(n)}$ is the refined pose after $n$ iterations of refinement. Eq.~\eqref{simpoe:eq:action} and \eqref{simpoe:eq:residual} describe a control generation unit $\mathcal{G}_\theta$ that maps the refined pose $\widetilde{\bs{q}}_{t+1}^{(n)}$, current pose $\bs{q}_t$ and velocities $\dot{\bs{q}}_t$ to the components of the mean action $\overline{\bs{a}}_t$. Specifically, the control generation unit  $\mathcal{G}_\theta$ includes a hand-crafted feature extraction layer, a normalization layer (based on running estimates of mean and variance) and another MLP $\mathcal{V}_\theta$, as illustrated in Fig.~\ref{simpoe:fig:overview}\,(c). As described in Eq.~\eqref{simpoe:eq:residual}, an important design of $\mathcal{G}_\theta$  is a residual connection that produces the mean PD controller target angles $\overline{\bs{u}}_t$ using the refined kinematic pose~$\widetilde{\bs{q}}_{t+1}^{(n)}$,
where we ignore the root angles and positions in $\widetilde{\bs{q}}_{t+1}^{(n)}$ for ease of notation. This design builds in proper inductive bias since $\widetilde{\bs{q}}_{t+1}^{(n)}$ provides a good guess for the desired next pose $\bs{q}_{t+1}$ and thus a good base value for $\overline{\bs{u}}_t$. It is important to note that the PD controller target angles $\bs{u}_t$ do not translate to the same next pose $\bs{q}_{t+1}$ of the character, \emph{i.e.,} $\bs{q}_{t+1} \neq \bs{u}_t$. The reason is that the character is subject to gravity and contact forces, and under these external forces the joint angles $\bs{q}_{t+1}$ will not be $\bs{u}_t$ when the PD controllers reach their equilibrium. As an analogy, since PD controllers act like springs, a spring will reach a different equilibrium position when you apply external forces to it. Despite this, the next pose $\bs{q}_{t+1}$ generally will not be far away from $\bs{u}_t$ and learning the residual $\delta\overline{\bs{u}}_t$ to $\widetilde{\bs{q}}_{t+1}^{(n)}$ is easier than learning from scratch as we will demonstrate in the experiments. This design also synergizes the kinematics of the character with its dynamics as the kinematic pose $\widetilde{\bs{q}}_{t+1}^{(n)}$ is now tightly coupled with the input of the character's PD controllers that control the character in the physics simulator.

\paragraph{Kinematic Refinement Unit.}
The kinematic refinement unit $\mathcal{R}_\theta$ is formed by an MLP $\mathcal{U}_\theta$ that maps a feature vector $\bs{z}$ (specific form will be described later) to a pose update:
\begin{align}
    \delta \widetilde{\bs{q}}_{t+1}^{(i)} &= \mathcal{U}_\theta\left(\bs{z}\right),\\
    \widetilde{\bs{q}}_{t+1}^{(i+1)} &= \widetilde{\bs{q}}_{t+1}^{(i)} + \delta \widetilde{\bs{q}}_{t+1}^{(i)}\,,
\end{align}
where $i$ denotes the $i$-th refinement iteration and $\widetilde{\bs{q}}_{t+1}^{(0)} = \widetilde{\bs{q}}_{t+1}$.
To fully leverage the 2D keypoints and kinematic pose at hand, we design the feature $\bs{z}$ to be the gradient of the keypoint reprojection loss with respect to current 3D joint positions, inspired by recent work~\cite{song2020human} on kinematic body fitting. The purpose of using the gradient is not to minimize the reprojection loss, but to use it as an informative kinematic feature to learn a pose update that eventually results in stable and accurate control of the character; there is no explicit minimization of the reprojection loss in our formulation.
Specifically, we first obtain the 3D joint positions $\widetilde{\bs{X}}_{t+1} = \text{FK}(\widetilde{\bs{q}}_{t+1}^{(i)})$ through forward kinematics and then compute the reprojection loss as
\begin{equation}
    \mathcal{L}(\widetilde{\bs{X}}_{t+1}) = \sum_{j=1}^J \left\| \Pi \left(\widetilde{\bs{X}}_{t+1}^j\right) - \widecheck{\bs{x}}_{t+1}^j\right\|^2\cdot c_{t+1}^j\,,
\end{equation}
where $\widetilde{\bs{X}}_{t+1}^j$ denotes the $j$-th joint position in $\widetilde{\bs{X}}_{t+1}$, $\Pi(\cdot)$ denotes the perspective camera projection, and $(\widecheck{\bs{x}}_{t+1}^j, c_{t+1}^j)$ are the $j$-th detected keypoint and its confidence. The gradient feature $\bs{z}\triangleq \partial \mathcal{L}/ \partial \widetilde{\bs{X}}_{t+1}$ is informative about the kinematic pose $\widetilde{\bs{q}}_{t+1}^{(i)}$ as it tells us how each joint should move to match the 2D keypoints $\widecheck{\bs{x}}_{t+1}^j$. It also accounts for keypoint uncertainty by weighting the loss with the keypoint confidence $c_{t+1}^j$. Note that $\bs{z}$ is converted to the character's root coordinate to be invariant of the character's orientation. The refinement unit integrates kinematics and dynamics as it utilizes a kinematics-based feature $\bs{z}$ to learn the update of a kinematic pose, which is used to produce dynamics-based control of the character. The joint learning of the kinematic refinement unit $\mathcal{R}_\theta$ and the control generation unit $\mathcal{G}_\theta$ ensures accurate and physically-plausible pose estimation.

\paragraph{Feature Extraction Layer.} After refinement, the control generation unit $\mathcal{G}_\theta$ needs to extract informative features from its input to output an action that advances the character from the current pose $\bs{q}_t$ to the next pose $\bs{q}_{t+1}$. To this end, the feature extraction layer uses information from both the current frame and next frame. Specifically, the extracted feature includes $\bs{q}_t$, $\dot{\bs{q}}_t$, the current 3D joint positions $\bs{X}_t$, the pose difference vector between $\bs{q}_t$ and the refined kinematic pose $\widetilde{\bs{q}}_{t+1}^{(n)}$, and the difference vector between $\bs{X}_t$ and the next-frame joint position $\widetilde{\bs{X}}_{t+1}$ computed from $\widetilde{\bs{q}}_{t+1}^{(n)}$. All features are converted to the character's root coordinate to be orientation-invariant and encourage robustness against variations in absolute pose encountered at test time.

\subsection{Meta-PD control}
\label{simpoe:sec:meta_pd}
PD controllers are essential in our approach as they relate the kinematics and dynamics of the character by converting target joint angles in pose space to joint torques. However, an undesirable aspect of PD controllers is the need to specify the parameters $\bs{k}_\texttt{p}$ and $\bs{k}_\texttt{d}$ for computing the joint torques $\bs{\tau}_t$ as described in Eq.~\eqref{simpoe:eq:pd}. It is undesirable because (i) manual parameter tuning requires significant domain knowledge and (ii) even carefully designed parameters can be suboptimal. The difficulty, here, lies in balancing the ratio between $\bs{k}_\texttt{p}$ and $\bs{k}_\texttt{d}$. Large ratios can lead to unstable and jittery motion while small values can result in motion that is too smooth and lags behind ground truth.

Motivated by this problem, we propose meta-PD control, a method that allows the policy to dynamically adjust $\bs{k}_\texttt{p}$ and $\bs{k}_\texttt{d}$ based on the state of the character. Specifically, given some initial values $\bs{k}'_\texttt{p}$ and $\bs{k}'_\texttt{d}$, the policy outputs $\lambda_\texttt{p}$ and $\lambda_\texttt{d}$ as additional elements of the action $\bs{a}_t$ that act to scale $\bs{k}'_\texttt{p}$ and $\bs{k}'_\texttt{d}$. Moreover, we take this idea one step further and let the policy output two sequences of scales $\bs{\lambda}_t^\texttt{p}=(\lambda_{t1}^\texttt{p}, \ldots, \lambda_{tm}^\texttt{p})$ and $\bs{\lambda}_t^\texttt{d}=(\lambda_{t1}^\texttt{d}, \ldots, \lambda_{tm}^\texttt{d})$ where $m=15$ corresponds to the number of PD controller (simulation) steps during a policy step. The PD controller parameters $\bs{k}_\texttt{p}$ and $\bs{k}_\texttt{d}$ at the $j$-th step of the 15 PD controller steps are then computed as follows:
\begin{equation}
    \bs{k}_\texttt{p} = \lambda_{tj}^\texttt{p}\bs{k}'_\texttt{p}, \quad \bs{k}_\texttt{d} = \lambda_{tj}^\texttt{d}\bs{k}'_\texttt{d}\,.
\end{equation}
Instead of using fixed $\bs{k}_\texttt{p}$ and $\bs{k}_\texttt{d}$, meta-PD control allows the policy to plan the scaling of $\bs{k}_\texttt{p}$ and $\bs{k}_\texttt{d}$ through the 15 PD controller steps to have more granular control over the torques produced by the PD controllers, which in turn enables a finer level of character control. With meta-PD control, the action $\bs{a}_t$ is now defined as $\bs{a}_t \triangleq (\bs{u}_t, \bs{\eta}_t, \bs{\lambda}_t^\texttt{p}, \bs{\lambda}_t^\texttt{d})$.

Since both the PD gains and target joint angles are output by the policy, it may seem that meta-PD control is similar to directly outputting joint torques at a finer time scale. However, the two approaches are still fundamentally different, because there are inherent constraints on the PD gains (e.g., cannot be negative) and the target joint angles (cannot be too far away from the current joint angles), which constrains the joint torques in a physically-plausible way.

\section{Experiments}

\paragraph{Datasets.}
We perform experiments on two large-scale human motion datasets. The first dataset is Human3.6M~\cite{ionescu2013human3}, which includes 7 annotated subjects captured at 50Hz and a total of 1.5 million training images. Following prior work~\cite{kolotouros2019learning,kocabas2020vibe,moon2020i2l}, we train our model on 5 subjects (S1, S5, S6, S7, S8) and test on the other 2 subjects (S9, S11). We subsample the dataset to 25Hz for both training and testing.
The second dataset we use is an in-house human motion dataset that also contains \emph{detailed finger motion}. It consists of 3 subjects captured at 30Hz performing various actions from free body motions to natural conversations. There are around 335k training frames and 87k test frames. Our in-house dataset has complex skeletons with twice more joints than the SMPL model, including fingers. The body shape variation among subjects is also greater than that of SMPL, which further evaluates the robustness of our approach.

\paragraph{Metrics.} We use both pose-based and \emph{physics-based} metrics for evaluation. To assess pose accuracy, we report mean per joint position error (MPJPE) and Procrustes-aligned mean per joint
position error (PA-MPJPE). We also use three physics-based metrics that measure jitter, foot sliding, and ground penetration, respectively. For jitter, we compute the difference in acceleration (Accel) between the predicted 3D joint and the ground-truth. For foot sliding (FS), we find body mesh vertices that contact the ground in two adjacent frames and compute their average displacement within the frames. For ground penetration (GP), we compute the average distance to the ground for mesh vertices below the ground. The units for these metrics are millimeters (mm) except for Accel (mm/frame$^2$). MPJPE, PA-MPJPE and Accel are computed in the root-centered coordinate.

\subsection{Implementation Details.}
 \paragraph{Character Models.}
 We use MuJoCo~\cite{todorov2012mujoco} as the physics simulator. For the character creation process in Sec.~\ref{simpoe:sec:character_creation}, we use VIBE~\cite{kocabas2020vibe} to recover an SMPL model for each subject in Human3.6M. Each MuJoCo character created from the SMPL model has 25 bones and 76 degrees of freedom (DoFs). For our in-house motion dataset, we use non-rigid ICP~\cite{amberg2007optimal} and linear blend skinning~\cite{kavan2007skinning} to reconstruct a skinned human mesh model for each subject. Each of these models has fingers and includes 63 bones and 114 DoFs.

\paragraph{Initialization.}
For Human3.6M, we use VIBE to provide the initial kinematic motion $\widetilde{\bs{q}}_{1:T}$. For our in-house motion dataset, since our skinned human models have more complex skeletons and meshes than the SMPL model, we develop our own kinematic pose estimator. To recover the global root position of the person, we assume the camera intrinsic parameters are calibrated and optimize the root position by minimizing the reprojection loss of 2D keypoints, similar to the kinematic initialization in~\cite{shimada2020physcap}.

\paragraph{Other Details.} The kinematic refinement unit in the policy network refines the kinematic pose $n=5$ times. To facilitate learning, we first pretrain the refinement unit with supervised learning using an MSE loss on the refined kinematic pose. The normalization layer in the policy computes the running average of the mean and variance of the input feature during training, and uses it to produce a normalized feature. Our learned policy runs at 38 FPS on a standard PC with an Intel Core i9 Processor.

\subsection{Comparison to state-of-the-art methods}
We compare SimPoE against state-of-the-art monocular 3D human pose estimation methods, including both kinematics-based (VIBE~\cite{kocabas2020vibe}, NeurGD~\cite{song2020human}) and physics-based (PhysCap~\cite{shimada2020physcap}, EgoPose~\cite{yuan2019ego}) approaches. The results of VIBE and EgoPose are obtained using their publicly released code and models. As PhysCap and NeurGD have not released their code, we directly use the reported results on Human3.6M from the PhysCap paper and implement our own version of NeurGD. Table~\ref{simpoe:table:quan} summarizes the quantitative results on Human3.6M and the in-house motion dataset. On Human3.6M, we can observe that our method, SimPoE, outperforms previous methods in pose accuracy as indicated by the smaller MPJPE and PA-MPJPE. In particular, SimPoE shows large pose accuracy improvements over state-of-the-art physics-based approaches (EgoPose~\cite{yuan2019ego} and PhysCap~\cite{shimada2020physcap}), reducing the MPJPE almost by half. For physics-based metrics (Accel, FS and GP), SimPoE also outperforms prior methods by large margins. It means that SimPoE significantly reduces the physical artifacts -- jitter (Accel), foot sliding (FS), and ground penetration (GP), which particularly deteriorate the results of kinematic methods (VIBE~\cite{kocabas2020vibe} and NeurGD~\cite{song2020human}). On the in-house motion dataset, SimPoE again outperforms previous methods in terms of both pose-based and physics-based metrics. In the table, KinPose denotes our own kinematic pose estimator used by SimPoE. We note that the large acceleration error (Accel) of EgoPose is due to the frequent falling of the character, which is a common problem in physics-based methods since the character can lose balance when performing agile motions. The learned policy of SimPoE is robust enough to stably control the character without falling, which prevents irregular accelerations.

\begin{figure*}
    \centering
    \includegraphics[width=\textwidth]{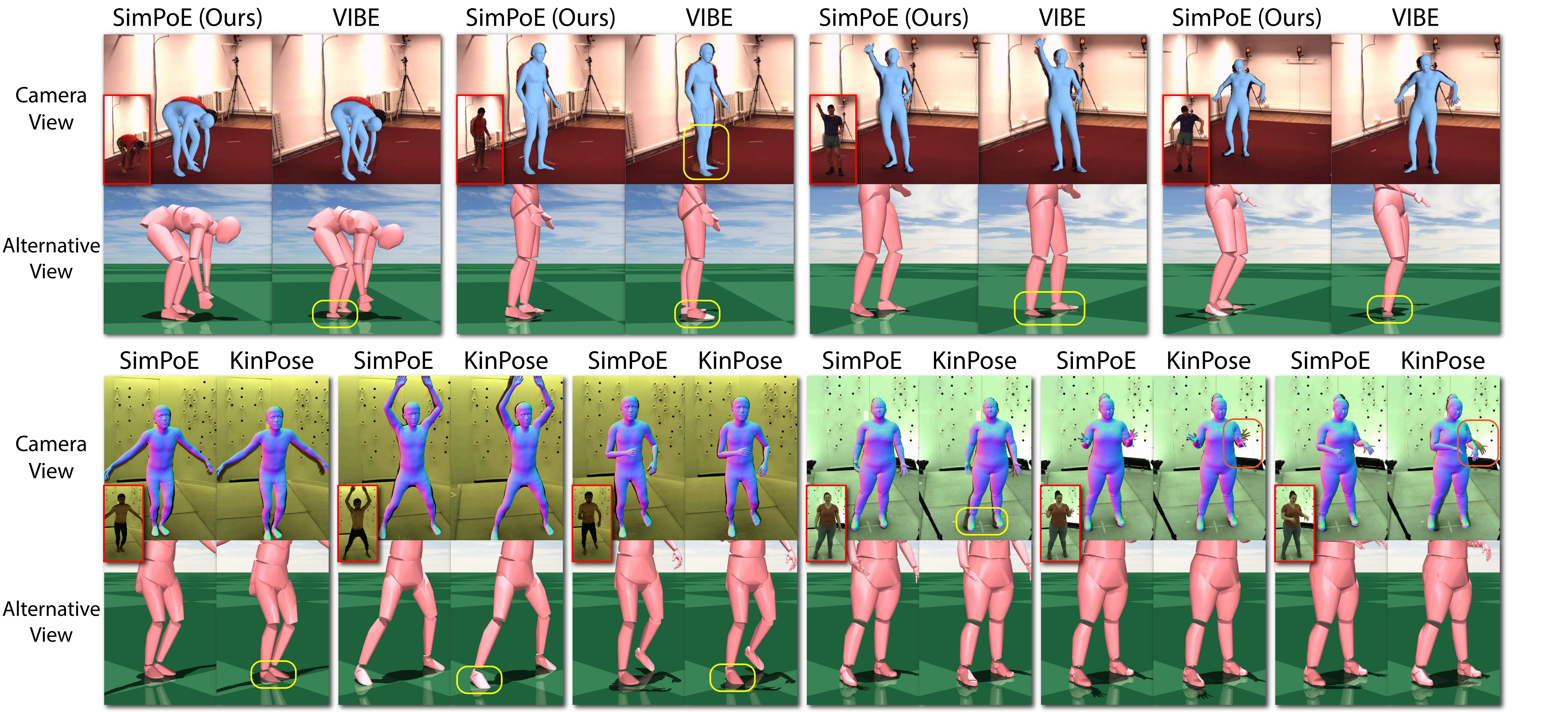}
    \caption{\textbf{Visualization} of estimated poses in the camera view and an alternative view. SimPoE estimates more accurate poses and foot contact. Pose mismatch and ground penetration are highlighted with boxes. Please see the \href{https://www.ye-yuan.com/simpoe}{supplementary video} for more comparisons.}
    \label{simpoe:fig:qual_res}
\end{figure*}

\setlength{\tabcolsep}{3pt}
\begin{table}[t]
\vspace{2mm}
\footnotesize
\centering
\begin{tabular}{@{\hskip 1mm}lcccccc@{\hskip 1mm}}
\toprule
\multicolumn{7}{c}{Human3.6M}\\
\midrule
Method & Physics  & MPJPE $\downarrow$ & PA-MPJPE $\downarrow$ & Accel $\downarrow$ & FS $\downarrow$ & GP $\downarrow$ \\ \midrule
VIBE~\cite{kocabas2020vibe} & \xmark & 61.3	& 43.1 & 15.2 & 15.1 & 12.6\\
NeurGD*~\cite{song2020human} & \xmark & 57.3 & 42.2 & 14.2 & 16.7 & 24.4\\
PhysCap~\cite{shimada2020physcap} & \cmark & 113.0 & 68.9 & - & - & - \\
EgoPose~\cite{yuan2019ego} & \cmark & 130.3 & 79.2 & 31.3 & 5.9  & 3.5 \\
SimPoE (Ours) & \cmark & \textbf{56.7} & \textbf{41.6} & \textbf{6.7} & \textbf{3.4} & \textbf{1.6}\setcounter{rownum}{0}\\
\midrule
\multicolumn{7}{c}{In-House Motion Dataset}\\
\midrule
Method & Physics  & MPJPE $\downarrow$ & PA-MPJPE $\downarrow$ & Accel $\downarrow$ & FS $\downarrow$ & GP $\downarrow$ \\ \midrule
KinPose  & \xmark & 49.7  & 40.4  & 12.8 & 6.4 & 3.9 \\
NeurGD*~\cite{song2020human} & \xmark & 36.7  & 30.9  & 16.2 & 7.7 & 3.6 \\
EgoPose~\cite{yuan2019ego} & \cmark & 202.2 & 131.4 & 32.6 & 2.2 & 0.5 \\
SimPoE (Ours) & \cmark & \textbf{26.6} & \textbf{21.2} & \textbf{8.4} & \textbf{0.5} & \textbf{0.1}\\
\bottomrule
\end{tabular}
\vspace{6mm}
\caption{Results of pose-based (MPJPE, PA-MPJPE) and physics-based (Accel, FS, GP) metrics on Human3.6M and our in-house motion dataset. Symbol ``-'' means results are not available and ``*'' means self-implementation (better results than the original paper).}
\label{simpoe:table:quan}
\end{table}

We also provide qualitative comparisons in Fig.~\ref{simpoe:fig:qual_res}, where we show the estimated poses in the camera view and the same poses rendered from an alternative view. The alternative view shows that SimPoE can estimate foot contact with the ground more accurately and without penetration. As the quality and physical plausibility of the estimated motions are best seen in videos, please refer to the \href{https://www.ye-yuan.com/simpoe}{supplementary video} for additional qualitative results and comparisons.

\subsection{Ablation Studies}

\begin{figure}
    \centering
    \includegraphics[width=0.5\textwidth]{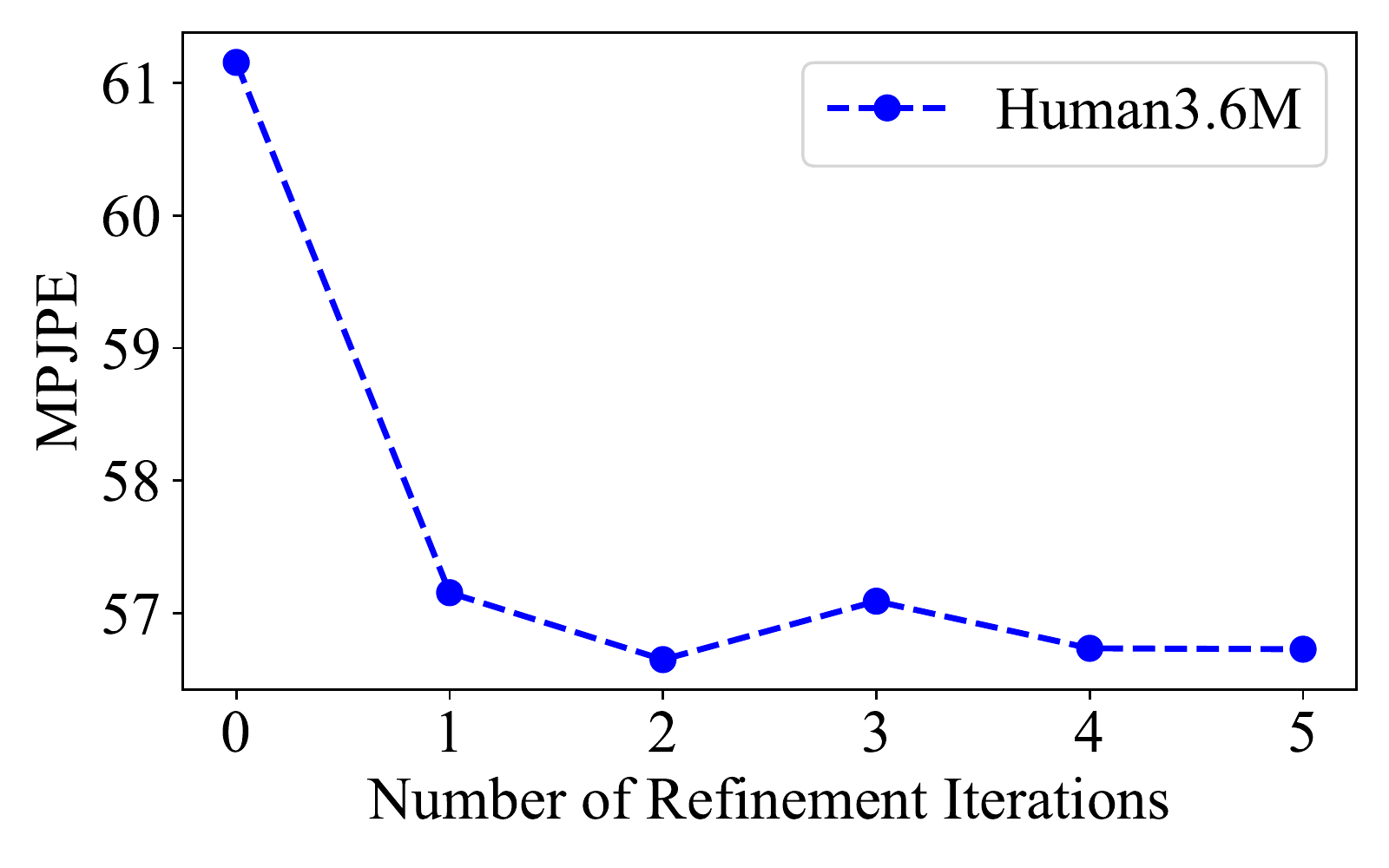}
    \vspace{2mm}
    \caption{Effect of refinement unit.}
    \label{simpoe:fig:plot}
\end{figure}

\setlength{\tabcolsep}{3pt}
\begin{table*}[t]
\footnotesize
\centering
\begin{tabular}{@{\hskip 1mm}lcccccccccccc@{\hskip 1mm}}
\toprule
\multirow{3}{*}[2pt]{Method} & \multicolumn{5}{c}{Human3.6M} & \multicolumn{5}{c}{In-House Motion Dataset} \\
\cmidrule(l{0mm}r{1mm}){2-6} \cmidrule(l{0.5mm}r{0mm}){7-11}
 & MPJPE $\downarrow$ & PA-MPJPE $\downarrow$ & Accel $\downarrow$ & FS $\downarrow$ & GP $\downarrow$ & MPJPE $\downarrow$ & PA-MPJPE $\downarrow$ & Accel $\downarrow$ & FS $\downarrow$ & GP $\downarrow$ \\ \midrule
w/o Meta-PD & 59.9 & 44.7 & \textbf{5.9} & \textbf{2.2} & \textbf{1.4} & 39.8  & 31.7  & 7.1  & \textbf{0.4} & \textbf{0.1} \\
w/o Refine & 61.2 & 43.5 & 8.0 & 3.4 & 2.0 & 47.9  & 38.9  & 9.6  & 0.6 & \textbf{0.1} \\
w/o ResAngle & 68.7 & 51.0 & 6.4 & 4.1 & 2.1 & 193.4 & 147.6 & \textbf{6.5}  & 0.9 & 0.3 \\
w/o ResForce & 115.2 & 65.1 & 23.5 & 6.1 & 3.2 & 48.4  & 31.3  & 12.5 & 0.9 & 0.3\\
w/o FeatLayer & 81.4 & 47.6	& 9.3 & 5.0	& 1.8 & 36.9  & 27.5  & 9.5  & 0.6 & \textbf{0.1} \\
SimPoE (Ours) & \textbf{56.7} & \textbf{41.6} & 6.7 & 3.4 & 1.6 & \textbf{26.6} & \textbf{21.2} & 8.4 & 0.5 & \textbf{0.1}\\
\bottomrule
\end{tabular}
\vspace{5mm}
\caption{Ablation studies on Human3.6M.}
\label{simpoe:table:abl}
\end{table*}

To further validate our proposed approach, we conduct extensive ablation studies to investigate the contribution of each proposed component to the performance. Table~\ref{simpoe:table:abl} summarizes the results where we train different variants of SimPoE by removing a single component each time. First, we can observe that both meta-PD control and the kinematic refinement unit contribute to better pose accuracy as indicated by the corresponding ablations (w/o Meta-PD and w/o Refine). Second, the ablation (w/o ResAngle) shows that it is important to have the residual connection in the policy network for producing the mean PD controller target angles $\overline{\bs{u}}_t$. Next, the residual forces $\bs{\eta}_t$ we use in action~$\bs{a}_t$ are also indispensable as demonstrated by the drop in performance of the variant (w/o ResForce). Without the residual forces, the policy is not robust and the character often falls down as indicated by the large acceleration error (Accel). Finally, it is evident from the ablation (w/o FeatLayer) that our feature extraction layer in the policy is also instrumental, because it extracts informative features of both the current frame and next frame to learn control that advances the character to the next pose. We also perform ablations to investigate how the number of refinement iterations in the policy affects pose accuracy. As shown in Fig.~\ref{simpoe:fig:plot}, the performance gain saturates around 5 refinement iterations.

\section{Discussion and Future Work}
In this work, we demonstrate that modeling both kinematics and dynamics improves the accuracy and physical plausibility of 3D human pose estimation from monocular video. Our approach, \mbox{SimPoE}, unifies kinematics and dynamics by integrating image-based kinematic inference and physics-based character control into a joint reinforcement learning framework. It runs in real-time, is compatible with advanced physics simulators, and addresses several drawbacks of prior physics-based approaches.

However, due to its physics-based formulation, \mbox{SimPoE} depends on 3D scene modeling to enforce contact constraints during motion estimation. This hinders direct evaluation on in-the-wild datasets, such as 3DPW~\cite{von2018recovering}, which includes motions such as climbing stairs or even trees. Future work may include integration of video-based 3D scene reconstruction to address this limitation.
\chapter{Global Occlusion-Aware Human Pose Estimation via Behavior Generation}
\label{chap:glamr}

\vspace{-5mm}
\section{Introduction}
\label{glamr:sec:intro}

\begin{figure*}[t]
    \centering
    \includegraphics[width=\linewidth]{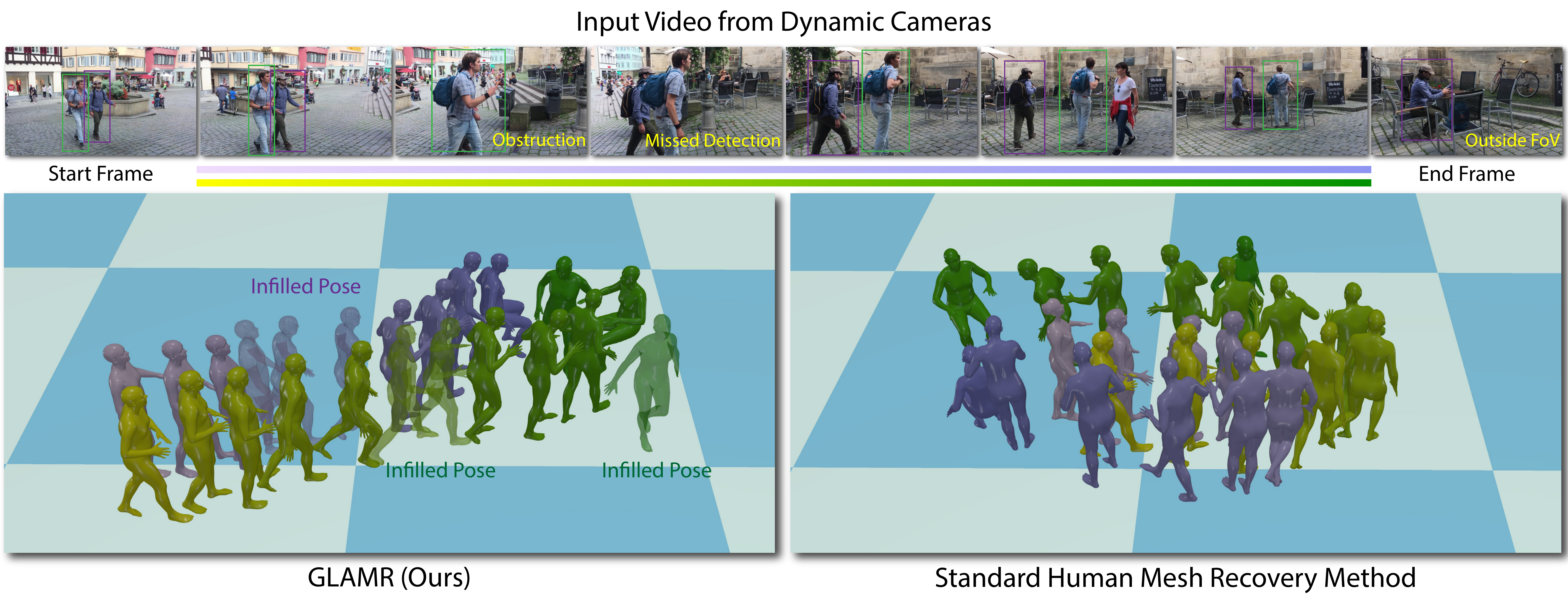}
    \vspace{-2mm}
    \caption{GLAMR (\textbf{Left}) recovers human meshes in consistent \emph{global} coordinates and \emph{infills missing poses} (transparent) due to various occlusions (obstruction, missed detection, outside field of view), while standard human mesh recovery methods (\textbf{Right}) fail to do so.}
    \label{glamr:fig:teaser}
\end{figure*}

Recovering fine-grained 3D human meshes from monocular videos is essential for understanding human behaviors and interactions, which can be the cornerstone for numerous applications including virtual or augmented reality, assistive living, autonomous driving, etc. Many of these applications use dynamic cameras to capture human behaviors yet also require estimating human motions in global coordinates consistent with their surroundings. For instance, assistive robots and autonomous vehicles need a holistic understanding of human behaviors and interactions in the world to safely plan their actions even when they are moving. Therefore, our goal in this paper is to tackle the important task of recovering global human meshes from monocular videos captured by dynamic cameras.

However, this task is highly challenging for two main reasons. First, dynamic cameras make it difficult to estimate human motions in \emph{consistent global coordinates}. Existing human mesh recovery methods estimate human meshes in the camera coordinates~\cite{moon2019camera,zhen2020smap} or even in the root-relative coordinates~\cite{kocabas2020vibe,moon2020i2l}. Hence, they can only recover global human meshes from dynamic cameras by using SLAM to estimate camera poses~\cite{liu20204d}. However, SLAM can often fail for in-the-wild videos due to moving and dynamic objects. It also has the problem of scale ambiguity, which often leads to camera poses that are inconsistent with the human motions. Second, videos captured by dynamic cameras often contain \emph{severe and long-term occlusions} of humans, which can be caused by missed detection, complete obstruction by objects and other people, or the person going outside the camera's field of view (FoV). These occlusions pose serious challenges to standard human mesh recovery methods, which rely on detections or visible parts to estimate human meshes. Only a few works have attempted to tackle the occlusion problem in human mesh recovery~\cite{jiang2020coherent,fieraru2020three}. However, these methods can only address partial occlusions of a person and fail to handle severe occlusions when the person is completely invisible for an extended period of time.

To tackle the above challenges, we propose Global Occlusion-Aware Human Mesh Recovery (GLAMR), which can handle severe occlusions and estimate human meshes in consistent global coordinates -- even for videos recorded with dynamic cameras. We start by using off-the-shelf methods (\eg, KAMA~\cite{iqbal2021kama} or SPEC~\cite{Kocabas_SPEC_2021}) to estimate the shape and pose sequences (motions) of visible people in the camera coordinates. These methods also rely on multi-object tracking and re-identification, which provide occlusion information, and the motion of occluded frames is not estimated. To tackle potentially severe occlusions, we propose a deep generative motion infiller that autoregressively infills the local body motions of occluded people based on visible motions. The motion infiller leverages human dynamics learned from a large motion database, AMASS~\cite{AMASS:ICCV:2019}. Next, to obtain global motions, we propose a global trajectory predictor that can generate global human trajectories based on local body motions. It is motivated by the observation that the global root trajectory of a person is highly correlated with the local body movements. Finally, using the predicted trajectories as anchors to constrain the solution space, we further propose a global optimization framework that jointly optimizes the global motions and camera poses to match the video evidence such as 2D keypoints.

The contributions of this paper are as follows: \textbf{(1)} We propose the first approach to address long-term occlusions and estimate global 3D human pose and shape from videos captured by dynamic cameras; \textbf{(2)} We propose a novel generative Transformer-based motion infiller that autoregressively infills long-term missing motions, which considerably outperforms state-of-the-art motion infilling methods; \textbf{(3)} We propose a method to generate global human trajectories from local body motions and use the generated trajectories as anchors to constrain global motion and camera optimization; \textbf{(4)} Extensive experiments on challenging indoor and in-the-wild datasets demonstrate that our approach outperforms prior state-of-the-art methods significantly in tackling occlusions and estimating global human meshes.

\section{Related Work}
\label{glamr:sec:related_work}
\paragraph{Camera-Relative Pose Estimation.}
3D human mesh recovery from RGB images or videos is an ill-posed problem due to the depth ambiguity. Most  existing methods simplify the problem by estimating human poses relative to the pelvis (root) of the human body~\cite{Akhter:CVPR:2015, bogo2016keep, lassner2017unite, hmrKanazawa18, pavlakos2018learning, guler2019holo, kolotouros2019spinmini, pavlakos2019texture, Rong_2019_ICCV, kolotouros2019convolutional, choutas2020expose, zanfir2020weakly, sun2019human, joo2021eft, choi2020pose, kundu2020mesh, pavlakos2019expressive, xu2019denserac, monototalcapture2019, song2020human, zhang2020object, zhou2021monocular, moon2020i2l, lin2021end, Mueller:CVPR:21, kolotouros2021prohmr, Zhang_2021_ICCV, Sun_2021_ICCV, kanazawa2018learning, kocabas2020vibe, luo20203d, choi2020beyond, rempe2021humor}. These methods assume an orthographic camera projection model and neglect the absolute 3D translation of the person \wrt the camera. To address the lack of translation, recent methods start to estimate human meshes in the camera coordinates~\cite{zanfir2018monocular, jiang2020coherent, Zanfir_2021_ICCV, ICG, Zhang_2021_CVPR, Xie_2021_ICCV, shimada2020physcap, liu20204d, li2020hybrik, iqbal2021kama, reddy2021tesstrack}. Several approaches recover the absolute translation of the person using an optimization framework~\cite{mono20173dhp, mehta2017vnect, XNect_SIGGRAPH2020,zanfir2018deep, rogez2017lcr}. A few methods exploit various scene constraints during the optimization process to improve depth prediction~\cite{zanfir2018monocular,Weng_2021_CVPR}. Alternatively, recent approaches use physics-based constraints to ensure the physical plausibility of the estimated poses~\cite{shimada2020physcap, Xie_2021_ICCV, GraviCap2021, yuan2021simpoe, isogawa2020optical}. Iqbal~\etal~\cite{iqbal2020learning} exploit a limb-length constraint to recover the absolute translation of the person using a 2.5D representation. Some approaches approximate the depth of the person using the bounding box size~\cite{jiang2020coherent, moon2019camera, Zhang_2021_CVPR}. HybrIK~\cite{li2020hybrik} and KAMA~\cite{iqbal2021kama} employ inverse kinematics to estimate human meshes with absolute translations in the camera coordinates. Several methods directly predict the absolute depth of each person using a heatmap representation~\cite{Fabbri_2020_CVPR,zhen2020smap}. Recently, SPEC~\cite{Kocabas_SPEC_2021} learns to predict the camera parameters from the image, which are used for absolute pose regression in the camera coordinates. THUNDR~\cite{Zanfir_2021_ICCV} also adopts a similar strategy but uses known camera parameters. While these methods show impressive results, they cannot estimate global human motions from videos captured by dynamic cameras. In contrast, our approach can recover human meshes in consistent global coordinates for dynamic cameras and handle severe and long-term occlusions.

\paragraph{Global Pose Estimation.}
Most existing methods that estimate 3D poses in world coordinates rely on calibrated, synchronized, and static multi-view capture setups~\cite{belagiannis20143d,joo2018total,reddy2021tesstrack, multiviewpose, zhang20204d, dong2021shape, zhang2021lightweight, zheng2021deepmulticap, huang2021dynamic, dong2021fastpami}. Huang~\etal~\cite{huang2021dynamic} use uncalibrated cameras but still assume time synchronization and static camera setups. Hasler~\etal~\cite{hasler2009markerless} handle unsynchronized moving cameras but assume multi-view input and rely on audio stream for synchronization.  More recently, Dong~\etal~\cite{dong2020motion} propose to recover 3D poses from unaligned internet videos of different actors performing the same activity from unknown cameras. However, they assume that multiple viewpoints of the same pose are available in the videos. Different from these methods, our approach estimates human meshes in global coordinates from \emph{monocular} videos recorded with dynamic cameras. Several methods rely on additional IMU sensors or pre-scanned environments to recover global human motions~\cite{von2018recovering,hps2021Vladmir}, which is unpractical for large-scale adoption. Recently, another line of work starts to focus on estimating accurate human-scene interaction~\cite{hassan2019resolving,luo2021dynamics,yi2022human,huang2022cap}. Liu~\etal~\cite{liu20204d} first obtain the camera poses and dense reconstruction of the scene from dynamic cameras using a SLAM algorithm, COLMAP~\cite{schonberger2016structure}. The camera poses are used for camera-to-world transformation, while the reconstructed scene is used to encourage human-scene contacts. However, SLAM can often fail for the in-the-wild videos and is prone to error propagation. In contrast, our approach does not require SLAM but instead uses global trajectory prediction to constrain the joint reconstruction of human motions and camera poses. Additionally, our approach can also handle severe and long-term occlusions common in dynamic camera setups.

\vspace{-5mm}
\paragraph{Occlusion-Aware Pose Estimation.}
Most existing human pose estimation methods assume the person is fully visible in the images and are not robust to strong occlusions. Only a few methods address the occlusion problem in pose estimation~\cite{zhang2020object,rockwell2020fullbody,fieraru2020three,rempe2021humor,pare2021kocabas}. While these methods show impressive results under partial occlusions, they do not address severe and long-term occlusions when people are completely obstructed or outside the camera's FoV for a long time. In contrast, our approach leverages deep generative human motion models to tackle severe and long-term occlusions.

\vspace{-2mm}
\paragraph{Human Motion Modeling.}
Extensive research has studied 3D human dynamics for various tasks including motion prediction and synthesis~\cite{fragkiadaki2015recurrent,jain2016structural,li2017auto,martinez2017human,villegas2017learning,pavllo2018quaternet,aksan2019structured,gopalakrishnan2019neural,yan2018mt,barsoum2018hp,yuan2019diverse,yuan2020dlow,yuan2020residual,cao2020long,petrovich2021action,hassan2021stochastic}. Recent human pose estimation methods start to leverage learned human dynamics models to improve the accuracy of estimated motions~\cite{kocabas2020vibe,rempe2021humor,zhang2021learning}. Several motion infilling approaches are also proposed to generate complete motions from partially observed motions~\cite{hernandez2019human,kaufmann2020convolutional,harvey2020robust,khurana2021detecting}. Additionally, recent work on motion capture shows that global human translations can be predicted from 3D local joint positions~\cite{schreiner2021global}. In contrast to prior work, our trajectory predictor does not require GT root orientations but can predict both global root translations and orientations. Furthermore, we also propose a novel generative autoregressive motion infiller that can use noisy poses as input instead of high-quality GT poses, and we demonstrate its effectiveness in tackling long-term occlusions in human pose estimation.
    
\section{Method}
\label{glamr:sec:method}

\begin{figure*}[t]
    \centering
    \includegraphics[width=\linewidth]{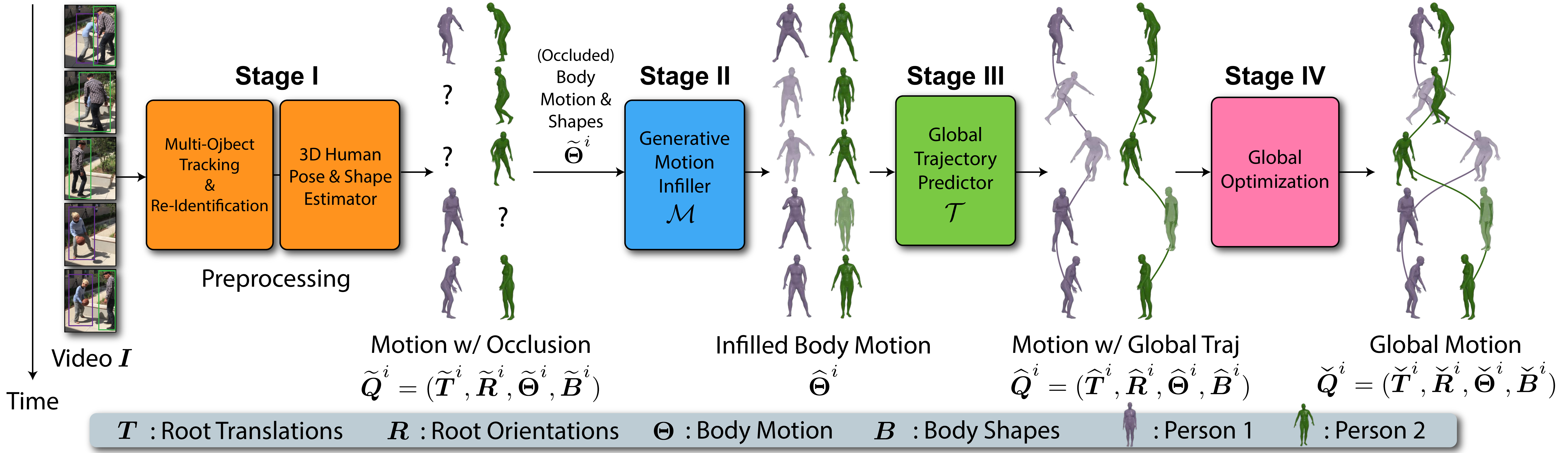}
    \caption{\textbf{Overview} of GLAMR. In \textbf{Stage I}, we preprocess the video with multi-object tracking, re-identification and human mesh recovery to extract each person's occluded motion $\bs{\wt{Q}}^i$ in the camera coordinates. In \textbf{Stage II}, we propose a generative motion infiller to infill the occluded body motion $\bs{\wt{\Theta}}^i$ to produce occlusion-free body motion $\bs{\wh{\Theta}}^i$. In \textbf{Stage III}, we propose a global trajectory predictor that uses the infilled body motion $\bs{\wh{\Theta}}^i$ to generate the global trajectory $(\bs{\wh{T}}^i,\bs{\wh{R}}^i)$ of each person and obtain their global motion $\bs{\wh{Q}}^i$. In \textbf{Stage IV}, we jointly optimize the global trajectories of all people and the camera parameters to produce global motions $\bs{\wc{Q}}^i$ consistent with the video.}
    \label{glamr:fig:overview}
    \vspace{-2mm}
\end{figure*}

The input to our framework is a video $\bs{I} = (\bs{I}_1, \ldots, \bs{I}_T)$ with $T$ frames, which is captured by a \emph{dynamic camera}, \ie, the camera poses can change every frame. Our goal is to estimate the global motion (pose sequence) $\{\bs{Q}^i\}_{i=1}^N$ of the $N$ people in the video in a \emph{consistent global coordinate} system. The global motion $\bs{Q}^i= (\bs{T}^i, \bs{R}^i, \bs{\Theta}^i, \bs{B}^i)$ for person $i$ consists of the root translations $\bs{T}^i = (\bs{\tau}^{i}_{s_i}, \ldots, \bs{\tau}^{i}_{e_i})$, root rotations $\bs{R}^i = (\bs{\rot}^{i}_{s_i}, \ldots, \bs{\rot}^{i}_{e_i})$, as well as the body motion $\bs{\Theta}^i = (\bs{\theta}^{i}_{s_i}, \ldots, \bs{\theta}^{i}_{e_i})$ and shapes $\bs{B}^i = (\bs{\beta}^{i}_{s_i}, \ldots, \bs{\beta}^{i}_{e_i})$, where the motion spans from the the first frame $s_i$ to the last frame $e_i$, when the person $i$ is relevant in the video.
In particular, each body pose $\bs{\theta}^{i}_t \in \mathbb{R}^{23 \times 3}$ and shape $\bs{\beta}^{i}_t \in \mathbb{R}^{10}$ corresponds to the pose parameters (excluding root rotation) and shape parameters of the SMPL model~\cite{loper2015smpl}. Using the root translation $\bs{\tau} \in \mathbb{R}^{3}$ and (axis-angle) rotation $\bs{\rot}\in \mathbb{R}^{3}$, SMPL represents a human body mesh with a linear function $\mathcal{S}(\bs{\tau}, \bs{\rot}, \bs{\theta}, \bs{\beta})$ that maps a global pose $\bs{q} = (\bs{\tau}, \bs{\rot}, \bs{\theta}, \bs{\beta})$ to an articulated triangle mesh $\bs{\Phi} \in \mathbb{R}^{K\times 3}$ with $K=6980$ vertices. We can therefore recover the global mesh sequence for each person from their global motion $\bs{Q}^i$ via SMPL.

As outlined in Fig.~\ref{glamr:fig:overview}, our framework consists of four stages. In \textbf{Stage I}, we first use multi-object tracking (MOT) and re-identification algorithms to obtain the bounding box sequence of each person, which is input to a human mesh recovery method (\eg, KAMA~\cite{iqbal2021kama} or SPEC~\cite{Kocabas_SPEC_2021}) to extract the motion $\bs{\wt{Q}}^i$ of each person (including translation) in the camera coordinates. The motion $\bs{\wt{Q}}^i$ may be incomplete due to various occlusions (\eg, obstruction, missed detection, going outside FoV), where bounding boxes from MOT are missing for some frames. In \textbf{Stage II} (Sec.~\ref{glamr:sec:motion_infill}), we propose a generative motion infiller to tackle the occlusions in the estimated body motion $\bs{\wt{\Theta}}^i$ and produce occlusion-free body motion $\bs{\wh{\Theta}}^i$. In \textbf{Stage III} (Sec.~\ref{glamr:sec:traj_pred}), we propose a global trajectory predictor that uses the infilled body motion $\bs{\wh{\Theta}}^i$ to generate the global trajectory (root translations and rotations) of each person and obtain their global motion $\bs{\wh{Q}}^i$. In \textbf{Stage IV} (Sec.~\ref{glamr:sec:global_opt}), we jointly optimize the global trajectories of all people and the camera parameters to produce global motions $\bs{\wc{Q}}^i$ consistent with the video evidence.

\subsection{Generative Motion Infiller}
\label{glamr:sec:motion_infill}
The task of the generative motion infiller $\mathcal{M}$ is to infill the occluded body motion $\bs{\wt{\Theta}}^i$ of each person to produce occlusion-free body motion $\bs{\wh{\Theta}}^i$. Here, we do not use the motion infiller $\mathcal{M}$ to infill other components in the estimated motion $\bs{\wh{Q}}^i$, \ie, root trajectory ($\bs{\wt{T}}^i,\bs{\wt{R}}^i$) and shapes $\bs{\wt{B}}^i$. This is because it is difficult to infill the root trajectory $(\bs{\wt{T}}^i, \bs{\wt{R}}^i)$ using learned human dynamics, since it resides in the camera coordinates rather than a consistent coordinate system due to the dynamic camera. In Sec.~\ref{glamr:sec:traj_pred}, we will use the proposed global trajectory predictor to generate occlusion-free global trajectory $(\bs{\wh{T}}^i,\bs{\wh{R}}^i)$ from the infilled body motion $\bs{\wh{\Theta}}^i$. The trajectory $(\bs{\wt{T}}^i, \bs{\wt{R}}^i)$ from the pose estimator is not discarded and will be used in the global optimization (Sec.~\ref{glamr:sec:global_opt}). We use linear interpolation to produce occlusion-free shapes $\bs{\wh{B}}^i$, which can be time-varying to be compatible with per-frame pose estimators such as KAMA.

Given a general occluded human body motion $\bs{\wt{\Theta}} = (\bs{\wt{\theta}}_1, \ldots, \bs{\wt{\theta}}_h)$ of $h$ frames and its visibility mask $\bs{V} = (V_1, \ldots, V_{h})$ as input, the motion infiller $\mathcal{M}$ outputs a complete occlusion-free motion $\bs{\wh{\Theta}} = (\bs{\wh{\theta}}_1, \ldots, \bs{\wh{\theta}}_h)$. The visibility mask $\bs{V}$ encodes the visibility of the occluded motion $\bs{\wt{\Theta}}$, where $V_{t} = 1$ if the body pose $\bs{\wt{\theta}}_t$ is visible in frame $t$ and $V_{t} = 0$ otherwise. Since the human pose for occluded frames can be highly uncertain and stochastic, we formulate the motion infiller $\mathcal{M}$ using the conditional variational autoencoder (CVAE)~\cite{kingma2013auto}:
\begin{align}
\label{glamr:eq:motion_infill}
\bs{\wh{\Theta}} = \mathcal{M} (\bs{\wt{\Theta}}, \bs{V}, \bs{z})\,,
\end{align}
where the motion infiller $\mathcal{M}$ corresponds to the CVAE decoder and $\bs{z}$ is a Gaussian latent code. We can obtain different occlusion-free motions $\bs{\wh{\Theta}}$ by varying $\bs{z}$.

\begin{figure*}[t]
    \centering
    \includegraphics[width=\linewidth]{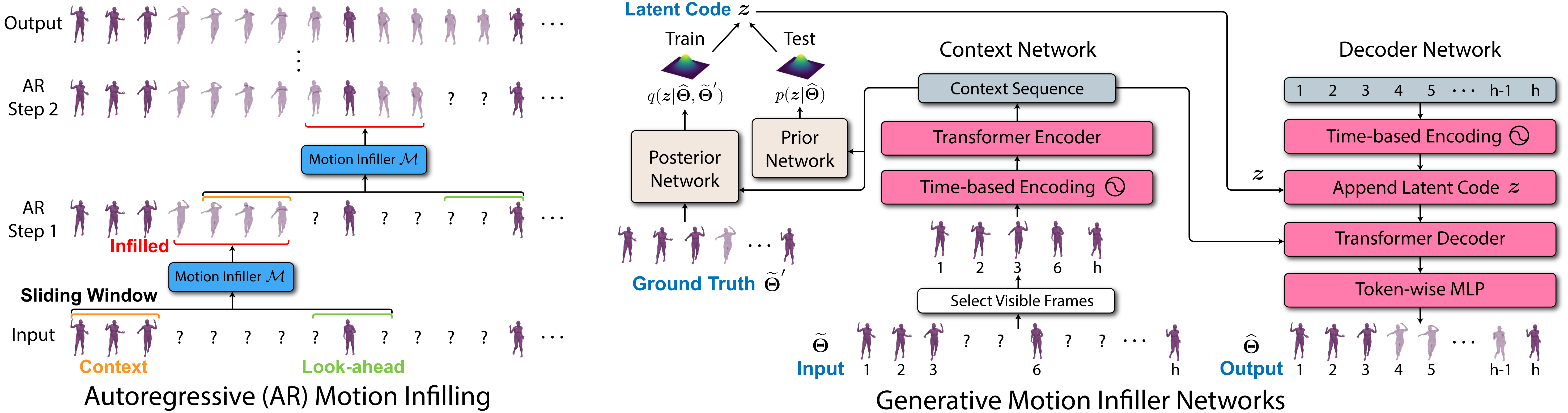}
    \caption{\textbf{Left:} We autoregressively infill the motion using a sliding window, where the first $h_\texttt{c}$ frames are already infilled to serve as \textcolor{orange}{context} and the last $h_\texttt{l}$ frames are \textcolor{grassgreen}{look-ahead} to guide the ending motion. Frames between the context and look-ahead are \textcolor{red}{infilled}. \textbf{Right:} The CVAE-based motion infiller adopts a Transformer-based seq2seq architecture, where we encode only the visible frames of occluded body motion $\bs{\wt{\Theta}}$ into a context sequence, which is used jointly with latent code $\bs{z}$ by a decoder network to generate occlusion-free motion~$\bs{\wh{\Theta}}$.}
    \label{glamr:fig:mfiller}
\end{figure*}

\paragraph{Autoregressive Motion Infilling.}
To ensure that the motion infiller $\mathcal{M}$ can handle much longer test motions than the training motions, we propose an autoregressive motion infilling process at test time as illustrated in Fig.~\ref{glamr:fig:mfiller} (Left). The key idea is to use a sliding window of $h$ frames, where we assume the first $h_\texttt{c}$ frames of motion are already occlusion-free or infilled and serve as \emph{context}, and we also use the last $h_\texttt{l}$ frames as \emph{look-ahead}. The look-ahead is essential to the motion infiller since it may contain visible poses that can guide the ending motion and avoid generating discontinuous motions. Excluding the context and look-ahead frames, only the middle $h_\texttt{o} = h - h_\texttt{c} - h_\texttt{l}$ frames of motion are infilled. We iteratively infill the motion using the sliding window and advance the window by $h_\texttt{o}$ frames every step.

\paragraph{Motion Infiller Network.}
The overall network design of the CVAE-based motion infiller is outlined in Fig.~\ref{glamr:fig:mfiller}~(Right). In particular, we employ a Transformer-based seq2seq architecture, which consists of three parts: (1) a \emph{context network} that uses a Transformer encoder to encode the visible poses from the occluded motion $\bs{\wt{\Theta}}$ into a context sequence, which serves as the condition for other networks; (2) a \emph{decoder network} that uses the latent code $\bs{z}$ and context sequence to generate occlusion-free motion $\bs{\wh{\Theta}}$ via a Transformer decoder and a multilayer perceptron (MLP); (3) \emph{prior and posterior networks} that generate the prior and posterior distributions for the latent code $\bs{z}$. In the networks, we adopt a time-based encoding that replaces the position in the original positional encoding~\cite{vaswani2017attention} with the time index. Unlike prior CNN-based methods~\cite{hernandez2019human,kaufmann2020convolutional}, our Transformer-based motion infiller does not require padding missing frames, but instead restricts its attention to visible frames to achieve effective temporal modeling.

\paragraph{Training.}
We train the motion infiller $\mathcal{M}$ using a large motion capture dataset, AMASS~\cite{AMASS:ICCV:2019}. To synthesize occluded motions $\bs{\wt{\Theta}}$, for any GT training motion $\bs{\wt{\Theta}}'$ of $h$ frames, we randomly occlude $H_\texttt{occ}$ consecutive frames of motion where $H_\texttt{occ}$ is uniformly sampled from $[H_\texttt{lb}, H_\texttt{ub}]$. Note that we do not occlude the first $h_\texttt{c}$ frames which are reserved as context. We use the standard CVAE objective to train the motion infiller $\mathcal{M}$:
\begin{align}
    L_\mathcal{M} = \sum_{t=1}^h \| \bs{\wt{\theta}}_t - \bs{\wt{\theta}}'_t \|_2^2 + L_\texttt{KL}^{\bs{z}}\,,
\end{align}
where $L_\texttt{KL}^{\bs{z}}$ is the KL divergence between the prior and posterior distributions of the CVAE latent code $\bs{z}$.

\subsection{Global Trajectory Predictor}
\label{glamr:sec:traj_pred}
After we obtain occlusion-free body motion $\bs{\wh{\Theta}}^i$ for each person using the motion infiller, a key problem still remains: the estimated trajectory $(\bs{\wt{T}}^i,\bs{\wt{R}}^i)$ of the person is still occluded and not in a consistent global coordinate system. To tackle this problem, we propose to learn a global trajectory predictor $\mathcal{T}$ that generates a person's occlusion-free global trajectory $(\bs{\wh{T}}^i, \bs{\wh{R}}^i)$ from the local body motion $\bs{\wh{\Theta}}^i$.

Given a general occlusion-free body motion ${\bs{\Theta}} = ({\bs{\theta}}_1, \ldots, {\bs{\theta}}_m)$ as input, the trajectory predictor $\mathcal{T}$ outputs its corresponding global trajectory $({\bs{T}}, {\bs{R}})$ including the root translations ${\bs{T}} = ({\bs{\tau}}_1, \ldots, {\bs{\tau}}_m)$ and rotations ${\bs{R}} = ({\bs{\rot}}_1, \ldots, {\bs{\rot}}_m)$. To address any potential ambiguity in the global trajectory, we also formulate the global trajectory predictor using the CVAE:
\begin{align}
\label{glamr:eq:traj_pred1}
\bs{{\Psi}} &= \mathcal{T} (\bs{{\Theta}}, \bs{v})\,, \\
\label{glamr:eq:traj_pred2}
(\bs{{T}}, \bs{{R}}) &= \texttt{EgoToGlobal}(\bs{{\Psi}})\,,
\end{align}
where the global trajectory predictor $\mathcal{T}$ corresponds to the CVAE decoder and $\bs{v}$ is the latent code for the CVAE. In Eq.~\eqref{glamr:eq:traj_pred1}, the immediate output of the global trajectory predictor $\mathcal{T}$ is an egocentric trajectory $\bs{{\Psi}} = (\bs{{\psi}}_1, \ldots, \bs{{\psi}}_m)$, which by design can be converted to a global trajectory $(\bs{{T}}, \bs{{R}})$ using a conversion function \texttt{EgoToGlobal}.

\paragraph{Egocentric Trajectory Representation.}
The egocentric trajectory $\bs{{\Psi}}$ is just an alternative representation of the global trajectory $(\bs{{T}}, \bs{{R}})$. It converts the global trajectory into relative local differences and represents rotations and translations in the heading coordinates ($y$-axis aligned with the heading, \ie, the person's facing direction). In this way, the egocentric trajectory representation is invariant of the absolute $xy$ translation and heading. It is more suitable for the prediction of long trajectories, since the network only needs to output the local trajectory change of every frame instead of the potentially large global trajectory offset.

The conversion from the global trajectory to the egocentric trajectory is given by another function: $\bs{{\Psi}} = \texttt{GlobalToEgo}(\bs{{T}}, \bs{{R}})$, which is the inverse of the function \texttt{EgoToGlobal}. In particular, the egocentric trajectory $\bs{{\psi}}_t = (\delta x_t, \delta y_t, z_t, \delta \phi_t, \bs{\eta}_t)$ at time $t$ is computed as:
\begin{align}
\label{glamr:eq:ego_traj1}
(\delta x_t, \delta y_t) &= \texttt{ToHeading}(\bs{\tau}_t^{xy} - \bs{\tau}_{t-1}^{xy})\,,\\
\label{glamr:eq:ego_traj2}
z_t &= \bs{\tau}_t^z, \quad \delta \phi_t = \bs{\rot}_t^{\phi} - \bs{\rot}_{t-1}^{\phi}\,,\\
\label{glamr:eq:ego_traj3}
\bs{\eta}_t &= \texttt{ToHeading}(\bs{\rot}_t)\,,
\end{align}
where $\bs{\tau}_t^{xy}$ is the $xy$ component of the translation $\bs{\tau}_t$, $\bs{\tau}_t^{z}$ is the $z$ component (height) of $\bs{\tau}_t$, $\bs{\rot}_t^{\phi}$ is the heading angle of the rotation $\bs{\rot}_t$, \texttt{ToHeading} is a function that converts translations or rotations to the heading coordinates defined by the heading $\bs{\rot}_t^{\phi}$, and $\bs{\eta}_t$ is the local rotation. As an exception, $(\delta x_0, \delta y_0)$ and $\delta \phi_0$ are used to store the initial $xy$ translation $\bs{\tau}_0^{xy}$ and heading $\bs{\tau}_0^{\phi}$. These initial values are set to the GT during training and arbitrary values during inference (as the trajectory can start from any position and heading). The inverse process of Eq. \eqref{glamr:eq:ego_traj1}-\eqref{glamr:eq:ego_traj3} defines the inverse conversion \texttt{EgoToGlobal} used in Eq.~\eqref{glamr:eq:traj_pred2}, which accumulates the egocentric trajectory to obtain the global trajectory. To correct potential drifts in the trajectory, in Sec.~\ref{glamr:sec:global_opt}, we will optimize the global trajectory of each person to match the video evidence, which also solves the trajectory's starting point $(\delta x_0, \delta y_0, \delta \phi_0)$.

\paragraph{Network and Training.}
The trajectory predictor adopts a similar network design as the motion infiller with one main difference:  we use LSTMs for temporal modeling instead of Transformers since the output of each frame is the local trajectory change in our egocentric trajectory representation, which mainly depends on the body motion of nearby frames and does not require long-range temporal modeling. We will show in Sec.~\ref{glamr:sec:eval_component} that the egocentric trajectory and use of LSTMs instead of Transformers are crucial for accurate trajectory prediction. We use the standard CVAE objective to train the trajectory predictor $\mathcal{T}$:
\begin{align}
    L_\mathcal{T} = \sum_{t=1}^m \left(\| \bs{{\tau}}_t - \bs{{\tau}}'_t \|_2^2 + \| \bs{\rot}_t \ominus \bs{\rot}'_t \|_a^2\right) + L_\texttt{KL}^{\bs{v}}\,,
\end{align}
where $\bs{{\tau}}'_t$ and $\bs{\rot}'_t$ denote the GT translation and rotation, $\ominus$ computes the relative rotation, $\|\cdot\|_a$ computes the rotation angle, and $L_\texttt{KL}^{\bs{v}}$ is the KL divergence between the prior and posterior distributions of the CVAE latent code $\bs{v}$. We again use AMASS~\cite{AMASS:ICCV:2019} to train the trajectory predictor $\mathcal{T}$.

\subsection{Global Optimization}
\label{glamr:sec:global_opt}

After using the generative motion infiller and global trajectory predictor, we have obtained an occlusion-free global motion $\wh{\bs{Q}}^i = (\wh{\bs{T}}^i, \wh{\bs{R}}^i, \wh{\bs{\Theta}}^i, \wh{\bs{B}}^i)$ for each person in the video. However, the global trajectory predictor generates trajectories for each person independently, which may not be consistent with the video evidence. To tackle this problem, we propose a global optimization process that jointly optimizes the global trajectories of all people and the extrinsic camera parameters to match the video evidence such as 2D keypoints. The final output of the global optimization and our framework is $\wc{\bs{Q}}^i = (\wc{\bs{T}}^i, \wc{\bs{R}}^i, \wc{\bs{\Theta}}^i, \wc{\bs{B}}^i)$ where $(\wc{\bs{\Theta}}^i, \wc{\bs{B}}^i) = (\wh{\bs{\Theta}}^i, \wh{\bs{B}}^i)$, \ie, we directly use the occlusion-free body motion and shapes from the previous stages.

\paragraph{Optimization Variables.}
The first set of variables we optimize is the egocentric representation $\{\wc{\bs{\Psi}}^i\}_{i=1}^N$ of the global trajectories $\{(\wc{\bs{T}}^i, \wc{\bs{R}}^i)\}_{i=1}^N$. We adopt the egocentric representation since it allows corrections of the translation and heading at one frame to propagate to all future frames. Therefore, it enables optimizing the trajectories of occluded frames since they will impact future visible frames. We will empirically demonstrate its effectiveness in Sec.~\ref{glamr:sec:eval_component}.

The second set of optimization variables is the extrinsic camera parameters $\bs{C} = (\bs{C}_1, \ldots, \bs{C}_T)$ where $\bs{C}_t \in \mathbb{R}^{4\times 4}$ is the camera extrinsic matrix at frame $t$ of the video.

\paragraph{Energy Function.}
The energy function we aim to minimize is defined as
\begin{equation}
\begin{aligned}
    E(\{\wc{\bs{\Psi}}^i\}_{i=1}^N, \bs{C}) &= \lambda_\texttt{2D} E_\texttt{2D} + \lambda_\texttt{traj} E_\texttt{traj} \\
    & \hspace{-4mm} + \lambda_\texttt{reg} E_\texttt{reg} + \lambda_\texttt{cam} E_\texttt{cam} + \lambda_\texttt{pen} E_\texttt{pen}\,,
\end{aligned}
\end{equation}
where we use five energy terms with their corresponding coefficients $\lambda_\texttt{2D},\lambda_\texttt{traj},\lambda_\texttt{reg},\lambda_\texttt{cam},\lambda_\texttt{pen}$.

The first term $E_\texttt{2D}$ measures the error between  the 2D projection $\wc{\bs{x}}_t^i$ of the optimized 3D keypoints $\wc{\bs{X}}_t^i \in \mathbb{R}^{J \times 3}$ and the estimated 2D keypoints $\wt{\bs{x}}_t^i$ from a keypoint detector: 
\begin{align}
    E_\texttt{2D} = \frac{1}{NTJ}\sum_{i=1}^N & \sum_{t=1}^T V_t^i \|\wc{\bs{x}}_t^i - \wt{\bs{x}}_t^i\|_F^2\,, \\
    \wc{\bs{x}}_t^i = \Pi \left(\wc{\bs{X}}_t^i, \bs{C}_t, \bs{K} \right), &\quad \wc{\bs{X}}_t^i = \mathcal{J}(\wc{\bs{\tau}}_t^i, \wc{\bs{\rot}}_t^i, \wc{\bs{\theta}}_t^i, \wc{\bs{\beta}}_t^i)
\end{align}
where $V_t^i$ is person $i$'s visibility at frame $t$, $\Pi$ is the camera projection with extrinsics $\bs{C}_t$ and approximated intrinsics $\bs{K}$, and $\wc{\bs{X}}_t^i$ is computed using the SMPL joint function $\mathcal{J}$ from the optimized global pose $\wc{\bs{q}}_t^i = (\wc{\bs{\tau}}_t^i, \wc{\bs{\rot}}_t^i, \wc{\bs{\theta}}_t^i, \wc{\bs{\beta}}_t^i) \in \wc{\bs{Q}}^i$.

The second term $E_\texttt{traj}$ measures the difference between the optimized global trajectory $(\wc{\bs{T}}^i, \wc{\bs{R}}^i)$ viewed in the camera coordinates and the trajectory $(\wt{\bs{T}}^i, \wt{\bs{R}}^i)$ output by the pose estimator (\eg, KAMA~\cite{iqbal2021kama}) in Stage I:
\begin{equation}
\begin{aligned}
    E_\texttt{traj} = \frac{1}{NT}\sum_{i=1}^N\sum_{t=1}^T V_t^i &\left(\|\Gamma(\wc{\bs{\rot}}_t^i, \bs{C}_t) \ominus \wt{\bs{\rot}}_t^i\|_a^2 \right. \\
    + &\left. w_t \|\Gamma(\wc{\bs{\tau}}_t^i, \bs{C}_t) - \wt{\bs{\tau}}_t^i\|_2^2\right),
\end{aligned}
\end{equation}
where the function $\Gamma(\cdot, \bs{C}_t)$ transforms the global rotation $\wc{\bs{\rot}}_t^i$ or translation $\wc{\bs{\tau}}_t^i$ to the camera coordinates defined by $\bs{C}_t$, and $w_t$ is a weighting factor for the translation term.

The third term $E_\texttt{reg}$ regularizes the egocentric trajectory $\wc{\bs{\Psi}}^i$ to stay close to the output $\wh{\bs{\Psi}}^i$ of the trajectory predictor:
\begin{align}
    E_\texttt{reg} = \frac{1}{NT}\sum_{i=1}^N\sum_{t=1}^T \left\|\bs{w}_\psi\circ\left(\wc{\bs{\psi}}_t^i - \wh{\bs{\psi}}_t^i\right)\right\|_2^2,
\end{align}
where $\circ$ denotes the element-wise product and $\bs{w}_\psi$ is a weighting vector for each element inside the egocentric trajectory. As an exception, we do not regularize each person's initial $xy$ position and heading $(\delta \wc{x}^i_0, \delta \wc{y}^i_0, \delta \wc{\phi}^i_0) \subset \wc{\bs{\psi}}_0^i$ as they need to be inferred from the video.

The fourth term $E_\texttt{cam}$ measures the smoothness of the camera parameters $\bs{C}$ and the uprightness of the camera:
\begin{equation}
\begin{aligned}
    E_\texttt{cam} & = \frac{1}{T}\sum_{t=1}^{T} \langle \bs{C}_t^y, \bs{Y} \rangle \\
    & \hspace{-5mm} +\frac{1}{T-1}\sum_{t=1}^{T-1} \left\|\bs{C}_{t+1}^\rot \ominus \bs{C}_t^\rot\right\|_a^2 + \left\|\bs{C}_{t+1}^\tau - \bs{C}_t^\tau\right\|_2^2,
\end{aligned}
\end{equation}
where $\langle\cdot,\cdot\rangle$ denotes the inner product, $\bs{C}_t^y$ is the $+y$ vector of the camera $\bs{C}_t$, and $\bs{Y}$ is the global up direction. $\bs{C}_t^\rot$ and $\bs{C}_t^\tau$ denote the rotation and translation of the camera $\bs{C}_t$.

The final term $E_\texttt{pen}$ is an signed distance field (SDF)-based inter-person penetration loss adopted from~\cite{jiang2020coherent}.

\section{Experiments}
\label{glamr:sec:exp}

\begin{figure}[t]
    \centering
    \includegraphics[width=0.75\linewidth]{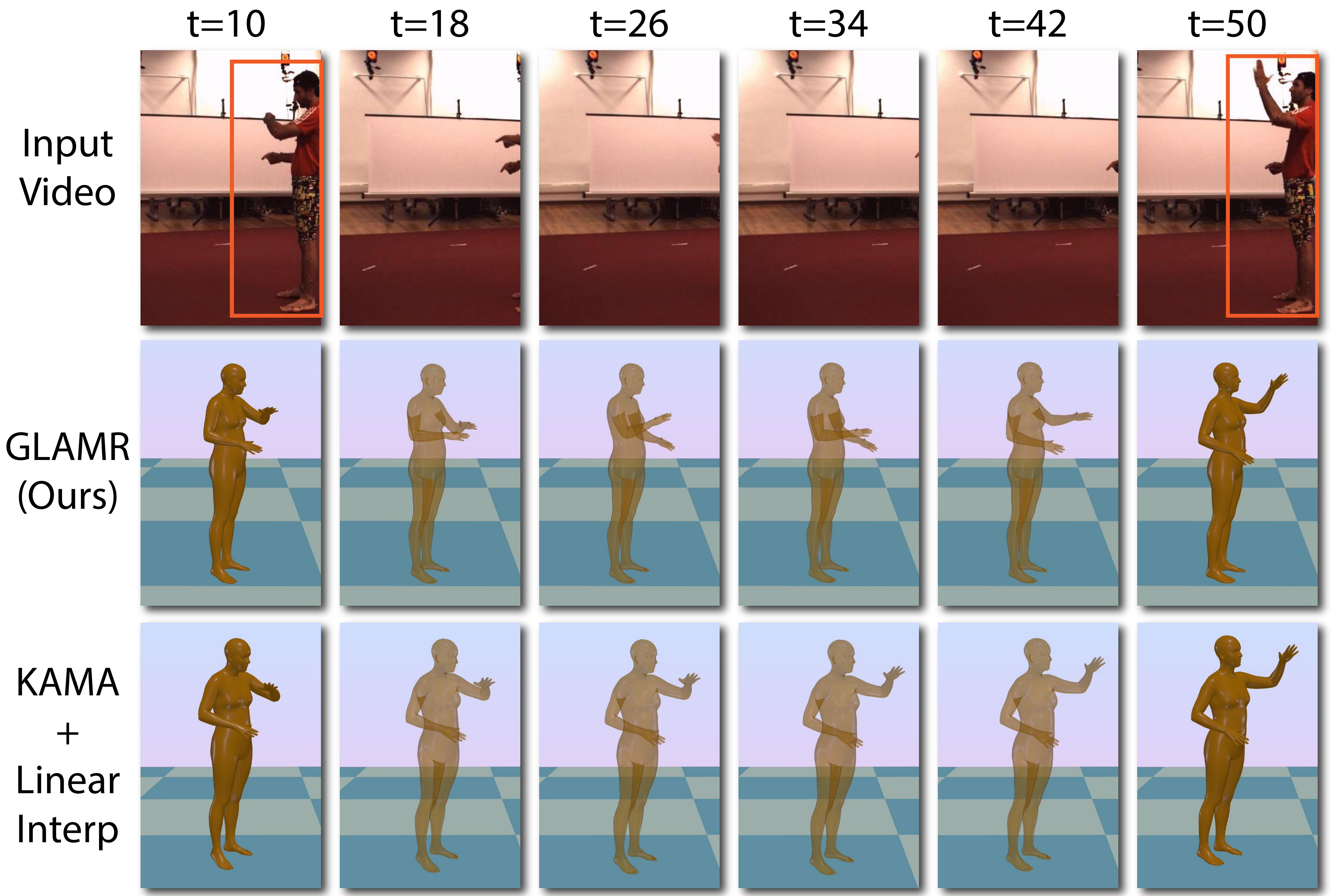}
    \vspace{4mm}
    \caption{Qualitative comparison of GLAMR on Dynamic Human3.6M.  GLAMR can generate natural hand motions for invisible frames instead of just doing linear interpolation.}
    \label{glamr:fig:results_vis_h36m}
\end{figure}

\paragraph{Datasets.}
We employ the following datasets in our experiments:
(1) \textbf{AMASS}~\cite{AMASS:ICCV:2019}, which is a large human motion database with 11000+ human motions. We use AMASS to train and evaluate the motion infiller and trajectory predictor.
(2)~\textbf{3DPW}~\cite{von2018recovering}, which is an \emph{in-the-wild} human motion dataset that uses videos and wearable IMU sensors to obtain GT poses, even when the person is occluded. We evaluate our approach using the test split of 3DPW.
(3)~\textbf{Dynamic Human3.6M} is a new benchmark for human pose estimation with dynamic cameras that we create from the Human3.6M dataset~\cite{ionescu2013human3}. We simulate dynamic cameras and occlusions by cropping each frame with a small view window that oscillates around the person (see Fig.~\ref{glamr:fig:results_vis_h36m}).

\paragraph{Evaluation Metrics.}
We use the following metrics for evaluation: (1) \textbf{G-MPJPE} and \textbf{G-PVE}, which extend the mean per joint position error (MPJPE) and per-vertex error (PVE) by computing the errors in the global coordinates. As errors in estimated global trajectories accumulate over time in our dynamic camera setting, we follow standard evaluations for open-loop reconstruction (\eg, SLAM~\cite{sturm2011towards} and inertial odometry~\cite{herath2020ronin}) to compute errors using a sliding window (10 seconds) and align the root translation and rotation with the GT at the start of the window. (2) \textbf{PA-MPJPE}, which is the Procrustes-aligned MPJPE for evaluating estimated body poses. For invisible poses, since there can be many plausible poses beside the GT, we follow prior work~\cite{aliakbarian2020stochastic,yuan2020dlow} to compute the best PA-MPJPE out of multiple samples for our probabilistic approach. (3) \textbf{Accel}, which computes the mean acceleration error of each joint and is commonly used to measure the jitter in estimated motions~\cite{yuan2021simpoe,kocabas2020vibe}. (4)~\textbf{FID}, which is an extension of the original Frechet Inception Distance that calculates the distribution distance between estimated motions and the GT. FID is a standard metric in motion generation literature to evaluate the quality of generated motions~\cite{li2020learning,valle2021transflower,huang2021,li2021ai}. Following prior work~\cite{li2021ai}, we compute FID using the well-designed kinetic motion feature extractor in the fairmotion library~\cite{gopinath2020fairmotion}.

\begin{figure*}[t]
\centering
\includegraphics[width=\linewidth]{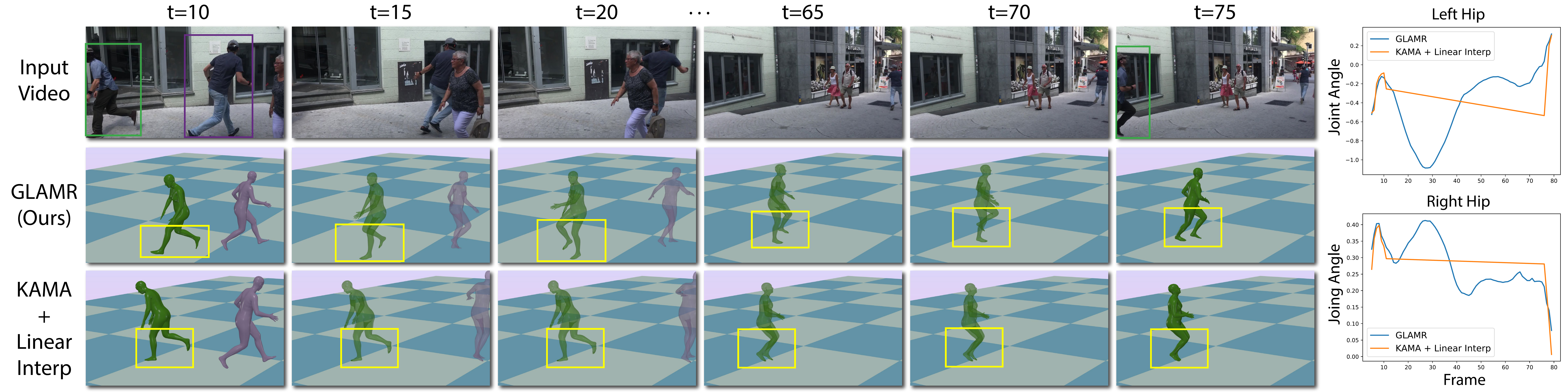}
\caption{Qualitative comparison of GLAMR with a strong baseline on 3DPW. The infilled motion (transparent) by GLAMR is more natural especially for the legs, while the baseline has very slow leg motions due to interpolation in a large window (frame 10 to 75).  On the \textbf{right}, we plot how the $x$-axis joint angles of left and right hips of the person (green) change over time for GLAMR and the baseline.}
\label{glamr:fig:results_vis}
\end{figure*}

\setlength{\tabcolsep}{3pt}
\begin{table}[t]
\footnotesize
\centering
\begin{tabular}{@{\hskip 1mm}lccccccc@{\hskip 1mm}}
\toprule
Method & \begin{tabular}{@{}c@{}}(All)  \\ G-MPJPE \end{tabular} & \begin{tabular}{@{}c@{}}(All)  \\ G-PVE \end{tabular} & \begin{tabular}{@{}c@{}}(Invisible)  \\ FID \end{tabular} & \begin{tabular}{@{}c@{}}(Invisible)  \\ PA-MPJPE \end{tabular} & \begin{tabular}{@{}c@{}}(Visible) \\ PA-MPJPE \end{tabular} & \begin{tabular}{@{}c@{}}(All)  \\ Accel \end{tabular} \\\midrule
KAMA~\cite{kaufmann2020convolutional} + Linear Interpolation & 1735.2 & 1744.1 & 30.2 & 74.8 & \textbf{47.4} & 8.0 \\
KAMA~\cite{kaufmann2020convolutional} + Last Pose & 1318.1 & 1330.3 & 36.7 & 88.8 & \textbf{47.4} & 12.3 \\
KAMA~\cite{kaufmann2020convolutional} + ConvAE~\cite{kaufmann2020convolutional} & 1737.8 & 1748.9 & 28.9 & 77.4 & 56.9 & 7.5 \\
SPEC~\cite{Kocabas_SPEC_2021} + Linear Interpolation & 2113.3 & 2119.5 & 29.7 & 78.7 & 55.7 & 14.2 \\
SPEC~\cite{Kocabas_SPEC_2021} + Last Pose & 1782.5 & 1790.9 & 36.2 & 92.6 & 55.7 & 18.8 \\
SPEC~\cite{Kocabas_SPEC_2021} + ConvAE~\cite{kaufmann2020convolutional} & 2113.3 & 2119.0 & 28.5 & 80.1 & 59.9 & 11.9 \\ \midrule
Ours (GLAMR w/ SPEC) & 899.1 & 913.7 & \textbf{8.2} & 72.8 & 55.0 & 6.6 \\
Ours (GLAMR w/ KAMA) & \textbf{806.2} & \textbf{824.1} & 11.4 & \textbf{67.7} & 47.6 & \textbf{6.0} \\
\bottomrule
\end{tabular}
\vspace{5mm}
\caption{Baseline comparison on Dynamic Human3.6M. We report results for visible, invisible (occluded), and all frames.}
\label{glamr:table:baseline_h36m}
\end{table}

\subsection{Evaluation of GLAMR}
\label{glamr:sec:eval_glamr}
\paragraph{Baselines.}
Since no prior methods can estimate global motions from dynamic cameras and address long-term occlusions, we design various baselines by combining state-of-the-art human mesh recovery methods (KAMA~\cite{iqbal2021kama} or SPEC~\cite{Kocabas_SPEC_2021}), motion infilling methods, and SLAM-based camera estimation (OpenSfM~\cite{opensfm2021}). In particular, we use the estimated camera parameters to convert estimated motions from the camera coordinates to the global coordinates.  For motion infilling, we use (1) linear interpolation, (2) last pose, \ie, replicating the last visible pose, and (3) a state-of-the-art CNN-based motion infilling method, ConvAE~\cite{kaufmann2020convolutional}. 

The results on Dynamic Human3.6M and 3DPW are summarized in Table~\ref{glamr:table:baseline_h36m} and \ref{glamr:table:baseline_3dpw} respectively. We only report G-MPJPE and G-PVE on Dynamic Human3.6M since they require accurate GT trajectories, which 3DPW does not provide. It is evident that our approach, GLAMR, outperforms the baselines in almost all metrics. In particular, GLAMR achieves significantly lower G-MPJPE and G-PVE, which demonstrates its strong ability to reconstruct global human motions. Furthermore, GLAMR attains considerably lower FID and PA-MPJPE (with ten samples) for occluded (invisible) poses. The lower FID means GLAMR can infill more humanlike motions, and the lower PA-MPJPE also shows GLAMR's probabilistic motion samples can cover the GT better. Finally, while GLAMR achieves almost the same PA-MPJPE for visible poses as the best method, it yields much smoother motions (smaller acceleration error). This is because our motion infiller leverages human dynamics learned from a large motion dataset to produce motions.

\setlength{\tabcolsep}{5pt}
\begin{table}[t]
\footnotesize
\centering
\begin{tabular}{@{\hskip 1mm}lccccc@{\hskip 1mm}}
\toprule
Method & \begin{tabular}{@{}c@{}}(Invisible)  \\ FID \end{tabular} & \begin{tabular}{@{}c@{}}(Invisible)  \\ PA-MPJPE \end{tabular} & \begin{tabular}{@{}c@{}}(Visible) \\ PA-MPJPE \end{tabular} & \begin{tabular}{@{}c@{}}(All)  \\ Accel \end{tabular} \\ \midrule
KAMA~\cite{kaufmann2020convolutional} + Linear Interpolation & 30.7 & 87.5 & \textbf{50.8} & 24.2 \\
KAMA~\cite{kaufmann2020convolutional} + Last Pose & 40.3 & 96.3 & \textbf{50.8} & 25.4 \\
KAMA~\cite{kaufmann2020convolutional} + ConvAE~\cite{kaufmann2020convolutional} & 32.0 & 84.5 & 56.4 & 19.6 \\
SPEC~\cite{Kocabas_SPEC_2021} + Linear Interpolation & 33.6 & 85.6 & 53.3 & 33.1 \\
SPEC~\cite{Kocabas_SPEC_2021} + Last Pose & 39.5 & 92.4 & 53.3 & 34.2 \\
SPEC~\cite{Kocabas_SPEC_2021} + ConvAE~\cite{kaufmann2020convolutional} & 35.4 & 86.9 & 59.3 & 24.0 \\ \midrule
Ours (GLAMR w/ SPEC) & 24.8 & 79.1 & 54.9 & 9.5 \\
Ours (GLAMR w/ KAMA) & \textbf{22.6} & \textbf{73.6} & 51.1 & \textbf{8.9} \\
\bottomrule
\end{tabular}
\vspace{5mm}
\caption{Baseline comparison on 3DPW. G-MPJPE and G-PVE are not reported since 3DPW does not provide accurate GT global human trajectories. See also the caption of Table~\ref{glamr:table:baseline_h36m}.}
\label{glamr:table:baseline_3dpw}
\end{table}

\paragraph{Qualitative Results.}
Fig.~\ref{glamr:fig:results_vis} and \ref{glamr:fig:results_vis_h36m} show qualitative comparisons of GLAMR against the strong baseline, KAMA + Linear Interpolation. Additionally, we provide abundant qualitative results on the \href{https://nvlabs.github.io/GLAMR}{project page}.

\subsection{Evaluation of Key Components}
\label{glamr:sec:eval_component}

\setlength{\tabcolsep}{5pt}
\begin{table}[t]
\footnotesize
\centering
\begin{tabular}{@{\hskip 1mm}lccccc@{\hskip 1mm}}
\toprule
Method & \begin{tabular}{@{}c@{}}(Sampled) \\ PA-MPJPE \end{tabular} & \begin{tabular}{@{}c@{}}(Reconstructed) \\ PA-MPJPE \end{tabular} & \begin{tabular}{@{}c@{}}(Sampled) \\ FID \end{tabular}\\ \midrule
Linear Interpolation & 83.5 & 83.5 & 35.3 \\
Last Pose & 104.4 & 104.4 & 41.6 \\
ConvAE~\cite{kaufmann2020convolutional} & 72.8	& 72.8 & 31.4 \\
Ours & \textbf{61.4} & \textbf{36.1} & \textbf{16.7}\\
\bottomrule
\end{tabular}
\vspace{5mm}
\caption{Benchmarking motion infiller on AMASS.}
\label{glamr:table:mfiller}
\end{table}

\setlength{\tabcolsep}{5pt}
\begin{table}[t]
\footnotesize
\centering
\begin{tabular}{@{\hskip 1mm}lcccc@{\hskip 1mm}}
\toprule
Method & G-MPJPE & G-PVE & Accel \\ \midrule
Transformer & 660.1 & 678.6 & 121.9 \\
Ours w/o Ego Trajectory &  763.0 & 780.6 & 8.7 \\
Ours & \textbf{466.9} & \textbf{472.5} & \textbf{5.8} \\
\bottomrule
\end{tabular}
\vspace{5mm}
\caption{Benchmarking trajectory predictor on AMASS.}
\label{glamr:table:traj_pred}
\end{table}

\setlength{\tabcolsep}{5pt}
\begin{table}[t]
\footnotesize
\centering
\begin{tabular}{@{\hskip 1mm}lcccc@{\hskip 1mm}}
\toprule
Method & G-MPJPE & G-PVE & Accel \\ \midrule
Ours w/o Trajectory Predictor & 1750.8 & 1761.4 & 12.6 \\
Ours w/o Opt Ego Trajectory &  877.3 & 895.0 & 15.5 \\
Ours (GLAMR) & \textbf{806.2} & \textbf{824.1} & \textbf{6.0} \\
\bottomrule
\end{tabular}
\vspace{5mm}
\caption{Global optimization ablations on Dynamic Human3.6M.}
\label{glamr:table:global_opt}
\end{table}

\paragraph{Benchmarking Motion Infiller.}
We evaluate the proposed generative motion infiller on the test split of the AMASS dataset~\cite{AMASS:ICCV:2019}. We compare against three motion infilling baselines: linear interpolation, replicating the last pose, and \mbox{ConvAE}~\cite{kaufmann2020convolutional}. As shown in Table~\ref{glamr:table:mfiller}, our generative motion infiller achieves significantly better PA-MPJPE for both the sampled motions (with five samples) and reconstructed motion for the infilled frames. Our approach also achieves considerably better FID, reducing the FID of ConvAE~\cite{kaufmann2020convolutional} by half, which indicates that the infilled motions by our approach are much closer to real human motions.

\paragraph{Benchmarking Trajectory Predictor.}
We also evaluate our global trajectory predictor against two variants on the AMASS test set: (1) ``Transformer'', which replaces the LSTMs in the trajectory predictor with Transformers; (2)~``Ours w/o Ego Trajectory'', which does not use the egocentric trajectory but instead directly outputs the 6-DoF global trajectory. As shown in Table~\ref{glamr:table:traj_pred}, both variants lead to worse global trajectory prediction (higher best-of-five G-MPJPE and G-PVE). We believe the reasons are: (1) the positional encoding in Transformers may not generalize well to longer motions compared to the LSTMs in our approach; (2) directly predicting the 6-DoF global trajectory offsets instead of egocentric trajectories from local body motions is also hard to generalize since the global offsets can be large.

\paragraph{Ablations for Global Optimization.}
We further perform ablation studies on the effect of key components in our global optimization. Specifically, we design two variants: (1) ``Ours w/o Trajectory Predictor'', which does not use our trajectory predictor to generate the global human trajectories and uses camera parameters from OpenSfM~\cite{opensfm2021} to obtain global trajectories instead; (2) ``Ours w/o Opt Ego Trajectory'', which does not employ the egocentric trajectory representation and directly optimizes the 6-DoF root trajectory instead. As shown in Table~\ref{glamr:table:global_opt}, both variants lead to significantly worse global trajectory reconstruction with large increases in G-MPJPE, G-PVE, and Accel. This demonstrates that both the global trajectory predictor and egocentric trajectory representation are vital in our approach.

\section{Discussion and Futurework}
In this paper, we proposed an approach for 3D human mesh recovery in consistent global coordinates from videos captured by dynamic cameras. We first proposed a novel Transformer-based generative motion infiller to address severe occlusions that often come with dynamic cameras. To resolve ambiguity in the joint reconstruction of global human motions and camera poses, we proposed a new solution by predicting global human trajectories from local body motions. Finally, we proposed a global optimization framework to refine the predicted trajectories, which serve as anchors for camera optimization. Our method achieves SOTA results on challenging datasets and marks a significant step towards global human mesh recovery in the wild.

As the first paper on this new problem, our method has a few limitations that are important for future research to address. First, our approach has five stages that are sequentially dependent. Therefore, errors in early stages can propagate to late stages, which may lead to inaccurate global pose estimation. Future work could integrate these stages together to form an end-to-end learnable framework. Second, like many works in human mesh recovery, our approach can only recover the SMPL parameters which omit the fine details of human meshes such as clothing. Integrating neural articulated shapes such as~\cite{deng2020nasa} into our approach could potentially address this problem. Next, our approach is not real-time due to the batch processing and global optimization. Future work could explore a causal version of our approach where only a small window around the incoming frame is optimized, which could substantially improve computational efficiency. Additionally, the generative motion infiller and global trajectory predictor in our approach operate for each person independently. Therefore, the generated motions and trajectories may not capture potentially complex and nuanced interactions between occluded people such as hugging or dancing. Future work could address this limitation by employing new generative models that produce interaction-aware motions of multiple people.

An important problem for future work is how to model more context in the motion infiller, e.g., interactions between multiple people, objects, and the scene. Although the motion infiller is learned using a large motion database, the training motions are often short, contain a single person, and lack human-object interaction. To learn a more context-dependent motion prior, we need large 3D human motion datasets containing long-term human behaviors, multi-person interactions, and human-object interactions. Such 3D datasets are often difficult to collect. One way to address this is to use our simulated-based behavior modeling framework for physically-plausible motion capture from videos and IMU sensors. Another problem for future work is that the motion infiller cannot model idiosyncratic behaviors since the motion prior is the same for every person in the scene. One potential way to address this problem is to learn a hierarchical latent variable model, where a new high-level latent code is introduced to control the behavior style of each individual. Once the model is learned, we can sample the high-level latent code to allow each person to have its own behavior style. Again, data is important here since we need long-term behavior data from diverse individuals to successfully learn the model.

Due to the already complicated pipeline, we did not use physics simulation in GLAMR and left it for future work. Unsurprisingly, the lack of physics leads to noticeable physical artifacts such as foot sliding, jitter, and ground penetration in the pose estimation results as shown \href{https://nvlabs.github.io/GLAMR}{here}. Although not trivial, it is possible to integrate our simulation-based behavior modeling framework into GLAMR. Specifically, we need to first learn a universal humanoid control (UHC) policy~\cite{luo2021dynamics} from a large motion database such as AMASS~\cite{AMASS:ICCV:2019}. The UHC policy learns to imitate an input kinematic motion sequence and output a physically-plausible version of the input via physics simulation. Once the UHC policy is learned, we can use it to process the output human motions of GLAMR into physically-plausible motions. One caveat is that, for dynamic motions such as dancing, the UHC policy may not be able to perfectly imitate the global trajectory of each person from GLAMR and its resulting trajectory can gradually drift away. To tackle this problem, we can have another trajectory optimization stage where the UHC policy is further finetuned to try to match the global trajectory output by GLAMR.

\part{Generation of Human Behavior}
\label{part:generation}
\chapter{Deterministic Simulation-Based Egocentric Human Motion Generation}
\label{chap:egopose19}

\section{Introduction}

With a single wearable camera, our goal is to estimate and forecast a person's pose sequence for a variety of complex motions. Estimating and forecasting complex human motions with egocentric cameras can be the cornerstone of many useful applications. In medical monitoring, the inferred motions can help physicians remotely diagnose patients' condition in motor rehabilitation. In virtual or augmented reality, anticipating motions can help allocate limited computational resources to provide better responsiveness. For athletes, the forecasted motions can be integrated into a coaching system to offer live feedback and reinforce good movements. In all these applications, human motions are very complex, as periodical motions (e.g., walking, running) are often mixed with non-periodical motions (e.g., turning, bending, crouching). It is challenging to estimate and forecast such complex human motions from egocentric videos due to the multi-modal nature of the data.

It has been shown that if the task of pose estimation can be limited to a single mode of action such as running or walking, it is possible to estimate a physically-valid pose sequence. Recent work by Yuan and Kitani~\cite{yuan20183d} has formulated egocentric pose estimation as a Markov decision process (MDP): a humanoid agent driven by a control policy with visual input to generate a pose sequence inside a physics simulator. They use generative adversarial imitation learning (GAIL~\cite{ho2016generative}) to solve for the optimal control policy. By design, this approach guarantees that the estimated pose sequence is physically-valid. However, their method focuses on a single action modality (i.e., simple periodical motions including walking and running). The approach also requires careful segmentation of the demonstrated motions, due to the instability of adversarial training when the data is multi-modal. To address these issues, we propose an ego-pose estimation approach that can learn a motion policy directly from unsegmented multi-modal motion demonstrations.

Unlike the history of work on egocentric pose estimation, there has been no prior work addressing the task of egocentric pose forecasting. Existing works on 3D pose forecasting not based on egocentric sensing take a pose sequence as input and uses recurrent models to output a future pose sequence by design~\cite{fragkiadaki2015recurrent, jain2016structural, butepage2017deep, li2017auto}. Even with the use of a 3D pose sequence as a direct input, these methods tend to produce unrealistic motions due to error accumulation (covariate shift~\cite{quionero2009dataset}) caused by feeding predicted pose back to the network without corrective interaction with the learning environment. More importantly, these approaches often generate physically-invalid pose sequences as they are trained only to mimic motion kinematics, disregarding causal forces like the laws of physics or actuation constraints. In this work, we propose a method that directly takes noisy observations of past egocentric video as input to forecast stable and physically-valid future human motions.

\begin{figure}
    \centering
    \includegraphics[width=\linewidth]{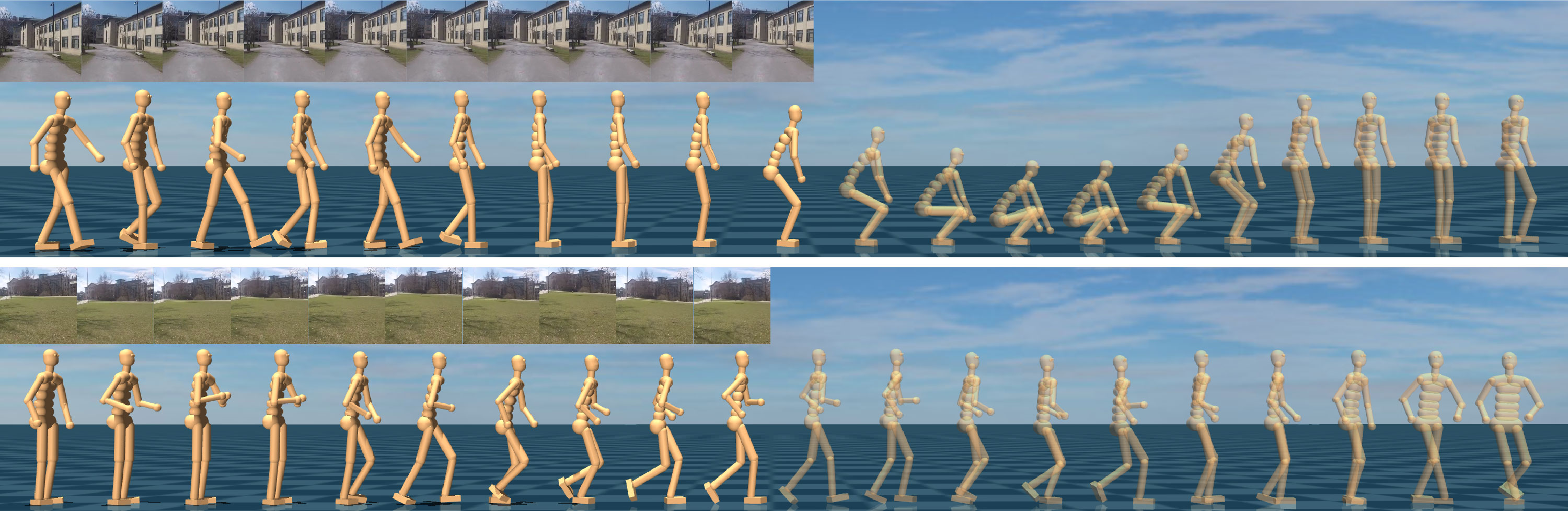}
    \caption{Proposed method estimates camera wearer's 3D poses (solid) and forecasts future poses (translucent) in real-time..}
    \label{egopose19:fig:teaser}
\end{figure}

We formulate both egocentric pose estimation and forecasting as a MDP. The humanoid control policy takes as input the current state of the humanoid for both inference tasks. Additionally, the visual context from the entire video is used as input for the pose estimation task. In the case of the forecasting task, only the visual input observed up to the current time step is used. For the action space of the policy, we use target joint positions of proportional-derivative (PD) controllers~\cite{tan2011stable} instead of direct joint torques. The PD controllers act like damped springs and compute the torques to be applied at each joint. This type of action design is more capable of actuating the humanoid to perform highly dynamic motions~\cite{peng2018deepmimic}. As deep reinforcement learning (DeepRL) based approaches for motion imitation~\cite{peng2018deepmimic, peng2018sfv} have proven to be more robust than GAIL based methods~\cite{yuan20183d, merel2017learning, wang2017robust}, we utilize DeepRL to encourage the motions generated by the control policy to match the ground-truth. However, reward functions designed for motion imitation methods are not suited for our task because they are tailored to learning locomotions from short segmented motion clips, while our goal is to learn to estimate and forecast complex human motions from unsegmented multi-modal motion data. Thus, we propose a new reward function that is specifically designed for this type of data. For forecasting, we further employ a decaying reward function to focus on forecasting for frames in the near future. Since we only take past video frames as input and the video context is fixed during forecasting, we use a recurrent control policy to better encode the phase of the human motion.

A unique problem encountered by the control-based approach taken in this work is that the humanoid being actuated in the physics simulator can fall down. Specifically, extreme domain shifts in the visual input at test time can cause irregular control actions. As a result, this irregularity in control actions causes the humanoid to lose balance and fall in the physics environment, preventing the method from providing any pose estimates. The control-based method proposed in~\cite{yuan20183d} prevented falling by fine-tuning the policy at test time as a batch process. As a result, this prohibits its use in streaming or real-time applications. Without fine-tuning, their approach requires that we reset the humanoid state to some reasonable starting state to keep producing meaningful pose estimates. However, it is not clear when to re-estimate the state. To address this issue of the humanoid falling in the physics simulator at test time, we propose a fail-safe mechanism based on a value function estimate used in the policy gradient method. The mechanism can anticipate falling much earlier and stabilize the humanoid before producing bad pose estimates.

We validate our approach for egocentric pose estimation and forecasting on a large motion capture (MoCap) dataset and an in-the-wild dataset consisting of various human motions (jogging, bending, crouching, turning, hopping, leaning, motion transitions, etc.). Experiments on pose estimation show that our method can learn directly from unsegmented data and outperforms state-of-the-art methods in terms of both quantitative metrics and visual quality of the motions. Experiments on pose forecasting show that our approach can generate intuitive future motions and is also more accurate compared to other baselines. Our in-the-wild experiments show that our method transfers well to real-world scenarios without the need for any fine-tuning. Our time analysis show that our approach can run at 30 FPS, making it suitable for many real-time applications.

In summary, our contributions are as follows: (1) We propose a DeepRL-based method for egocentric pose estimation that can learn from unsegmented MoCap data and estimate accurate and physically-valid pose sequences for complex human motions. (2) We are the first to tackle the problem of egocentric pose forecasting and show that our method can generate accurate and stable future motions. (3) We propose a fail-safe mechanism that can detect instability of the humanoid control policy, which prevents generating bad pose estimates. (4) Our model trained with MoCap data transfers well to real-world environments without any fine-tuning. (5) Our time analysis show that our pose estimation and forecasting algorithms can run in real-time.

\section{Related Work}
\paragraph{3D Human Pose Estimation.} Third-person pose estimation has long been studied by the vision community~\cite{liu2015survey, sarafianos20163d}. Existing work leverages the fact that the human body is visible from the camera. Traditional methods tackle the depth ambiguity with strong priors such as shape models~\cite{zhou2017sparse, bogo2016keep}. Deep learning based approaches~\cite{zhou2016sparseness, pavlakos2017coarse, mehta2017vnect, tung2017self} have also succeeded in directly regressing images to 3D joint locations with the help of large-scale MoCap datasets~\cite{ionescu2013human3}. To achieve better performance for in-the-wild images, weakly-supervised methods~\cite{zhou2017towards, rogez2016mocap, kanazawa2018end} have been proposed to learn from images without annotations. Although many of the state-of-art approaches predict pose for each frame independently, several works have utilized video sequences to improve temporal consistency~\cite{tekin2016direct, xu2018monoperfcap, dabral2017structure, kanazawa2018learning}.

A limited amount of research has looked into egocentric pose estimation. Most existing methods only estimate the pose of visible body parts~\cite{li2013model,li2013pixel,ren2010figure,arikan2003motion,rogez2015first}. Other approaches utilize 16 or more body-mounted cameras to infer joint locations via structure from motion~\cite{shiratori2011motion}. Specially designed head-mounted rigs have been used for markerless motion capture~\cite{rhodin2016egocap, xu2019mo, tome2019xr}, where~\cite{tome2019xr} utilizes photorealistic synthetic data. Conditional random field based methods~\cite{jiang2017seeing} have also been proposed to estimate a person's full-body pose with a wearable camera. The work most related to ours is~\cite{yuan20183d} which formulates egocentric pose estimation as a Markov decision process to enforce physics constraints and solves it by adversarial imitation learning. It shows good results on simple periodical human motions but fails to estimate complex non-periodical motions. Furthermore, they need fine-tuning at test time to prevent the humanoid from falling. In contrast, we propose an approach that can learn from unsegmented MoCap data and estimate various complex human motions in real-time without fine-tuning.

\paragraph{Human Motion Forecasting.}
Plenty of work has investigated third-person~\cite{Xie2013InferringM, ma2017forecasting, ballan2016knowledge, kitani2012activity, alahi2016social, robicquet2016learning, yagi2018future} and first-person~\cite{soo2016egocentric} trajectory forecasting, but this line of work only forecasts a person's future positions instead of poses. There are also works focusing on predicting future motions in image space~\cite{finn2016unsupervised, walker2016uncertain, walker2017pose, denton2017unsupervised, xue2016visual, li2018flow, gao2018im2flow}. Other methods use past 3D human pose sequence as input to predict future human motions~\cite{fragkiadaki2015recurrent, jain2016structural, butepage2017deep, li2017auto}. Recently,~\cite{kanazawa2018learning, chao2017forecasting} forecast a person's future 3D poses from third-person static images, which require the person to be visible. Different from previous work, we propose to forecast future human motions from egocentric videos where the person can hardly be seen.

\paragraph{Humanoid Control from Imitation.} The idea of using reference motions has existed for a long time in computer animation. Early work has applied this idea to bipedal locomotions with planar characters~\cite{sharon2005synthesis, sok2007simulating}. Model-based methods~\cite{yin2007simbicon, muico2009contact, lee2010data} generate locomotions with 3D humanoid characters by tracking reference motions. Sampling-based control methods~\cite{liu2010sampling, liu2016guided, liu2017learning} have also shown great success in generating highly dynamic humanoid motions. DeepRL based approaches have utilized reference motions to shape the reward function~\cite{peng2017deeploco, peng2017learning}. Approaches based on GAIL~\cite{ho2016generative} have also been proposed to eliminate the need for manual reward engineering~\cite{merel2017learning, wang2017robust, yuan20183d}. The work most relevant to ours is DeepMimic~\cite{peng2018deepmimic} and its video variant~\cite{peng2018sfv}. DeepMimic has shown beautiful results on human locomotion skills with manually designed reward and is able to combine learned skills to achieve different tasks. However, it is only able to learn skills from segmented motion clips and relies on the phase of motion as input to the policy. In contrast, our approach can learn from unsegmented MoCap data and use the visual context as a natural alternative to the phase variable.

\section{Approach}
\begin{figure*}
    \centering
    \includegraphics[width=\textwidth]{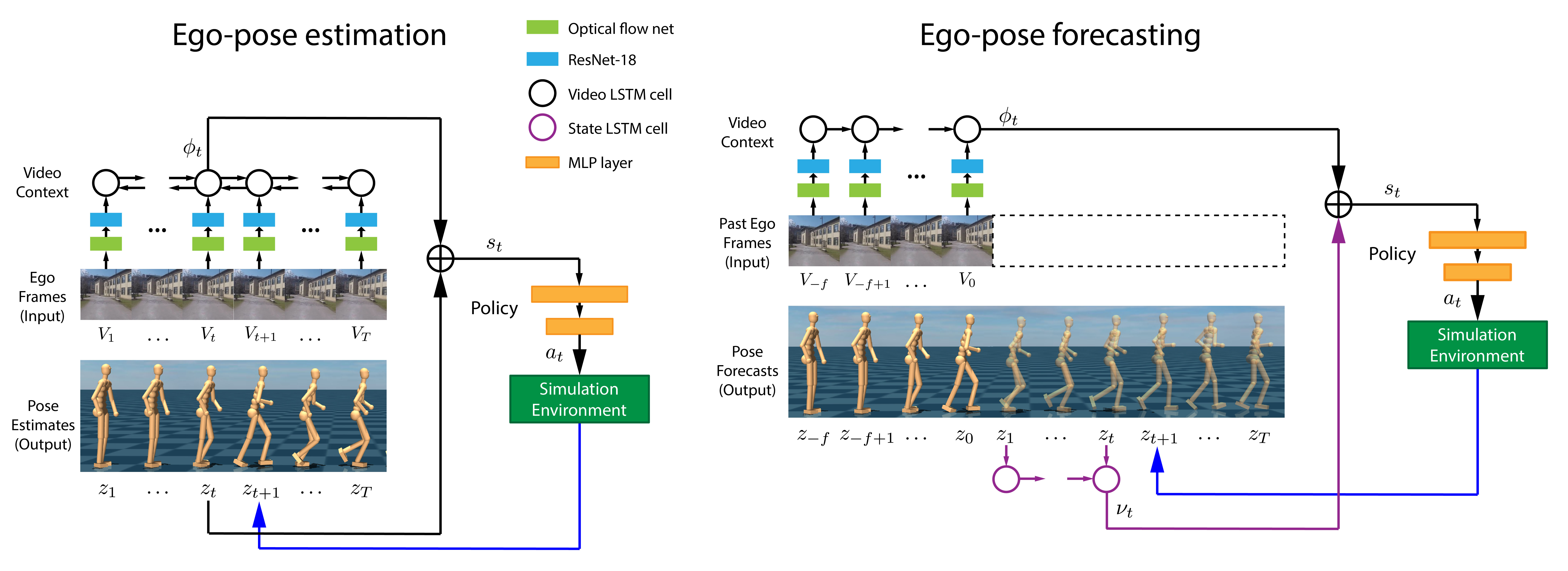}
    \caption{Overview for ego-pose estimation and forecasting. The policy takes in the humanoid state $z_t$ (estimation) or recurrent state feature $\nu_t$ (forecasting) and the visual context $\phi_t$ to output the action $a_t$, which generates the next humanoid state $z_{t+1}$ through physics simulation. \textbf{Left:} For ego-pose estimation, the visual context $\phi_t$ is computed from the entire video $V_{1:T}$ using a Bi-LSTM to encode CNN features. \textbf{Right:} For ego-pose forecasting, $\phi_t$ is computed from past frames $V_{-f:0}$ using a forward LSTM and is kept fixed for all $t$.}
    \label{egopose19:fig:overview}
\end{figure*}

We choose to model human motion as the result of the optimal control of a dynamical system governed by a cost (reward) function, as control theory provides mathematical machinery necessary to explain human motion under the laws of physics. In particular, we use the formalism of the Markov Decision process (MDP). The MDP is defined by a tuple $\mathcal{M} = \left(S, A, P, R, \gamma\right)$ of states, actions, transition dynamics, a reward function, and a discount factor. 

\paragraph{State.} The state $s_t$ consists of both the state of the humanoid $z_t$ and the visual context $\phi_t$. The humanoid state $z_t$ consists of the pose $p_t$ (position and orientation of the root, and joint angles) and velocity $v_t$ (linear and angular velocities of the root, and joint velocities). All features are computed in the humanoid's local heading coordinate frame which is aligned with the root link's facing direction. The visual context $\phi_t$ varies depending on the task (pose estimation or forecasting) which we will address in Sec.~\ref{egopose19:sec:estimation} and~\ref{egopose19:sec:forecast}.

\paragraph{Action.} The action $a_t$ specifies the target joint angles for the Proportional-Derivative (PD) controller at each degree of freedom (DoF) of the humanoid joints except for the root. For joint DoF $i$, the torque to be applied is computed as
\begin{equation}
    \tau^{i} = k_p^i(a_t^i - p_t^i) - k_d^i v_t^i\,,
\end{equation}
where $k_p$ and $k_d$ are manually-specified gains.
Our policy is queried at $30$Hz while the simulation is running at $450$Hz, which gives the PD-controllers 15 iterations to try to reach the target positions. Compared to directly using joint torques as the action, this type of action design increases the humanoid's capability of performing highly dynamic motions~\cite{peng2018deepmimic}.

\paragraph{Policy.} The policy $\pi_\theta(a_t|s_t)$ is represented by a Gaussian distribution with a fixed diagonal covariance matrix $\Sigma$. We use a neural network with parameter $\theta$ to map state $s_t$ to the mean $\mu_t$ of the distribution. We use a multilayer perceptron (MLP) with two hidden layers $(300, 200)$ and ReLU activation to model the network. Note that at test time we always choose the mean action from the policy to prevent performance drop from the exploration noise.

\paragraph{Solving the MDP.} At each time step, the humanoid agent in state $s_t$ takes an action $a_t$ sampled from a policy $\pi(a_t|s_t)$, and the environment generates the next state $s_{t+1}$ through physics simulation and gives the agent a reward $r_t$ based on how well the humanoid motion aligns with the ground-truth. This process repeats until some termination condition is triggered such as when the time horizon is reached or the humanoid falls. To solve this MDP, we apply policy gradient methods (e.g., PPO~\cite{schulman2017proximal}) to obtain the optimal policy $\pi^\star$ that maximizes the expected discounted return $\mathbb{E}\left[\sum_{t=1}^T \gamma^{t-1} r_t\right]$. At test time, starting from some initial state $s_1$, we rollout the policy $\pi^\star$ to generate state sequence $s_{1:T}$, from which we extract the output pose sequence $p_{1:T}$.

\subsection{Ego-pose Estimation}
\label{egopose19:sec:estimation}
The goal of egocentric pose estimation is to use video frames $V_{1:T}$ from a wearable camera to estimate the person's pose sequence $p_{1:T}$. To learn the humanoid control policy $\pi(a_t|z_t,\phi_t)$ for this task, we need to define the procedure for computing the visual context $\phi_t$ and the reward function $r_t$. As shown in Fig.~\ref{egopose19:fig:overview} (Left), the visual context $\phi_t$ is computed from the video $V_{1:T}$. Specifically, we calculate the optical flow for each frame and pass it through a CNN to extract visual features $\psi_{1:T}$. Then we feed $\psi_{1:T}$ to a bi-directional LSTM to generate the visual context $\phi_{1:T}$, from which we obtain per frame context $\phi_t$. For the starting state $z_1$, we set it to the ground-truth $\hat{z}_1$ during training. To encourage the pose sequence $p_{1:T}$ output by the policy to match the ground-truth $\hat{p}_{1:T}$, we define our reward function as
\begin{equation}
\label{egopose19:eq:reward}
    r_t = w_q r_q + w_e r_e + w_p r_p + w_v r_v\,,
\end{equation}
where $w_q, w_e, w_p, w_v$ are weighting factors.

The pose reward $r_q$ measures the difference between pose $p_t$ and the ground-truth $\hat{p}_t$ for non-root joints. We use $q_t^j$ and $\hat{q}_t^j$ to denote the local orientation quaternion of joint $j$ computed from $p_t$ and $\hat{p}_t$ respectively. We use $q_1 \ominus q_2$ to denote the relative quaternion from $q_2$ to $q_1$, and $\|q\|$ to compute the rotation angle of $q$.
\begin{equation}
    r_q = \exp\left[-2\left(\sum_{j}\|q_t^j\ominus \hat{q}_t^j\|^2\right)\right]\,.
\end{equation}
The end-effector reward $r_e$ evaluates the difference between local end-effector vector $e_t$ and the ground-truth $\hat{e}_t$. For each end-effector $e$ (feet, hands, head), $e_t$ is computed as the vector from the root to the end-effector.
\begin{equation}
    r_e = \exp\left[-20\left(\sum_{e}\|e_t- \hat{e}_t\|^2\right)\right]\,.
\end{equation}
The root pose reward $r_p$ encourages the humanoid's root joint to have the same height $h_t$ and orientation quaternion $q_t^r$ as the ground-truth $\hat{h}_t$ and $\hat{q}_t^r$.
\begin{equation}
    r_p = \exp\left[-300\left((h_t - \hat{h}_t)^2 + \|q_t^r\ominus \hat{q}_t^r\|^2\right) \right]\,.
\end{equation}
The root velocity reward $r_v$ penalizes the deviation of the root's linear velocity $l_t$ and angular velocity $\omega_t$ from the ground-truth $\hat{l}_t$ and $\hat{\omega}_t$. The ground-truth velocities can be computed by the finite difference method.
\begin{equation}
    r_v = \exp\left[-\|l_t- \hat{l}_t\|^2 -0.1 \|\omega_t^r-\hat{\omega}_t^r\|^2 \right]\,.
\end{equation}
Note that all features are computed inside the local heading coordinate frame instead of the world coordinate frame, which is crucial to learn from unsegmented MoCap data for the following reason: when imitating an unsegmented motion demonstration, the humanoid will drift from the ground-truth motions in terms of global position and orientation because the errors made by the policy accumulate; if the features are computed in the world coordinate, their distance to the ground-truth quickly becomes large and the reward drops to zero and stops providing useful learning signals. Using local features ensures that the reward is well-shaped even with large drift. To learn global motions such as turning with local features, we use the reward $r_v$ to encourage the humanoid's root to have the same linear and angular velocities as the ground-truth.

\paragraph{Initial State Estimation.} As we have no access to the ground-truth humanoid starting state $z_1$ at test time, we need to learn a regressor $\mathcal{F}$ that maps video frames $V_{1:T}$ to their corresponding state sequence $z_{1:T}$. $\mathcal{F}$ uses the same network architecture as ego-pose estimation (Fig.~\ref{egopose19:fig:overview} (Left)) for computing the visual context $\phi_{1:T}$ . We then pass $\phi_{1:T}$ through an MLP with two hidden layers (300, 200) to output the states. We use the mean squared error (MSE) as the loss function: $L(\zeta) = \frac{1}{T}\sum_{t=1}^T \|\mathcal{F}(V_{1:T})_t - z_t\|^2$, where $\zeta$ is the parameters of $\mathcal{F}$. The optimal $\mathcal{F}^\star$ can be obtained by an SGD-based method.

\subsection{Ego-pose Forecasting}
\label{egopose19:sec:forecast}
For egocentric pose forecasting, we aim to use past video frames $V_{-f:0}$ from a wearable camera to forecast the future pose sequence $p_{1:T}$ of the camera wearer. We start by defining the visual context $\phi_t$ used in the control policy $\pi$. As shown in Fig.~\ref{egopose19:fig:overview} (Right), the visual context $\phi_t$ for this task is computed from past frames $V_{-f:0}$ and is kept fixed for all time $t$ during a policy rollout.  We compute the optical flow for each frame and use a CNN to extract visual features $\psi_{-f:0}$. We then use a forward LSTM to summarize $\psi_{-f:0}$ into the visual context $\phi_t$. For the humanoid starting state $z_1$, we set it to the ground-truth $\hat{z}_1$, which at test time is provided by ego-pose estimation on $V_{-f:0}$. Now we define the reward function for the forecasting task. Due to the stochasticity of human motions, the same past frames can correspond to multiple future pose sequences. As the time step $t$ progresses, the correlation between pose $p_t$ and past frames $V_{-f:0}$ diminishes. This motivates us to use a reward function that focuses on frames closer to the starting frame:
\begin{equation}
    \tilde{r}_t = \beta r_t\,,
\end{equation}
where $\beta = (T - t) / T$ is a linear decay factor and $r_t$ is defined in  Eq.~\ref{egopose19:eq:reward}. Unlike ego-pose estimation, we do not have new video frame coming as input for each time step $t$, which can lead to ambiguity about the motion phase, such as whether the human is standing up or crouching down. To better encode the phase of human motions, we use a recurrent policy $\pi(a_t|\nu_t, \phi_t)$ where $\nu_t \in \mathbb{R}^{128}$ is the output of a forward LSTM encoding the state forecasts $z_{1:t}$ so far.

\subsection{Fail-safe Mechanism}
\begin{figure}
    \centering
    \includegraphics[width=\linewidth]{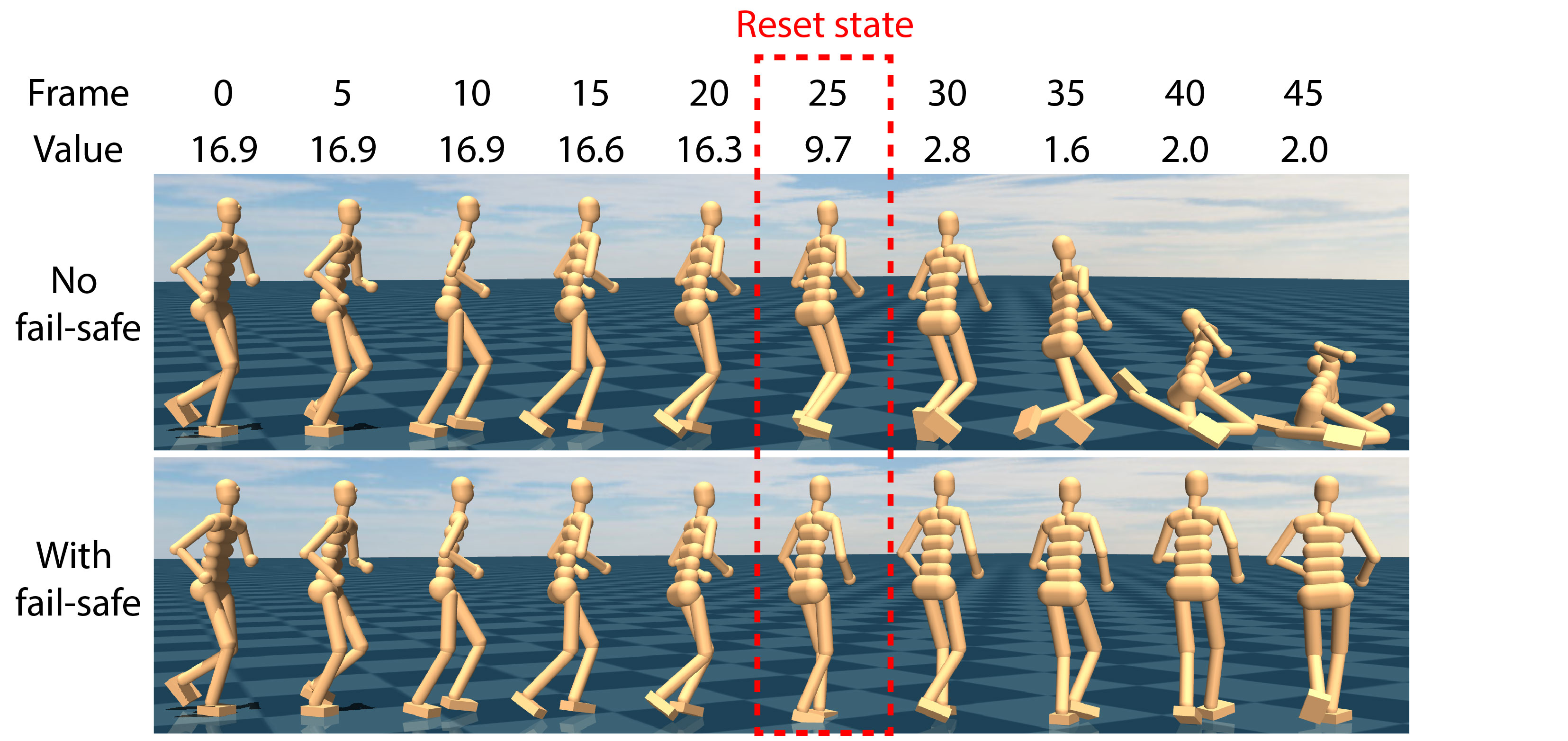}
    \caption{\textbf{Top:} The humanoid at unstable state falls to the ground and the value of the state drops drastically during falling. \textbf{Bottom:} At frame 25, the instability is detected by our fail-safe mechanism, which triggers the state reset and allows our method to keep producing good pose estimates.}
    \label{egopose19:fig:failsafe}
\end{figure}
When running ego-pose estimation at test time, even though the control policy $\pi$ is often robust enough to recover from errors, the humanoid can still fall due to irregular actions caused by extreme domain shifts in the visual input. When the humanoid falls, we need to reset the humanoid state to the output of the state regressor $\mathcal{F}$ to keep producing meaningful pose estimates. However, it is not clear when to do the reset. A naive solution is to reset the state when the humanoid falls to the ground, which will generate a sequence of bad pose estimates during falling (Fig.~\ref{egopose19:fig:failsafe} (Top)). We propose a fail-safe mechanism that can detect the instability of current state before the humanoid starts to fall, which enables us to reset the state before producing bad estimates (Fig.~\ref{egopose19:fig:failsafe} (Bottom)). Most policy gradient methods have an actor-critic structure, where they train the policy $\pi$ alongside a value function $\mathcal{V}$ which estimates the expected discounted return of a state $s$:
\begin{equation}
    \mathcal{V}(s) = \mathbb{E}_{s_1 = s,\,a_t \sim \pi} \left[\sum_{t=1}^T \gamma^{t-1} r_t\right]\,.
\end{equation}
Assuming that $1 / (1 - \gamma) \ll T$, and for a well-trained policy, $r_t$ varies little across time steps, the value function can be approximated as
\begin{equation}
    \mathcal{V}(s)\approx \sum_{t=1}^{\infty} \gamma^{t-1} \bar{r}_s = \frac{1}{1-\gamma}\bar{r}_s\,,
\end{equation}
where $\bar{r}_s$ is the average reward received by the policy starting from state $s$. During our experiments, we find that for state $s$ that is stable (not falling), its value $\mathcal{V}(s)$ is always close to $1/(1-\gamma)\bar{r}$ with little variance, where $\bar{r}$ is the average reward inside a training batch. But when the humanoid begins falling, the value starts dropping significantly (Fig.~\ref{egopose19:fig:failsafe}).
This discovery leads us to the following fail-safe mechanism: when executing the humanoid policy $\pi$, we keep a running estimate of the average state value $\bar{\mathcal{V}}$ and reset the state when we find the value of current state is below $\kappa \bar{\mathcal{V}}$, where $\kappa$ is a coefficient determining how sensitive this mechanism is to instability. We set $\kappa$ to 0.6 in our experiments.

\section{Experimental Setup}
\subsection{Datasets}
\label{egopose19:sec:dataset}
The main dataset we use to test our method is a large MoCap dataset with synchronized egocentric videos. It includes five subjects and is about an hour long. Each subject is asked to wear a head-mounted GoPro camera and perform various complex human motions for multiple takes. The motions consist of walking, jogging, hopping, leaning, turning, bending, rotating, crouching and transitions between these motions. Each take is about one minute long, and we do not segment or label the motions. To further showcase our method's utility, we also collected an in-the-wild dataset where two new subjects are asked to perform similar actions to the MoCap data. It has 24 videos each lasting about 20s. Both indoor and outdoor videos are recorded in different places. Because it is hard to obtain ground-truth 3D poses in real-world environment, we use a third-person camera to capture the side-view of the subject, which is used for evaluation based on 2D keypoints.

\subsection{Baselines}
For ego-pose estimation, we compare our method against three baselines:
\begin{itemize}[leftmargin=*]
\setlength\itemsep{-3pt}
    \item \textbf{VGAIL~\cite{yuan20183d}:} a control-based method that uses joint torques as action space, and learns the control policy with video-conditioned GAIL.
    \item \textbf{PathPose:} an adaptation of a CRF-based method~\cite{jiang2017seeing}. We do not use static scene cues as the training data is from MoCap.
    \item \textbf{PoseReg:} a method that uses our state estimator $\mathcal{F}$ to output the kinematic pose sequence directly. We integrate the linear and angular velocities of the root joint to generate global positions and orientations.
\end{itemize}

For ego-pose forecasting, no previous work has tried to forecast future human poses from egocentric videos, so we compare our approach to methods that forecast future motions using past poses, which at test time is provided by our ego-pose estimation algorithm:
\begin{itemize}[leftmargin=*]
\setlength\itemsep{-3pt}
    \item \textbf{ERD~\cite{fragkiadaki2015recurrent}:} a method that employs an encoder-decoder structure with recurrent layers in the middle, and predicts the next pose using current ground-truth pose as input. It uses noisy input at training to alleviate drift.
    \item \textbf{acLSTM~\cite{li2017auto}:} a method similar to ERD with a different training scheme for more stable long-term prediction: it schedules fixed-length fragments of predicted poses as input to the network.
\end{itemize}

\subsection{Metrics}
To evaluate both the accuracy and physical correctness of our approach, we use the following metrics:
\begin{itemize}[leftmargin=*]
\setlength\itemsep{-3pt}
	\item \textbf{Pose Error} ($\mathbf{E}_{\textrm{pose}}$): a pose-based metric that measures the Euclidean distance between the generated pose sequence $p_{1:T}$ and the ground-truth pose sequence $\hat{p}_{1:T}$. It is calculated as $\frac{1}{T}\sum_{t=1}^T||p_t - \hat{p}_t||_2$.
	
	\item \textbf{2D Keypoint Error} ($\mathbf{E}_{\textrm{key}}$): a pose-based metric used for our in-the-wild dataset. It can be calculated as $\frac{1}{TJ}\sum_{t=1}^T\sum_{j=1}^J||x_t^j - \hat{x}_t^j||_2$, where $x_t^j$ is the $j$-th 2D keypoint of our generated pose and $\hat{x}_t^j$ is the ground truth extracted with OpenPose~\cite{cao2017realtime}. We obtain 2D keypoints for our generated pose by projecting the 3D joints to an image plane with a side-view camera. For both generated and ground-truth keypoints, we set the hip keypoint as the origin and scale the coordinate to make the height between shoulder and hip equal 0.5.
	
	\item \textbf{Velocity Error} ($\mathbf{E}_{\textrm{vel}}$): a physics-based metric that measures the Euclidean distance between the generated velocity sequence $v_{1:T}$ and the ground-truth $\hat{v}_{1:T}$. It is calculated as $\frac{1}{T}\sum_{t=1}^T||v_t - \hat{v}_t||_2$. $v_t$ and $\hat{v}_t$ can be computed by the finite difference method.
	
	\item \textbf{Average Acceleration} ($\mathbf{A}_{\textrm{accl}}$): a physics-based metric that uses the average magnitude of joint accelerations to measure the smoothness of the generated pose sequence. It is calculated as $\frac{1}{TG}\sum_{t=1}^T||\dot{v}_t||_1$ where $\dot{v}_t$ denotes joint accelerations and $G$ is the number of actuated DoFs. 
	
	\item \textbf{Number of Resets} ($\mathbf{N}_{\textrm{reset}}$): a metric for control-based methods (Ours and VGAIL) to measure how frequently the humanoid becomes unstable.
\end{itemize}

\subsection{Implementation Details}
\paragraph{Simulation and Humanoid.}
We use MuJoCo~\cite{todorov2012mujoco} as the physics simulator.
The humanoid model is constructed from the BVH file of a single subject and is shared among other subjects. The humanoid consists of 58 DoFs and 21 rigid bodies with proper geometries assigned. Most non-root joints have three DoFs except for knees and ankles with only one DoF. We do not add any stiffness or damping to the joints, but we add 0.01 armature inertia to stabilize the simulation. We use stable PD controllers~\cite{tan2011stable} to compute joint torques. The gains $k_p$ ranges from 50 to 500 where joints such as legs and spine have larger gains while arms and head have smaller gains. Preliminary experiments showed that the method is robust to a wide range of gains values. $k_d$ is set to $0.1k_p$. We set the torque limits based on the gains.

\paragraph{Networks and Training.} For the video context networks, we use PWC-Net~\cite{sun2018pwc} to compute optical flow and ResNet-18~\cite{he2016deep} pretrained on ImageNet to generate the visual features $\psi_t \in \mathbb{R}^{128}$. To accelerate training, we precompute $\psi_t$ for the policy using the ResNet pretrained for initial state estimation. We use a BiLSTM (estimation) or LSTM (forecasting) to produce the visual context $\phi_t \in \mathbb{R}^{128}$. For the policy, we use online z-filtering to normalize humanoid state $z_t$, and the diagonal elements of the covariance matrix $\Sigma$ are set to {$0.1$}. When training for pose estimation, for each episode we randomly sample a data fragment of 200 frames ({6.33}s) and pad 10 frames of visual features $\psi_t$ on both sides to alleviate border effects when computing $\phi_t$. When training for pose forecasting, we sample 120 frames and use the first 30 frames as context to forecast 90 future frames. We terminate the episode if the humanoid falls or the time horizon is reached. For the reward weights $(w_q, w_e, w_p, w_v)$, we set them to (0.5, 0.3, 0.1, 0.1) for estimation and (0.3, 0.5, 0.1, 0.1) for forecasting. We use PPO~\cite{schulman2017proximal} with a clipping epsilon of 0.2 for policy optimization. The discount factor $\gamma$ is 0.95. We collect trajectories of 50k timesteps at each iteration. We use Adam~\cite{kingma2014adam} to optimize the policy and value function with learning rate 5e-5 and 3e-4 respectively. The policy typically converges after 3k iterations, which takes about 2 days on a GTX 1080Ti.

\begin{figure}
    \centering
    \includegraphics[width=0.95\linewidth]{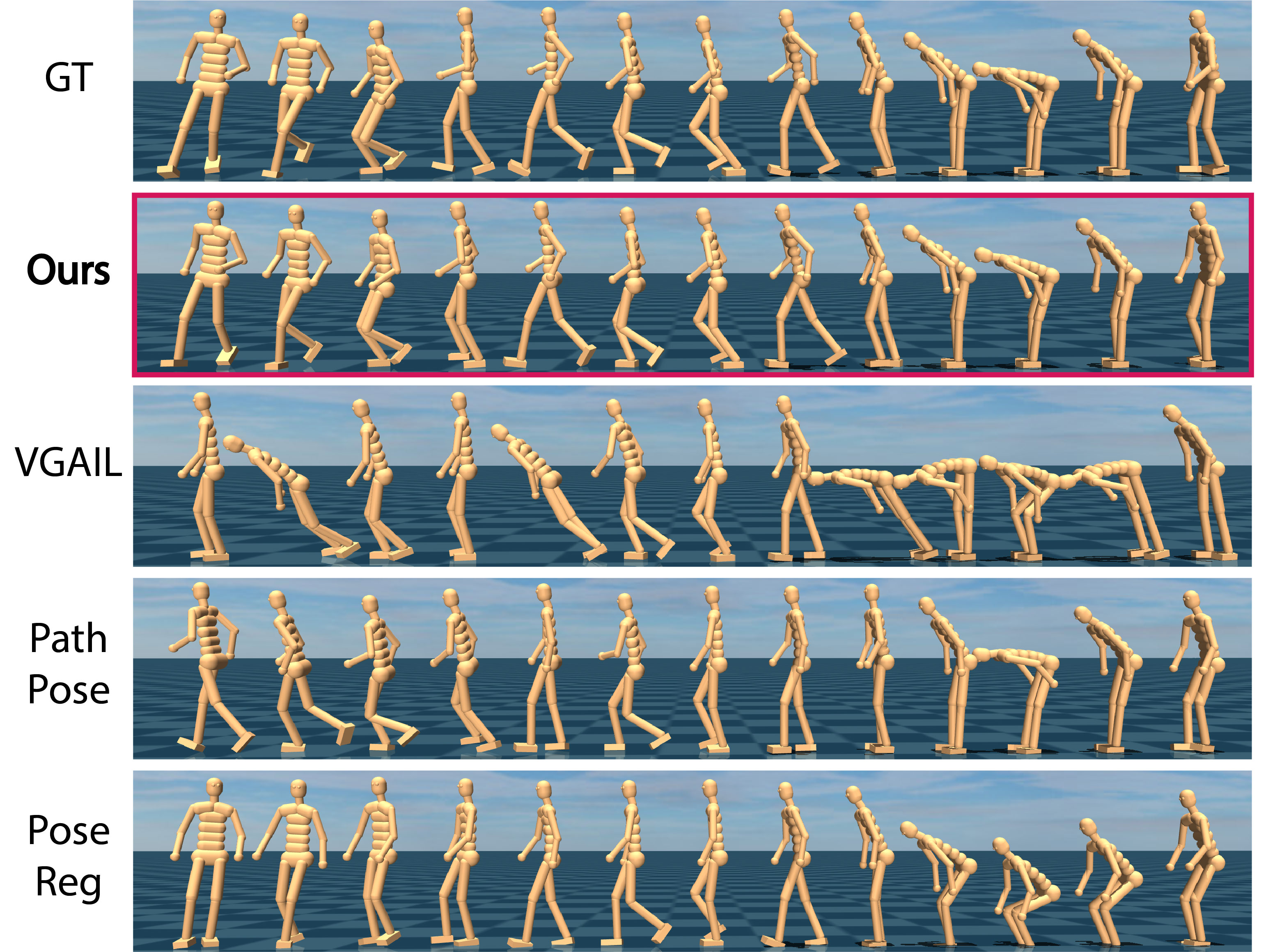}
    \vspace{5mm}
    \caption{Single-subject ego-pose estimation results.}
    \label{egopose19:fig:single_estimate}
\end{figure}

\begin{figure}
    \centering
    \includegraphics[width=0.95\linewidth]{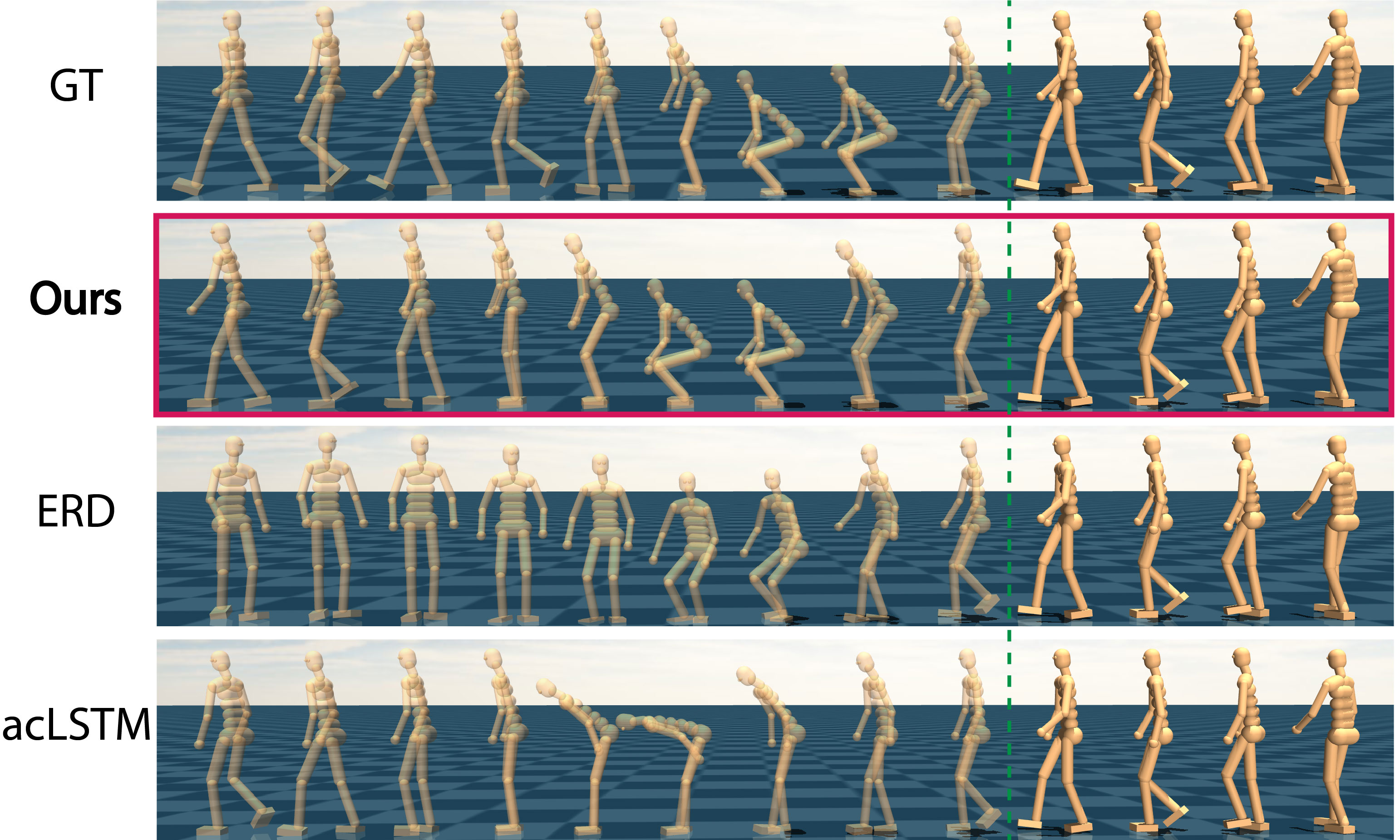}
    \vspace{5mm}
    \caption{Single-subject ego-pose forecasting results.}
    \label{egopose19:fig:single_forecast}
\end{figure}

\begin{figure}
    \centering
    \includegraphics[width=\linewidth]{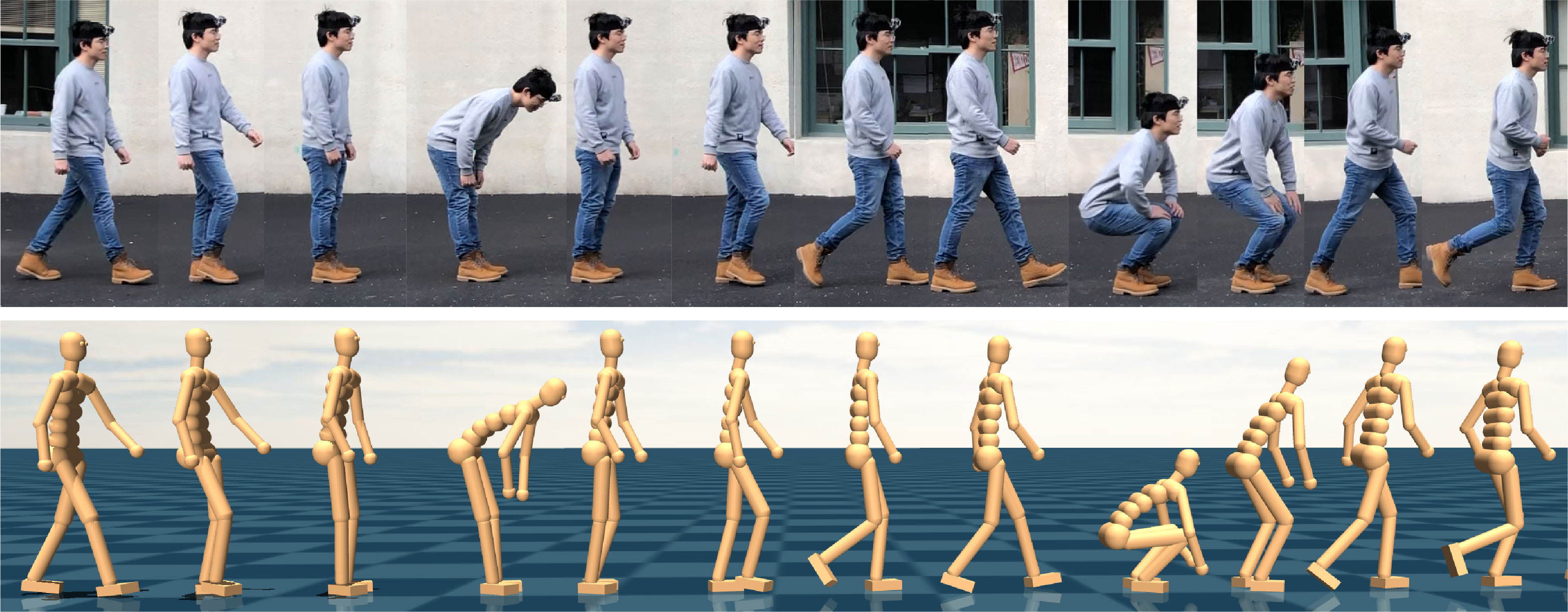}
    \caption{In-the-wild ego-pose estimation results.}
    \label{egopose19:fig:wild_estimate}
\end{figure}

\begin{figure}
    \centering
    \includegraphics[width=\linewidth]{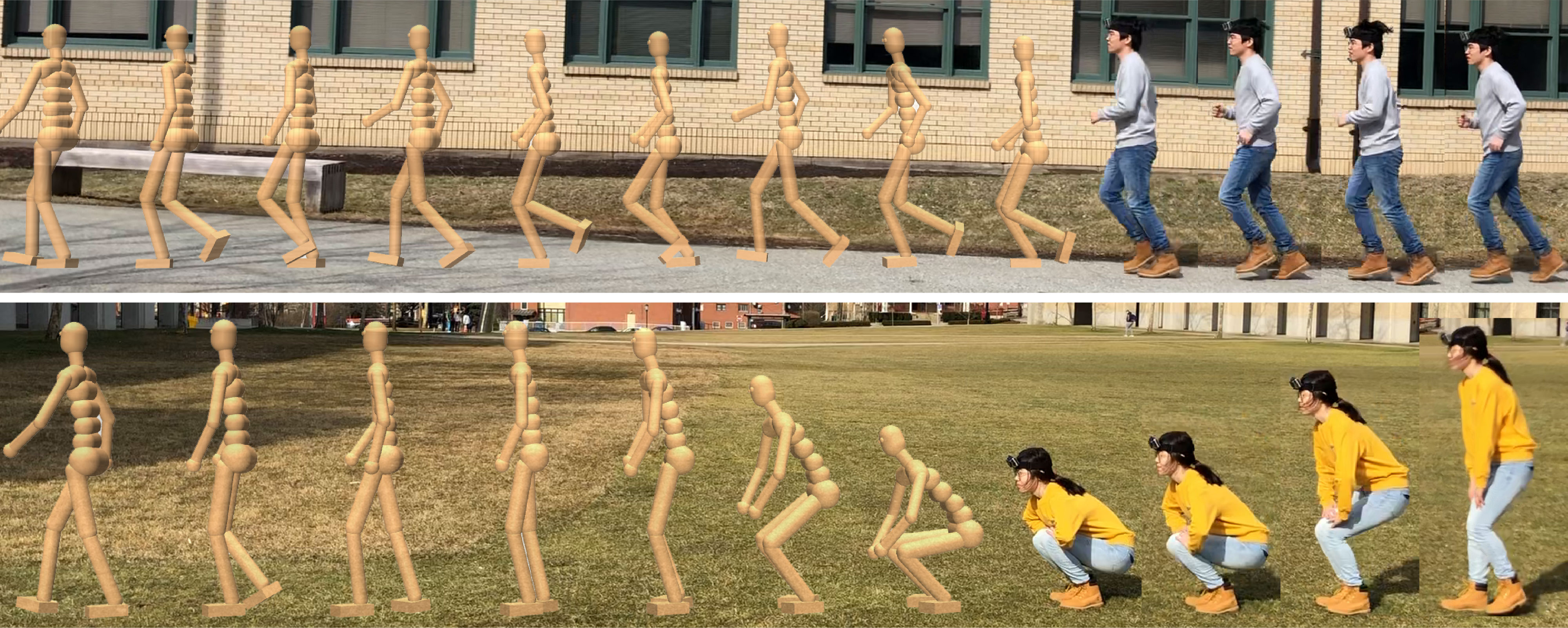}
    \caption{In-the-wild ego-pose forecasting results.}
    \label{egopose19:fig:wild_forecast}
\end{figure}

\section{Results}
\begin{table*}[ht]
\footnotesize
\centering
\begin{tabular}{@{\hskip 1mm}lr@{\hskip -0mm}rrrrrrrrrrrr@{\hskip 1mm}}
\toprule
 \multicolumn{14}{c}{\textsc{\textbf{Ego-pose Estimation}}}\\ \midrule
 && \multicolumn{4}{c}{Single Subject} && \multicolumn{4}{c}{Cross Subjects} & & \multicolumn{2}{c}{In the Wild} \\ \cmidrule{3-6} \cmidrule{8-11} \cmidrule{13-14}
Method && $\mathbf{E}_{\textrm{pose}}$ & $\mathbf{N}_{\textrm{reset}}$ & $\mathbf{E}_{\textrm{vel}}$ & $\mathbf{A}_{\textrm{accl}}$ && $\mathbf{E}_{\textrm{pose}}$ & $\mathbf{N}_{\textrm{reset}}$ & $\mathbf{E}_{\textrm{vel}}$ & $\mathbf{A}_{\textrm{accl}}$ && $\mathbf{E}_{\textrm{key}}$ & $\mathbf{A}_{\textrm{accl}}$ \\ \midrule
Ours   && \textbf{0.640} & \textbf{1.4} & \textbf{4.469} & \textbf{5.002} && \textbf{1.183} & \textbf{4} & \textbf{5.645} & \textbf{5.260} && \textbf{0.099} & \textbf{5.795}\\  
VGAIL~\cite{yuan20183d}  && 0.978 & 94 & 6.561 & 9.631 && 1.316 & 418 & 7.198 & 8.837 && 0.175 & 9.278\\
PathPose~\cite{jiang2017seeing} && 1.035 & -- & 19.135 & 63.526 && 1.637 & -- & 32.454 & 117.499 && 0.147 & 125.406\\
PoseReg && 0.833 & -- & 5.450 & 7.733 && 1.308 & -- & 6.334 & 8.281 && 0.109 & 7.611 \\
\midrule
\midrule
 \multicolumn{14}{c}{\textsc{\textbf{Ego-pose Forecasting}}}\\ \midrule
 && \multicolumn{4}{c}{Single Subject} && \multicolumn{4}{c}{Cross Subjects} & & \multicolumn{2}{c}{In the Wild} \\ \cmidrule{3-6} \cmidrule{8-11} \cmidrule{13-14}
Method && $\mathbf{E}_{\textrm{pose}}$ & $\mathbf{E}_{\textrm{pose}}$(3s) & $\mathbf{E}_{\textrm{vel}}$ & $\mathbf{A}_{\textrm{accl}}$ && $\mathbf{E}_{\textrm{pose}}$ & $\mathbf{E}_{\textrm{pose}}$(3s) & $\mathbf{E}_{\textrm{vel}}$ & $\mathbf{A}_{\textrm{accl}}$ & & $\mathbf{E}_{\textrm{key}}$ & $\mathbf{A}_{\textrm{accl}}$ \\ \midrule
Ours   && \textbf{0.833} & \textbf{1.078} & \textbf{5.456} & \textbf{4.759} && \textbf{1.179} & \textbf{1.339} & \textbf{6.045} & \textbf{4.210} && \textbf{0.114} & \textbf{4.515}\\  
ERD~\cite{fragkiadaki2015recurrent}  && 0.949 & 1.266 & 6.242 & 5.916 && 1.374 & 1.619 & 7.238 & 6.419 && 0.137 & 7.021\\
acLSTM~\cite{li2017auto} && 0.861 & 1.232 & 6.010 & 5.855 && 1.314 & 1.511 & 7.454 & 7.123 && 0.134 & 8.177 \\
\bottomrule
\end{tabular}
\vspace{7mm}
\caption{Quantitative results for egocentric pose estimation and forecasting. For forecasting, by default the metrics are computed inside the first 1s window, except that $\mathbf{E}_{\textrm{pose}}$(3s) are computed in the first 3s window.}
\label{egopose19:table:big}
\end{table*}

\begin{table}[ht]
\footnotesize
\centering
\begin{tabular}{@{\hskip 1mm}lrrrr@{\hskip 1mm}}
\toprule
Method & $\mathbf{N}_{\textrm{reset}}$ & $\mathbf{E}_{\textrm{pose}}$ & $\mathbf{E}_{\textrm{vel}}$ & $\mathbf{A}_{\textrm{accl}}$ \\ \midrule
(a) Ours   & \textbf{4} & \textbf{1.183} & \textbf{5.645} & \textbf{5.260}\\  
(b) Partial reward $r_q + r_e$ & 55 & 1.211 & 5.730 & 5.515\\
(c) Partial reward $r_q$  & 14 & 1.236 & 6.468 & 8.167\\
(d) DeepMimic reward~\cite{peng2018deepmimic} & 52 & 1.515 & 7.413 & 17.504 \\
(e) No fail-safe & 4 & 1.206 & 5.693 & 5.397 \\
\bottomrule
\end{tabular}
\vspace{5mm}
\caption{Ablation study for ego-pose estimation.}
\label{egopose19:table:ablation_estimate}
\end{table}

To comprehensively evaluate performance, we test our method against other baselines in three different experiment settings: (1) single subject in MoCap; (2) cross subjects in MoCap; and (3) cross subjects in the wild. We further conduct an extensive ablation study to show the importance of each technical contributon of our approach. Finally, we show time analysis to validate that our approach can run in real-time.

\paragraph{Subject-Specific Evaluation.}
In this setting, we train an estimation model and a forecasting model for each subject. We use a 80-20 train-test data split. For forecasting, we test every 1s window to forecast poses in the next 3s. The quantitative results are shown in Table~\ref{egopose19:table:big}. For ego-pose estimation, we can see our approach outperforms other baselines in terms of both pose-based metric (pose error) and physics-based metrics (velocity error, acceleration, number of resets). We find that VGAIL~\cite{yuan20183d} is often unable to learn a stable control policy from the training data due to frequent falling, which results in the high number of resets and large acceleration. For ego-pose forecasting, our method is more accurate than other methods for both short horizons and long horizons. We also present qualitative results in Fig.~\ref{egopose19:fig:single_estimate} and~\ref{egopose19:fig:single_forecast}. Our method produces pose estimates and forecasts closer to the ground-truth than any other baseline. 

\paragraph{Cross-Subject Evaluation.} To further test the robustness of our method, we perform cross-subject experiments where we train our models on four subjects and test on the remaining subject. This is a challenging setting since people have very unique style and speed for the same action. As shown in Table~\ref{egopose19:table:big}, our method again outperforms other baselines in all metrics and is surprisingly stable with only a small number of resets. For forecasting, we also show in Table~\ref{egopose19:table:forecast_horizon} how pose error changes across different forecasting horizons. We can see our forecasting method is accurate for short horizons ($<1$s) and even achieves comparable results as our pose estimation method (Table~\ref{egopose19:table:big}).

\paragraph{In-the-Wild Cross-Subject.} To showcase our approach's utility in real-world scenarios, we further test our method on the in-the-wild dataset described in Sec.~\ref{egopose19:sec:dataset}. Due to the lack of 3D ground truth, we make use of accompanying third-person videos and compute 2D keypoint error as the pose metric. As shown in Table~\ref{egopose19:table:big}, our approach is more accurate and smooth than other baselines for real-world scenes. We also present qualitative results in Fig.~\ref{egopose19:fig:wild_estimate} and~\ref{egopose19:fig:wild_forecast}. For ego-pose estimation (Fig.~\ref{egopose19:fig:wild_estimate}), our approach produces very accurate poses and the phase of the estimated motion is synchronized with the ground-truth motion. For ego-pose forecasting (Fig.~\ref{egopose19:fig:wild_forecast}), our method generates very intuitive future motions, as a person jogging will keep jogging forward and a person crouching will stand up and start to walk. %

\paragraph{Ablative Analysis.} The goal of our ablation study is to evaluate the importance of our reward design and fail-safe mechanism. We conduct the study in the cross-subject setting for the task of ego-pose estimation. We can see from Table~\ref{egopose19:table:ablation_estimate} that using other reward functions will reduce performance in all metrics. We note that the large acceleration in (b) and (c) is due to jittery motions generated from unstable control policies. Furthermore, by comparing (e) to (a) we can see that our fail-safe mechanism can improve performance even though the humanoid seldom becomes unstable (only 4 times).

\paragraph{Time analysis.}
We perform time analysis on a mainstream CPU with a GTX 1080Ti using PyTorch implementation of ResNet-18 and PWCNet\footnote{\url{https://github.com/NVlabs/PWC-Net}}.
The breakdown of the processing time is: optical flow 5ms, CNN 20ms, LSTM + MLP 0.2ms, simulation 3ms. The total time per step is $\sim30$ms which translates to 30 FPS. To enable real-time pose estimation which uses a bi-directional LSTM, we use a 10-frame look-ahead video buffer and only encode these 10 future frames with our backward LSTM, which corresponds to a fixed latency of 1/3s. For pose forecasting, we use multi-threading and run the simulation on a separate thread. Forecasting is performed every 0.3s to predict motion 3s (90 steps) into the future. To achieve this, we use a batch size of 5 for the optical flow and CNN (cost is 14ms and 70ms with batch size 1).

\begin{table}
\footnotesize
\centering
\begin{tabular}{@{\hskip 1mm}lrrrrrrrrr@{\hskip 1mm}}
\toprule
Method & 1/3s & 2/3s & 1s & 2s & 3s \\ \midrule
Ours   & \textbf{1.140} & \textbf{1.154} & \textbf{1.179} & \textbf{1.268} & \textbf{1.339}\\  
ERD~\cite{fragkiadaki2015recurrent} & 1.239 & 1.309 & 1.374 & 1.521 & 1.619 \\
acLSTM~\cite{li2017auto}  & 1.299 & 1.297 & 1.314 & 1.425 & 1.511 \\
\bottomrule
\end{tabular}
\vspace{5mm}
\caption{Cross-subject $\mathbf{E}_{\textrm{pose}}$ for different forecasting horizons.}
\label{egopose19:table:forecast_horizon}
\end{table}

\section{Conclusion}
We have proposed the first approach to use egocentric videos to both estimate and forecast 3D human poses. Through the use of a PD control based policy and a reward function tailored to unsegmented human motion data, we showed that our method can estimate and forecast accurate poses for various complex human motions. Experiments and time analysis showed that our approach is robust enough to transfer directly to real-world scenarios and can run in real-time.
\chapter{Stochastic Human Motion Generation with Determinantal Point Processes}
\label{chap:dsf}

\section{Introduction}
\begin{figure}
    \centering
    \includegraphics[width=\linewidth]{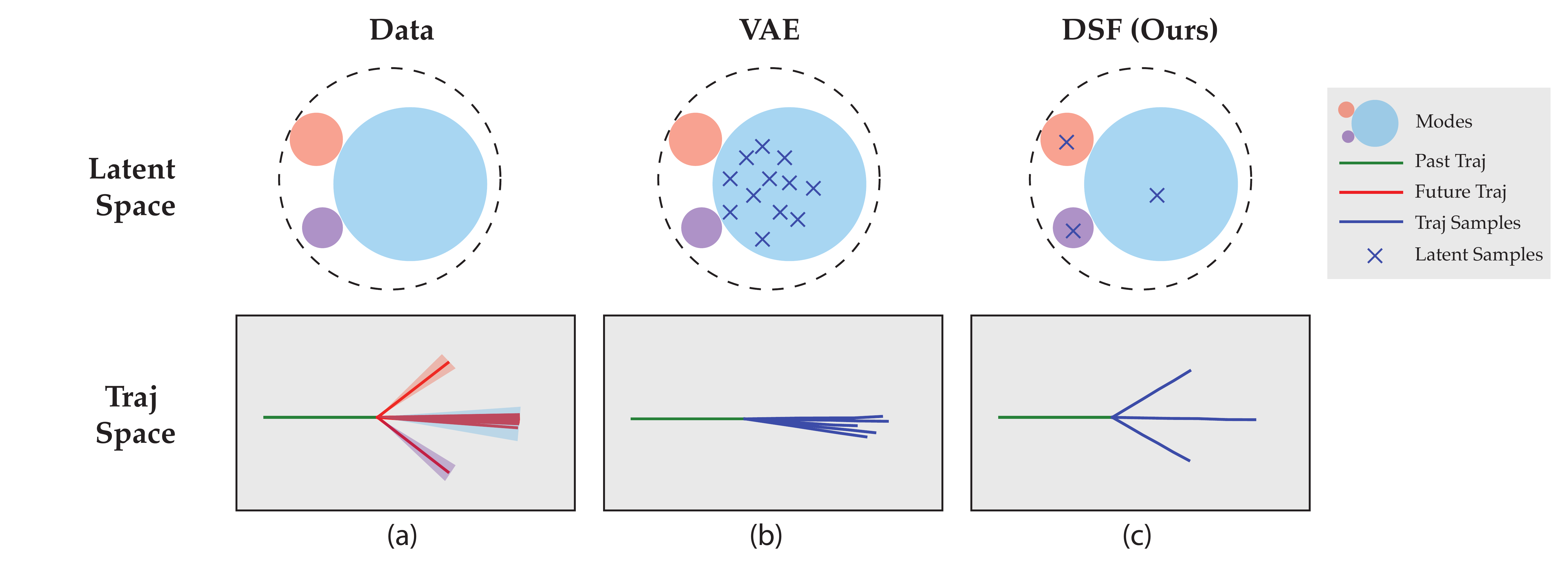}
    \caption{ A toy trajectory forecasting example. (a) The three modes (pink, blue, purple) of the future trajectory distribution are shown in both the trajectory space and the latent space of a learned VAE model. The data distribution is imbalanced, where the blue mode has most data and covers most of the latent space. (b) Random samples from the VAE only cover the major (blue) mode. (c) Our proposed DSF generates a diverse set of future trajectories covering all three modes.}
    \label{dsf:fig:intro}
\end{figure}

Forecasting future trajectories of vehicles and human has many useful applications in autonomous driving, virtual reality and assistive living. What makes trajectory forecasting challenging is that the future is uncertain and multi-modal -- vehicles can choose different routes and people can perform different future actions. In many safety-critical applications, it is important to consider a diverse set of possible future trajectories, even those that are less likely, so that necessary preemptive actions can be taken. For example, an autonomous vehicle should understand that a neighboring car can merge into its lane even though the car is most likely to keep driving straight. To address this requirement, we need to take a generative approach to trajectory forecasting that can fully characterize the multi-modal distribution of future trajectories. To capture all modes of a data distribution, variational autoencoders (VAEs) are well-suited generative models. However, random samples from a learned VAE model with Gaussian latent codes are not guaranteed to be diverse for two reasons. First, the sampling procedure is stochastic and the VAE samples can fail to cover some minor modes even with a large number of samples. Second, since VAE sampling is based on the implicit likelihood function encoded in the training data, if most of the training data is centered around a specific mode while other modes have less data (Fig.~\ref{dsf:fig:intro} (a)), the VAE samples will reflect this bias and concentrate around the major mode (Fig.~\ref{dsf:fig:intro} (b)). To tackle this problem, we propose to learn a diversity sampling function (DSF) that can reliably generate a diverse set of trajectory samples (Fig.~\ref{dsf:fig:intro} (c)).

The proposed DSF is a deterministic parameterized function that maps forecasting context features (\emph{e.g.,} past trajectories) to a set of latent codes. The latent codes are decoded by the VAE docoder into a set of future trajectory samples, denoted as the DSF samples. In order to optimize the DSF, we formulate a diversity loss based on a determinantal point process (DPP) \cite{macchi1975coincidence} to evaluate the diversity of the DSF samples. The DPP defines the probability of choosing a random subset from the set of trajectory samples. It models the negative correlations between samples: the inclusion of a sample reduces the probability of including a similar sample. This makes the DPP an ideal tool for modeling the diversity within a set. In particular, we use the expected cardinality of the DPP as the diversity measure, which is defined as the expected size of a random subset drawn from the set of trajectory samples according to the DPP. Intuitively, since the DPP inhibits selection of similar samples, if the set of trajectory samples is more diverse, the random subset is more likely to select more samples from the set. The expected cardinality of the DPP is easy to compute and differentiable, which allows us to use it as the objective to optimize the DSF to enable diverse trajectory sampling.

Our contributions are as follows: (1) We propose a new forecasting approach that learns a diversity sampling function to produce a diverse set of future trajectories; (2) We propose a novel application of DPPs to optimize a set of items (trajectories) in continuous space with a DPP-based diversity measure; (3) Experiments on synthetic data and human motion validate that our method can reliably generate a more diverse set of future trajectories compared to state-of-the-art generative models.

\section{Related Work}
\paragraph{Trajectory Forecasting\hspace{-0.5em}} has recently received significant attention from the vision community. A large portion of previous work focuses on forecasting 2D future trajectories for pedestrians~\cite{kitani2012activity, ma2017forecasting,  ballan2016knowledge, Xie2013InferringM} or vehicles ~\cite{jain2016recurrent}. Some approaches use deterministic trajectory modeling and only forecast one future trajectory~\cite{alahi2016social, yagi2018future, robicquet2016learning}. As there are often multiple plausible future trajectories, several approaches have tried to forecast distributions over trajectories~\cite{lee2017desire, galceran2015multipolicy, gupta2018social}. Recently, \cite{rhinehart2018r2p2, rhinehart2019precog} propose a generative model that can accurately forecast multi-modal trajectories for vehicles. \cite{soo2016egocentric} also use egocentric videos to predict the future trajectories of the camera wearer. Some work has investigated forecasting higher dimensional trajectories such as the 3D full-body pose sequence of human motions. Most existing work takes a deterministic approach and forecasts only one possible future motion from past 3D poses \cite{fragkiadaki2015recurrent, butepage2017deep, li2017auto, jain2016structural}, static images \cite{chao2017forecasting, kanazawa2018learning} or egocentric videos \cite{yuan2019ego}. Differently, some probabilistic approaches \cite{habibie2017recurrent, yan2018mt} use conditional variational autoencoders (cVAEs) to generate multiple future motions. In constrast to previous work, our approach can generate a \emph{diverse} set of future motions with a limited number of samples.

\paragraph{Diverse Solutions\hspace{-0.5em}} have been sought after in a number of problems in computer vision and machine learning. A branch of these methods aiming for diversity stems from the M-Best MAP problem~\cite{nilsson1998efficient, seroussi1994algorithm}, including diverse M-Best solutions~\cite{batra2012diverse} and multiple choice learning~\cite{guzman2012multiple, lee2016stochastic}. Alternatively, previous work has used submodular function maximization to select a diverse subset of garments from fashion images~\cite{hsiao2018creating}. Determinantal point processes (DPPs)~\cite{macchi1975coincidence, kulesza2012determinantal} are efficient probabilistic models that can measure both the diversity and quality of items in a subset, which makes it a natural choice for the diverse subset selection problem. DPPs have been applied for document and video summarization~\cite{kulesza2011k, gong2014diverse}, recommendation systems~\cite{gillenwater2014expectation}, object detection~\cite{azadi2017learning}, and grasp clustering~\cite{huang2015we}. \cite{elfeki2018gdpp} have also used DPPs to mitigate mode collapse in generative adversarial networks (GANs). The work most related ours is~\cite{gillenwater2014expectation}, which also uses the cardinality of DPPs as a proxy for user engagement. However, there are two important differences between our approach and theirs. First, the context is different as they use the cardinality for a subset selection problem while we apply the cardinality as an objective of a continuous optimization problem in the setting of generative models. Second, their main motivation behind using the cardinality is that it aligns better with the user engagement semantics, while our motivation is that using cardinality as a diversity loss for deep neural networks is more stable due to its tolerance of similar trajectories, which are often produced by deep neural networks during stochastic gradient descent.

\section{Background}
\subsection{Variational Autoencoders}
\label{dsf:sec:vae}
The aim of multi-modal trajectory forecasting is to learn a generative model over future trajectories. Variational autoencoders (VAEs) are a popular choice of generative models for trajectory forecasting~\cite{lee2017desire, walker2016uncertain} because it can effectively capture all possible future trajectories by explicitly mapping each data point to a latent code. VAEs model the joint distribution $p_\theta(\mathbf{x}, \mathbf{z})= p(\mathbf{z})p_\theta(\mathbf{x} | \mathbf{z})$ of each data sample $\mathbf{x}$ (\emph{e.g.}, a future trajectory) and its corresponding latent code $\mathbf{z}$, where $p(\mathbf{z})$ denotes some prior distribution (\emph{e.g.}, Gaussians) and $p_\theta(\mathbf{x} | \mathbf{z})$ denotes the conditional likelihood model. To calculate the marginal likelihood $p_{\theta}(\mathbf{x}) = p_\theta(\mathbf{x}, \mathbf{z}) / p_\theta(\mathbf{z}|\mathbf{x})$, one needs to compute the posterior distribution $p_\theta(\mathbf{z}|\mathbf{x})$ which is typically intractable. To tackle this, VAEs use variational inference~\cite{jordan1999introduction} which introduces an approximate posterior $q_\phi(\mathbf{z}|\mathbf{x})$ and decomposes the marginal log-likelihood as
\begin{equation}
    \log p_{\theta}(\mathbf{x})=\operatorname{KL}\left(q_{\phi}(\mathbf{z} | \mathbf{x}) \| p_{\theta}(\mathbf{z} | \mathbf{x})\right)+\mathcal{L}(\mathbf{x} ; \theta, \phi)\,,
\end{equation}
where $\mathcal{L}(\mathbf{x} ; \theta, \phi)$ is the evidence lower bound (ELBO) defined as
\begin{equation}
    \mathcal{L}(\mathbf{x} ; \theta, \phi)=\mathbb{E}_{q_{\phi}(\mathbf{z} | \mathbf{x})}\left[\log p_{\theta}(\mathbf{x} | \mathbf{z})\right]-\operatorname{KL}\left(q_{\phi}(\mathbf{z} | \mathbf{x}) \| p(\mathbf{z})\right).
\end{equation}
During training, VAEs jointly optimize the recognition model (encoder) $q_{\phi}(\mathbf{z} | \mathbf{x})$ and the likelihood model (decoder) $p_{\theta}(\mathbf{x} | \mathbf{z})$ to maximize the ELBO. In the context of multi-modal trajectory forecasting, one can generate future trajectories from $p(\mathbf{x})$ by drawing a latent code $\mathbf{z}$ from the prior $p(\mathbf{z})$ and decoding $\mathbf{z}$ with the decoder $p_{\theta}(\mathbf{x} | \mathbf{z})$ to produce a corresponding future trajectory~$\mathbf{x}$.

\subsection{Determinantal Point Processes}
\label{dsf:sec:dpp}

Our core technical innovation is a method to learn a \emph{diversity} sampling function (DSF) that can generate a diverse set of future trajectories. To achieve this, we must equip ourselves with a tool to evaluate the diversity of a set of trajectories. To this end, we make use of determinantal point processes (DPPs) to model the diversity within a set. DPPs promote diversity within a set because the inclusion of one item makes the inclusion of a similar item less likely if the set is sampled according to a DPP.

Formally, given a set of items (\emph{e.g.,} data points) $\mathcal{Y} = \{\mathbf{x}_1, \ldots, \mathbf{x}_N\}$, a point process $\mathcal{P}$ on the ground set $\mathcal{Y}$ is a probability measure on $2^{\mathcal{Y}}$, \emph{i.e.}, the set of all subsets of $\mathcal{Y}$. $\mathcal{P}$ is called a determinantal point process if a random subset $\boldsymbol{Y}$ drawn according to $\mathcal{P}$ has
\begin{equation}
    \mathcal{P}_{\mathbf{L}}(\boldsymbol{Y}=Y)=\frac{\det\left(\mathbf{L}_{Y}\right)}{\sum_{Y \subseteq \mathcal{Y}} \det\left(\mathbf{L}_{Y}\right)}=\frac{\det\left(\mathbf{L}_{Y}\right)}{\det(\mathbf{L}+\mathbf{I})}\,,
\end{equation}
where $Y \subseteq \mathcal{Y}$, $\mathbf{I}$ is the identity matrix, $\mathbf{L} \in \mathbb{R}^{N\times N}$ is the DPP kernel, a symmetric positive semidefinite matrix, and $\mathbf{L}_Y \in \mathbb{R}^{|Y|\times |Y|}$ is a submatrix of $\mathbf{L}$ indexed by elements of $Y$.

The DPP kernel $\mathbf{L}$ is typically constructed by a \emph{similarity} matrix $\mathbf{S}$, where $\mathbf{S}_{ij}$ defines the similarity between two items $\mathbf{x}_i$ and $\mathbf{x}_j$. If we use the inner product as the similarity measure, $\mathbf{L}$ can be written in the form of a Gram matrix $\mathbf{L} = \mathbf{S} =  \mathbf{X}^T\mathbf{X}$ where $\mathbf{X}$ is the stacked feature matrix of $\mathcal{Y}$. As a property of the Gram matrix, $\det\left(\mathbf{L}_{Y}\right)$ equals the squared volume spanned by vectors $\mathbf{x}_i \in Y$. With this geometric interpretation in mind, one can observe that diverse sets are more probable because their features are more orthogonal, thus spanning a larger volume.

In addition to set diversity encoded in the similarity matrix $\mathbf{S}$, it is also convenient to introduce a \textit{quality} vector $\mathbf{r} = [r_1, \ldots, r_N]$ to weigh each item according to some unary metric. For example, the quality weight might be derived from the likelihood of an item. To capture both diversity and quality of a subset, the DPP kernel $\mathbf{L}$ is often decomposed in the more general form:
\begin{equation}
    \mathbf{L} = \text{Diag}(\mathbf{r})\cdot \mathbf{S} \cdot\text{Diag}(\mathbf{r})\,.
\end{equation}
With this decomposition, we can see that both the quality vector $\mathbf{r}$ and similarity matrix $\mathbf{S}$ contribute to the DPP probability of a subset $Y$:
\begin{equation}
    \mathcal{P}_{L}(\boldsymbol{Y}=Y) \propto \det\left(\mathbf{L}_{Y}\right)=\left(\prod_{\mathbf{x}_i \in Y} r_i^2\right) \det\left(\mathbf{S}_Y\right).
\end{equation}
Due to its ability to capture the global diversity and quality of a set of items, we choose DPPs as the probabilistic approach to evaluate and optimize the diversity of the future trajectories drawn by our proposed diversity sampling function.

\section{Approach}
Safety-critical applications often require that the system can maintain a diverse set of outcomes covering all modes of a predictive distribution and not just the most likely one. To address this requirement, we propose to learn a diversity sampling function (DSF) to draw deterministic trajectory samples by generating a set of latent codes in the latent space of a conditional variational autoencoder (cVAE) and decoding them into trajectories using the cVAE decoder. The DSF trajectory samples are evaluated with a DPP-based diversity loss, which in turn optimizes the parameters of the DSF for more diverse trajectory samples.

Formally, the future trajectory $\mathbf{x} \in \mathbb{R}^{T\times D}$ is a random variable denoting a $D$ dimensional feature over a future time horizon $T$ (\emph{e.g.,} a vehicle trajectory or a sequence of humanoid poses). The context $\boldsymbol{\psi} =\{\mathbf{h},\mathbf{f}\}$ provides the information to infer the future trajectory $\mathbf{x}$, and it contains the past trajectory $\mathbf{h} \in \mathbb{R}^{H\times D}$ of last $H$ time steps and optionally other side information $\mathbf{f}$, such as an obstacle map. In the following, we first describe how we learn the future trajectory model $p_\theta(\mathbf{x}|\boldsymbol{\psi})$ with a cVAE. Then, we introduce the DSF and the DPP-based diversity loss used to optimize the DSF.

\subsection{Learning a cVAE for Future Trajectories}

 In order to generate a diverse set of future trajectory samples, we need to learn a generative trajectory forecasting model $p_\theta(\mathbf{x}|\boldsymbol{\psi})$ that can cover all modes of the data distribution. Here we use cVAEs (other proper generative models can also be used), which explicitly map data $\mathbf{x}$ with the encoder $q_\phi(\mathbf{z}|\mathbf{x}, \boldsymbol{\psi})$ to its corresponding latent code $\mathbf{z}$ and reconstruct the data from the latent code using the decoder $p_\theta(\mathbf{x}|\mathbf{z}, \boldsymbol{\psi})$. By maintaining this one-on-one mapping between the data and the latent code, cVAEs attempt to capture all modes of the data. As discussed in Sec.~\ref{dsf:sec:vae}, cVAEs jointly optimize the encoder and decoder to maximize the variational lower bound:
\begin{equation}
\label{dsf:eq:cvae}
    \mathcal{L}(\mathbf{x}, \boldsymbol{\psi} ; \theta, \phi)=\mathbb{E}_{q_{\phi}(\mathbf{z} | \mathbf{x}, \boldsymbol{\psi})}\left[\log p_{\theta}(\mathbf{x} | \mathbf{z}, \boldsymbol{\psi})\right]-\operatorname{KL}\left(q_{\phi}(\mathbf{z} | \mathbf{x}, \boldsymbol{\psi}) \| p(\mathbf{z})\right).
\end{equation}
We use multivariate Gaussians for the prior, encoder and decoder: $p(\mathbf{z})=\mathcal{N}(\mathbf{z} ; \mathbf{0}, \mathbf{I})$,  $q_{\phi}(\mathbf{z} | \mathbf{x}, \boldsymbol{\psi}) = \mathcal{N}(\mathbf{z} ; \boldsymbol{\mu}, \boldsymbol{\sigma}^2\mathbf{I})$, and $p_\theta(\mathbf{x}|\mathbf{z}, \boldsymbol{\psi}) = \mathcal{N}(\mathbf{x} ; \tilde{\mathbf{x}}, \alpha\mathbf{I})$.
Both the encoder and decoder are implemented as neural networks.
The encoder network $f_\phi$ outputs the parameters of the posterior distribution: $(\boldsymbol{\mu}, \boldsymbol{\sigma}) = f_\phi(\mathbf{x}, \boldsymbol{\psi})$. The decoder network $g_\theta$ outputs the reconstructed future trajectory $\tilde{\mathbf{x}}$: $\tilde{\mathbf{x}} = g_\theta(\mathbf{z}, \boldsymbol{\psi})$. Based on the Gaussian parameterization of the cVAE, the objective in Eq.~\ref{dsf:eq:cvae} can be rewritten as
\begin{equation}
    \mathcal{L}_{cvae}(\mathbf{x}, \boldsymbol{\psi} ; \theta, \phi) = -\frac{1}{V}\sum_{v=1}^V\|\tilde{\mathbf{x}}_v - \mathbf{x}\|^2 + \beta \cdot \frac{1}{D_z}\sum_{j=1}^{D_z} \left(1+2\log \sigma_j-\mu_{j}^{2}-\sigma_{j}^{2}\right),
\end{equation}
where we take $V$ samples from the posterior $q_{\phi}(\mathbf{z} | \mathbf{x}, \boldsymbol{\psi})$, $D_z$ is the number of latent dimensions, and $\beta = 1 / \alpha$ is a weighting factor. Once the cVAE model is trained, sampling from the learned future trajectory model $p_\theta(\mathbf{x}|\boldsymbol{\psi})$ is efficient: we can sample a latent code $\mathbf{z}$ according to the prior $p(\mathbf{z})$ and use the decoder $p_\theta(\mathbf{x}|\mathbf{z}, \boldsymbol{\psi})$ to decode it into a future trajectory $\mathbf{x}$.

\begin{algorithm}
\caption{Training the diversity sampling function (DSF) $\mathcal{S}_\gamma(\boldsymbol{\psi})$}
\label{alg:dsf}
\begin{algorithmic}[1]
\State \textbf{Input:} Training data $\{\mathbf{x}^{(i)}, \boldsymbol{\psi}^{(i)}\}_{i=1}^M$, cVAE decoder network $g_\theta(\mathbf{z}, \boldsymbol{\psi})$
\State \textbf{Output:} DSF $\mathcal{S}_\gamma(\boldsymbol{\psi})$
\State Initialize $\gamma$ randomly
\While{not converged}
\For{ each $\boldsymbol{\psi}^{(i)}$ }
    \State Generate latent codes $\mathcal{Z} = \{\mathbf{z}_1, \ldots, \mathbf{z}_N\}$ with the DSF $\mathcal{S}_\gamma(\boldsymbol{\psi})$
    \State Generate the trajectory ground set $\mathcal{Y} = \{\mathbf{x}_1, \ldots, \mathbf{x}_N\}$ with the decoder $g_\theta(\mathbf{z}, \boldsymbol{\psi})$
    \State Compute the similarity matrix $\mathbf{S}$ and quality vector $\mathbf{r}$ with
    Eq.~\ref{dsf:eq:sim} and ~\ref{dsf:eq:qual}
    \State Compute the DPP kernel $\mathbf{L}(\gamma) = \text{Diag}(\mathbf{r})\cdot \mathbf{S} \cdot\text{Diag}(\mathbf{r})$
    \State Calculate the diversity loss $\mathcal{L}_{diverse}$
    \State Update $\gamma$ with the gradient $\nabla \mathcal{L}_{diverse}$
\EndFor
\EndWhile
\end{algorithmic}
\end{algorithm}

\subsection{Diversity Sampling Function (DSF)}
\label{dsf:sec:dsf}
As mentioned before, randomly sampling from the learned cVAE model according to the implicit likelihood function $p_\theta(\mathbf{x}|\boldsymbol{\psi})$, \emph{i.e.,} sampling latent codes from the prior $p(\mathbf{z})$, does not guarantee that the trajectory samples are diverse: major modes (those having more data) with higher likelihood will produce most of the samples while minor modes with lower likelihood will have almost no sample. This prompts us to devise a new sampling strategy that can reliably generate a diverse set of samples covering both major and minor modes. We propose to learn a diversity sampling function (DSF) $\mathcal{S}_\gamma(\boldsymbol{\psi})$ that maps context $\boldsymbol{\psi}$ to a set of latent codes $\mathcal{Z} = \{\mathbf{z}_1, \ldots, \mathbf{z}_N\}$. The DSF is implemented as a $\gamma$-parameterized neural network which takes $\boldsymbol{\psi}$ as input and outputs a vector of length $N \cdot D_z$. The latent codes $\mathcal{Z}$ are decoded into a diverse set of future trajectories $\mathcal{Y} = \{\mathbf{x}_1, \ldots, \mathbf{x}_N\}$, which are denoted as the DSF trajectory samples. We note that $N$ is the sampling budget. To solve for the parameters of the DSF, we propose a diversity loss based on a DPP defined over $\mathcal{Y}$. In this section, we first describe how the DPP kernel $\mathbf{L}$ is defined, which involves the construction of the similarity matrix $\mathbf{S}$ and quality vector $\mathbf{r}$. We then discuss how we use the DPP kernel $\mathbf{L}$ to formulate a diversity loss to optimize the parameters of the DSF.

Recall that the DPP kernel is defined as $\mathbf{L} = \text{Diag}(\mathbf{r})\cdot \mathbf{S} \cdot\text{Diag}(\mathbf{r})$, where $\mathbf{r}$ defines the quality of each trajectory and $\mathbf{S}$ measures the similarity between two trajectories. 
The DPP kernel $\mathbf{L}(\gamma)$ is a function of $\gamma$ as it is defined over the ground set $\mathcal{Y}$ output by the DSF $\mathcal{S}_\gamma(\boldsymbol{\psi})$.

\paragraph{Similarity.}
We measure the similarity $\mathbf{S}_{ij}$ between two trajectories $\mathbf{x}_i$ and $\mathbf{x}_j$ as
\begin{equation}
\label{dsf:eq:sim}
    \mathbf{S}_{ij} = \exp\left(-k \cdot d^2_\mathbf{x}(\mathbf{x}_i, \mathbf{x}_j)\right),
\end{equation}
where $d_\mathbf{x}$ is the Euclidean distance and $k$ is a scaling factor. This similarity design ensures that $0 \leq \mathbf{S}_{ij} \leq 1$ and $\mathbf{S}_{ii} = 1$. It also makes $\mathbf{S}$ positive definite since the Gaussian kernel we use is a positive definite kernel.

\paragraph{Quality.}
It may be tempting to use $p(\mathbf{x}|\boldsymbol{\psi})$ to define the quality of each trajectory sample. However, this likelihood-based measure will clearly favor major modes that have higher probabilities, making it less likely to generate samples from minor modes. This motivates us to design a quality metric that treats all modes equally. To this end, unlike the similarity metric which is defined in the trajectory space, the quality of each sample is measured in the latent space and is defined as
\begin{equation}
\label{dsf:eq:qual}
    r_i = 
    \begin{cases}
    \omega, & \text{if } \|\mathbf{z}_i\| \leq R\\
     \omega\exp\left(-\mathbf{z}_i^T\mathbf{z}_i + R^2\right),              & \text{otherwise}
    \end{cases}
\end{equation}
Geometrically, let $R$ be the radius of a sphere $\Phi$ containing most samples from the Gaussian prior $p(\mathbf{z})$. We treat samples inside $\Phi$ equally and only penalize samples outside $\Phi$. In this way, samples from major modes are not preferred over those from minor modes as long as they are inside $\Phi$, while samples far away from the data manifold are heavily penalized as they are outside $\Phi$. The radius $R$ is determined by where $\rho$ percent of the Gaussian samples lie within, and we set $\rho=90$. To compute $R$, we use the percentage point function of the chi-squared distribution which models the distribution over the sum of squares of independent standard normal variables. The base quality $\omega$ is a hyperparameter which we set to 1 during training in our experiments. At test time, we can use a larger $\omega$ to encourage the DPP to select more items from the ground set $\mathcal{Y}$. The hyperparameter $\rho$ (or $R$) allows for the trade-off between diversity and quality. When $R$ is small, the quality metric is reduced to a pure likelihood-based metric (proportional to the latent likelihood), which will prefer samples with high likelihood and result in a less diverse sample set. When $R$ is large, most samples will have the same quality, and the resulting samples will be highly diverse but less likely. In practice, the choice of $R$ should be application dependent, as one could imagine autonomous vehicles would need to consider more diverse scenarios including those less likely ones to ensure robustness. We note that after the diverse samples are obtained, it is possible to reassign the quality score for each sample based on its likelihood to allow users to prioritize more likely samples.

\paragraph{Diversity Loss.} To optimize the DSF $\mathcal{S}_\gamma(\boldsymbol{\psi})$, we need to define a diversity loss that measures the diversity of the trajectory ground set $\mathcal{Y}$ generated by $\mathcal{S}_\gamma(\boldsymbol{\psi})$. An obvious choice for the diversity loss would be the negative log likelihood $-\log \mathcal{P}_{\mathbf{L}(\gamma)}(\boldsymbol{Y} = \mathcal{Y}) = -\log \det(\mathbf{L}(\gamma)) + \log \det(\mathbf{L}(\gamma) + \mathbf{I})$. However, there is a problem with directly using the DPP log likelihood. The log likelihood heavily penalizes repeated items: if two trajectories inside $\mathcal{Y}$ are very similar, their corresponding rows in $\mathbf{L}$ will be almost identical, making $\det(\mathbf{L}(\gamma)) = \lambda_1\lambda_2\ldots \lambda_N \approx 0$ ($\lambda_n$ is the $n$-th eigenvalue). In practice, if the number of modes in the trajectory distribution $p(\mathbf{x}|\boldsymbol{\psi})$ is smaller than $|\mathcal{Y}|$, $\mathcal{Y}$ will always have similar trajectories, thus making $\det(\mathbf{L}(\gamma))$ always close to zero. In such cases, optimizing the negative log likelihood causes numerical issues, which is observed in our early experiments.

Instead, the expected cardinality of the DPP is a better measure for the diversity of $\mathcal{Y}$, which is defined as $\mathbb{E}_{\boldsymbol{Y}\sim \mathcal{P}_{\mathbf{L}(\gamma)}}[|\boldsymbol{Y}|]$. Intuitively, since the DPP discourages selection of similar items, if $\mathcal{Y}$ is more diverse, a random subset $\boldsymbol{Y}$ drawn according to the DPP is more likely to select more items from $\mathcal{Y}$, thus having larger cardinality. The expected cardinality can be computed as (Eq. 15 and 34 in~\cite{kulesza2012determinantal}): 
\begin{equation}
    \mathbb{E}[|\boldsymbol{Y}|] = \sum_{n=1}^{N} \frac{\lambda_{n}}{\lambda_{n}+1} = \text{tr}\left(\mathbf{I} - (\mathbf{L}(\gamma) + \mathbf{I})^{-1}\right).
\end{equation}
The main advantage of the expected cardinality is that it is well defined even when the ground set $\mathcal{Y}$ has duplicated items, since it does not require all eigenvalues of $\mathbf{L}$ to be non-zero as the log likelihood does. Thus, our diversity loss is defined as %
\begin{equation}
\label{dsf:eq:div}
    \mathcal{L}_{diverse}(\gamma) = -\text{tr}\left(\mathbf{I} - (\mathbf{L}(\gamma) + \mathbf{I})^{-1}\right).
\end{equation}
The training procedure for $\mathcal{S}_\gamma(\boldsymbol{\psi})$ is outlined in Alg.~\ref{alg:dsf}.

\begin{algorithm}
\caption{Inference with the DSF $\mathcal{S}_\gamma(\boldsymbol{\psi})$}
\label{alg:dsf_inf}
\begin{algorithmic}[1]
\State \textbf{Input:} Context $\boldsymbol{\psi}$, DSF $\mathcal{S}_\gamma(\boldsymbol{\psi})$, cVAE decoder network $g_\theta(\mathbf{z}, \boldsymbol{\psi})$
\State \textbf{Output:} Forecasted trajectory set $Y_f$
\State Generate latent codes $\mathcal{Z} = \{\mathbf{z}_1, \ldots, \mathbf{z}_N\}$ with the DSF $\mathcal{S}_\gamma(\boldsymbol{\psi})$
\State Generate the trajectory ground set $\mathcal{Y} = \{\mathbf{x}_1, \ldots, \mathbf{x}_N\}$ with the decoder $g_\theta(\mathbf{z}, \boldsymbol{\psi})$ 
\State Compute the DPP kernel $\mathbf{L} = \text{Diag}(\mathbf{r})\cdot \mathbf{S} \cdot\text{Diag}(\mathbf{r})$
\State $Y_f \leftarrow \emptyset, U \leftarrow \mathcal{Y}$
\While{$U$ is not empty}
\State $\mathbf{x}^\ast \leftarrow \argmax_{\mathbf{x}\in U} \> \log \operatorname{det}\left(\mathbf{L}_{Y_f \cup\left\{\mathbf{x}\right\}}\right)$
\If{$\log \operatorname{det}\left(\mathbf{L}_{Y_f \cup\left\{\mathbf{x^\ast}\right\}}\right) - \log \operatorname{det}\left(\mathbf{L}_{Y_f}\right) < 0$}
\State \textbf{break}
\EndIf
\State $Y_f \leftarrow Y_f \cup\left\{\mathbf{x}^\ast\right\}$
\State $U \leftarrow U \setminus \left\{\mathbf{x}^\ast\right\}$
\EndWhile
\end{algorithmic}
\end{algorithm}

\paragraph{Inference.}
At test time, given current context $\boldsymbol{\psi}, $we use the learned DSF $\mathcal{S}_\gamma(\boldsymbol{\psi})$ to generate the future trajectory ground set $\mathcal{Y}$. In some cases, $\mathcal{Y}$ may still contain some trajectories that are similar to others. In order to obtain a diverse set of trajectories without repetition, we aim to perform MAP inference for the DPP to find the most diverse subset $Y^\ast = \argmax_{Y\in \mathcal{Y}} \mathcal{P}_{\mathbf{L}(\gamma)}(Y)$. A useful property of DPPs is that the log-probability function is submodular~\cite{gillenwater2012near}. Even though submodular maximization is NP-hard, we use a greedy algorithm~\cite{nemhauser1978analysis} which is a popular optimization procedure that works well in practice. As outlined in Alg.~\ref{alg:dsf_inf}, the output set $Y_{f}$ is initialized as $\emptyset$, and at each iteration, the trajectory which maximizes the log probability
\begin{equation}
    \mathbf{x}^\ast = \argmax_{\mathbf{x}\in\mathcal{Y} \setminus Y_f} \>\> \log \operatorname{det}\left(\mathbf{L}_{Y_f \cup\left\{\mathbf{x}\right\}}\right)
\end{equation}
is added to $Y_f$, until the marginal gain becomes negative or $Y_{f} = \mathcal{Y}$.

\section{Experiments}
The primary focus of our experiments is to answer the following questions: (1) Are trajectory samples generated with our diversity sampling function more diverse than samples from the cVAE and other baselines? (2) How does our method perform on both balanced and imbalanced data? (3) Is our method general enough to perform well on both low-dimensional and high-dimensional tasks?

\begin{wrapfigure}{r}{0.45\textwidth}
  \begin{center}
    \includegraphics[width=0.45\textwidth]{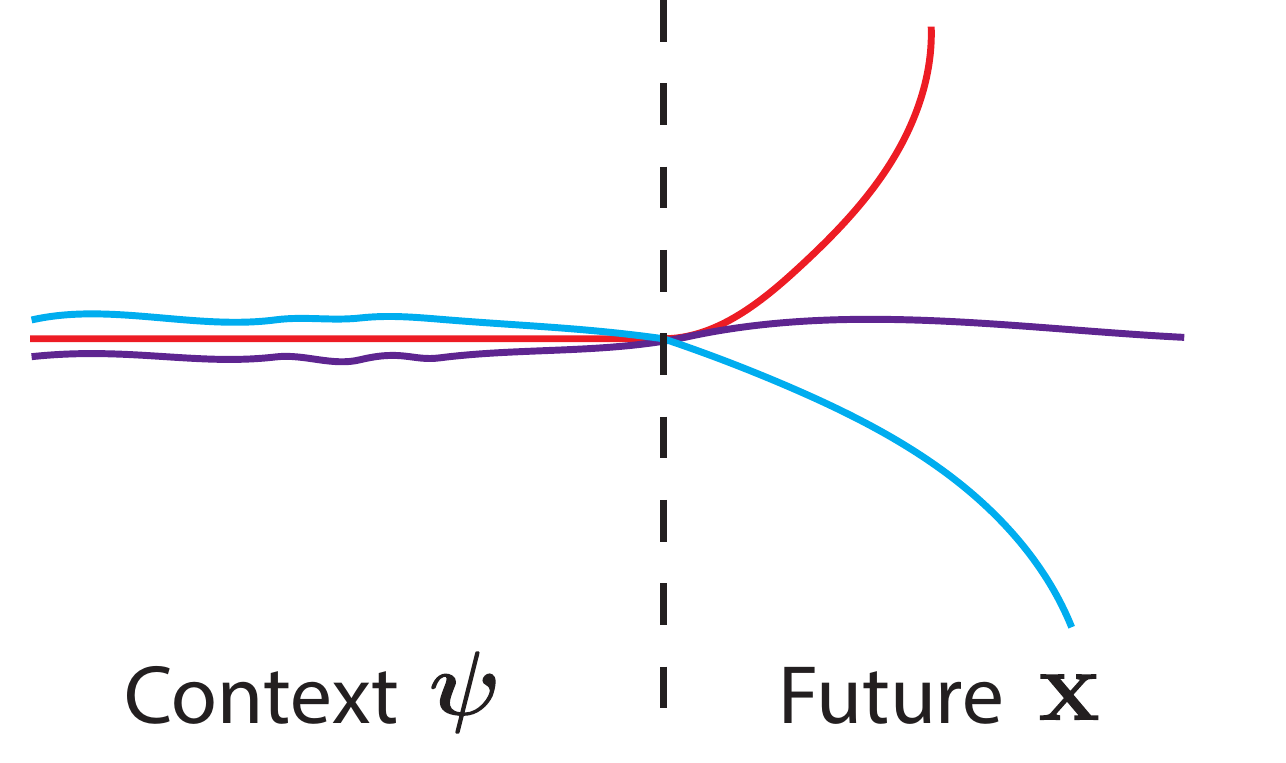}
  \end{center}
  \caption{In real data, contexts (past trajectories) are seldom the same due to noise.}
  \label{dsf:fig:metric}
\end{wrapfigure}
\paragraph{Metrics.}
\label{para:metrics}
A problem with trajectory forecasting evaluation is that in real data each context $\boldsymbol{\psi}^{(i)}$ usually only has one future trajectory $\mathbf{x}^{(i)}$, which means we only have one sample from a multi-modal distribution. Let us consider a scenario of three data examples $\{\mathbf{x}^{(i)}, \boldsymbol{\psi}^{(i)}\}_{i=1}^3$ as shown in Fig.~\ref{dsf:fig:metric} (red, purple, blue). The contexts (past trajectories) of the three examples are instances of the same trajectory but they are slightly different due to noise. As these three contexts have the same semantic meaning, they should share the future trajectories, \emph{e.g.}, the purple and blue future trajectories are also valid for the red context. If we evaluate each example $(\mathbf{x}^{(i)}, \boldsymbol{\psi}^{(i)})$ only with its own future trajectory $\mathbf{x}^{(i)}$, a method can achieve high scores by only forecasting the mode corresponding to $\mathbf{x}^{(i)}$ and dropping other modes. This is undesirable because we want a good method to capture all modes of the future trajectory distribution, not just a single mode. To allow for multi-modal evaluation, we propose collecting multiple future trajectories for each example by clustering examples with similar contexts. Specifically, we augment each data example $(\mathbf{x}^{(i)}, \boldsymbol{\psi}^{(i)})$ with a future trajectory set $\mathcal{X}^{(i)} = \{\mathbf{x}^{(j)} | \|\boldsymbol{\psi}^{(j)} - \boldsymbol{\psi}^{(i)}\| \leq \varepsilon,\, j=1, \ldots, M\}$ and metrics are calculated based on $\mathcal{X}^{(i)}$ instead of $\mathbf{x}^{(i)}$, \emph{i.e.,} we compute metrics for each $\mathbf{x} \in \mathcal{X}^{(i)}$ and average the results.

We use the following metrics for evaluation:
    (1)~\textbf{Average Displacement Error (ADE)}: average mean square error (MSE) over all time steps between the ground truth future trajectory $\mathbf{x}$ and the closest sample $\mathbf{\tilde{x}}$ in the forecasted set of trajectories $Y_f$.
    (2)~\textbf{Final Displacement Error (FDE)}: MSE between the final ground truth position $\mathbf{x}^T$ and the closest sample's final position $\mathbf{\tilde{x}}^T$.
    (3)~\textbf{Average Self Distance (ASD)}: average $L2$ distance over all time steps between a forecasted sample $\tilde{\mathbf{x}}_i$ and its closest neighbor $\mathbf{\tilde{x}}_j$ in $Y_f$.
    (4)~\textbf{Final Self Distance (FSD)}: $L2$ distance between the final position of a sample $\tilde{\mathbf{x}}_i^T$ and its closest neighbor's final position $\tilde{\mathbf{x}}_j^T$.
The ADE and FDE are common metrics used in prior work on trajectory forecasting \cite{alahi2016social, lee2017desire, rhinehart2018r2p2, gupta2018social}. However, these two metrics do not penalize repeated samples. To address this, we introduce two new metrics ASD and FSD to evaluate the similarity between samples in the set of forecasted trajectories. Larger ASD and FSD means the forecasted trajectories are more non-repetitive and diverse.

\paragraph{Baselines.} We compare our \textbf{Diversity Sampler Function (DSF)} with the following baselines:
    (1)~\textbf{cVAE}: a method that follows the original sampling scheme of cVAE by sampling latent codes from a Gaussian prior $p(\mathbf{z})$.
    (2)~\textbf{MCL}: an approach that uses multiple choice learning ~\cite{lee2016stochastic} to optimize the sampler $\mathcal{S}_\gamma(\boldsymbol{\psi})$ with the following loss: $\mathcal{L}_{\text{mcl}} = \min_{\tilde{\mathbf{x}} \in \mathcal{Y}}\| \tilde{\mathbf{x}} - \mathbf{x}\|^2$, where $\mathbf{x}$ is the ground truth future trajectory.
    (3)~\textbf{R2P2}: a method proposed in \cite{rhinehart2018r2p2} that uses a reparametrized pushforward policy to improve modeling of multi-modal distributions for vehicle trajectories.
    (4)~\textbf{cGAN}: generative adversarial networks \cite{goodfellow2014generative} conditioned on the forecasting context.
We implement all baselines using similar networks and perform hyperparameter search for each method for fair comparisons. For methods whose samples are stochastic, we use 10 random seeds and report the average results for all metrics.

\subsection{Synthetic 2D Trajectory Data}
\begin{figure}
    \centering
    \includegraphics[width=\linewidth]{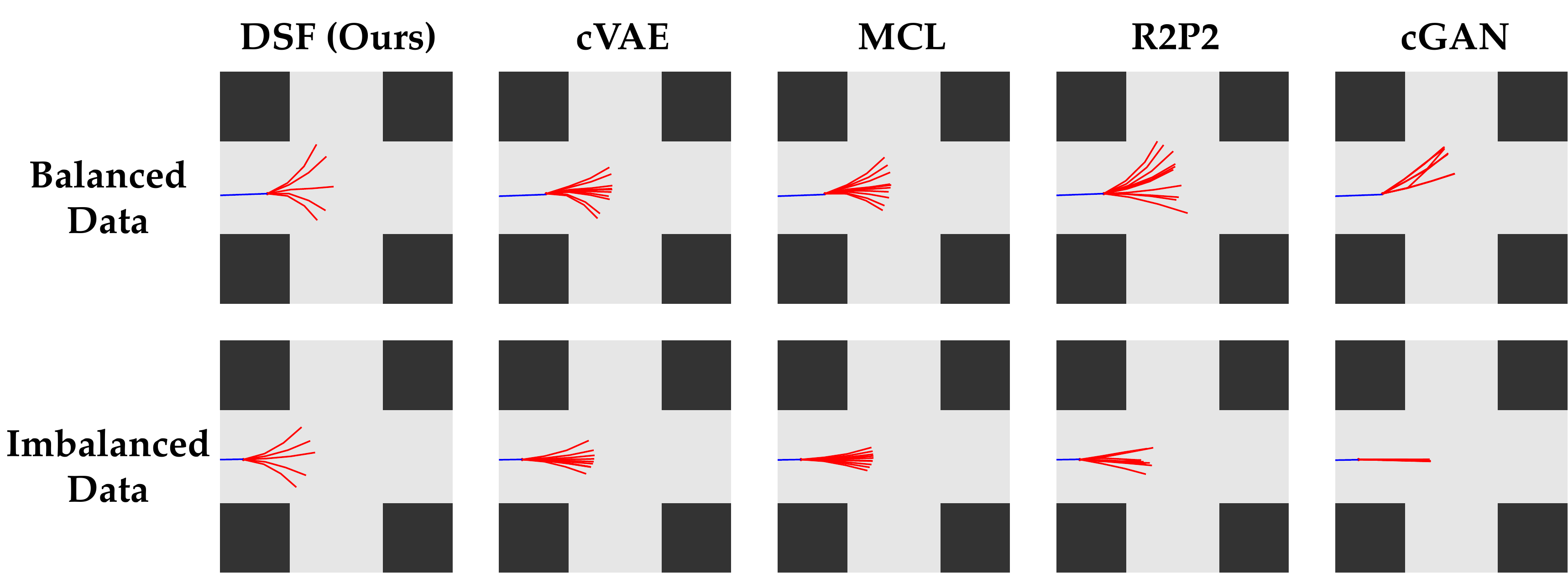}
    \caption{Qualitative results on synthetic data for both balanced and imbalanced data distribution when $N = 10$. Blue represents the past trajectory and red represents forecasted future trajectories.}
    \label{dsf:fig:synthetic}
\end{figure}

\begin{table}
\footnotesize
\centering
\begin{tabular}{@{\hskip 1mm}lrrrrrrrrr@{\hskip 1mm}}
\toprule
& \multicolumn{4}{c}{Balanced Data} & & \multicolumn{4}{c}{Imbalanced Data} \\ \cmidrule{2-5} \cmidrule{7-10}
Method & ADE $\downarrow$ & FDE $\downarrow$ & ASD $\uparrow$ & FSD $\uparrow$ & & ADE $\downarrow$ & FDE $\downarrow$ & ASD $\uparrow$ & FSD $\uparrow$\\ \midrule
DSF (Ours)   & \textbf{0.182} & \textbf{0.344} & \textbf{0.147} & \textbf{0.340} & &\textbf{0.198} &\textbf{0.371} &\textbf{0.207} &\textbf{0.470}\\  
cVAE & 0.262 & 0.518 & 0.022 & 0.050 & &0.332 & 0.662 & 0.021 & 0.050 \\
MCL  & 0.276 & 0.548 & 0.012 & 0.030 & &0.457 & 0.938 & 0.005 & 0.010\\
R2P2  & 0.211 & 0.361 & 0.047 & 0.080 & &0.393 & 0.776 & 0.019 & 0.030 \\
cGAN  & 0.808 & 1.619 & 0.018 & 0.010 & &1.784 & 3.744 & 0.006 & 0.001  \\
\bottomrule
\end{tabular}
\vspace{6mm}
\caption{Quantitative results on synthetic data (numbers scaled by 10) when $N = 10$.}
\label{dsf:table:synthetic}
\end{table}

We first use synthetic data to evaluate our method's performance for low-dimensional tasks. We design a virtual 2D traffic scene where a vehicle comes to a crossroad and can choose three different future routes: forward, left, and right. We consider two types of synthetic data: (1) Balanced data, which means the probabilities of the vehicle choosing one of the three routes are the same; (2) Imbalanced data, where the probabilities of the vehicle going forward, left and right are 0.8, 0.1, 0.1, respectively. We synthesize trajectory data by simulating the vehicle's behavior and adding Gaussian noise to vehicle velocities. Each data example $(\mathbf{x}^{(i)}, \boldsymbol{\psi}^{(i)})$ contains future trajectories of 3 steps and past trajectories of 2 steps. We also add an obstacle map around the current position to the context $\boldsymbol{\psi}^{(i)}$. In total, we have around 1100 training examples and 1000 test examples.

Table~\ref{dsf:table:synthetic} summarizes the quantitative results for both balanced and imbalanced data when the sampling budget $N$ is 10. We can see that our method DSF outperforms the baselines in all metrics in both test settings. As shown in Fig.~\ref{dsf:fig:synthetic}, our method generates more diverse trajectories and is less affected by the imbalanced data distribution. The trajectory samples of our method are also less repetitive, a feature afforded by our DPP formulation. Fig.~\ref{dsf:fig:plot} shows how ADE changes as a function of the sampling budget $N$.

\begin{figure}
    \centering
    \includegraphics[width=\linewidth]{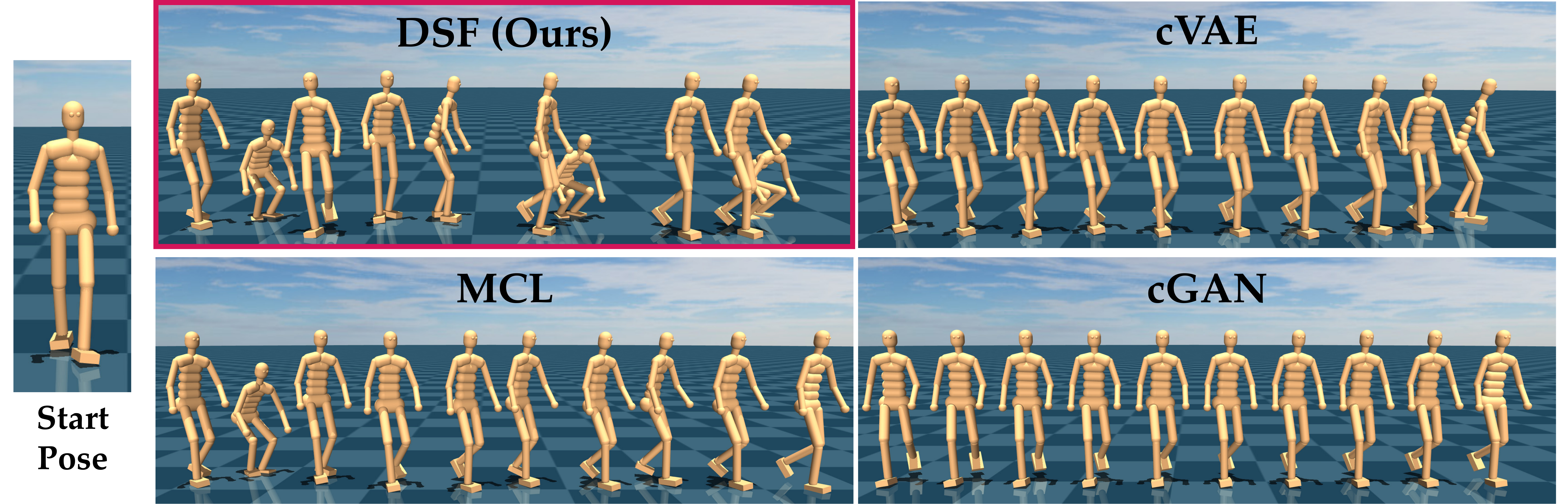}
    \caption{Qualitative results for human motion forecasting when $N=10$. The left shows the starting pose, and the right shows for each method the final pose of all 10 forecasted motion samples.}
    \label{dsf:fig:human}
\end{figure}

\subsection{Diverse Human Motion Forecasting}
\begin{wraptable}{r}{0.5\textwidth}
\footnotesize
\centering
\begin{tabular}{@{\hskip 1mm}l@{\hskip -0pt}rrrrr@{\hskip 1mm}}
\toprule
Method & ADE $\downarrow$ & FDE $\downarrow$ & ASD $\uparrow$ & FSD $\uparrow$ \\  \midrule
DSF (Ours) & \textbf{0.259} & \textbf{0.421} & \textbf{0.115} & \textbf{0.282}\\  
cVAE  & 0.332 & 0.642 & 0.034 & 0.098\\
MCL  & 0.344 & 0.674 & 0.036 & 0.122\\
cGAN & 0.652 & 1.296 & 0.001 & 0.003 \\
\bottomrule
\end{tabular}
\vspace{6mm}
\caption{Quantitative results for human motion forecasting when $N=10$.}
\label{dsf:table:human}
\end{wraptable}

To further evaluate our method's performance for more complex and high-dimensional tasks, we apply our method to forecast future human motions (pose sequences). We use motion capture to obtain 10 motion sequences including different types of motions such as walking, turning, jogging, bending, and crouching. Each sequence is about 1 minute long and each pose consists of 59 joint angles. We use past 3 poses (0.1s) to forecast next 30 poses (1s). There are around 9400 training examples and 2000 test examples where we use different sequences for training and testing.

\begin{figure}
	\centering
	\includegraphics[width=\linewidth]{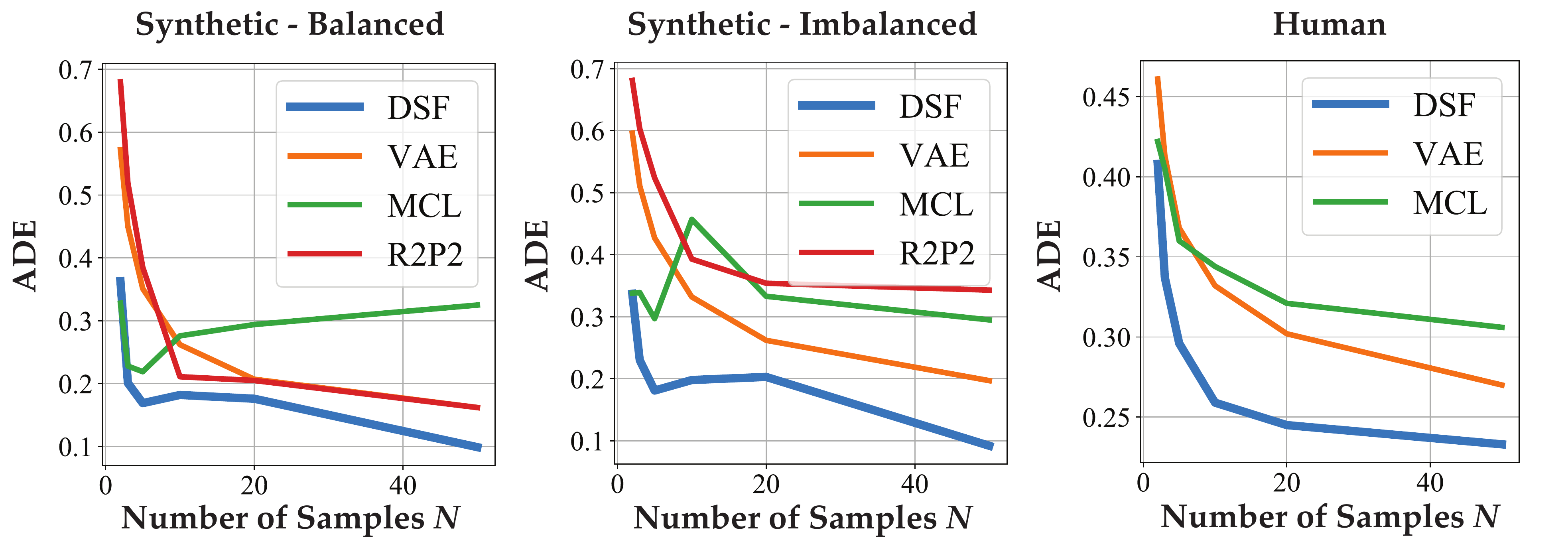}
	\caption{ADE vs. $N$ for synthetic data and human motion forecasting. cGAN is not shown in this plot as it is much worse than other methods due to mode collapse.}
	\label{dsf:fig:plot}
\end{figure}

We present quantitative results in Table~\ref{dsf:table:human} and we can see that our approach outperforms other methods in all metrics. As the dynamics model used in R2P2~\cite{rhinehart2018r2p2} does not generalize well for high-dimensional human motion, we find the model fails to converge and we do not compare with it in this experiment. Fig.~\ref{dsf:fig:plot} shows that our method achieves large improvement when the sampling budget is big ($N$ = 50). We also present qualitative results in Fig.~\ref{dsf:fig:human}, where we show the starting pose and the final pose of all 10 forecasted motion samples for each method. We can clearly see that our method generates more diverse future human motions than the baselines. Please refer to our \href{https://youtu.be/5i71SU_IdS4}{video} for additional qualitative results.

\subsection{Additional Experiments with Diversity-Based Baselines}

\begin{table}
\footnotesize
\centering
\begin{tabular}{@{\hskip 1mm}lrrrrrrrrr@{\hskip 1mm}}
\toprule
& \multicolumn{4}{c}{$N = 10$} & & \multicolumn{4}{c}{$N = 50$} \\ \cmidrule{2-5} \cmidrule{7-10}
Method & ADE $\downarrow$ & FDE $\downarrow$ & ASD $\uparrow$ & FSD $\uparrow$ & & ADE $\downarrow$ & FDE $\downarrow$ & ASD $\uparrow$ & FSD $\uparrow$\\ \midrule
DSF (Ours) & \textbf{0.340} & \textbf{0.521} & 0.381 & 0.621 & & \textbf{0.236} & \textbf{0.306} & 0.313 & 0.415 \\
DSF-NLL        & \textbf{0.335} & \textbf{0.514} & 0.343 & 0.496 & & X     & X     & X     & X     \\
DSF-COS    & 2.588 & 1.584 & \textbf{5.093} & \textbf{5.718} & & 0.978 & 0.891 & \textbf{2.007} & \textbf{1.968} \\
cVAE       & 0.363 & 0.549 & 0.235 & 0.360 & & 0.276 & 0.369 & 0.160 & 0.220 \\
cVAE-LDPP  & 0.373 & 0.554 & 0.280 & 0.426 & & 0.277 & 0.365 & 0.176 & 0.240 \\
\bottomrule
\end{tabular}
\vspace{6mm}
\caption{Quantitative results on Human3.6M~\cite{ionescu2013human3} for $N=10$ and $N=50$. X means the method is unable to learn a model due to numerical instability.}
\label{dsf:table:add}
\end{table}

In this section, we perform additional experiments on a large human motion dataset (3.6 million frames), Human3.6M~\cite{ionescu2013human3}, to evaluate the generalization ability of our approach. We predict future motion of 2 seconds based on observed motion of 0.5 seconds. We also use a new selection of baselines including several variants of our method (DSF) and the cVAE to validate several design choices of our method, including the choice of the expected cardinality over the negative log likelihood (NLL) of the DPP as the diversity loss. Specifically, we use the following new baselines: (1)~\textbf{DSF-NLL}: a variant of DSF that uses NLL as the diversity loss instead of the expected cardinality. (2)~\textbf{DSF-COS}: a DSF variant that uses cosine similarity to build the similarity matrix $\mathbf{S}$ for the DPP kernel $\mathbf{L}$. (3)~\textbf{DSF-NLL}: a variant of the cVAE that samples 100 latent codes and performs DPP MAP inference on the latent codes to obtain a diverse set of latent codes, which are then decoded into trajectory samples.

We present quantitative results in Table~\ref{dsf:table:add} when the number of samples $N$ is 10 and 50. The baseline DSF-COS is able to achieve very high diversity (ASD and FSD) but its samples are overly diverse and have poor quality which is indicated by the large ADE and FDE. Compared with DSF-NLL, our method achieves better diversity (ASD and FSD) and similar ADE and FDE when the number of samples is small ($N = 10$). For a larger number of samples ($N=50$), NLL becomes unstable even with a large $\epsilon$ (1e-3) added to the diagonal. This behavior of NLL, \emph{i.e.,} stable for small $N$ but unstable for large $N$, matches our intuition that NLL becomes unstable when samples become similar (as discussed in Sec.~\ref{dsf:sec:dsf}), because when there are more samples, it is easier to have similar samples during the SGD updates of the DSF network. The baseline cVAE-LDPP also performs worse than DSF in all metrics even though it is able to outperfom the cVAE. We believe the reason is that diversity in sample space may not be well reflected in the latent space due to the non-linear mapping from latent codes to samples induced by deep neural networks.

\section{Conclusion}
We proposed a novel forecasting approach using a DSF to optimize over the sample space of a generative model. Our method learns the DSF with a DPP-based diversity measure to generate a diverse set of trajectories. The diversity measure is a novel application of DPPs to optimize a set of items in continuous space. Experiments have shown that our approach can generate more diverse vehicle trajectories and human motions compared to state-of-the-art baseline forecasting approaches.
\chapter{Stochastic Human Motion Generation with Diversifying Latent Flows}
\label{chap:dlow}

\section{Introduction}

Human motion prediction, i.e., predicting the future 3D poses of a person based on past poses, is an important problem in computer vision and has many useful applications in autonomous driving~\cite{paden2016survey}, human robot interaction~\cite{koppula2013anticipating} and healthcare~\cite{troje2002decomposing}. It is a challenging problem because the future motion of a person is potentially diverse and multi-modal due to the complex nature of human behavior. For many safety-critical applications, it is important to predict a diverse set of human motions instead of just the most likely one. For examples, an autonomous vehicle should be aware that a nearby pedestrian can suddenly cross the road even though the pedestrian will most likely remain in place. This diversity requirement calls for a generative approach that can fully characterize the multi-modal distribution of future human motion.

\begin{figure}[t]
	\centering
	\includegraphics[width=\textwidth]{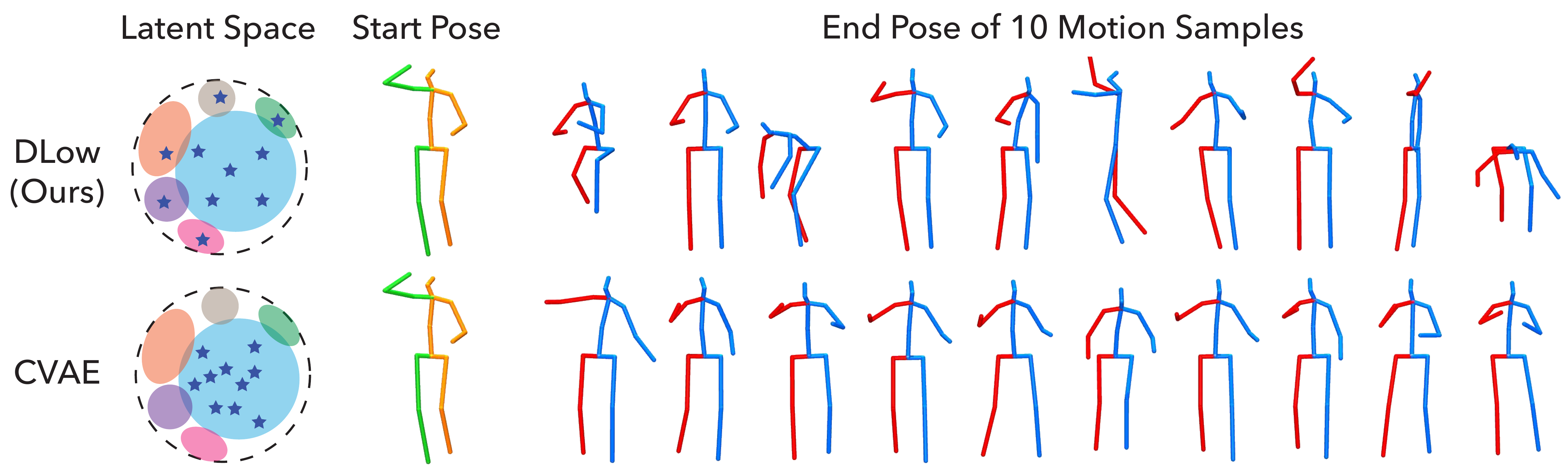}
	\caption{In the latent space of a conditional variational autoencoder (CVAE), samples (stars) from our method DLow are able to cover more modes (colored ellipses) than the CVAE samples. In the motion space, DLow generates a diverse set of future human motions while the CVAE only produces perturbations of the motion of the major mode.}
	\label{dlow:fig:teaser}
\end{figure}

Deep generative models, e.g., variational autoencoders (VAEs)~\cite{kingma2013auto}, are effective tools to model multi-modal data distributions. Most existing work~\cite{walker2017pose,lin2018human,barsoum2018hp,ruiz2018human,kundu2019bihmp,yan2018mt,aliakbarian2020stochastic} using deep generative models for human motion prediction is focused on the design of the generative model to allow it to effectively learn the data distribution. After the generative model is learned, little attention has been paid to the sampling method used to produce \emph{motion samples} (predicted future motions) from the \emph{pretrained} generative model (weights kept fixed). Most of prior work predicts a set of motions by randomly sampling a set of latent codes from the latent prior and decoding them with the generator into motion samples. We argue that such a sampling strategy is not guaranteed to produce a diverse set of samples for two reasons: (1) The samples are independently drawn, which makes it difficult to enforce diversity; (2) The samples are drawn based on likelihood only, which means many samples may concentrate around the major modes (which have more observed data) of the data distribution and fail to cover the minor modes (as shown in Fig.~\ref{dlow:fig:teaser} (Bottom)). The poor sample efficiency of random sampling means that one needs to draw a large number of samples in order to cover all the modes which is computationally expensive and can lead to high latency, making it unsuitable for real-time applications such as autonomous driving and virtual reality.  This prompts us to address an overlooked aspect of diverse human motion prediction --- the sampling strategy.

We propose a novel sampling method, Diversifying Latent Flows (DLow), to obtain a diverse set of samples from a pretrained deep generative model. For this work, we use a conditional variational autoencoder (CVAE) as our pretrained generative model but other generative models can also be used with our approach. DLow is inspired by the two previously mentioned problems with random (independent) sampling. To tackle problem (1) where sample independence limits model diversity, we introduce a new random variable and a set of learnable deterministic mapping functions to correlate the motion samples. We first transform the random variable with the mappings functions to generate a set of correlated latent codes which are then decoded into motion samples using the generator. As all motion samples are generated from a common random factor, this formulation allows us to model the joint sample distribution and offers us the opportunity to impose diversity on the samples by optimizing the parameters of the mapping functions. To address problem (2) where likelihood-based sampling limits diversity, we introduce a diversity-promoting prior (loss function) on the samples during the training of DLow. The prior follows an energy-based formulation using an energy function based on pairwise sample distance. We optimize the mapping functions during training to minimize the cross entropy between the joint sample distribution and diversity-promoting prior to increase sample diversity. To strike a balance between diversity and likelihood, we add a KL term to the optimization to enhance the likelihood of each sample. The relative weights between the prior term and the KL term represent the trade-off between the diversity and likelihood of the generated motion samples.
Furthermore, our approach is highly flexible in that by designing different forms of the diversity-promoting prior we can impose a variety of structures on the samples besides diversity. For example, we can design the prior to ask the motion samples to cover the ground truth better to achieve higher sample accuracy. Additionally, other designs of the prior can enable new applications, such as controllable motion prediction, where we generate diverse motion samples that share some common features (e.g., similar leg motion but diverse upper-body motion).

The contributions of this work are the following: 
(1) We propose a novel perspective for addressing sample diversity in deep generative models --- designing sampling methods for a \emph{pretrained} generative model. 
(2) We propose a principled sampling method, DLow, which formulates diversity sampling as a constrained optimization problem over a set of learnable mapping functions using a diversity-promoting prior on the samples and KL constraints on the latent codes, which allows us to balance between sample diversity and likelihood.
(3)~Our approach allows for flexible design of the diversity-promoting prior to obtain more accurate samples or enable new applications such as controllable motion prediction. 
(4)~We demonstrate through human motion prediction experiments that our approach outperforms state-of-the-art baseline methods in terms of sample diversity and accuracy.

\section{Related Work}
\paragraph{Human Motion Prediction.} Most previous work takes a deterministic approach to modeling human motion and regress a single future motion from past 3D poses \cite{fragkiadaki2015recurrent,jain2016structural,butepage2017deep,li2017auto,ghosh2017learning,martinez2017human,pavllo2018quaternet,chiu2019action,gopalakrishnan2019neural,aksan2019structured,wang2019imitation,mao2019learning} or video frames \cite{chao2017forecasting,zhang2019predicting,yuan2019ego}. While these approaches are able to predict the most likely future motion, they fail to model the multi-modal nature of human motion, which is essential for safety-critical applications. More related to our work, stochastic human motion prediction methods start to gain popularity with the development of deep generative models. These methods~\cite{walker2017pose,lin2018human,barsoum2018hp,ruiz2018human,kundu2019bihmp,yan2018mt,aliakbarian2020stochastic,yuan2020residual} often build upon popular generative models such as conditional generative adversarial networks (CGANs;~\cite{goodfellow2014generative}) or conditional variational autoencoders (CVAEs;~\cite{kingma2013auto}). The aforementioned methods differ in the design of their generative models, but at test time they follow the same sampling strategy --- randomly and independently sampling trajectories from the pretrained generative model without considering the correlation between samples. In this work, we propose a principled sampling method that can produce a diverse set of samples, thus improving sample efficiency compared to the random sampling typically used in prior work.

\paragraph{Diverse Inference.} Producing a diverse set of solutions has been investigated in numerous problems in computer vision and machine learning. A branch of these diversity-driven methods stems from the M-Best MAP problem~\cite{nilsson1998efficient,seroussi1994algorithm}, including diverse M-Best solutions~\cite{batra2012diverse} and multiple choice learning~\cite{guzman2012multiple,lee2016stochastic}. Alternatively, submodular function maximization has been applied to select a diverse subset of garments from fashion images~\cite{hsiao2018creating}. Another type of methods~\cite{kulesza2011k,gong2014diverse,gillenwater2014expectation,huang2015we,azadi2017learning,yuan2019diverse,weng2021ptp} seeks diversity using determinantal point processes (DPPs;~\cite{macchi1975coincidence,kulesza2012determinantal}) which are efficient probabilistic models that can measure the global diversity and quality within a set. Similarly, Fisher information~\cite{rissanen1996fisher} has been used for diverse feature~\cite{gu2012generalized} and data~\cite{sourati2017probabilistic} selection. Diversity has also been a key aspect in generative modeling. A vast body of work has tried to alleviate the mode collapse problem in GANs~\cite{che2016mode,chen2016infogan,srivastava2017veegan,arjovsky2017wasserstein,gulrajani2017improved,elfeki2018gdpp,lin2018pacgan,yang2019diversity} and the posterior collapse problem in VAEs~\cite{zhao2017infovae,tolstikhin2017wasserstein,kim2018semi,bhattacharyya2018accurate,liu2019cyclical,he2019lagging}. Normalizing flows~\cite{rezende2015variational} have also been used to promote diversity in trajectory forecasting~\cite{rhinehart2018r2p2,guan2020generative}. This line of work aims to improve the diversity of the data distribution learned by deep generative models. We address diversity from a different angle by improving the strategy for producing samples from a pretrained deep generative model. 
\section{Diversifying Latent Flows (DLow)}
\label{dlow:sec:dlow}

For many existing methods on generative vision tasks such as multi-modal human motion prediction, the primary focus is to learn a good generative model that can capture the multi-modal distribution of the data. In contrast, once the generative model is learned, little attention has been paid to devising sampling strategies for producing diverse samples from the \emph{pretrained} generative model.

In this section, we will introduce our method, Diversifying Latent Flows (DLow), as a principled way for drawing a diverse and likely set of samples from a pretrained generative model (weights kept fixed). To provide the proper context, we will first start with a brief review of deep generative models and how traditional methods produce samples from a pretrained generative model.

\paragraph{Background: Deep Generative Models.} 
Let $\mathbf{x} \in \mathcal{X}$ denote data (e.g., human motion) drawn from a data distribution $p(\mathbf{x}|\mathbf{c})$ where $\mathbf{c}$ is some conditional information (e.g., past motion). One can reparameterize the data distribution by introducing a latent variable $\mathbf{z} \in \mathcal{Z}$ such that $p(\mathbf{x}|\mathbf{c}) = \int_\mathbf{z} p(\mathbf{x}|\mathbf{z}, \mathbf{c}) p(\mathbf{z})d\mathbf{z}$, where $p(\mathbf{z})$ is a Gaussian prior distribution. Deep generative models learn $p(\mathbf{x}|\mathbf{c})$ by modeling the conditional distribution $p(\mathbf{x}|\mathbf{z}, \mathbf{c})$, and the generative process can be described as sampling $\mathbf{z}$ and mapping them to data samples $\mathbf{x}$ using a deterministic \emph{generator} function $G_\theta: \mathcal{Z} \rightarrow \mathcal{X}$ as
\begin{align}
\label{dlow:eq:p_z}
&\mathbf{z} \sim p(\mathbf{z})\,, \\
\label{dlow:eq:G_theta}
&\mathbf{x} = G_\theta(\mathbf{z}, \mathbf{c})\,,
\end{align}
where the generator $G_\theta$ is instantiated as a deep neural network parametrized by $\theta$. This generative process produces samples from the implicit sample distribution $p_\theta(\mathbf{x}|\mathbf{c})$ of the generative model, and the goal of generative modeling is to learn a generator $G_\theta$ such that $p_\theta(\mathbf{x}|\mathbf{c}) \approx p(\mathbf{x}|\mathbf{c})$. There are various approaches for learning the generator function $G_\theta$, which yield different types of deep generative models such as variational autoencoders (VAEs;~\cite{kingma2013auto}), normalizing flows (NFs;~\cite{rezende2015variational}), and generative adversarial networks (GANs;~\cite{goodfellow2014generative}). Note that even though the discussion in this work is focused on conditional generative models, our method can be readily applied to the unconditional case.

\paragraph{Random Sampling.} 
Once the generator function $G_\theta$ is learned, traditional approaches produce samples from the learned data distribution $p_\theta(\mathbf{x}|\mathbf{c})$ by first randomly sampling a set of latent codes $Z = \{\mathbf{z}_1, \ldots, \mathbf{z}_K\}$ from the latent prior $p(\mathbf{z})$ (Eq.~\eqref{dlow:eq:p_z}) and decode $Z$ with the generator $G_\theta$ into a set of data samples $X = \{\mathbf{x}_1, \ldots, \mathbf{x}_K\}$ (Eq.~\eqref{dlow:eq:G_theta}). We argue that such a sampling strategy may result in a less diverse sample set for two reasons: (1) Independent sampling cannot model the repulsion between samples within a diverse set; (2) The sampling is only based on the data likelihood and many samples can concentrate around a small number of modes that have more training data. As a result, random sampling can lead to low sample efficiency because many samples are similar to one another and fail to cover other modes in the data distribution.

\begin{figure*}[t]
    \centering
    \includegraphics[width=\textwidth]{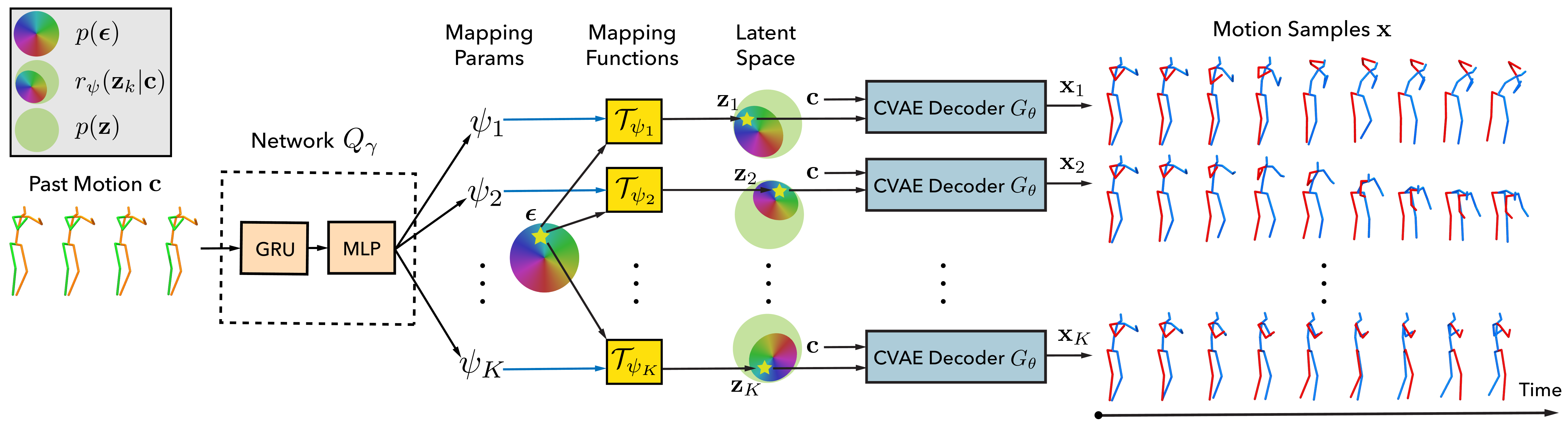}
    \caption{\textbf{Overview of our DLow framework applied to diverse human motion prediction.} The network $Q_\gamma$ takes past motion $\mathbf{c}$ as input and outputs the parameters of the mapping functions $\mathcal{T}_{\psi_1}, \ldots, \mathcal{T}_{\psi_K}$. Each mapping $\mathcal{T}_{\psi_k}$ transforms the random variable $\boldsymbol{\epsilon}$ to a different latent code $\mathbf{z}_k$ and also warps the density $p(\boldsymbol{\epsilon})$ to the latent code density $r_\psi(\mathbf{z}_k|\mathbf{c})$. Each latent code $\mathbf{z}_k$ is decoded by the CVAE decoder into a motion sample $\mathbf{x}_k$.}
    \label{dlow:fig:overview}
\end{figure*}

\paragraph{DLow Sampling.} To address the above issues with the random sampling approach, we propose an alternative sampling method, Diversifying Latent Flows (DLow), that can generate a diverse and likely set of samples from a pretrained deep generative model. Again, we stress that the weights of the generative model are kept fixed for DLow. We later apply DLow to the task of human motion prediction in Sec.~\ref{dlow:sec:motion_pred} to demonstrate DLow's ability to improve sample diversity.

Instead of sampling each latent code $\mathbf{z}_k \in Z$ independently according to $p(\mathbf{z})$, we introduce a random variable $\boldsymbol{\epsilon}$ and conditionally generate the latent codes $Z$ and data samples $X$ as follows:
\begin{align}
& \boldsymbol{\epsilon} \sim p(\boldsymbol{\epsilon})\,, \\
& \mathbf{z} _k = \mathcal{T}_{\psi_k}(\boldsymbol{\epsilon}) \,,\;\, \quad \quad 1 \leq k \leq K\,, \\
& \mathbf{x} _k = G_\theta(\mathbf{z}_k, \mathbf{c})\,,	\,\quad 1 \leq k \leq K\,,
\end{align}
where $p(\boldsymbol{\epsilon})$ is a Gaussian distribution, $\mathcal{T}_{\psi_1}, \ldots, \mathcal{T}_{\psi_K}$ are latent mapping functions with parameters $\psi = \{ \psi_1, \ldots, \psi_K \}$, and each $\mathcal{T}_{\psi_k}$ maps $\boldsymbol{\epsilon}$ to a different latent code $\mathbf{z}_k$.
The above generative process defines a joint distribution $r_\psi(X,Z|\mathbf{c})=p_\theta(X|Z, \mathbf{c})r_\psi(Z|\mathbf{c})$ over the samples $X$ and latent codes $Z$, where $p_\theta(X|Z,\mathbf{c})$ is the conditional distribution induced by the generator $G_\theta(\mathbf{z}, \mathbf{c})$. Notice that in our setup, $r_\psi(X,Z|\mathbf{c})$ depends only on $\psi$ as the generator parameters $\theta$ are learned in advance and are kept fixed. The data samples $X$ can be viewed as a sample from the joint sample distribution $r_\psi(X|\mathbf{c})=\int r_\psi(X,Z|\mathbf{c})dZ$ and the latent codes $Z$ can be regarded as a sample from the joint latent distribution $r_\psi(Z|\mathbf{c})$ induced by warping $p(\boldsymbol{\epsilon})$ through $\mathcal{T}_{\psi_1}, \ldots, \mathcal{T}_{\psi_K}$. If we further marginalize out all variables except for $\mathbf{x}_k$ from $r_\psi(X|\mathbf{c})$, we obtain the marginal sample distribution $r_\psi(\mathbf{x}_k|\mathbf{c})$ from which each sample $\mathbf{x}_k$ is drawn. Similarly, each latent code $\mathbf{z}_k \in Z$ can be viewed as a latent sample from the marginal latent distribution $r_\psi(\mathbf{z}_k|\mathbf{c})$.

The above distribution reparametrizations are illustrated in Fig.~\ref{dlow:fig:overview}. We can see that all latent codes $Z$ and data samples $X$ are correlated as they are uniquely determined by $\boldsymbol{\epsilon}$, and by sampling $\boldsymbol{\epsilon}$ one can easily produce $Z$ and $X$ from the joint latent distribution $r_\psi(Z|\mathbf{c})$ and joint sample distribution $r_\psi(X|\mathbf{c})$. Because $r_\psi(Z|\mathbf{c})$ and $r_\psi(X|\mathbf{c})$ are controlled by the latent mapping functions $\mathcal{T}_{\psi_1}, \ldots, \mathcal{T}_{\psi_K}$, we can impose structural constraints on $r_\psi(Z|\mathbf{c})$ and $r_\psi(X|\mathbf{c})$ by optimizing the parameters $\psi$ of the latent mapping functions.

To encourage the diversity of samples $X$, we introduce a diversity-promoting prior $p(X)$ (specific form defined later) and formulate a constrained optimization problem:
\begin{align}
\label{dlow:eq:obj}
\min_\psi & \quad -\mathbb{E}_{X \sim r_\psi(X|\mathbf{c})}[\log p(X)]\,, \\
\label{dlow:eq:constraint}
\text{s.t.} & \quad \text{KL} (r_\psi(\mathbf{z}_k|\mathbf{c}) \| p(\mathbf{z}_k)) = 0\,, \quad 1 \leq k \leq K\,,
\end{align}
where we minimize the cross entropy between the sample distribution $r_\psi(X|\mathbf{c})$ and the diversity-promoting prior $p(X)$.
However, the objective in Eq.~\eqref{dlow:eq:obj} alone can result in very low-likelihood samples $\mathbf{x}_k$ corresponding to latent codes $\mathbf{z}_k$ that are far away from the Gaussian prior $p(\mathbf{z}_k)$.
To ensure that each sample $\mathbf{x}_k$ also has high likelihood under the generative model $p_\theta(\mathbf{x}|\mathbf{c})$,  we add constraints in Eq.~\eqref{dlow:eq:constraint} on the KL divergence between $r_\psi(\mathbf{z}_k|\mathbf{c})$ and the Gaussian prior $p(\mathbf{z}_k)$ (same as $p(\mathbf{z})$) to make $r_\psi(\mathbf{z}_k|\mathbf{c}) = p(\mathbf{z}_k)$ and thus $r_\psi(\mathbf{x}_k|\mathbf{c}) = p_\theta(\mathbf{x}_k|\mathbf{c})$ where $r_\psi(\mathbf{x}_k|\mathbf{c})= \int p_\theta(\mathbf{x}_k|\mathbf{z}_k, \mathbf{c})r_\psi(\mathbf{z}_k|\mathbf{c})d\mathbf{z}_k$ and $ p_\theta(\mathbf{x}_k|\mathbf{c})= \int p_\theta(\mathbf{x}_k|\mathbf{z}_k, \mathbf{c})p(\mathbf{z}_k)d\mathbf{z}_k$. To optimize this constrained objective, we soften the constraints with the Lagrangian function:
\begin{equation}
\label{dlow:eq:dlow_opt}
\min_\psi \, -\mathbb{E}_{X \sim r_\psi(X|\mathbf{c})}[\log p(X)] +\beta\sum_{k=1}^K\text{KL} (r_\psi(\mathbf{z}_k|\mathbf{c}) \| p(\mathbf{z}_k))\,,
\end{equation}
where we use the same Lagrangian multiplier $\beta$ for all constraints. Despite having similar form, the above objective is very \emph{different} from the objective function of $\beta$-VAE~\cite{higgins2017beta} in many ways: (1) our goal is to learn a diverse sampling distribution $r_\psi(X|\mathbf{c})$ for a pretrained generative model rather than learning the generative model itself; (2) The first part in our objective is a diversifying term instead of a reconstruction term; (3) Our objective function applies to most deep generative models, not just VAEs. In this objective, the softening of the hard KL constraints allows for the trade-off between the diversity and likelihood of the samples $X$. For small $\beta$, $r_\psi(\mathbf{z}_k|\mathbf{c})$ is allowed to deviate from $p(\mathbf{z}_k)$ so that $r_\psi(\mathbf{z}_1|\mathbf{c}), \ldots, r_\psi(\mathbf{z}_K|\mathbf{c})$ can potentially attend to different regions in the latent space as shown in Fig.~\ref{dlow:fig:overview} (latent space) to further improve sample diversity. For large $\beta$, the objective will focus on minimizing the KL term so that $r_\psi(\mathbf{z}_k|\mathbf{c})\approx p(\mathbf{z}_k)$ and $r_\psi(\mathbf{x}_k|\mathbf{c})\approx p_\theta(\mathbf{x}_k|\mathbf{c})$, and thus the sample $\mathbf{x}_k$ will have high likelihood under $p_\theta(\mathbf{x}_k|\mathbf{c})$.

The overall DLow objective is defined as:
\begin{equation}
\label{dlow:eq:dlow_overall}
L_\text{DLow} = L_{\text{prior}}  + \beta L_{\text{KL}}\,,
\end{equation}
where $L_{\text{prior}}$ and $L_{\text{KL}}$ are the first and second term in Eq.~\eqref{dlow:eq:dlow_opt} respectively.
In the following, we will discuss in detail how we design the latent mapping functions $\mathcal{T}_{\psi_1}, \ldots, \mathcal{T}_{\psi_K}$ and the diversity-promoting prior $p(X)$.

\paragraph{Latent Mapping Functions.}
Each latent mapping $\mathcal{T}_{\psi_k}$ transforms the Gaussian distribution $p(\boldsymbol{\epsilon})$ to the marginal latent distribution $r_\psi(\mathbf{z}_k|\mathbf{c})$ for latent code $\mathbf{z}_k$ where $\mathcal{T}_{\psi_k}$ is also conditioned on $\mathbf{c}$. As $r_\psi(\mathbf{z}_k|\mathbf{c})$ should stay close to the Gaussian latent prior $p(\mathbf{z}_k)$, it would be ideal if the mapping $\mathcal{T}_{\psi_k}$ makes $r_\psi(\mathbf{z}_k|\mathbf{c})$ also a Gaussian. Thus, we design $\mathcal{T}_{\psi_k}$ to be an invertible affine transformation:
\begin{equation}
\label{dlow:eq:mapping}
\mathcal{T}_{\psi_k}(\boldsymbol{\epsilon}) = \mathbf{A}_k(\mathbf{c})\boldsymbol{\epsilon} + \mathbf{b}_k(\mathbf{c}) \,,
\end{equation}
where the mapping parameters $\psi_k = \{\mathbf{A}_k(\mathbf{c}), \mathbf{b}_k(\mathbf{c})\}$, $\mathbf{A}_k \in \mathbb{R}^{n_z \times n_z}$ is a nonsingular matrix, $\mathbf{b}_k \in \mathbb{R}^{n_z}$ is a vector, and $n_z$ is the number of dimensions for $\mathbf{z}_k$ and $\boldsymbol{\epsilon}$. As shown in Fig.~\ref{dlow:fig:overview} and Fig.~\ref{dlow:fig:network} (Right), we use a $K$-head network $Q_\gamma(\mathbf{c})$ to output $\psi_1, \ldots, \psi_K$, and the parameters $\gamma$ of the network $Q_\gamma(\mathbf{c})$ are the parameters to be optimized with the DLow objective in Eq.~\eqref{dlow:eq:dlow_overall}.

Under the invertible affine transformation $\mathcal{T}_{\psi_k}$, $r_\psi(\mathbf{z}_k|\mathbf{c})$ becomes a Gaussian distribution $\mathcal{N}(\mathbf{b}_k, \mathbf{A}_k\mathbf{A}_k^T)$. This allows us to compute the KL divergence terms in $L_\text{KL}$ analytically:
\begin{equation}
\label{dlow:eq:dlow_kl}
    \text{KL} (r_\psi(\mathbf{z}_k|\mathbf{c})\| p(\mathbf{z}_k)) = \frac{1}{2}\left(\operatorname{tr}\left(\mathbf{A}_k\mathbf{A}_k^T\right)+\mathbf{b}_k^T\mathbf{b}_k -n_z - \log\det\left(\mathbf{A}_k\mathbf{A}_k^T\right) \right).
\end{equation}
The KL divergence is minimized when $r_\psi(\mathbf{z}_k|\mathbf{c}) = p(\mathbf{z}_k)$ which implies that $\mathbf{A}_k\mathbf{A}_k^T = \mathbf{I}$ and $\mathbf{b}_k = \mathbf{0}$. Geometrically, this means that $\mathbf{A}_k$ is in the orthogonal group $O(n_z)$, which includes all rotations and reflections in an $n_z$-dimensional space. This means any mapping $\mathcal{T}_{\psi_k}$ that is a rotation or reflection operation will minimize the KL divergence. As mentioned before, there is a trade-off between diversity and likelihood in Eq.~\eqref{dlow:eq:dlow_overall}. To improve sample diversity (minimize $L_\text{prior}$) without compromising likelihood (KL divergence), we can optimize $\mathcal{T}_{\psi_1}, \ldots, \mathcal{T}_{\psi_K}$ to be different rotations or reflections to map $\boldsymbol{\epsilon}$ to different feasible points $\mathbf{z}_1, \ldots, \mathbf{z}_k$ in the latent space. This geometric understanding sheds light on the mapping space admitted by the hard KL constraints. In practice, we use soft KL constraints in the DLow objective to further enlarge the feasible mapping space which allows us to achieve lower $L_\text{prior}$ and better sample diversity.

\paragraph{Diversity-Promoting Prior.} In the DLow objective, a diversity-promoting prior $p(X)$ on the joint sample distribution is used to guide the optimization of the latent mapping functions $\mathcal{T}_{\psi_1}, \ldots, \mathcal{T}_{\psi_K}$.  
With an energy-based formulation, the prior $p(X)$ can be defined using an energy function $E(X)$:
\begin{equation}
\label{dlow:eq:dlow_energy}
p(X) = \exp(-E(X)) / \mathcal{S}\,,
\end{equation}
where $\mathcal{S}$ is a normalizing constant.  Dropping the constant $\mathcal{S}$, the first term in Eq.~\eqref{dlow:eq:dlow_opt} can be rewritten as
\begin{equation}
\label{dlow:eq:dlow_diverse}
L_{\text{prior}} = \mathbb{E}_{X \sim r_\psi(X|\mathbf{c})}[E(X)]\,.
\end{equation}
 To promote sample diversity of $X$, we design an energy function $E := E_d$ based on a pairwise distance metric $\mathcal{D}$:
\begin{equation}
\label{dlow:eq:e_diverse}
E_d(X) = \frac{1}{K(K-1)}\sum_{i=1}^K\sum_{j\neq i}^K \exp\left(-\frac{\mathcal{D}^2(\mathbf{x}_i, \mathbf{x}_j)} {\sigma_d}\right),
\end{equation}
where we use the Euclidean distance for $\mathcal{D}$ and an RBF kernel with scale $\sigma_d$. Minimizing $L_{\text{prior}}$ moves the samples towards a lower-energy (diverse) configuration.
$L_{\text{prior}}$ can be evaluated efficiently with the reparametrization trick~\cite{kingma2013auto}.

Up to this point, we have described the proposed sampling method, DLow, for generating a diverse set of samples from a pretrained generative model $p_\theta(\mathbf{x}|\mathbf{c})$. By introducing a common random variable $\boldsymbol{\epsilon}$, DLow allows us to generate correlated samples $X$. Moreover, by introducing learnable mapping functions $\mathcal{T}_{\psi_k}$, we can model the joint sample distribution $r_\psi(X|\mathbf{c})$ and impose structural constraints, such as diversity, on the sample set $X$ which cannot be modeled by random sampling from the generative model.

\section{Diverse Human Motion Generation}
\label{dlow:sec:motion_pred}

Equipped with a method to generate diverse samples from a pretrained deep generative model, we now turn our attention to the task of diverse human motion prediction.
Suppose the pose of a person is a $V$-dimensional vector consisting of 3D joint positions, we use $\mathbf{c} \in \mathbb{R}^{H \times V}$ to denote the past motion of $H$ time steps and $\mathbf{x} \in \mathbb{R}^{T \times V}$ to denote the future motion over a future time horizon of $T$. Given a past motion $\mathbf{c}$, the goal of diverse human motion prediction is to generate a diverse set of future motions $X = \{\mathbf{x}_1, \ldots, \mathbf{x}_K\}$.

\begin{figure}[t]
    \centering
    \includegraphics[width=\linewidth]{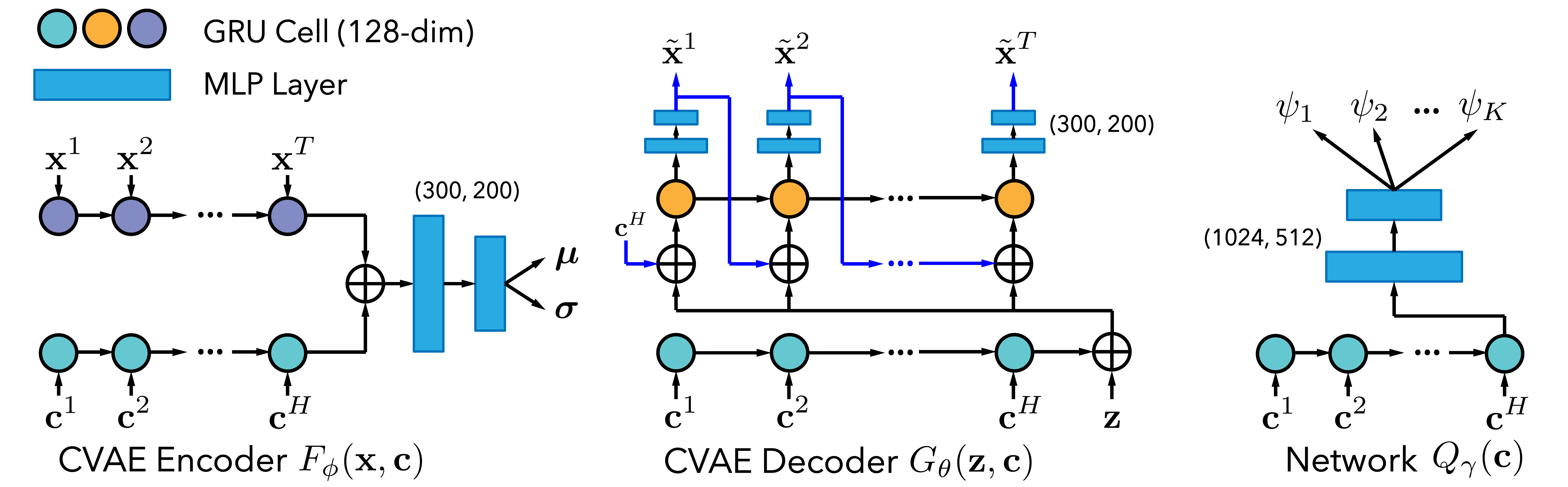}
    \caption{\textbf{Network architectures} for the CVAE and DLow. We use GRUs~\cite{chung2014empirical} to extract motion features. $\mathbf{x}^t$ and $\mathbf{c}^t$ denotes the $t$-th pose in $\mathbf{x}$ and $\mathbf{c}$ respectively.}
    \label{dlow:fig:network}
\end{figure}

To capture the multi-modal distribution of the future trajectory $\mathbf{x}$, we take a generative approach and use a conditional variational autoencoder (CVAE) to learn the future trajectory distribution $p_\theta(\mathbf{x}|\mathbf{c})$. Here we use the CVAE for its stability over other popular approaches such as CGANs, but other suitable deep generative models could also be used. The CVAE uses a varitional lower bound~\cite{jordan1999introduction} as a surrogate for the intractable true data log-likelihood:
\begin{equation}
\label{dlow:eq:cvae}
	\mathcal{L}(\mathbf{x} ; \theta, \phi)= \; \mathbb{E}_{q_{\phi}(\mathbf{z} | \mathbf{x}, \mathbf{c})}\left[\log p_{\theta}(\mathbf{x} | \mathbf{z}, \mathbf{c})\right]
	-\operatorname{KL}\left(q_{\phi}(\mathbf{z} | \mathbf{x}, \mathbf{c}) \| p(\mathbf{z})\right),
\end{equation}
where $q_{\phi}(\mathbf{z} | \mathbf{x}, \mathbf{c})$ is an $\phi$-parametrized approximate posterior distribution.
We use multivariate Gaussians for the prior, posterior (encoder distribution) and likelihood (decoder distribution): $p(\mathbf{z})=\mathcal{N}(\mathbf{0}, \mathbf{I})$,  $q_{\phi}(\mathbf{z} | \mathbf{x}, \mathbf{c}) = \mathcal{N}(\boldsymbol{\mu}, \text{Diag}(\boldsymbol{\sigma}^2))$, and $p_\theta(\mathbf{x}|\mathbf{z}, \mathbf{c}) = \mathcal{N}(\tilde{\mathbf{x}}, \alpha\mathbf{I})$ where $\alpha$ is a hyperparameter.
Both the encoder and decoder are implemented as recurrent neural networks (RNNs). As shown in Fig.~\ref{dlow:fig:network}, the encoder network $F_\phi$ outputs the parameters of the posterior distribution: $(\boldsymbol{\mu}, \boldsymbol{\sigma}) = F_\phi(\mathbf{x}, \mathbf{c})$; the decoder network $G_\theta$ outputs the reconstructed future trajectory $\tilde{\mathbf{x}} = G_\theta(\mathbf{z}, \mathbf{c})$. The CVAE is learned via jointly optimizing the encoder and decoder with Eq.~\eqref{dlow:eq:cvae}.

\subsection{Diversity Sampling with DLow}
Once the CVAE is learned, we follow the DLow framework proposed in Sec.~\ref{dlow:sec:dlow} to optimize the network $Q_\gamma$ (Fig.~\ref{dlow:fig:network} (Right)) and learn the latent mapping functions $\mathcal{T}_{\psi_1}, \ldots, \mathcal{T}_{\psi_K}$. Before doing this, to fully leverage the DLow framework, we will look at one of DLow's key feature, i.e., the design of the diversity-promoting prior $p(X)$ in $L_\text{prior}$ can be flexibly changed by modifying the underlying energy function $E(X)$. This allows us to impose various structural constraints besides diversity on the sample set $X$. Below, we will provide two examples of such prior designs that (1) improve sample accuracy or (2) enable new applications such as controllable motion prediction.

\paragraph{Reconstruction Energy.}
To ensure that the sample set $X$ is both diverse and accurate, i.e., the ground truth future motion $\hat{\mathbf{x}}$ is close to one of the samples in $X$, we can modify the prior's energy function $E$ in Eq.~\eqref{dlow:eq:dlow_energy} by adding a reconstruction term $E_r$:
\begin{align}
\label{dlow:eq:human_prior}
&E(X) = E_d(X) + \lambda_r E_r(X)\,,\\
\label{dlow:eq:e_recon}
&E_r(X) = \min_k \mathcal{D}^2(\mathbf{x}_k, \hat{\mathbf{x}})\,,
\end{align}
where $\lambda_r$ is a weighting factor and we use Euclidean distance as the distance metric $\mathcal{D}$. As DLow produces a correlated set of samples $X$ instead of independent samples, the network $Q_\gamma$ can learn to distribute samples in a way that are both diverse and accurate, covering the ground truth better. We use this prior design for our main experiments.

\paragraph{Controllable Motion Prediction.} 
Another possible design of the diversity-promoting prior $p(X)$ is one that promotes diversity in a certain subspace of the sample space. In the context of human motion prediction, we may want certain body parts to move similarly but other parts to move differently. For example, we may want leg motion to be similar but upper-body motion to be diverse across motion samples.
We call this task controllable motion prediction, i.e., finding a set of diverse samples that share some common features, which can allow users or down-stream systems to explore variations of a certain type of samples.

Formally, we divide the human joints into two sets, $J_s$ and $J_d$, and ask samples in $X$ to have similar motions for joints $J_s$ but diverse motions for joints $J_d$. 
We can slice a motion sample $\mathbf{x}_k$ into two parts: $\mathbf{x}_k = \left(\mathbf{x}_k^s, \mathbf{x}_k^d\right)$ where $\mathbf{x}_k^s$ and $\mathbf{x}_k^d$ correspond to $J_s$ and $J_d$ respectively. Similarly, we can slice the sample set $X$ into two sets: $X_s = \{\mathbf{x}_1^s, \ldots, \mathbf{x}_K^s\}$ and $X_d = \{\mathbf{x}_1^d, \ldots, \mathbf{x}_K^d\}$. We then define a new energy function $E$ for the prior $p(X)$:
\begin{align}
\label{dlow:eq:human_ctrl}
&E(X) = E_d(X_d) + \lambda_s E_s(X_s) + \lambda_r E_r(X)\,,\\
&E_s(X_s) = \frac{1}{K(K-1)}\sum_{i=1}^K\sum_{j\neq i}^K \mathcal{D}^2(\mathbf{x}_i^s, \mathbf{x}_j^s)\,,
\end{align}
where we add another energy term $E_s$ weighted by $\lambda_s$ to minimize the motion distance between samples for joints $J_s$, and we only compute the diversity-promoting term $E_d$ using motions of joints $J_d$.
After optimizing $Q_\gamma$ using the DLow objective with the new energy $E$, we can produce diverse samples $X$ that have similar motions for joints $J_s$.

Furthermore, we may also want to use a reference motion sample $\mathbf{x}_\text{ref}$ to provide the desired features. To achieve this, we can treat $\mathbf{x}_\text{ref}$ as the first sample $\mathbf{x}_1$ in $X$. We first find its corresponding latent code $\mathbf{z}_1 := \mathbf{z}_\text{ref}$ using the CVAE encoder: $\mathbf{z}_\text{ref} =F_\phi^{\boldsymbol{\mu}}(\mathbf{x}_\text{ref}, \mathbf{c})$. We can then find the common variable $\boldsymbol{\epsilon}_\text{ref}$ for generating $X$ using the inverse mapping $\mathcal{T}_{\psi_1}^{-1}$:
\begin{equation}
\boldsymbol{\epsilon_\text{ref}} = \mathcal{T}_{\psi_1}^{-1}(\mathbf{z}_\text{ref}) = \mathbf{A}_1^{-1}(\mathbf{z}_\text{ref} - \mathbf{b}_1)\,.
\end{equation}
With $\boldsymbol{\epsilon}_\text{ref}$ known, we can generate $X$ that includes $\mathbf{x}_\text{ref}$.
In practice, we force $\mathcal{T}_{\psi_1}$ to be an identity mapping to enforce $r_\psi(\mathbf{z}_1|\mathbf{c}) = p(\mathbf{z}_1)$ so that $r_\psi(\mathbf{z}_1|\mathbf{c})$ covers the posterior distribution of $\mathbf{z}_\text{ref}$. Otherwise, if $\mathbf{z}_\text{ref}$ lies outside of the high density region of $r_\psi(\mathbf{z}_1|\mathbf{c})$, it may lead to low-likelihood $\boldsymbol{\epsilon}_\text{ref}$ after the inverse mapping.

\section{Experiments}

\paragraph{Datasets.}
We perform evaluation on two public motion capture datasets: Human3.6M~\cite{ionescu2013human3} and HumanEva-I~\cite{sigal2010humaneva}. Human3.6M is a large-scale dataset with 11 subjects (7 with ground truth) and 3.6 million video frames in total. Each subject performs 15 actions and the human motion is recorded at 50 Hz. Following previous work \cite{martinez2017simple,luvizon20182d,yang20183d,pavllo20193d}, we adopt a 17-joint skeleton and train on five subjects (S1, S5, S6, S7, S8) and test on two subjects (S9 and S11). HumanEva-I is a relatively small dataset, containing only three subjects recorded at 60 Hz. We adopt a 15-joint skeleton~\cite{pavllo20193d} and use the same train/test split provided in the dataset. By using both a large dataset with more variation in motion and a small dataset with less variation, we can better evaluate the generalization of our method to different types of data. For Human3.6M, we predict future motion for 2 seconds based on observed motion of 0.5 seconds. For HumanEva-I, we forecast future motion for 1 second given observed motion of 0.25 seconds.

\paragraph{Baselines.}
To fully evaluate our method, we consider three types of baselines: (1) Deterministic motion prediction methods, including \textbf{ERD}~\cite{fragkiadaki2015recurrent} and \textbf{acLSTM}~\cite{li2017auto}; (2) Stochastic motion prediction methods, including CVAE based methods, \textbf{Pose-Knows}~\cite{walker2017pose} and \textbf{MT-VAE}~\cite{yan2018mt}, as well as a CGAN based method, \textbf{HP-GAN}~\cite{barsoum2018hp}; (3) Diversity-promoting methods for generative models, including \textbf{Best-of-Many}~\cite{bhattacharyya2018accurate}, \textbf{GMVAE}~\cite{dilokthanakul2016deep}, \textbf{DeLiGAN}~\cite{gurumurthy2017deligan}, and \textbf{DSF}~\cite{yuan2019diverse}.

\paragraph{Metrics.} We use the following metrics to measure both sample \emph{diversity} and \emph{accuracy}. (1) \textbf{Average Pairwise Distance (APD)}: average $L2$ distance between all pairs of motion samples to measure diversity within samples, which is computed as $\frac{1}{K(K-1)}\sum_{i=1}^K \sum_{j\neq i}^K \|\mathbf{x}_i - \mathbf{x}_j\|$. (2) \textbf{Average Displacement Error (ADE)}: average $L2$ distance over all time steps between the ground truth motion $\hat{\mathbf{x}}$ and the closest sample, which is computed as $\frac{1}{T}\min_{\mathbf{x} \in X} \|\hat{\mathbf{x}} - \mathbf{x}\|$. (3) \textbf{Final Displacement Error (FDE)}: $L2$ distance between the final ground truth pose $\mathbf{x}^T$ and the closest sample's final pose, which is computed as $\min_{\mathbf{x} \in X} \|\hat{\mathbf{x}}^T - \mathbf{x}^T\|$. (4) \textbf{Multi-Modal ADE (MMADE)}: the multi-modal version of ADE that obtains multi-modal ground truth future motions by grouping similar past motions. (5) \textbf{Multi-Modal FDE (MMFDE)}: the multi-modal version of FDE.

In these metrics, APD has been used to measure sample diversity~\cite{aliakbarian2020stochastic}. ADE and FDE are common metrics for evaluating sample accuracy in trajectory forecasting literature~\cite{alahi2016social,lee2017desire,gupta2018social}. MMADE and MMFDE~\cite{yuan2019diverse} are metrics used to measure a method's ability to produce multi-modal predictions.

\paragraph{Implementation Details.} We use a batch size of 64 and set the latent dimensions $n_z$ to 128 in all experiments. For the CVAE, we sample 5000 training examples every epoch and train the networks for 500 epochs using Adam~\cite{kingma2014adam} and a learning rate of~\mbox{1e-3}. The DLow objective in Eq.~\eqref{dlow:eq:dlow_overall} can be rewritten as: $L(\psi) = \beta L_\text{KL} + \lambda_d E_d + \lambda_r E_r$. We set $(\beta, \lambda_d, \lambda_r)$ to $(1, 25, 2)$ for Human3.6M and $(1, 50, 2)$ for HumanEva-I. For the mappings $T_{\psi_k}$, we specify $\mathbf{A}_k$ to be diagonal to reduce the output size of $Q_\gamma$. This design is mainly for computational efficiency, as we do find that using a full parametrization of $\mathbf{A}_k$ improves performance. The RBF kernel scale $\sigma_d$ is set to 100 for Human3.6M and 20 for HumanEva-I. For both datasets, we sample 5000 training examples every epoch and train $Q_\gamma$ for 500 epochs using Adam with a learning rate of 1e-4.

\begin{table}[ht]
\footnotesize
\centering
\resizebox{\columnwidth}{!}{
\begin{tabular}{@{\hskip 0mm}lcccclcccccl@{\hskip 0mm}}
\toprule
& \multicolumn{5}{c}{Human3.6M~\cite{ionescu2013human3}} & & \multicolumn{5}{c}{HumanEva-I~\cite{sigal2010humaneva}} \\ \cmidrule{2-6} \cmidrule{8-12}
Method & APD $\uparrow$ & ADE $\downarrow$ & FDE $\downarrow$ & MMADE $\downarrow$ & MMFDE $\downarrow$ & & APD $\uparrow$ & ADE $\downarrow$ & FDE $\downarrow$ & MMADE $\downarrow$ & MMFDE $\downarrow$ \\ \midrule
DLow (Ours) & \textbf{11.741} & \textbf{0.425} & \textbf{0.518} & \textbf{0.495} & \textbf{0.531} &  & \textbf{4.855} & \textbf{0.251} & \textbf{0.268} & \textbf{0.362} & \textbf{0.339} \\
 ERD \cite{fragkiadaki2015recurrent}                     & 0  & 0.722 & 0.969 & 0.776 & 0.995 &  & 0 & 0.382 & 0.461 & 0.521 & 0.595 \\
 acLSTM \cite{li2017auto}                      & 0  & 0.789 & 1.126 & 0.849 & 1.139 &  & 0 & 0.429 & 0.541 & 0.530 & 0.608 \\
 Pose-Knows \cite{walker2017pose}              & 6.723  & 0.461 & 0.560 & 0.522 & 0.569 &  & 2.308 & 0.269 & 0.296 & 0.384 & 0.375 \\
 MT-VAE \cite{yan2018mt}           & 0.403  & 0.457 & 0.595 & 0.716 & 0.883 &  & 0.021 & 0.345 & 0.403 & 0.518 & 0.577 \\
 HP-GAN \cite{barsoum2018hp}           & 7.214  & 0.858 & 0.867 & 0.847 & 0.858 &  & 1.139 & 0.772 & 0.749 & 0.776 & 0.769 \\
 Best-of-Many \cite{bhattacharyya2018accurate} & 6.265  & 0.448 & 0.533 & 0.514 & 0.544 &  & 2.846 & 0.271 & 0.279 & 0.373 & 0.351 \\
 GMVAE \cite{dilokthanakul2016deep}            & 6.769  & 0.461 & 0.555 & 0.524 & 0.566 &  & 2.443 & 0.305 & 0.345 & 0.408 & 0.410 \\
 DeLiGAN \cite{gurumurthy2017deligan}          & 6.509  & 0.483 & 0.534 & 0.520 & 0.545 &  & 2.177 & 0.306 & 0.322 & 0.385 & 0.371 \\
 DSF \cite{yuan2019diverse}                    & 9.330  & 0.493 & 0.592 & 0.550 & 0.599 &  & 4.538 & 0.273 & 0.290 & 0.364 & 0.340 \\
\bottomrule
\end{tabular}
}
\vspace{4mm}
\caption{\textbf{Quantitative results} on Human3.6M and HumanEva-I.}
\label{dlow:table:quan}
\end{table}

\begin{table}[ht]
\footnotesize
\centering
\resizebox{\columnwidth}{!}{
\begin{tabular}{@{\hskip 1mm}ccc@{\hskip 2mm}cccclcccccl@{\hskip 1mm}}
\toprule
\multicolumn{2}{c}{Energy} & & \multicolumn{5}{c}{Human3.6M~\cite{ionescu2013human3}} & & \multicolumn{5}{c}{HumanEva-I~\cite{sigal2010humaneva}} \\ \cmidrule{1-2} \cmidrule{4-8} \cmidrule{10-14}
$E_d$ & $E_r$& & APD $\uparrow$ & ADE $\downarrow$ & FDE $\downarrow$ & MMADE $\downarrow$ & MMFDE $\downarrow$ & & APD $\uparrow$ & ADE $\downarrow$ & FDE $\downarrow$ & MMADE $\downarrow$ & MMFDE $\downarrow$ \\ \midrule
\cmark & \cmark & & 11.741 & \textbf{0.425} & \textbf{0.518} & \textbf{0.495} & \textbf{0.531} & & 4.855 & \textbf{0.251} & \textbf{0.268} & \textbf{0.362} & \textbf{0.339}\\  
\cmark & \xmark & & \textbf{13.091} & 0.546 & 0.663 & 0.599 & 0.669 & & \textbf{4.927} & 0.263 & 0.281 & 0.368  & 0.347\\  
\xmark & \cmark & & 6.844 & 0.432 & 0.525 & 0.500 & 0.539 & & 2.355 & 0.252 & 0.277 & 0.376 & 0.366 \\  
\xmark & \xmark & & 6.383 & 0.520 & 0.629 & 0.577 & 0.638 & & 2.247 & 0.281 & 0.317 & 0.395 & 0.393 \\  
\bottomrule
\end{tabular}
}
\vspace{4mm}
\caption{\textbf{Ablation study} on Human3.6M and HumanEva-I.}
\label{dlow:table:ablation}
\end{table}

\subsection{Quantitative Results}
We summarize the quantitative results on Human3.6M and HumanEva-I in Table~\ref{dlow:table:quan}. The metrics are computed with the sample set size $K = 50$. For both datasets, we can see that our method, DLow, outperforms all baselines in terms of both sample diversity (APD) and accuracy (ADE, FDE) as well as covering multi-modal ground truth (MMADE, MMFDE). Determinstic methods like ERD~\cite{fragkiadaki2015recurrent} and acLSTM~\cite{li2017auto} do not perform well because they only predict one future trajectory which can lead to mode averaging. Methods like MT-VAE~\cite{yan2018mt} produce trajectories samples that lack diversity so they fail to cover the multi-modal ground-truth (indicated by high MMADE and MMFDE) despite having decently low ADE and FDE. We would also like to point out the closest competitor DSF~\cite{yuan2019diverse} can only generate one deterministic set of samples, while our method can produce multiple diverse sets by sampling $\boldsymbol{\epsilon}$.

\paragraph{Ablation Study.}
We further perform an ablation study (Table~\ref{dlow:table:ablation}) to analyze the effects of the two energy terms $E_d$ and $E_r$ in Eq.~\eqref{dlow:eq:human_prior}. First, without the reconstruction term $E_r$, the DLow variant is able to achieve higher diversity (APD) at the cost of sample accuracy (ADE, FDE, MMADE, MMFDE). This is expected because the network only optimizes the diversity term $E_d$ and focuses solely on diversity. Second, for the variant without $E_d$, both sample diversity and accuracy decrease. It is intuitive to see why the diversity (APD) decreases. To see why the sample accuracy (ADE, FDE, MMADE, MMFDE) also decreases, we should consider the fact that a more diverse set of samples have a better chance at covering the ground truth. Finally, when we remove both $E_d$ and $E_r$ (i.e., only optimize $L_\text{KL}$), the results are the worst, which is expected.

\begin{figure}[t]
    \centering
    \includegraphics[width=\linewidth]{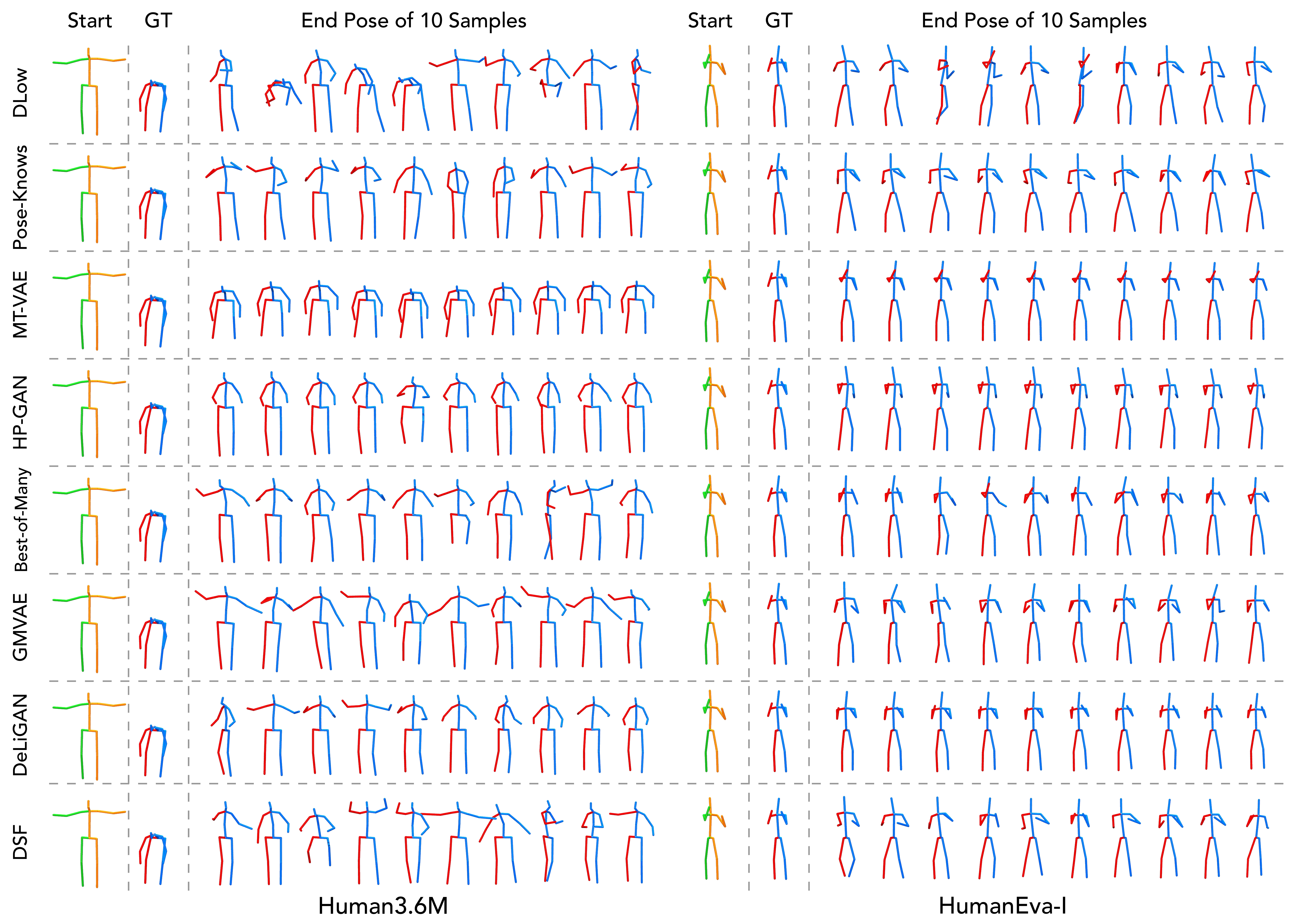}
    \caption{\textbf{Qualitative Results} on Human3.6M and HumanEva-I.}
    \label{dlow:fig:comp_base}
\end{figure}

\begin{figure}[t]
    \centering
    \includegraphics[width=\linewidth]{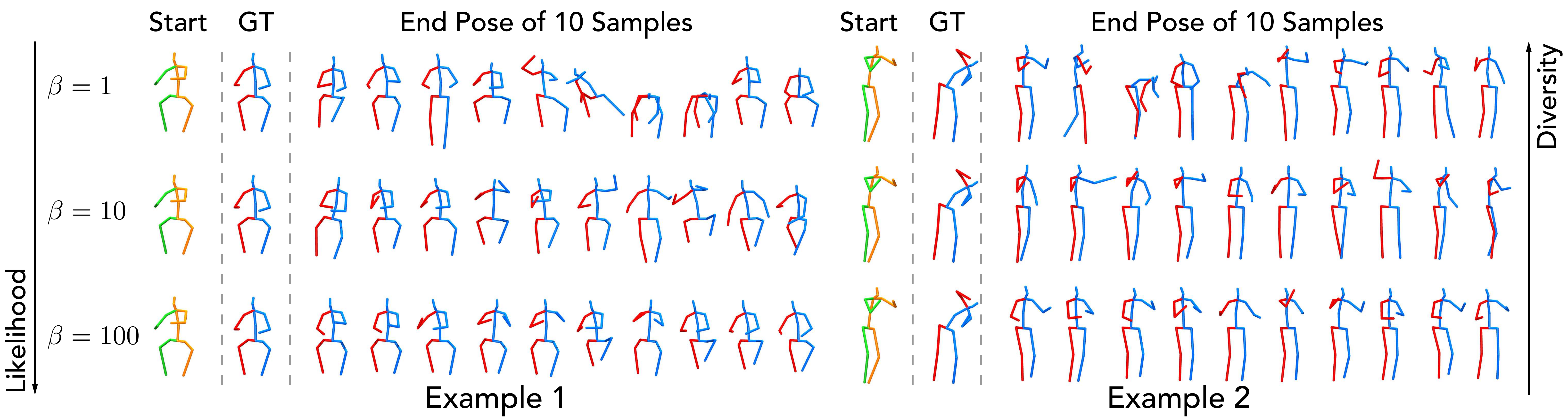}
    \caption{\textbf{Varying $\beta$ in DLow} allows us to balance between diversity and likelihood. }
    \label{dlow:fig:beta}
\end{figure}

\subsection{Qualitative Results}
To visually evaluate the diversity and accuracy of each method, we present a qualitative comparison in Fig.~\ref{dlow:fig:comp_base} where we render the start pose, the end pose of the ground truth future motion, and the end pose of 10 motion samples. Note that we do not model the global translation of the person, which is why some sitting motions appear to be floating. For Human3.6M, we can see that our method DLow can predict a wide array of future motions, including standing, sitting, bending, crouching, and turning, which cover the ground truth bending motion. In contrast, the baseline methods mostly produce perturbations of a single motion --- standing. For HumanEva-I, we can see that DLow produces interesting variations of the fighting motion, while the baselines produce almost identical future motions.

\begin{figure}[t]
    \centering
    \includegraphics[width=\linewidth]{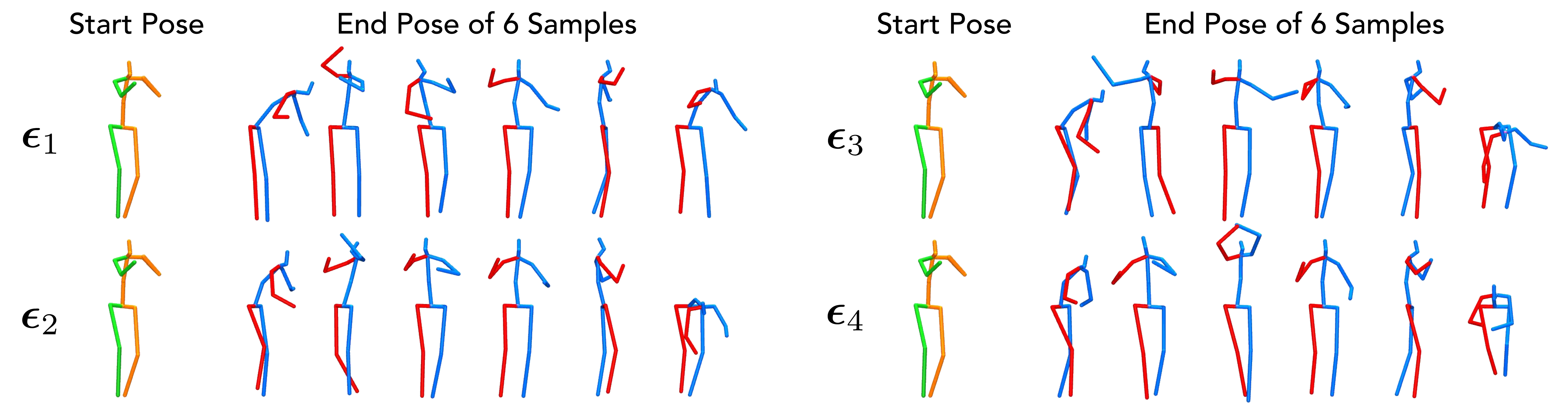}
    \caption{\textbf{Effect of varying $\boldsymbol{\epsilon}$} on motion samples. }
    \label{dlow:fig:var}
\end{figure}

\paragraph{Diversity vs. Likelihood.} As discussed in the approach section, the $\beta$ in Eq.~\eqref{dlow:eq:dlow_opt} represents the trade-off between sample diversity and likelihood. To verify this, we trained three DLow models with different $\beta$ (1, 10, 100) and visualize the motion samples generated by each model in Fig.~\ref{dlow:fig:beta}. We can see that a larger $\beta$ leads to less diverse samples which correspond to the major mode of the generator distribution, while a smaller $\beta$ can produce more diverse motion samples covering other plausible yet less likely future motions.

\paragraph{Effect of varying $\boldsymbol{\epsilon}$.} A key difference between our method and DSF~\cite{yuan2019diverse} is that we can generate multiple diverse sets of samples while DSF can only produce a fixed diverse set. To demonstrate this, we show in Fig.~\ref{dlow:fig:var} how the motion samples of DLow change with different $\boldsymbol{\epsilon}$. By comparing the four sets of motion samples, one can conclude that changing $\boldsymbol{\epsilon}$ varies each set of samples but preserves the main structure of each motion.

\begin{figure}[t]
    \centering
    \includegraphics[width=\linewidth]{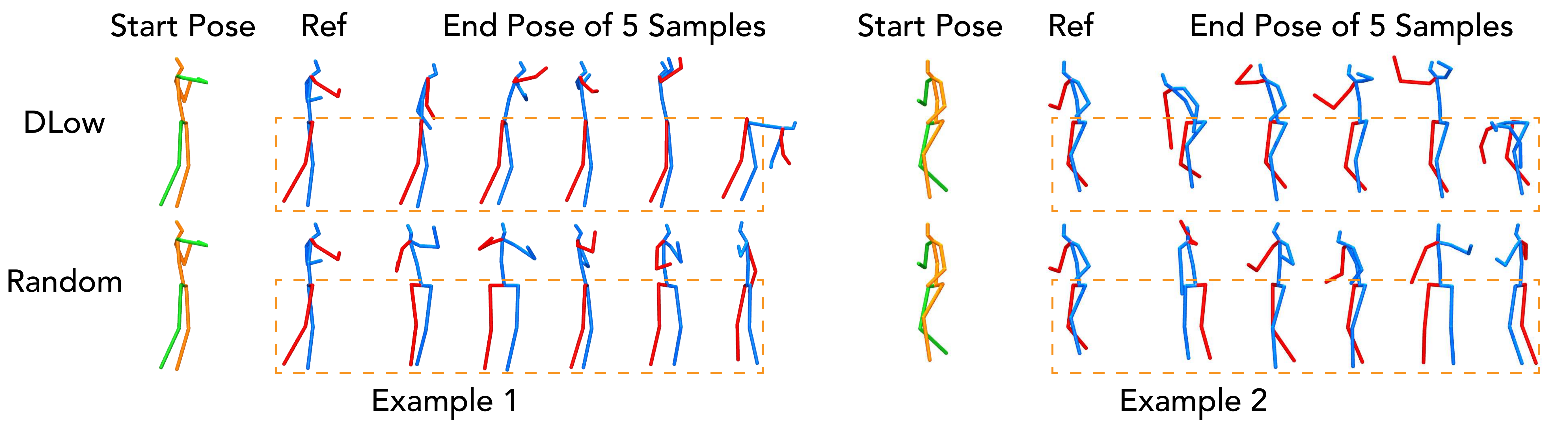}
    \caption{\textbf{Controllable Motion Prediction.} DLow enables samples to have more similar leg motion to the reference.}
    \label{dlow:fig:control}
\end{figure}

\paragraph{Controllable Motion Prediction.}
As highlighted before, the flexible design of the diversity-promoting prior enables a new application, controllable motion prediction, where we predict diverse motions that share some common features. We showcase this application by conducting an experiment using the energy function defined in Eq.~\eqref{dlow:eq:human_ctrl}. 
The network is trained so that the leg motion of the motion samples is similar while the upper-body motion is diverse.
The results are shown in Fig.~\ref{dlow:fig:control}. We can see that given a reference motion, our method can generate diverse upper-body motion and preserve similar leg motion, while random samples from the CVAE cannot enforce similar leg motion.

\section{Conclusion}
We have proposed a novel sampling strategy, DLow, for deep generative models to obtain a diverse set of future human motions. We introduced learnable latent mapping functions which allowed us to generate a set of correlated samples, whose diversity can be optimized by a diversity-promoting prior. Experiments demonstrated superior performance in generating diverse motion samples. Moreover, we showed that the flexible design of the diversity-promoting prior further enables new applications, such as controllable human motion prediction. We hope that our exploration of deep generative models through the lens of diversity will encourage more work towards understanding the complex nature of modeling and predicting future human behavior.

\chapter{Multi-Agent Stochastic Trajectory Generation with Transformers}
\label{chap:aformer}

\section{Introduction}
\label{aformer:sec:intro}
The safe planning of autonomous systems such as self-driving vehicles requires forecasting accurate future trajectories of surrounding agents (\eg, pedestrians, vehicles). However, multi-agent trajectory forecasting is challenging since the social interaction between agents, \ie, behavioral influence of an agent on others, is a complex process. The problem is further complicated by the uncertainty of each agent's future behavior, \ie, each agent has its latent intent unobserved by the system (\eg, turning left or right) that governs its future trajectory and in turn affects other agents. Therefore, a good multi-agent trajectory forecasting method should effectively model (1) the complex social interaction between agents and (2) the latent intent of each agent's future behavior and its social influence on other agents. 

\begin{figure}
    \centering
    \includegraphics[width=0.85\linewidth]{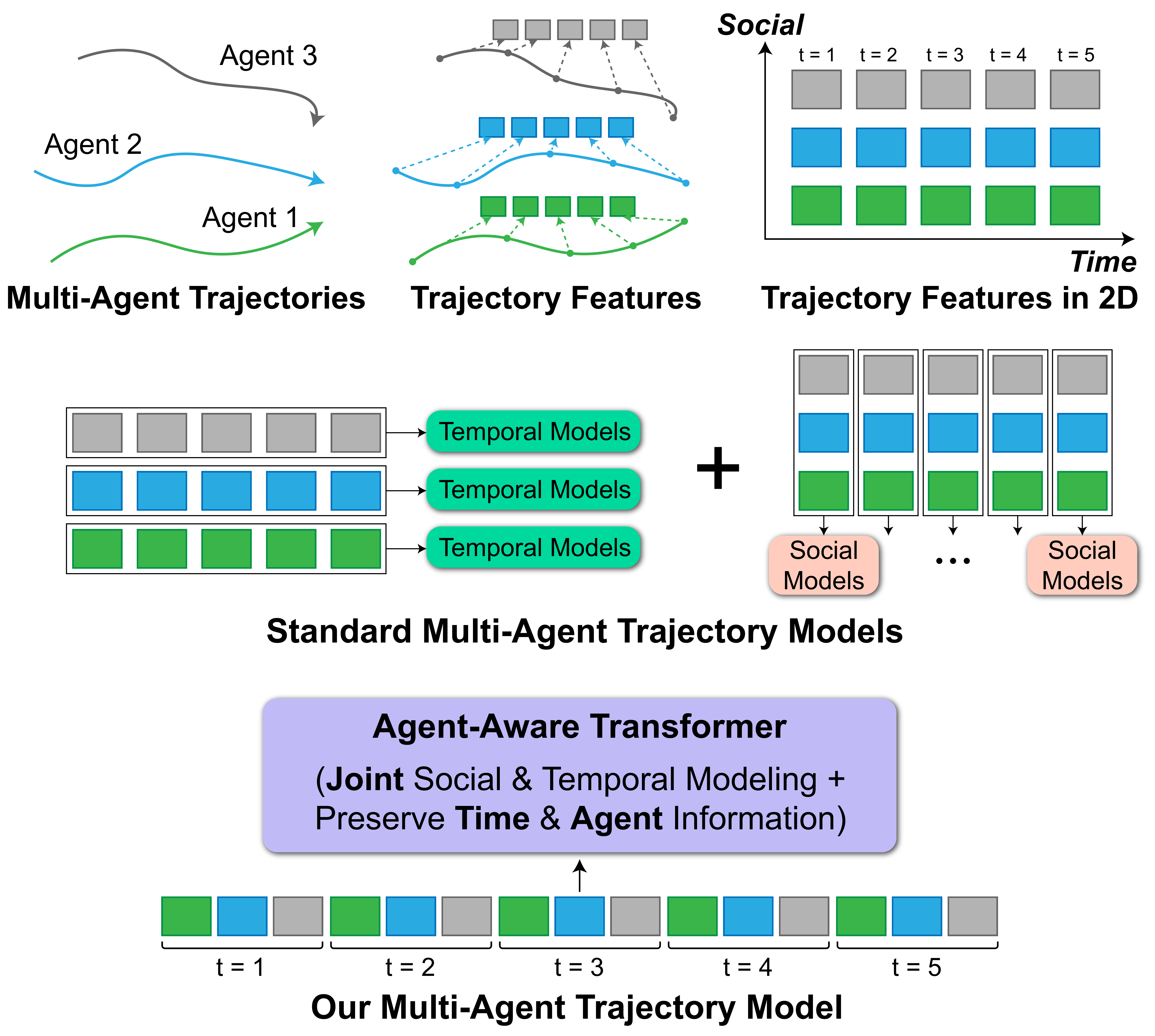}
    \vspace{5mm}
    \caption{Different from standard approaches that model multi-agent trajectories in the time and social dimensions separately, our AgentFormer\ allows for joint modeling of the time and social dimensions while preserving time and agent information. }
    \label{aformer:fig:teaser}
\end{figure}

Multi-agent social interaction modeling involves two key dimensions as illustrated in Fig.~\ref{aformer:fig:teaser} (Top): (1)~\textbf{\emph{time dimension}}, where we model how past agent states (positions and velocities) influence future agent states; (2)~\textbf{\emph{social dimension}}, where we model how each agent's state affects the state of other agents. Most prior multi-agent trajectory forecasting methods model these two dimensions separately (see Fig.~\ref{aformer:fig:teaser}\,(Middle)). Approaches like~\cite{kosaraju2019social,alahi2016social,gupta2018social} first use temporal models (\eg, LSTMs~\cite{hochreiter1997long} or Transformers~\cite{vaswani2017attention}) to summarize trajectory features over time for each agent independently and then input the summarized temporal features to social models (\eg, graph neural networks~\cite{kipf2016semi}) to capture social interaction between agents. Alternatively, methods like~\cite{salzmann2020trajectron++,huang2019stgat} first use social models to produce social features for each agent at each independent timestep and then apply temporal models over the social features. In this work, we argue that modeling the time and social dimensions separately can be suboptimal since the independent feature encoding over either the time or social dimension is not informed by features across the other dimension, and the encoded features may not contain the necessary information for modeling the other dimension.

To tackle this problem, we propose a new Transformer model, termed AgentFormer, that simultaneously learns representations from both the time and social dimensions. \mbox{AgentFormer}\ allows an agent's state at one time to affect another agent's state at a future time \emph{directly} instead of through intermediate features encoded over one dimension. As Transformers require sequences as input, we leverage a sequence representation of multi-agent trajectories by flattening trajectory features across time and agents (see Fig.~\ref{aformer:fig:teaser}\,(Bottom)). However, directly applying standard Transformers to these multi-agent sequences will result in a loss of \emph{time} and \emph{agent} information since standard attention operations discard the timestep and agent identity associated with each element in the sequence. We solve the loss of time information using a time encoder that appends a timestamp feature to each element. However, the loss of agent identity is a more complicated problem: unlike time, there is no innate ordering between agents, and assigning an agent index-based encoding will break the required permutation invariance of agents and create artificial dependencies on agent indices in the model. Instead, we propose a novel agent-aware attention mechanism to preserve agent information. Specifically, agent-aware attention generates two sets of keys and queries via different linear transformations; one set of keys and queries is used to compute inter-agent attention  (agent to agent) while the other set is designated for intra-agent attention (agent to itself). This design allows agent-aware attention to attend to elements of the same agent differently than elements of other agents, thus keeping the notion of agent identity. Agent-aware attention can be implemented efficiently via masked operations. Furthermore, AgentFormer\ can also encode rule-based connectivity between agents (\eg, based on distance) by masking out the attention weights between unconnected agents.

Based on AgentFormer, which allows us to model social interaction effectively, we propose a multi-agent trajectory prediction framework that also models the social influence of each agent's future trajectory on other agents. The probabilistic formulation of the model follows the conditional variational autoencoder (CVAE~\cite{kingma2013auto}) where we model the generative future trajectory distribution conditioned on context (\eg, past trajectories, semantic maps). We introduce a latent code for each agent to represent its latent intent. To model the social influence of each agent's future behavior (governed by latent intent) on other agents, the latent codes of all agents are jointly inferred from the future trajectories of all agents during training, and they are also jointly used by a trajectory decoder to output socially-aware multi-agent future trajectories. Thanks to AgentFormer, the trajectory decoder can attend to features of any agent at any previous timestep when inferring an agent's future position. To improve the diversity of sampled trajectories and avoid similar samples caused by random sampling, we further adopt a multi-agent trajectory sampler that can generate diverse and plausible multi-agent trajectories by mapping context to various configurations of all agents' latent codes.

We evaluate our method on well-established pedestrian datasets, ETH~\cite{pellegrini2009you} and UCY~\cite{lerner2007crowds}, and an autonomous driving dataset, nuScenes~\cite{caesar2020nuscenes}. On ETH/UCY and nuScenes, we outperform state-of-the-art multi-agent prediction methods with substantial performance improvement. We further conduct extensive ablation studies to show the superiority of AgentFormer\ over various combinations of social and temporal models. We also demonstrate the efficacy of agent-aware attention against agent encoding.

To summarize, the main contributions of this paper are:
(1) We propose a new Transformer that simultaneously models the time and social dimensions of multi-agent trajectories with a sequence representation. (2) We propose a novel agent-aware attention mechanism that preserves the agent identity of each element in the multi-agent trajectory sequence. (3) We present a multi-agent forecasting framework that models the latent intent of all agents jointly to produce socially-plausible future trajectories. (4) Our approach substantially improves the state of the art on well-established pedestrian and autonomous driving datasets.

\vspace{-3mm}
\section{Related Work}

\paragraph{Sequence Modeling.} Sequences are an important representation of data such as video, audio, price, etc. Historically, RNNs (\eg, LSTMs~\cite{hochreiter1997long}, GRUs~\cite{chung2014empirical}) have achieved remarkable success in sequence modeling, with applications to speech recognition~\cite{xiong2018microsoft,miao2015eesen}, image captioning~\cite{xu2015show}, machine translation~\cite{luong2015effective}, human pose estimation~\cite{yuan20183d,kocabas2020vibe}, etc. In particular, RNNs have been the preferred temporal models for trajectory and motion forecasting. Many RNN-based methods model the trajectory pattern of pedestrians to predict their 2D future locations~\cite{alahi2016social,ivanovic2019trajectron,zhang2019sr}. Prior work has also used RNNs to model the temporal dynamics of 3D human pose~\cite{fragkiadaki2015recurrent,yuan2019ego,yuan2020residual}. With the invention of Transformers and positional encoding~\cite{vaswani2017attention}, many works start to adopt Transformers for sequence modeling due to their strong ability to capture long-range dependencies. Transformers have first dominated the natural language processing (NLP) domain across various tasks~\cite{devlin2018bert,lan2019albert,yang2019xlnet}. Beyond NLP, numerous visual Transformers have been proposed to tackle vision tasks, such as image classification~\cite{dosovitskiy2020image}, object detection~\cite{carion2020end}, and instance segmentation~\cite{wang2020end}. Recently, Transformers have also been used for trajectory forecasting. Transformer-TF~\cite{giuliari2020transformer} applies the standard Transformer to predict the future trajectories of each agent independently. STAR~\cite{yu2020spatio} uses separate temporal and spatial Transformers to forecast multi-agent trajectories. Interaction Transformer~\cite{li2020end} combines RNNs and Transformers for multi-agent trajectory modeling.
Different from prior work, Our AgentFormer\ leverages a sequence representation of multi-agent trajectories and a novel agent-aware attention mechanism to preserve time and agent information in the sequence.

\vspace{-1mm}
\paragraph{Trajectory Prediction.}
Early work on trajectory prediction adopts a deterministic approach using models such as social forces~\cite{helbing1995social}, Gaussian process (GP)~\cite{wang2007gaussian}, and RNNs~\cite{alahi2016social,morton2016analysis,vemula2018social}. A thorough review of these deterministic methods is provided in~\cite{rudenko2020human}. As the future trajectory of an agent is uncertain and often multi-modal, recent trajectory prediction methods start to model the trajectory distribution with deep generative models~\cite{kingma2013auto,goodfellow2014generative,rezende2015variational} such as conditional variational autoencoders (CVAEs)~\cite{lee2017desire,yuan2019diverse,ivanovic2019trajectron,tang2019multiple,weng2021ptp,salzmann2020trajectron++}, generative adversarial networks (GANs)~\cite{gupta2018social,sadeghian2019sophie,kosaraju2019social,zhao2019multi}, and normalizing flows (NFs)~\cite{rhinehart2018r2p2,rhinehart2019precog,guan2020generative}. Most of these methods follow a seq2seq structure~\cite{bahdanau2014neural,cho2014learning} and predict future trajectories using intermediate features of past trajectories. In contrast, our AgentFormer-based trajectory prediction framework can directly attend to features of any agent at any previous timestep when inferring an agent's future position. Moreover, our approach models the future trajectories of all agents jointly to predict socially-aware trajectories.

\vspace{-1mm}
\paragraph{Social Interaction Modeling.} Methods for social interaction modeling can be categorized based on how they model the time and social dimensions. While RNNs~\cite{hochreiter1997long,chung2014empirical} and Transformers~\cite{vaswani2017attention} are the prefered temporal models~\cite{huang2019stgat,alahi2016social,yu2020spatio}, graph neural networks (GNNs)~\cite{kipf2016semi,li2015gated} are often employed as the social models for interaction modeling~\cite{kipf2018neural,li2020evolvegraph,kosaraju2019social}. One popular type of methods~\cite{kosaraju2019social,alahi2016social,gupta2018social} first uses temporal models to summarize trajectory features over time for each agent independently and then feeds the temporal features to social models to obtain socially-aware agent features. Alternatively, approaches like~\cite{salzmann2020trajectron++,huang2019stgat} first use social models to produce social features of each agent at each independent timestep and then apply temporal models to summarize the social features over time for each agent. One common characteristic of these prior works is that they model the time and social dimensions on separate levels. This can be suboptimal since it prevents an agent's feature at one time from directly interacting with another agent's feature at a different time, thus limiting the model's ability to capture long-range dependencies. Instead, our method models both the time and social dimensions simultaneously, allowing direct feature interaction across time and agents.

\vspace{-2mm}
\section{Approach}
\label{aformer:sec:approach}
We formulate multi-agent trajectory prediction as modeling the generative future trajectory distribution of $N$ (variable) agents conditioned on their past trajectories. For observed timesteps $t \leq 0$, we represent the joint state of all $N$ agents at time $t$ as $\mb{X}^t = (\mb{x}_1^t, \mb{x}_2^t, \ldots, \mb{x}_N^t)$, where $\mb{x}_n^t \in \mathbb{R}^{d_s}$ is the state of agent $n$ at time $t$, which includes the position, velocity and (optional) heading angle of the agent. We denote the history of all agents as $\mb{X} = \left(\mb{X}^{-H}, \mb{X}^{-H+1}, \ldots, \mb{X}^{0}\right)$ which includes the joint agent state at all $H+1$ observed timesteps. Similarly, the joint state of all $N$ agents at future time $t$ ($t > 0$) is denoted as $\mb{Y}^t = (\mb{y}_1^t, \mb{y}_2^t, \ldots, \mb{y}_N^t)$, where $\mb{y}_n^t \in \mathbb{R}^{d_p}$ is the future position of agent $n$ at time $t$. We denote the future trajectories of all $N$ agents over $T$ future timesteps as $\mb{Y} = \left(\mb{Y}^{1}, \mb{Y}^{2}, \ldots, \mb{Y}^{T}\right)$. Depending on the data, optional contextual information $\mb{I}$ may also be given, such as a semantic map around the agents (annotations of sidewalks, road boundaries, etc.). Our goal is to learn a generative model $p_\theta(\mb{Y}|\mb{X}, \mb{I})$ where $\theta$ are the model parameters.

In the following, we first introduce the proposed agent-aware Transformer, AgentFormer, for joint modeling of socio-temporal relations. We then present a stochastic multi-agent trajectory prediction framework that jointly models the latent intent of all agents. %

\subsection{AgentFormer: Agent-Aware Transformers}
\label{aformer:sec:agent_former}
Our agent-aware Transformer, AgentFormer, is a model that learns representations from multi-agent trajectories over both time and social dimensions simultaneously, in contrast to standard approaches that model the two dimensions in separate stages. AgentFormer\ has two types of modules -- encoders and decoders, which follow the encoder and decoder design of the original Transformer~\cite{vaswani2017attention} but with two major differences: (1) it replaces positional encoding with a time encoder; (2) it uses a novel agent-aware attention mechanism instead of the scaled dot-product attention. As we will discuss below, these two modifications are motivated by a sequence representation of multi-agent trajectories that is suitable for Transformers.

\paragraph{Multi-Agent Trajectories as a Sequence.}
The past multi-agent trajectories $\mb{X}$ can be denoted as a sequence $\mb{X} = \left(\mb{x}_1^{-H}, \ldots, \mb{x}_N^{-H}, \mb{x}_1^{-H+1}, \ldots, \mb{x}_N^{-H+1}, \ldots, \mb{x}_1^{0}, \ldots,  \mb{x}_N^{0}\right)$ of length $L_p=N\times(H + 1)$. Similarly, the future multi-agent trajectories can also be represented as a sequence $\mb{Y} = \left(\mb{y}_1^{1}, \ldots, \mb{y}_N^{1}, \mb{y}_1^{2}, \ldots, \mb{y}_N^{2}, \ldots, \mb{y}_1^{T}, \ldots,  \mb{y}_N^{T}\right)$ of length $L_f = N\times T$. We adopt this sequence representation to be compatible with Transformers. At first glance, it may seem that we can directly apply standard Transformers to these sequences to model temporal and social relations. However, there are \emph{two problems} with this approach:
(1)~\textbf{loss of \emph{time} information}, as Transformers have no notion of time when computing attention for each element (\eg, $\mb{x}_n^{t}$) \emph{w.r.t.} other elements in the sequence; for instance, $\mb{x}_n^{t}$ does not know $\mb{x}_{m}^{t}$ is a feature of the same timestep while $\mb{x}_n^{t+1}$ is a feature of the next timestep;
(2)~\textbf{loss of \emph{agent} information}, since Transformers do not consider agent identities when applying attention to each element, and elements of the same agent are not distinguished from elements of other agents; for example, when computing attention for $\mb{x}_n^{t}$, both $\mb{x}_n^{t+1}$ and $\mb{x}_{m}^{t+1}$ are treated the same, disregarding the fact that $\mb{x}_n^{t+1}$ is from the same agent while $\mb{x}_{m}^{t+1}$ is from a different agent. Below, we present the solutions to these two problems -- (1)~time encoder and (2) agent-aware attention.

\paragraph{Time Encoder.} To inform AgentFormer\ about the timestep associated with each element in the trajectory sequence, we employ a time encoder similar to the positional encoding in the original Transformer. Instead of encoding the position of each element based on its index in the sequence, we compute a timestamp feature based on the timestep $t$ of the element. The timestamp uses the same sinusoidal design as the positional encoding. Let us take the past trajectory sequence $\mb{X}$ as an example. For each element $\mathbf{x}_n^t$, the timestamp feature $\bs{\tau}_n^t \in \mathbb{R}^{d_\tau}$ is defined as
\begin{equation*}
\bs{\tau}_n^t(k) =
\begin{cases}
    \sin((t + H)/10000^{k/d_\tau}),              & k\text{ is even}\\
    \cos((t + H)/10000^{(k-1)/d_\tau}),          & k\text{ is odd}
\end{cases}
\end{equation*}
where $\bs{\tau}_n^t(k)$ denotes the $k$-th feature of $\bs{\tau}_n^t$ and $d_\tau$ is the feature dimension of the timestamp. The time encoder outputs a timestamped sequence $\bar{\mb{X}}$ and each element $\bar{\mb{x}}_n^t \in \mathbb{R}^{d_\tau}$ in $\bar{\mb{X}}$ is computed as $ \bar{\mb{x}}_n^t = \mb{W}_2(\mb{W}_1\mb{x}_n^t\oplus \bs{\tau}_n^t)$ where $\mb{W}_1 \in \mathbb{R}^{d_\tau\times d_s}$ and $\mb{W}_2 \in \mathbb{R}^{d_\tau\times 2d_\tau}$ are weight matrices and $\oplus$ denotes concatenation.

\begin{figure}[t]
    \centering
    \includegraphics[width=\linewidth]{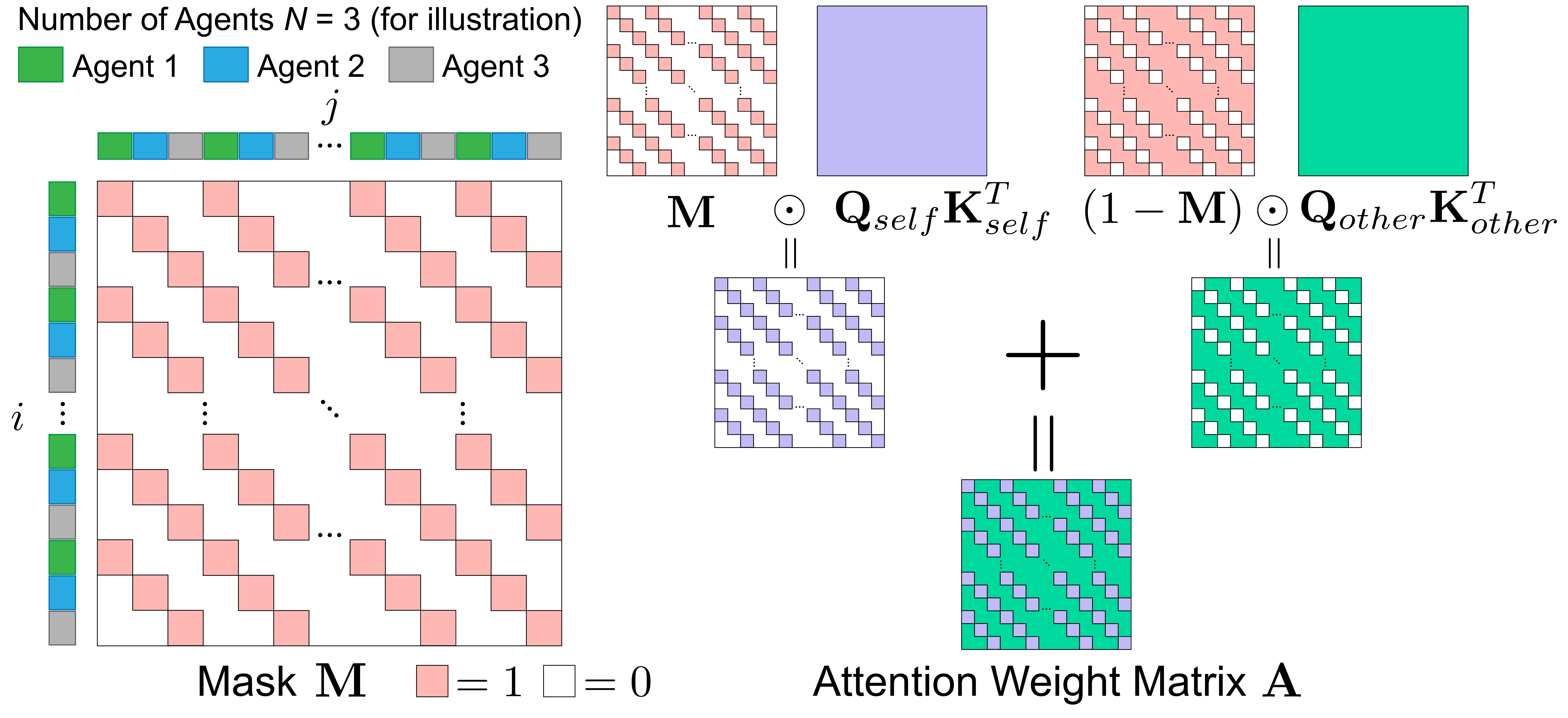}
    \caption{\textbf{Illustration of agent-aware attention.} The mask $\mb{M}$ allows the attention weights in $\mb{A}$ to be computed differently based on whether the $i$-th query and $j$-th key belong to the same agent. }
    \label{aformer:fig:attention}
\end{figure}

\paragraph{Agent-Aware Attention.} To preserve agent information in the trajectory sequence, it may be tempting to employ a similar strategy to the time encoder, such as an agent encoder that assigns an agent index-based encoding to each element in the sequence. However, using such agent encoding is not effective as we will show in the experiments. The reason is that, different from time which is naturally ordered, there is no innate ordering between agents, and assigning encodings based on agent indices will break the required permutation invariance of agents and create artificial dependencies on agent indices in the model.

We tackle the loss of agent information from a different angle by proposing a novel agent-aware attention mechanism. The agent-aware attention takes as input keys $\mb{K}$, queries $\mb{Q}$ and values $\mb{V}$, each of which uses the sequence representation of multi-agent trajectories. As an example, let the keys $\mb{K}$ and values $\mb{V}$ be the past trajectory sequence $\mb{X}\in \mathbb{R}^{L_p \times d_s}$, and let the queries $\mb{Q}$ be the future trajectory sequence $\mb{Y}\in \mathbb{R}^{L_f \times d_p}$. Recall that $\mb{X}$ is of length $L_p = N\times(H+1)$ as $\mb{X}$ contains the trajectory features of $N$ agents of $H+1$ past timesteps; $\mb{Y}$ is of length $L_f = N\times T$ containing trajectory features of $T$ future timesteps. The output of agent-aware attention is computed as
\begin{align}
&\text{AgentAwareAttention}(\mb{Q}, \mb{K}, \mb{V}) = \text{softmax}\left(\frac{\mb{A}}{\sqrt{d_k}}\right)V \\
\label{aformer:eq:attn}
&\mb{A} = \mb{M} \odot (\mb{Q}_{self} \mb{K}_{self}^T) + (1 - \mb{M}) \odot (\mb{Q}_{other}\mb{K}_{other}^T) \\
&\mb{Q}_{self} = \mb{Q}\mb{W}_{self}^Q, \quad\;\; \mb{K}_{self} = \mb{K}\mb{W}_{self}^K\\
&\mb{Q}_{other} = \mb{Q}\mb{W}_{other}^Q, \quad \mb{K}_{other} = \mb{K}\mb{W}_{other}^K
\end{align}
where $\odot$ denotes element-wise product and we use two sets of projections $\{\mb{W}_{self}^Q,\mb{W}_{self}^K\}$ and $\{\mb{W}_{other}^Q,\mb{W}_{other}^K\}$ to generate projected keys $\mb{K}_{self},\mb{K}_{other}\in \mathbb{R}^{L_p\times d_k}$ and queries $\mb{Q}_{self},\mb{Q}_{other}\in \mathbb{R}^{L_f\times d_k}$ with key (query) dimension $d_k$. Each element $A_{ij}$ in the attention weight matrix $\mb{A}$ represents the attention weight between the $i$-th query $\mb{q}_i$ and \mbox{$j$-th} key $\mb{k}_j$. As illustrated in Fig.~\ref{aformer:fig:attention}, when computing the attention weight matrix $\mb{A} \in \mathbb{R}^{L_f\times L_p}$, we also use a mask $\mb{M} \in \mathbb{R}^{L_f\times L_p}$ which is defined as
\begin{equation}
M_{ij} = \mathbbm{1}  (i \text{ mod } N = j \text{ mod } N)
\end{equation}
where $M_{ij}$ denotes each element inside the mask $\mb{M}$ and $\mathbbm{1}(\cdot)$ denotes the indicator function. As $\cdot$ mod $N$ computes the agent index of a query/key, $M_{ij}$ equals to one if the $i$-th query $\mb{q}_i$ and $j$-th key $\mb{k}_j$ belongs to the same agent, and $M_{ij}$ equals to zero otherwise, as shown in Fig.~\ref{aformer:fig:attention}. Using the mask $\mb{M}$, Eq.~\eqref{aformer:eq:attn} computes each element $A_{ij}$ of the attention weight matrix $\mb{A}$ differently based on the agreement of agent identity: If $\mb{q}_i$ and $\mb{k}_j$ have the same agent identity, $A_{ij}$ is computed using the projected queries $\mb{Q}_{self}$ and keys $\mb{K}_{self}$ designated for intra-agent attention (agent to itself); If $\mb{q}_i$ and $\mb{k}_j$ have different agent identities, $A_{ij}$ is computed using the projected queries $\mb{Q}_{other}$ and keys $\mb{K}_{other}$ designated for inter-agent attention (agent to other agents). In this way, the agent-aware attention learns to attend to elements of the same agent in the sequence differently than elements of other agents, thus preserving the notion of agent identity. Note that AgentFormer\ only uses agent-aware attention to replace the scaled dot-product attention in the original Transformer and still allows multi-head attention to learn distributed representations.

\paragraph{Encoding Agent Connectivity.}
AgentFormer\ can also encode rule-based agent connectivity information by masking out the attention weights between unconnected agents. Specifically, we define that two agents $n$ and $m$ are connected if their distance $D_{nm}$ at the current time ($t=0$) is smaller than a threshold $\eta$. If agents $n$ and $m$ are not connected, we set the attention weight $A_{ij} = -\infty$ between any query $\mb{q}_i$ of agent $n$ and any key $\mb{k}_j$ of agent $m$.

\begin{figure*}[t]
    \centering
    \includegraphics[width=\textwidth]{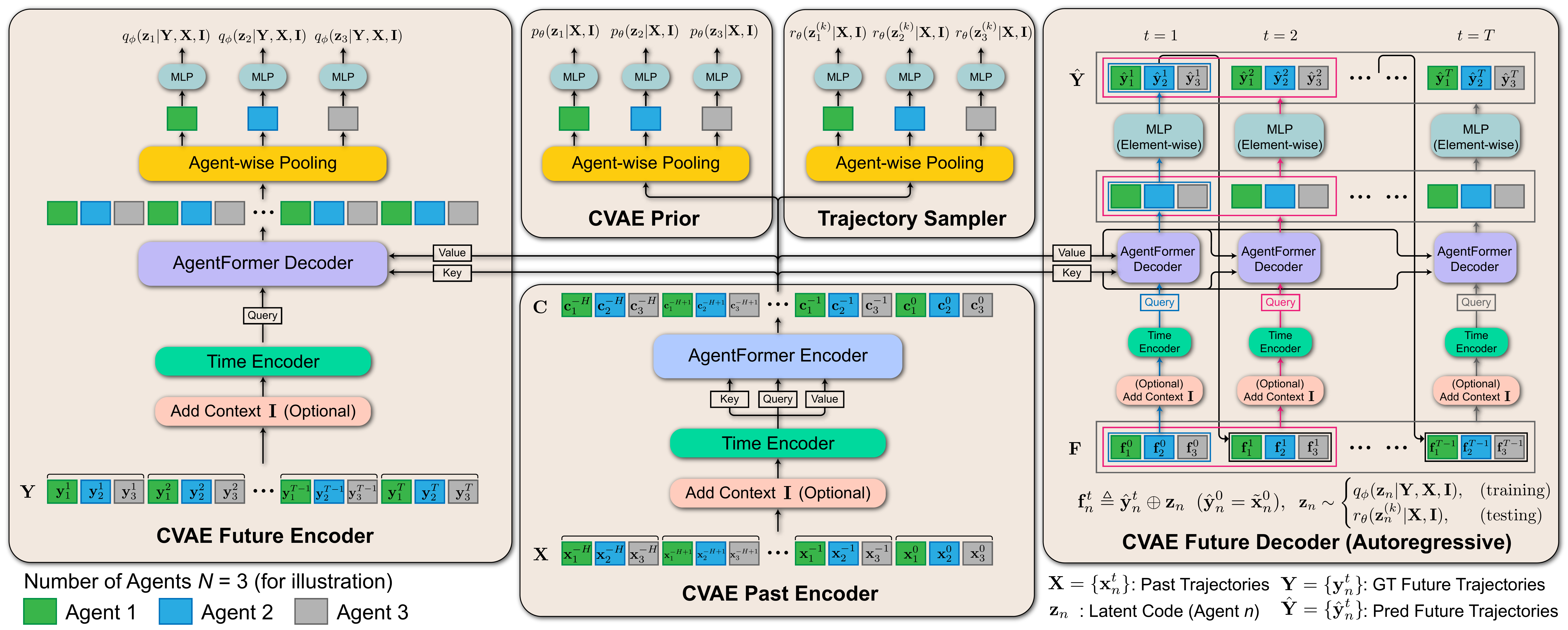}
    \caption{\textbf{Overview} of our AgentFormer-based multi-agent trajectory prediction framework.}
    \label{aformer:fig:overview}
\end{figure*}

\subsection{Multi-Agent Prediction with AgentFormer}
\label{aformer:sec:cvae}
Having introduced AgentFormer\ for modeling temporal and social relations, we are now ready to apply it in our multi-agent trajectory prediction framework based on CVAEs. As discussed at the start of Sec.~\ref{aformer:sec:approach}, the goal of multi-agent trajectory prediction is to model the future trajectory distribution $p_\theta(\mb{Y}|\mb{X},\mb{I})$ conditioned on past trajectories $\mb{X}$ and contextual information $\mb{I}$. To account for stochasticity and multi-modality in each agent's future behavior, we introduce latent variables $\mb{Z} = \{\mb{z}_1, \ldots, \mb{z}_N\}$ where $\mb{z}_n \in \mathbb{R}^{d_z}$ represents the latent intent of agent $n$. We can then rewrite the future trajectory distribution as
\begin{equation}
\label{aformer:eq:cvae}
p_\theta(\mb{Y}|\mb{X},\mb{I}) = \int p_\theta(\mb{Y}|\mb{Z},\mb{X},\mb{I}) p_\theta(\mb{Z}|\mb{X},\mb{I}) d\mb{Z}\,,
\end{equation}
where $p_\theta(\mb{Z}|\mb{X},\mb{I}) = \prod_{n=1}^N p_\theta(\mb{z}_n|\mb{X},\mb{I})$ is a conditional Gaussian prior factorized over agents and $p_\theta(\mb{Y}|\mb{Z},\mb{X},\mb{I})$ is a conditional likelihood model. To tackle the intractable integral in Eq.~\eqref{aformer:eq:cvae}, we use the negative evidence lower bound (ELBO) $\mathcal{L}_{elbo}$ in the CVAE as our loss function:
\begin{equation}
\label{aformer:eq:elbo}
\begin{aligned}
\mathcal{L}_{elbo} = & -\mathbb{E}_{q_\phi(\mb{Z}|\mb{Y},\mb{X},\mb{I})}[\log p_\theta(\mb{Y}|\mb{Z},\mb{X},\mb{I})] \\
& + \text{KL}(q_\phi(\mb{Z}|\mb{Y},\mb{X},\mb{I})\| p_\theta(\mb{Z}|\mb{X},\mb{I}))\,,
\end{aligned}
\end{equation}
where $q_\phi(\mb{Z}|\mb{Y},\mb{X},\mb{I}) = \prod_{n=1}^N q_\phi(\mb{z}_n|\mb{Y},\mb{X},\mb{I})$ is an approximate posterior distribution factorized over agents and parametrized by $\phi$. In our probabilistic formulation, the latent codes $\mb{Z}$ of all agents in the posterior $q_\phi(\mb{Z}|\mb{Y},\mb{X},\mb{I})$ are jointly inferred from the future trajectories $\mb{Y}$ of all agents; similarly, the future trajectories $\mb{Y}$ in the conditional likelihood $p_\theta(\mb{Y}|\mb{Z},\mb{X},\mb{I})$ are modeled using the latent codes~$\mb{Z}$ of all agents. This design allows each agent's latent intent represented by $\mb{z}_n$ to affect not just its own future trajectory but also the future trajectories of other agents, which enables us to generate socially-aware multi-agent trajectories. Having described the probabilistic formulation, we now introduce the detailed model architecture as outlined in Fig.~\ref{aformer:fig:overview}.

\paragraph{Encoding Context (Semantic Map).}
As aforementioned, our model can optionally take as input contextual information $\mb{I}$ if provided by the data. Here, we assume $\mb{I}\in \mathbb{R}^{H_0\times W_0\times C}$ is a semantic map around the agents at the current timestep ($t=0$) with annotated semantic information (\eg, sidewalks, crosswalks, and road boundaries). For each agent $n$, we rotate $\mb{I}$ to align with the agent's heading angle and crop an image patch $\mb{I}_n \in \mathbb{R}^{H\times W\times C}$ around the agent. We use a hand-designed convolutional neural network (CNN) to extract visual features $\mb{v}_n$ from $\mb{I}_n$, which will later be used by other modules in the model.

\paragraph{CVAE Past Encoder.}
The past encoder starts with the multi-agent past trajectory sequence $\mb{X}$. If the semantic map $\mb{I}$ is provided, the past encoder concatenates each element $\mb{x}_n^t \in \mb{X}$ with the corresponding visual feature $\mb{v}_n$ of agent $n$. The new sequence is then fed into the time encoder to obtain a timestamped sequence, which is then input to the AgentFormer\ encoder as keys, queries, and values. The output of the encoder is a past feature sequence $\mb{C} =  \left(\mb{c}_1^{-H}, \ldots \mb{c}_N^{-H}, \mb{c}_1^{-H+1}, \ldots \mb{c}_N^{-H+1}, \ldots, \mb{c}_1^{0}, \ldots,  \mb{c}_N^{0}\right)$ that summarizes the past agent trajectories $\mb{X}$ and context $\mb{I}$.

\paragraph{CVAE Prior.}
The prior module first performs an agent-wise pooling that computes a mean agent feature $\mb{C}_n$ from the past features across timesteps: $\mb{C}_n = \text{mean}(\mb{c}_n^{-H}, \ldots, \mb{c}_n^{0})$. We then use a multilayer perceptron (MLP) to map $\mb{C}_n$ to the Gaussian parameters $(\bs{\mu}_n^p, \bs{\sigma}_n^p)$ of the prior distribution $p_\theta(\mb{z}_n|\mb{X},\mb{I}) = \mathcal{N}(\bs{\mu}_n^p,\text{Diag}( \bs{\sigma}_n^p)^2)$.

\paragraph{CVAE Future Encoder.}
Given the multi-agent future trajectory sequence $\mb{Y}$, similar to the past encoder, the future encoder appends visual features from the semantic map $\mb{I}$ to $\mb{Y}$ and feeds the resulting sequence to the time encoder to produce a timestamped sequence. The timestamped sequence is then input as queries to the AgentFormer\ decoder along with the past feature sequence $\mb{C}$ which serves as both keys and values. We use the AgentFormer\ decoder here because it allows the feature extraction of $\mb{Y}$ to condition on $\mb{X}$ through $\mb{C}$, thus effectively modeling the $\mb{X}$-conditioning in the posterior $q_\phi(\mb{Z}|\mb{Y},\mb{X},\mb{I})$. We then perform an agent-wise mean pooling across timesteps on the output sequence of the AgentFormer\ decoder to extract a feature for each agent. Each agent feature is then input to an MLP to obtain the Gaussian parameters $(\bs{\mu}_n^q, \bs{\sigma}_n^q)$ of the approximate posterior distribution $q_\phi(\mb{z}_n|\mb{Y},\mb{X},\mb{I}) = \mathcal{N}(\bs{\mu}_n^q,\text{Diag}( \bs{\sigma}_n^q)^2)$.

\paragraph{CVAE Future Decoder.}
Unlike the original Transformer decoder, our future trajectory decoder is autoregressive, which means it outputs trajectories one step at a time and feeds the currently generated trajectories back into the model to produce the trajectories of the next timestep. This design mitigates compounding errors during test time at the expense of training speed. Starting from an initial sequence $(\hat{\mb{y}}_1^0,\ldots,\hat{\mb{y}}_N^0)$ where $\hat{\mb{y}}_n^0 = \tilde{\mb{x}}_n^0$ ($\tilde{\mb{x}}_n^0$ is the position feature inside $\mb{x}_n^0$), the future decoder module maps an input sequence $(\hat{\mb{y}}_1^0,\ldots,\hat{\mb{y}}_N^0, \ldots, \hat{\mb{y}}_1^{t'},\ldots,\hat{\mb{y}}_N^{t'})$ to an output sequence $(\hat{\mb{y}}_1^1,\ldots,\hat{\mb{y}}_N^1, \ldots, \hat{\mb{y}}_1^{t'+1},\ldots,\hat{\mb{y}}_N^{t'+1})$. By concatenating the last $N$ elements of the output, it grows the input sequence into $(\hat{\mb{y}}_1^0,\ldots,\hat{\mb{y}}_N^0, \ldots, \hat{\mb{y}}_1^{t'+1},\ldots,\hat{\mb{y}}_N^{t'+1})$. By autoregressively applying the decoder $T$ times, we obtain the output sequence $\hat{\mb{Y}} = (\hat{\mb{y}}_1^1,\ldots,\hat{\mb{y}}_N^1, \ldots, \hat{\mb{y}}_1^{T},\ldots,\hat{\mb{y}}_N^{T}) $. Inside the future decoder module (Fig.~\ref{aformer:fig:overview}\,(Right)), we first form a feature sequence $\mb{F} = (\mb{f}_1^0,\ldots,\mb{f}_N^0, \ldots,\mb{f}_1^{t'},\ldots,\mb{f}_N^{t'})$ where $\mb{f}_n^{t} = \hat{\mb{y}}_n^{t} \oplus \mb{z}_n$, thus concatenating the currently generated trajectories with the corresponding latent codes. The latent codes are sampled from the approximate posterior during training but from the trajectory sampler (as discussed below) at test time. The feature sequence $\mb{F}$ is then concatenated with the semantic map features and timestamped before being input as queries to the AgentFormer\ decoder alongside the past feature sequence $\mb{C}$ which serves as keys and values. The AgentFormer\ decoder enables the future trajectories to directly attend to features of any agent at any previous timestep (\eg, $\mb{c}_3^{-H}$ or $\hat{\mb{y}}_2^{1}$), allowing the model to effectively infer future trajectories based on the whole agent history. We use proper masking inside the AgentFormer\ decoder to enforce causality of the decoder output sequence. Each element of the output sequence is then passed through an MLP to generate the decoded future agent position $\hat{\mb{y}}_n^{t}$. As we use a Gaussian to model the conditional likelihood $p_\theta(\mb{Y}|\mb{Z},\mb{X},\mb{I}) = \mathcal{N}(\hat{\mb{Y}},I/\beta)$, where $I$ is the identity matrix and $\beta$ is a weighting factor, the first term in Eq.~\eqref{aformer:eq:elbo} equals the mean squred error (MSE): $\mathcal{L}_{mse} = \frac{\beta}{2}\|\mb{Y} - \hat{\mb{Y}}\|^2$.

\paragraph{Trajectory Sampler.}
We adapt a diversity sampling technique, DLow~\cite{yuan2020dlow}, to our multi-agent trajectory prediction setting and employ a trajectory sampler to produce diverse and plausible trajectories once our CVAE model is trained. The trajectory sampler generates $K$ sets of latent codes $\{\mb{Z}^{(1)}, \ldots, \mb{Z}^{(K)}\}$ where each set $\mb{Z}^{(k)} = \{\mb{z}_1^{(k)}, \ldots, \mb{z}_N^{(k)}\}$ contains the latent codes of all agents and can be decoded by the CVAE decoder into a multi-agent future trajectory sample $\hat{\mb{Y}}^{(k)}$. Each latent code $\mb{z}_n^{(k)} \in \mb{Z}^{(k)}$ is generated by a linear transformation of a Gaussian noise $\bs{\epsilon}_n \in \mathbb{R}^{d_z}$:
\begin{equation}
\label{aformer:eq:samp}
\mb{z}_n^{(k)} = \mb{A}_n^{(k)}\bs{\epsilon}_n + \mb{b}_n^{(k)}, \quad \bs{\epsilon}_n \sim \mathcal{N}(\mb{0}, I),
\end{equation}
where $\mb{A}_n^{(k)} \in \mathbb{R}^{d_z \times d_z}$ is a non-singular matrix and $\mb{b}_n^{(k)} \in \mathbb{R}^{d_z}$ is a vector. Eq.~\eqref{aformer:eq:samp} induces a Gaussian sampling distribution $r_\theta(\mb{z}_n^{(k)}|\mb{X}, \mb{I})$ over $\mb{z}_n^{(k)}$. The distribution is conditioned on $\mb{X}$ and $\mb{I}$ because its inner parameters $\{\mb{A}_n^{(k)},\mb{b}_n^{(k)}\}$ are generated by the trajectory sampler module (Fig.~\ref{aformer:fig:overview}) through agent-wise pooling of the past feature sequence $\mb{C}$ and an MLP. The trajectory sampler loss is defined as
\begin{equation}
\begin{aligned}
\label{aformer:eq:samp_loss}
&\mathcal{L}_{samp} = \min_k \|\hat{\mb{Y}}^{(k)} - \mb{Y}\|^2\\
& + \sum_{n=1}^N\text{KL}(r_\theta(\mb{z}_n^{(k)}|\mb{X}, \mb{I})\| p_\theta(\mb{z}_n|\mb{X}, \mb{I})) \\
& + \frac{1}{K(K-1)} \sum_{k_1=1}^{K} \sum_{k_1 \neq k_2}^{K} \exp \left(-\frac{\|\hat{\mb{Y}}^{(k_1)} - \hat{\mb{Y}}^{(k_2)}\|^2}{\sigma_{d}}\right),
\end{aligned}
\end{equation}
where $\sigma_d$ is a scaling factor. The first term encourages the future trajectory samples $\hat{\mb{Y}}^{(k)}$ to cover the ground truth $\mb{Y}$. The second KL term encourages each latent code $\mb{z}_n^{(k)}$ to follow the prior and be plausible; the KL can be computed analytically as both distributions inside are Gaussians. The third term encourages diversity among the future trajectory samples $\hat{\mb{Y}}^{(k)}$ by penalizing small pairwise distance. When training the trajectory sampler with Eq.~\eqref{aformer:eq:samp_loss}, we freeze the weights of the CVAE modules. At test time, we sample latent codes $\{\mb{Z}^{(1)}, \ldots, \mb{Z}^{(K)}\}$ using the trajectory sampler instead of sampling from the CVAE prior and decode the latent codes into trajectory samples $\{\hat{\mb{Y}}^{(1)}, \ldots, \hat{\mb{Y}}^{(K)}\}$.

\section{Experiments}
\label{aformer:sec:exp}

\paragraph{Datasets.}
We evaluate our method on well-established public datasets: the ETH~\cite{pellegrini2009you}, UCY~\cite{lerner2007crowds}, and nuScenes~\cite{caesar2020nuscenes} datasets. The ETH/UCY datasets are the major benchmark for pedestrian trajectory prediction. There are five datasets in ETH/UCY, each of which contains pedestrian trajectories captured at 2.5Hz in multi-agent social scenarios with rich interaction. nuScenes is a recent large-scale autonomous driving dataset, which consists of 1000 driving scenes with each scene annotated at 2Hz. nuScenes also provides HD semantic maps with 11 semantic classes.

\paragraph{Metrics.}
We report the minimum average displacement error $\text{ADE}_K$ and final displacement error $\text{FDE}_K$ of $K$ trajectory samples of each agent compared to the ground truth: $\text{ADE}_K = \frac{1}{T}\min_{k=1}^{K}\sum_{t=1}^T\|\hat{\mb{y}}_n^{t,(k)} - \mb{y}_n^{t}\|^2, \quad \text{FDE}_K = \min_{k=1}^{K}\|\hat{\mb{y}}_n^{T,(k)} - \mb{y}_n^{T}\|^2$, where  $\hat{\mb{y}}_n^{t,(k)}$ denotes the future position of agent $n$ at time $t$ in the $k$-th sample and $\mb{y}_n^{T}$ is the corresponding ground truth. $\text{ADE}_K$ and $\text{FDE}_K$ are the standard metrics for trajectory prediction~\cite{gupta2018social,sadeghian2019sophie,salzmann2020trajectron++,phan2020covernet,chai2020multipath}.

\paragraph{Evaluation Protocol.}
For the ETH/UCY datasets, we adopt a leave-one-out strategy for evaluation, following prior work~\cite{gupta2018social,sadeghian2019sophie,salzmann2020trajectron++,mangalam2020not,yu2020spatio}. We forecast 2D future trajectories of 12 timesteps (4.8s) based on observed trajectories of 8 timesteps (3.2s). Similar to most prior works, we do not use any semantic/visual information for ETH/UCY for fair comparisons. All metrics are computed with $K=20$ samples. For the nuScenes dataset, following prior work~\cite{phan2020covernet,chai2020multipath,cui2019multimodal,ma2020diverse}, we use the vehicle-only train-val-test split provided by the nuScenes prediction challenge and predict 2D future trajectories of 12 timesteps (6s) based on observed trajectories of 4 timesteps (2s). We report results with metrics computed using $K=1, 5 \text{ and } 10$ samples.

\paragraph{Implementation Details.}
For all datasets, we represent trajectories in a scene-centered coordinate where the origin is the mean position of all agents at $t=0$. The future decoder in Fig.~\ref{aformer:fig:overview} outputs the offset to the agent's current position $\tilde{\mb{x}}_n^0$, so $\tilde{\mb{x}}_n^0$ is added to obtain $\hat{\mb{y}}_n^t$ for each element in the output sequence. Following prior work~\cite{salzmann2020trajectron++,yu2020spatio}, random rotation of the scene is adopted for data augment.
Our multi-agent prediction model (Fig.~\ref{aformer:fig:overview}) uses two stacks (defined in \cite{vaswani2017attention}) of identical layers in each AgentFormer\ encoder/decoder with 0.1 dropout rate. The dimensions $d_k,d_v,d_\tau$ of keys, queries, and timestamps in AgentFormer\ are all set to 256, and the hidden dimension of feedforward layers is 512. The number of heads for multi-head agent-aware attention is 8. All MLPs in the model have hidden dimensions (512, 256). For the CVAE, the latent code dimension $d_z$ is 32, the coefficient $\beta$ of the MSE loss equals 1, and we clip the maximum value of the KL term in $L_{elbo}$ (Eq.~\eqref{aformer:eq:elbo}) down to 2. We also use the variety loss in SGAN~\cite{gupta2018social} in addition to $L_{elbo}$. The agent connectivity threshold $\eta$ is set to 100. We train the CVAE model using the Adam optimizer~\cite{kingma2014adam} for 100 epochs on ETH/UCY and nuScenes. We use an initial learning rate of $10^{-4}$ and halve the learning rate every 10 epochs.

\setlength{\tabcolsep}{3pt}
\begin{table}[t]
\footnotesize
\centering
\begin{tabular}{@{\hskip 1mm}l@{\hskip 1mm}|ccccc|@{\hskip 1mm}c@{\hskip 1mm}}
\toprule
\multicolumn{1}{c|}{\multirow{3}{*}[2pt]{Method}} & \multicolumn{6}{c}{$\text{ADE}_{20}/\text{FDE}_{20} \downarrow$ (m), $K=20$ Samples} \\
\cmidrule(l{0.8mm}r{0.5mm}){2-7}
 & ETH & Hotel & Univ &  Zara1 & Zara2 & Average\\ \midrule%
SGAN~\cite{gupta2018social} & 0.81/1.52 & 0.72/1.61 & 0.60/1.26 & 0.34/0.69 & 0.42/0.84 & 0.58/1.18 \\
SoPhie~\cite{sadeghian2019sophie} & 0.70/1.43 & 0.76/1.67 & 0.54/1.24 & 0.30/0.63 & 0.38/0.78 & 0.54/1.15 \\
Transformer-TF~\cite{giuliari2020transformer} & 0.61/1.12 & 0.18/0.30 & 0.35/0.65 & 0.22/0.38 & 0.17/0.32 & 0.31/0.55 \\
STAR~\cite{yu2020spatio} & 0.36/0.65 & 0.17/0.36 & 0.31/0.62 & 0.26/0.55 & 0.22/0.46 & 0.26/0.53 \\
PECNet~\cite{mangalam2020not} & 0.54/0.87 & 0.18/0.24 & 0.35/0.60 & 0.22/0.39 & 0.17/0.30 & 0.29/0.48 \\
Trajectron++~\cite{salzmann2020trajectron++} & \textbf{0.39}/0.83 & \textbf{0.12}/\textbf{0.21} & \textbf{0.20}/\textbf{0.44} & \textbf{0.15}/0.33 & \textbf{0.11}/0.25 & \textbf{0.19}/0.41 \\
Ours (AgentFormer) & 0.45/\textbf{0.75} & 0.14/0.22 & 0.25/0.45 & 0.18/\textbf{0.30} & 0.14/\textbf{0.24} & 0.23/\textbf{0.39} \\
\bottomrule
\end{tabular}
\vspace{5mm}
\caption{\textbf{Baseline comparisons} on the ETH/UCY datasets.}
\label{aformer:table:eth}
\end{table}
\setlength{\tabcolsep}{4pt}
\begin{table}[t]
\footnotesize
\centering
\begin{tabular}{@{\hskip 1mm}l@{\hskip 1mm}|cccc}
\toprule
\multicolumn{1}{c|}{\multirow{3}{*}[2pt]{Method}} & \multicolumn{2}{c}{$K=5$ Samples} & \multicolumn{2}{c}{$K=10$ Samples} \\
\cmidrule(l{0.8mm}r{0.8mm}){2-3}
\cmidrule(l{0.8mm}r{0.8mm}){4-5}
 & $\text{ADE}_5\downarrow$ & $\text{FDE}_5\downarrow$ & $\text{ADE}_{10}\downarrow$ & $\text{FDE}_{10}\downarrow$ \\ \midrule
MTP~\cite{cui2019multimodal} & 2.93 & - & 2.93 & - \\
MultiPath~\cite{chai2020multipath} & 2.32 & - & 1.96 & - \\
CoverNet~\cite{phan2020covernet} & 1.96 & - & 1.48 & - \\
DSF-AF~\cite{ma2020diverse} & 2.06 & 4.67 & 1.66 & 3.71 \\
DLow-AF~\cite{yuan2020dlow} & 2.11 & 4.70 & 1.78 & 3.58 \\
Trajectron++~\cite{salzmann2020trajectron++} & 1.88 & - & 1.51 & - \\
Ours (AgentFormer) & \textbf{1.86} & \textbf{3.89} & \textbf{1.45} & \textbf{2.86}\\
\bottomrule
\end{tabular}
\vspace{5mm}
\caption{\textbf{Baseline comparisons} on the nuScenes dataset.}
\label{aformer:table:nuscene}
\end{table}

\subsection{Results}
\paragraph{Baseline Comparisons.}
On the ETH/UCY datasets, we compare our approach with current state-of-the-art methods -- Trajectron++~\cite{salzmann2020trajectron++}, PECNet~\cite{mangalam2020not}, STAR~\cite{yu2020spatio}, and Transformer-TF~\cite{giuliari2020transformer} -- as well as common baselines -- SGAN~\cite{gupta2018social} and Sophie~\cite{sadeghian2019sophie}. The performance of all methods is summarized in Table~\ref{aformer:table:eth}, where we use officially-reported results for the baselines. We can observe that our AgentFormer\ achieves very competitive performance and attains the best FDE. Particularly, our approach significantly outperforms prior Transformer-based methods, Transformer-TF~\cite{giuliari2020transformer} and STAR~\cite{yu2020spatio}. As FDE measures the final displacement error of predicted trajectories, it places more emphasis on a method's ability to predict distant futures than ADE. We believe the strong performance of our method in FDE can be attributed to the design of AgentFormer, which can model long-range trajectory dependencies effectively by directly attending to features of any agent at any previous timestep when inferring an agent's future position.

Compared to ETH/UCY, the trajectories in nuScenes are much longer as we evaluate with a longer time horizon (6s) and vehicles are much faster than pedestrians. Thus, nuScenes presents a different challenge for multi-agent prediction methods. On the nuScenes dataset, we evaluate our approach against state-of-the-art vehicle prediction methods -- Trajectron++~\cite{salzmann2020trajectron++}, MTP~\cite{cui2019multimodal}, MultiPath~\cite{chai2020multipath}, CoverNet~\cite{phan2020covernet}, DSF-AF~\cite{ma2020diverse}, and DLow-AF~\cite{yuan2020dlow}. We report the performance of all methods in Table~\ref{aformer:table:nuscene}, where the results of Trajectron++ are taken from the nuScenes prediction challenge leaderboard, the performance of DLow-AF is from ~\cite{ma2020diverse}, and we also use the officially-reported results for the other baselines. The FDE of some baselines is not available since the number has not been reported. We can see that our approach, AgentFormer, outperforms the baselines, especially the strong model Trajectron++~\cite{salzmann2020trajectron++}, consistently in ADE and FDE for both 5 and 10 sample settings.

\setlength{\tabcolsep}{3pt}
\begin{table}[t]
\footnotesize
\centering
\begin{tabular}{@{\hskip 1mm}cc|ccccc|@{\hskip 1mm}c@{\hskip 1mm}}
\toprule
\multicolumn{2}{c|}{Model} & \multicolumn{6}{c}{$\text{ADE}_{20}/\text{FDE}_{20} \downarrow$ (m), $K=20$ Samples} \\
\cmidrule(l{0mm}r{0.5mm}){1-2}\cmidrule(l{0.5mm}r{1mm}){3-8}
 \hspace{1.5mm} Social \hspace{1.5mm} & Temporal & ETH & Hotel & Univ &  Zara1 & Zara2 & Average\\ \midrule%
GCN & LSTM & 0.57/0.90 & 0.20/0.34 & 0.29/0.52 & 0.24/0.44 & 0.23/0.42 & 0.31/0.52 \\
GCN & TF & 0.56/0.93 & 0.15/0.28 & 0.28/0.51 & 0.24/0.45 & 0.19/0.35 & 0.28/0.50 \\
TF & LSTM & 0.55/0.91 & 0.18/0.31 & 0.28/0.50 & 0.24/0.44 & 0.21/0.39 & 0.29/0.51 \\
TF & TF & 0.50/0.82 & 0.15/0.27 & 0.28/0.52 & 0.22/0.42 & 0.16/0.31 & 0.26/0.47 \\
\midrule
\multicolumn{2}{c|}{Joint Socio-Temporal} & ETH & Hotel & Univ &  Zara1 & Zara2 & Average\\ \midrule%
\multicolumn{2}{l|}{Ours w/o joint latent}  & 0.49/0.77 & 0.15/0.25 & 0.29/0.52 & 0.22/0.41 & 0.18/0.33 & 0.27/0.46 \\
\multicolumn{2}{l|}{Ours w/o AA attention} & 0.49/0.80 & 0.15/0.25 & 0.31/0.54 & 0.23/0.41 & 0.19/0.34 & 0.27/0.47 \\
\multicolumn{2}{l|}{Ours w/ agent encoding} & 0.48/0.78 & \textbf{0.14}/0.23 & 0.32/0.55 & 0.22/0.40 & 0.19/0.34 & 0.27/0.46 \\
\multicolumn{2}{l|}{Ours (AgentFormer)} & \textbf{0.45}/\textbf{0.75} & \textbf{0.14}/\textbf{0.22} & \textbf{0.25}/\textbf{0.45} & \textbf{0.18}/\textbf{0.30} & \textbf{0.14}/\textbf{0.24} & \textbf{0.23}/\textbf{0.39} \\
\bottomrule
\end{tabular}
\vspace{5mm}
\caption{\textbf{Ablation studies} on the ETH/UCY datasets. ``TF'' means Transformer and ``AA Attention'' denotes agent-aware attention.}
\label{aformer:table:eth_abl}
\end{table}
\setlength{\tabcolsep}{4pt}
\begin{table}[t]
\footnotesize
\centering
\begin{tabular}{@{\hskip 1mm}cc|cccc}
\toprule
\multicolumn{2}{c|}{Model} & \multicolumn{2}{c}{$K=5$ Samples} & \multicolumn{2}{c}{$K=10$ Samples}\\
\cmidrule(l{0mm}r{0.5mm}){1-2}
\cmidrule(l{0.5mm}r{0.8mm}){3-4}
\cmidrule(l{0.8mm}r{0.8mm}){5-6}
 \hspace{1.5mm} Social \hspace{1.5mm} & Temporal & $\text{ADE}_5\downarrow$ & $\text{FDE}_5\downarrow$ & $\text{ADE}_{10}\downarrow$ & $\text{FDE}_{10}\downarrow$\\ \midrule%
GCN & LSTM & 2.17 & 4.42 & 1.57 & 3.09 \\
GCN & TF & 2.03 & 4.36 & 1.52 & 2.95 \\
TF & LSTM & 2.12 & 4.48 & 1.69 & 3.31 \\
TF & TF & 1.99 & 4.12 & 1.54 & 3.07 \\
\midrule
\multicolumn{2}{c|}{Joint Socio-Temporal} &  $\text{ADE}_5\downarrow$ & $\text{FDE}_5\downarrow$ & $\text{ADE}_{10}\downarrow$ & $\text{FDE}_{10}\downarrow$\\ \midrule%
\multicolumn{2}{l|}{Ours w/o semantic map}  & 1.97 & 4.21 & 1.58 & 3.14 \\
\multicolumn{2}{l|}{Ours w/o joint latent}  & 1.95 & 3.98 & 1.50 & 2.92 \\
\multicolumn{2}{l|}{Ours w/o AA attention} & 2.02 & 4.29 & 1.55 & 2.91 \\
\multicolumn{2}{l|}{Ours w/ agent encoding} & 2.01 & 4.28 & 1.63 & 3.11 \\
\multicolumn{2}{l|}{Ours (AgentFormer)} & \textbf{1.86} & \textbf{3.89} & \textbf{1.45} & \textbf{2.86}\\
\bottomrule
\end{tabular}
\vspace{5mm}
\caption{\textbf{Ablation studies} on the nuScenes dataset. ``TF'' means Transformer and ``AA Attention'' denotes agent-aware attention.}
\label{aformer:table:nuscenes_abl}
\end{table}

\paragraph{Ablation Studies.}
We further perform extensive ablation studies on ETH/UCY and nuScenes to investigate the contribution of key technical components in our method. The first ablation study explores variants of our method that use separate social and temporal models to replace our joint socio-temporal model, AgentFormer, in our multi-agent prediction framework. We choose GCN~\cite{kipf2016semi} or Transformer (TF) as the social model, and LSTM or Transformer as the temporal model. In total, there are 4 ($2\times2$) combinations of social and temporal models. The ablation results are summarized in the first group of Table~\ref{aformer:table:eth_abl} and~\ref{aformer:table:nuscenes_abl}. It is evident that all combinations of separate social and temporal models lead to inferior performance compared to our method which models the social and temporal dimensions jointly. 

The second ablation study investigates the role of (1) joint latent intent modeling, (2) agent-aware attention, and (3) semantic maps, and we denote the corresponding variants as ``w/o joint latent'', ``w/o AA attention'', and ``w/o semantic map''. We further test a variant ``w/ agent encoding'' where we replace agent-aware attention with agent encoding. The results are reported in the second group of Table~\ref{aformer:table:eth_abl} and~\ref{aformer:table:nuscenes_abl}. We can see that all variants lead to considerably worse performance compared to our full method. In particular, the variants ``w/o AA attention'' and ``w/ agent encoding'' result in pronounced performance drop, which indicates that agent-aware attention is essential in our method and alternatives like agent encoding are not effective.

\begin{figure}
    \centering
    \includegraphics[width=\linewidth]{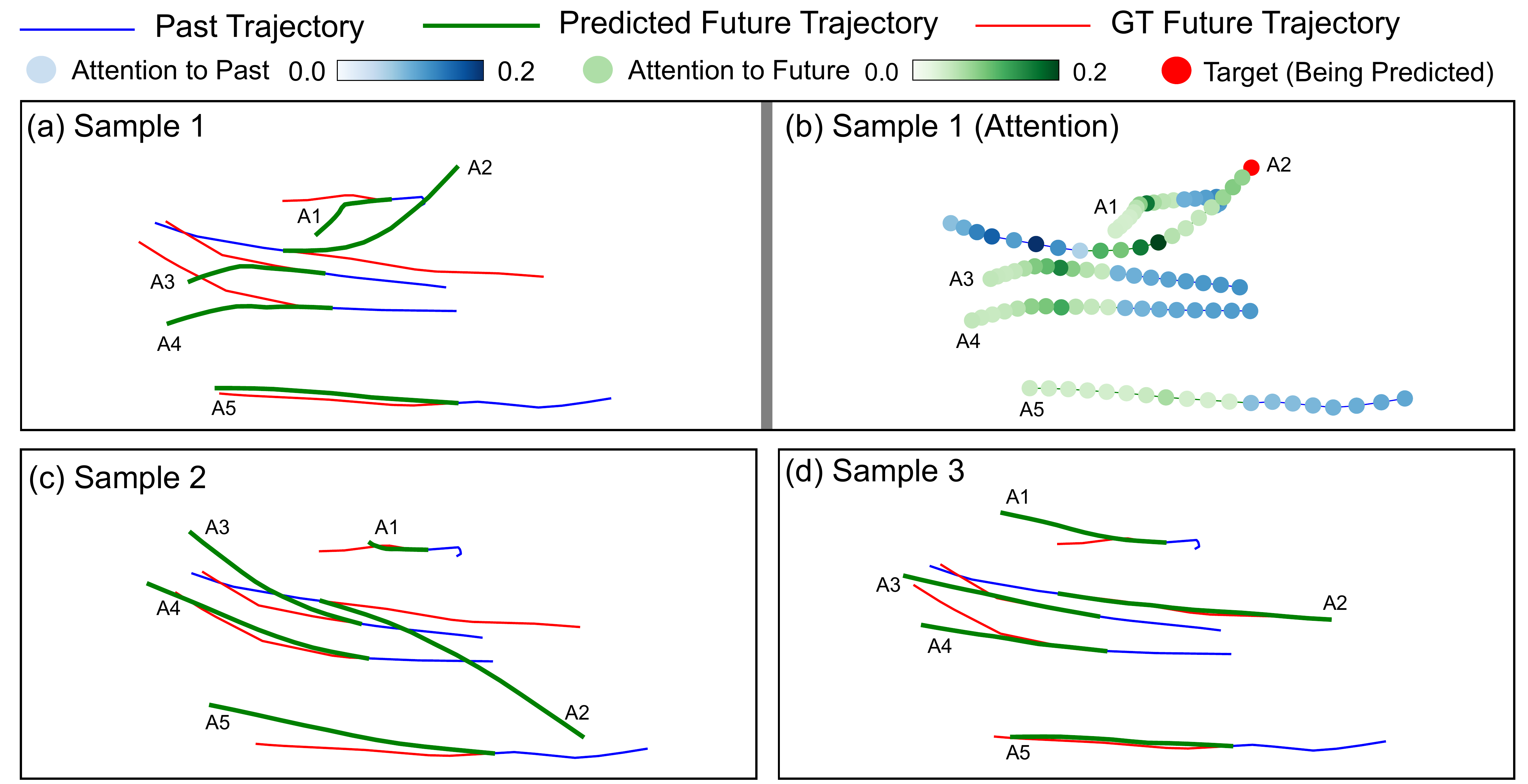}
    \caption{\textbf{(a,c,d)} Three samples of forecasted multi-agent futures (green) via our method, which exhibit social behaviors like following (A3 \& A4) and collision avoidance (A1 \& A2 in (a), A2 \& A3 in (c)). \textbf{(b)} Attention visualization for sample 1. When predicting the target (red), the model pays more attention (darker color) to key timesteps (turning point) of adjacent agents and spreads out attention to the target's past timesteps to reason about dynamics.}
    \label{aformer:fig:vis}
\end{figure}

\paragraph{Trajectory Visualization.}
Fig.~\ref{aformer:fig:vis}\,(a,c,d) shows three samples of forecasted multi-agent futures of the same scene via our method. We can see that the samples correspond to different modes of socially-aware and non-colliding trajectories, and exhibit behaviors like following (A3 \& A4) and collision avoidance (A1 \& A2 in (a), A2 \& A3 in (c)). Fig.~\ref{aformer:fig:vis}\,(b) visualizes the attention of sample 1 and shows that, when predicting the target (red), the model pays more attention to key timesteps (turning point) of adjacent agents and also spreads out attention to the target's past timesteps to reason about the dynamics and curvature of its trajectory.

\section{Conclusion}
In this paper, we proposed a new Transformer, AgentFormer, that can simultaneously model the time and social dimensions of multi-agent trajectories using a sequence representation. To preserve agent identities in the sequence, we proposed a novel agent-aware attention mechanism that can attend to features of the same agent differently than features of other agents. Based on AgentFormer, we presented a stochastic multi-agent trajectory prediction framework that jointly models the latent intent of all agents to produce diverse and socially-aware multi-agent future trajectories. Experiments demonstrated that our method substantially improved state-of-the-art performance on challenging pedestrian and autonomous driving datasets.

\part{Conclusion and Future Work}
\label{part:conclusion}
\chapter{Conclusion and Future Work}
\label{chap:conclusion}

In this chapter, we first conclude the contributions and discuss open problems and future research directions in each of the three aspects --- simulation, perception, and generation --- of human behavior modeling. We will then discuss the main lessons learned from our work towards unifying the three aspects and share our vision on what is next for human behavior modeling.

\section{Simulation of Human Behavior}
\label{sec:conclusion:simulation}

In this thesis, we made two main contributions to simulation of human behavior: (1) we proposed a robust approach, called residual force control (RFC), for simulating human behavior in physics simulation, which is crucial for downstream perception and generation tasks; (2) we proposed a new approach for automatic and efficient design of simulated agents, which can create more performant simulated agents (e.g., humanoids) than those designed by experts. However, there is still much work to be done in simulation of human behavior. Below we discuss two important open problems and future research directions.

\subsection{Efficient Learning of Robust Humanoid Control Policy}
We achieved robust humanoid control in this thesis mainly through the RFC approach discussed in Ch.~\ref{chap:rfc}. However, the external residual forces employed in RFC can also comprise the physical accuracy of the simulated human behavior, since the humanoid can perform super-human actions with the help of large residual forces. This is fine as a temporary solution to unblock application of simulation to downstream perception and generation tasks, but ideally we want to learn a humanoid control policy that is robust even without RFC.

A main hurdle in learning a robust policy is the sample-inefficient RL algorithms, which may not be able to find the optimal control policy under a standard computing budget. While we can hope that RL research will eventually come up with more efficient algorithms, another direction we can look into is to use differentiable physics simulator, which provides us with analytic gradients of the physics simulator. This opens up many opportunties for sample-efficient learning of humanoid control policies. For example, if the reference motion data is given, we can use supervised learning to learn a imitative control policy. Instead, if only some objective function is given, gradient-based trajectory optimization can also be used to efficiently synthesize simulated humanoid behavior or learn a control policy. Recently, many differentiable simulators start to emerge such as Brax~\cite{freeman2021brax} and Nimble~\cite{werling2021fast}. It would be interesting to implement the simulation-based behavior modeling framework in these differentiable simulators and train the policy with supervised learning. A caveat is that it is still not clear how reliable or stable the gradients of these simulators will be, especially when the simulated environment is highly dynamic and discontinuous in nature. Along this line, another way to use supervised learning is to learn a neural simulator to mimic the physics simulation, which is similar to model-based RL. Recent work on control-based character animation already showed that this is a viable approach~\cite{fussell2021supertrack}, but it remains unclear how the neural simulator generalizes to more complicated environments with terrains and objects.

\subsection{Better Modeling of Humans and Environments in Simulation}
Another possible cause for the difficulty in robust humanoid control could be the over-simplified modeling of humans in the physics simulator, which causes the dynamics mismatch between the humanoid and real humans as discussed in Ch.~\ref{chap:rfc}. Although in Ch.~\ref{chap:transform2act}, we proposed a method, Transform2Act, for automatic design of simulated agents, it only works within the confines of the design space provided by the physics simulator. For instance, it can adjust the motor strength of each joint or the shape and size of the feet, but it cannot make the feet soft if the physics simulator does not support soft-tissue simulation. Although there are works that use more advanced models to simulate humans, they are still far from physically-accurate and are often too computationally-inefficient to be used by RL. Therefore, an important open problem is how to model humans better and more efficiently in physics simulation.

Another underexplored aspect is the modeling of the environment in physics simulation, which is especially important for simulation-based perception and generation tasks. Our recent work~\cite{luo2021dynamics} tried to address this aspect by reconstructing human-object interaction in physics simulation. However, our method can only handle a few types of objects with primitive geometries. To accurately model human behavior in simulation, we need to simulate all types of objects in environments and their interaction with humans. For example, to simulate a human sitting on a sofa, we need to simulate the deformable materials of the sofa to be consistent with the real world. With potentially hundreds of objects in the environment, simulating them efficiently is also of crucial importance to enable learning algorithms.

\section{Perception of Human Behavior}
\label{sec:conclusion:perception}

In this thesis, we made the following contributions to perception of human behavior: (1) we tackled the highly under-constrained problem of first-person human pose estimation via the use of our simulation-based framework; (2) we improved the physical plausibility and pose accuracy of third-person human pose estimation using our simulation-based framework; (3) we leveraged behavior generation to tackle a new perception task, i.e., global occlusion-aware human pose estimation with dynamic cameras. Yet, there are still many remaining problems and challenges to be addressed by future research. Below we discuss two of the open problems.

\subsection{Modeling Humans as Embodied Agents Interacting with Environments}
\label{sec:conclusion:embody}
Human pose estimation has come a long way, progressing from 2D and 3D pose estimation to temporal and scene-aware pose estimation. However, much of the research still takes a third-person approach to human behavior modeling, where human behaviors are observed through a third-person camera and the focus is on modeling the geometric relationships of 2D and 3D keypoints or joint angles of the human. A drawback of this approach is that it lacks a fundamental understanding of how human behaviors are generated in the 3D environment, and it is sensitive to camera angles and occlusions. Instead, an embodied agent approach, i.e., modeling the human as an embodied agent interacting with the environment, could be better for generalization. For example, if a person sees a chair and adjusts their pose to sit down, the behavior is governed by the person's egocentric perception of the chair and the 3D spatial relationships between the person and the chair, and the behavior is invariant to any third-person cameras observing the behavior. In Ch.~\ref{chap:egopose18} and \ref{chap:egopose19} as well as our recent work~\cite{luo2021dynamics}, we have taken this embodied agent approach to tackle first-person human pose estimation and showed that it can improve generalization. However, there are still many open problems such as how to effectively incorporate 3D environment information into the state of the agent and how to simulate accurate human-object interaction in a physics simulator.

\subsection{Data Collection of 3D Human Behavior and Environment}
Data collection poses a major challenge to research on perception of human behavior. For first-person human pose estimation (Ch.~\ref{chap:egopose18} and \ref{chap:egopose19}), we had to use motion capture studios to collect paired data of first-person videos and 3D human poses. There are several drawbacks of using motion capture studios. First, the space is typically limited, which constrains the types of human behavior that can be captured. Second, the visual data collected in motion capture studios typically lacks diversity in appearance, which creates domain gaps between motion capture data and real-world data. Third, it is difficult to reproduce real-world environments in the studio, such as large furniture, stairs, and outdoor scenes.

To capture more flexible human poses, recent work started to use additional sensors such as IMUs~\cite{hps2021Vladmir} or RGB-D cameras~\cite{hassan2019resolving} to estimate human pose in the wild. These approaches also use SLAM to capture the environment to provide scene context for the pose data. A promising future direction along this line is to combine our simulation-based behavior modeling framework with these approaches to further improve the physical plausibility and fidelity of human and scene reconstruction.

While 3D human behavior data may be limited, we have abundant 2D human behavior data such as internet videos. It would be very useful to develop a weakly-supervised approach that leverages 2D human behavior data and physics simulation to reconstruct high-quality 3D human behavior. For example, we can learn to control the humanoid in physics simulation via RL to match the 2D keypoints from internet videos. An important challenge in this aspect is how to also estimate the ground plane or terrain in order to simulate human behavior for such in-the-wild videos.

Another important aspect is to capture multi-person interaction, which is often quite challenging due to occlusions and dynamic motions. Kinematic 3D pose fitting methods such as SPIN~\cite{kolotouros2019spinmini} can produce interpenetrating human poses due to the lack of physical constraints and fundamental understanding of how humans interact with each other. As discussed in Sec.~\ref{sec:conclusion:embody}, modeling multiple people as embodied agents in physics simulation could help address this problem and largely improve the realism of multi-person interaction.

\section{Generation of Human Behavior}
\label{sec:conclusion:generation}

In this thesis, we made the following contributions to simulation of human behavior: (1) we proposed a simulation-based human behavior generation approach that can forecast the future human motion from a first-person video; (2) we tackled stochastic human behavior generation and improved the sample diversity of deep generative models with determinantal point processes (DPPs) and latent normalizing flows; (3) extending from the single-agent setting, we further studied stochastic multi-agent trajectory generation and proposed a new agent-aware Transformer model that achieved state-of-the-art performance. Similar to simulation and perception, there are many remaining open problems and challenges in generation of human behavior. Below we discuss two important problems.

\subsection{Generating Out-of-Distribution Human Behavior}
The current paradigm for generating human behavior is mainly based on deep generative models, which learn to mimic the training data distribution via supervised learning. However, this also limits the generalization of the learned behavior generation model, since they are designed to only generate in-distribution human behavior and cannot generate out-of-distribution (OOD) behavior. For example, if we try to learn a generation model of a person sitting in a chair and the training data only contains the person starting in front of the chair, the learned model will have a hard time generating the sitting behavior if the person starts behind the chair. To successfully generate all types of sitting motion, the training data has to largely cover the approaching angles of the person to prevent extrapolation. Yet, real-world data does not often have sufficient coverage of all possible scenarios.

A possible solution to this problem is to use physics simulation and RL, which allows us to learn control policies for all kinds of basic human locomotions. These locomotion policies form the building blocks for physically-plausbile human behaviors, and they are robust to perturbations due to RL. To generate OOD behavior, such as the sitting example discussed previously, we can just define a reward function based on goal states (e.g., sitting in the chair) and learn to compose the locomotion policies to maximize rewards. The use of physics simulation allows physically-plausible OOD behaviors to emerge instead of any random behavior that can maximize rewards.

\subsection{Evaluation of Generated Human Behavior}
Another key problem is the effective evaluation of the generated human behaviors. Generation tasks such as human trajectory forecasting can often be formulated as learning a conditional distribution. However, for each condition (e.g., past human trajectory), there is often only one GT (e.g., future human trajectory) in the dataset, so it is difficult to approximate the GT conditional distribution. Many SOTA methods on trajectory forecasting are only evaluated by comparing the best-of-$N$ sample with the GT, which does not penalize implausible samples and can be overoptimistic. In Ch.~\ref{chap:dsf}, we proposed new metrics to alleviate the problem but the problem still exists. 

One potentially better way to evaluate generated human behaviors is to define a downstream task. We can then use the generated human behaviors to train the downstream task and score the behaviors by how much performance improvement is gained. There are two generic tasks that are suitable for most human behavior data. The first task is action recognition, which is for behavior data that comes with semantic labels such as actions. In this case, we can require the behavior generation model to additionally generate the semantic labels, which together with the generated behaviors can be used to train an action recognition model. The second task is behavior forecasting, which can be used for almost any human behavior data even without semantic labels. We can define different behavior forecasting settings with different observation and forecasting horizons, and train behavior forecasting models under these settings. We can use the average performance of different settings to evaluate the quality of generated behaviors.

\section{Lessons Learned and Outlook}
Over the course of the extensive research in human behavior modeling, we have learned many lessons that have helped us tremendously. Below are the most important ones:
\begin{enumerate}
    \item \textbf{Physics simulation does not need to be perfect in order to be useful.} Computer vision researchers sometimes dismiss the use of physics simulation due to its inaccuracies and oversimplified modeling. However, what we found in this thesis is that imperfect physics simulation, when coupled with data-driven methods, can still be very useful for perception and generation tasks such as human pose estimation and human motion generation. The main reason is that the data-driven part of the model (such as the control policy in SimPoE~\cite{yuan2021simpoe}) can absorb the errors in the simulation model and produce the desired behavior. That said, it is still important to have a moderately accurate simulation model so that the errors will not be too difficult for the data-driven part to correct.
    \item \textbf{Generalization ability of simulation-based model at test time.} In most of the time, using physics simulation can improve the model's generalization at test time. This is because simulation ensures that \emph{at test time the laws of physics are still observed}. For example, when a human behavior generation model extrapolates, simulation ensures that the generated behaviors still maintain proper contact with the ground while purely-kinematic methods may have lots of foot sliding. Sometimes, simulation can also lead to worse generalization at test time, which is because the model (such as the control policy in SimPoE) cannot generalize to unseen state space of the simulation. For example, if the control policy has never learned to do backflips in simulation, it will have a hard time imitating the backflips in a test video. To improve generalization in this aspect, the training data needs to have sufficient coverage of the simulation state space of humanoid dynamics. This does not mean that the training data has to cover every human behavior, but it at least needs to have similar dynamic motions in order for the policy to learn backflips.
    \item \textbf{Learning a neural physics simulator is still difficult.} At various stages of the thesis research, we have tried to learn a neural physics simulator, since it could provide us with analytic gradients to help learning. We found that the learned neural simulator, even for humanoid locomotion, has several problems that prevent it from being useful. First, the learned neural simulator does not generalize well outside the training data and is often unstable. This is not surprising since the neural network can overfit to the training data without learning the basic laws of physics. Using physics-inspired network design (such as~\cite{sanchez2020learning} for fluid simulation) could potentially alleviate this problem. Second, the learned neural simulator does not capture the nuanced aspects of physics simulation such as proper contact and friction. This directly makes the simulator less useful since it cannot eliminate physical artifacts such as foot sliding or ground penetration. While it is possible to keep improving approaches for learning a neural simulator, a more promising direction is to develop differentiable physics simulators that are not learned directly from data. These differentiable simulators are typically based on laws of physics and dynamics equations, which allow them to almost always generalize to different parts of state space.
    \item \textbf{The devils are in the details.} Our simulation-based humanoid control framework may seem quite straightforward to implement, but we have had many struggles in making it work and every time the problem was hidden in the details. Here are three useful tips. First, \emph{state normalization matters}. It is extremely important to normalize the states in RL since they can be quite different in a training batch. Traditional normalization techniques such as batch normalization are not very suitable for RL since its normalization process is different for training and sampling. A less-known technique, called running normalization (i.e., using running estimates of mean and standard deviation) is better since it gradually updates the normalization factors while avoiding different normalization for training and sampling. Switching to running normalization almost always significantly improves performance. Second, \emph{predicting residuals is better}. When the policy needs to predict certain actions such as the target PD controller angles, it is wise to make use of some good baselines, such as the input kinematic pose, and only predict the residual to it. This makes policy learning much easier since it already has a good guess and only needs to refine it. Finally, \emph{use egocentric coordinates}. By representing every feature in the state space in the egocentric coordinate of the humanoid, the policy is invariant to its global position and orientation, which allows experience collected under different positions and orientations to be consolidated and distilled into the policy. Moreover, it also prevents the policy from overfitting to specific positions or orientations.
\end{enumerate}
Looking forward, we believe that the three aspects of human behavior modeling --- simulation, perception, and generation will become increasingly integrated. As behavior modeling is moving in the direction of human-scene and human-object interaction, simulation of human behavior will become more and more important for perception and generation since it is key to producing physical-plausible interactions and diverse variations. The perception and generation aspects will also synergize with each other, where perception can provide the necessary context for generation while generation can fill in the blind spots of perception systems. Eventually, we believe that a unified system for simulation, perception, and generation of human behavior will emerge when machine learning, robotics, and computer vision come to fruition with more efficient learning algorithms, advanced differentiable physics simulators, and better perception networks for sensing human behaviors and environments.

\clearpage

\bibliographystyle{finitplain} 
{\small \bibliography{reference.bib}}

\end{document}